%% file: main.tex
\documentclass{article}
\usepackage{aa-report}
\title{\model{}: A Sovereign European Model on the Pareto Frontier}
\author{Aleph Alpha}
\date{October 3, 2026}
\AAAuthorNote{See \hyperref[sec:contributors]{Contributors} for the full author list.}

\AAReportType{Technical Report}
\AAWebsite{https://aleph-alpha.com/}
\AALogoWidth{0.20\linewidth}

\hypersetup{pdfcreationdate={D:20261003000000+02'00'}, pdfmoddate={D:20261003000000+02'00'}}
\ExplSyntaxOn
\AddToHook{begindocument/before}
  {
    \bool_lazy_and:nnT
      { ! \file_if_exist_p:n { \c_sys_jobname_str .aux } }
      { \file_if_exist_p:n { crossrefs.aux } }
      {
        \ior_open:Nn \g_tmpa_ior { crossrefs.aux }
        \iow_open:Nn \g_tmpa_iow { \c_sys_jobname_str .aux }
        \ior_str_map_inline:Nn \g_tmpa_ior { \iow_now:Nn \g_tmpa_iow {#1} }
        \iow_close:N \g_tmpa_iow
        \ior_close:N \g_tmpa_ior
      }
  }
\ExplSyntaxOff
\begin{document}

\maketitle

\begin{abstract}
  \setlength{\parindent}{0pt}\setlength{\parskip}{0.35em}
  \noindent\input{sections/00-abstract}\end{abstract}
\acresetall
\newpage
\clearpage

\tableofcontents
\clearpage

\input{sections/01-introduction}

\input{sections/02-pre-training/00-pre-training}

\input{sections/02-pre-training/01-model-architecture}

\input{sections/02-pre-training/02-optimisation}

\input{sections/02-pre-training/03-data-and-training-stages}
\input{sections/02-pre-training/04-base-model-evaluation}
\input{sections/03-post-training/00-post-training}

\input{sections/03-post-training/01-supervised-fine-tuning}

\input{sections/03-post-training/02-reinforcement-learning}

\input{sections/03-post-training/03-released-model-evaluation}
\input{sections/04-model-factory/00-model-factory}

\input{sections/04-model-factory/01-savanna-model-training-as-code}

\input{sections/04-model-factory/02-savanna-in-bienenelfe-training}
\input{sections/05-summary}

\input{sections/06-legal-due-diligence-example-deep-dive-into-copyright-law}

\input{sections/07-contributors}

\input{sections/08-acknowledgements}
\printbibliography[heading=backmatter]
\clearpage

\clearpage
\appendix

\input{sections/09-appendix/00-appendix}

\input{sections/09-appendix/A-quality-cost-analysis}

\input{sections/09-appendix/B-model-architecture}

\input{sections/09-appendix/C-optimisation}

\input{sections/09-appendix/D-data-and-training-stages}

\input{sections/09-appendix/E-base-model-evaluation}

\input{sections/09-appendix/F-i-don-t-know-evaluations}

\input{sections/09-appendix/G-reasoning-mode-at-long-context}
\input{sections/09-appendix/H-supervised-fine-tuning}
\input{sections/09-appendix/I-reinforcement-learning}

\end{document}

%% file: sections/00-abstract.tex
This report introduces \model{2}, an English--German mixture-of-experts transformer with
\num{\modelTwoTotalParams}B total parameters and \num{\modelTwoActiveParams}B active parameters per token,
released as open weights under the Apache 2.0 licence.\footnote{Weights and model card: \hfmodelcardurl}
We built \model{2} for sovereign and specialised deployment, with a particular focus on German and on regulated domains such as public administration, industry, and aerospace.
To support these settings, we control the full model-development process, including data, architecture, training infrastructure, post-training, and evaluation.

We train \model{2} on \num{24}T tokens across pre-training, mid-training, and long-context extension, with German accounting for more than \qty{20}{\percent} of the data mix by bytes.
To reach this share, we complement German datasets with more than \num{2}T German tokens that we curate from the web with our own pipeline and generate synthetically.
Post-training combines supervised fine-tuning on curated German and English data with reinforcement learning on more than \num{1.2}M internally curated tasks, which cover reasoning, tool use, instruction following, code, and retrieval.
The model activates \qty{\modelTwoActiveShare}{\percent} of its parameters per token, and among the open models evaluated in this report, it lies on the Pareto frontier of quality and serving cost in English and German.

\model{2} builds on \model{1}, an earlier iteration that validated our full pipeline at scale.
Both models come from our Model Factory, which connects data, experimentation, training, evaluation, and deployment through version-controlled workflows.

\begin{figure}[!b]
  \centering
  \includegraphics{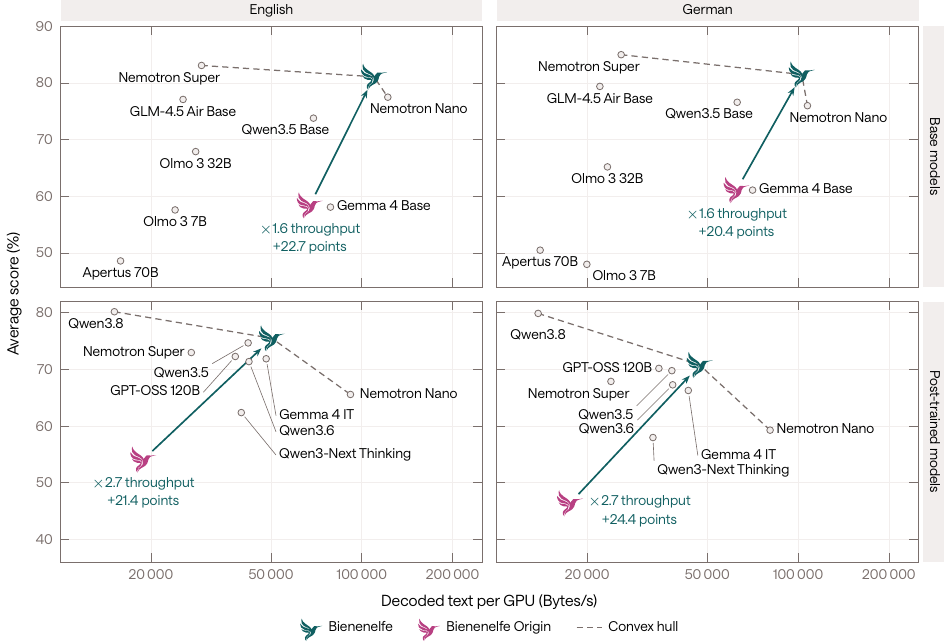}

  \caption{Average English and German benchmark scores of base (top) and post-trained (bottom) models
    against decoded bytes of text per second and GPU, where higher and further right is better.
    Every model runs in its fastest configuration on eight B200 GPUs at a sequence length of \num{4}k (base) or
    \num{16}k (post-trained).
    Scores from \cref{tab:pre-training-evals,tab:post-training-evals,tab:post-training-evals-2}; evals in \cref{sec:eval-suite,sec:post-training-eval-suite}; plot details in \cref{app:quality-cost}.
  }
  \label{fig:quality-cost}
\end{figure}

%% file: sections/01-introduction.tex
\clearpage
\part{Introduction}

Organisations in public administration, industry, and aerospace process sensitive data under regulation.
For these domains, a language model has to run on infrastructure that the organisation controls, rest on training in line with European values, and serve German as well as English.
In addition, the serving cost of the model limits the scale at which an organisation can deploy it.

\begin{figure}[tb]
  \centering
  \includegraphics{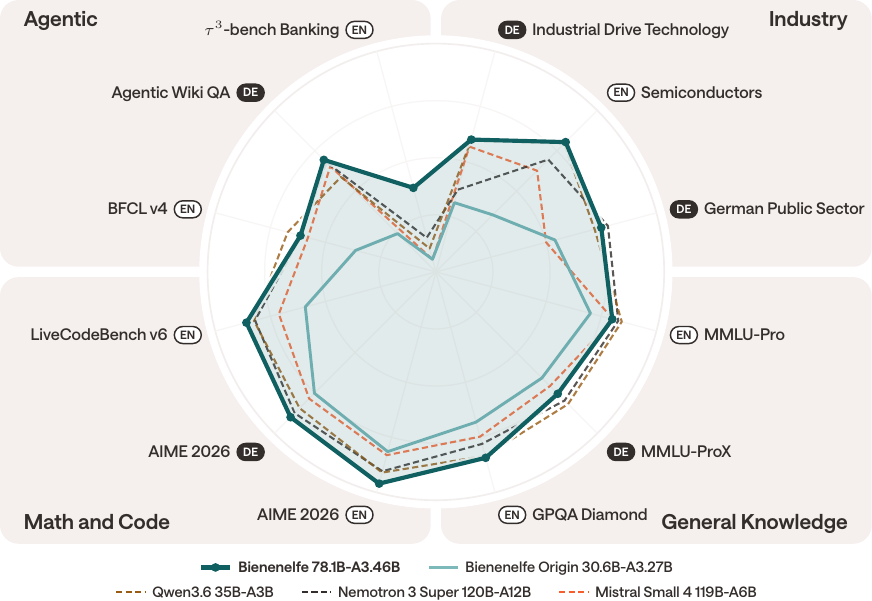}
  \caption{Post-training benchmark scores of \model{2}, \model{1} and three open models in four domains, from \cref{tab:post-training-evals,tab:post-training-evals-2}.
    Badges mark the English (EN) and German (DE) version of a benchmark; the scale runs from \qty{0}{\percent} at the centre to \qty{100}{\percent} at the outer ring.
  }
  \label{fig:intro-spider}
\end{figure}

\model{2},\footnote{\model{} is the German name of the bee hummingbird (\emph{Mellisuga helenae}), the smallest hummingbird.} which we built for these requirements, is an English--German \ac{MoE} transformer with \num{\modelTwoTotalParams}B total parameters, of which \num{\modelTwoActiveParams}B, or \qty{\modelTwoActiveShare}{\percent}, are active per token.
The Apache 2.0 licence of the open weights permits the deploying organisation to operate the model on its own infrastructure.
Among the open models we evaluate, \model{2} lies on the Pareto frontier of quality and serving cost in English and German, both as a base and as a post-trained model (\cref{fig:quality-cost}).
\Cref{fig:intro-spider} shows that \model{2} scores higher than three open models on most benchmarks across Agentic, Industry, Math, Code and General Knowledge.
The model processes contexts of up to \maxContextLengthShort{} tokens, and users set a reasoning effort per request to trade cost and latency against answer quality.

We designed the development pipeline of \model{2} with applicable regulation, such as the EU AI Act, its \acl*{GPAI} Code of Practice and the \acl*{GDPR}, in mind from the ground up.
Our data pipeline filters against illegal, harmful, and pirated content and redacts personal data from data sources.
In addition, we align the answers of the model with the values of the democratic consensus in Europe and train the model to abstain when the provided context does not support an answer.

Teams in Germany developed \model{2} end to end and trained it on infrastructure in Germany and Finland.
Training spans \num{24}T tokens across pre-training, mid-training, and long-context extension, with a German byte share of more than \qty{20}{\percent} and sequences of up to \recommendedContextLengthShort{} tokens in the final stage.
To reach this share, we complement German datasets with more than \num{2}T German tokens that we curate from the web with our own pipeline and generate synthetically.
Post-training combines \ac{SFT} on curated German and English data with \ac{RL} on more than \num{1.2}M internally curated tasks, which cover reasoning, tool use, instruction following, code, and retrieval.
The predecessor of \model{2}, \model{1}, is an internal model that we used to validate the full development pipeline at scale.
From \model{1} to \model{2}, we scaled the data, the model size, the context length, and post-training, and we refined the architecture and the training recipe (\cref{tab:model-comparison}).

\begin{table}[tb]
  \centering
  \small
  \begin{tabular}{lll}
    \toprule
    Property
    & \model{2} & \model{1} \\
    \midrule
    Size
    & \num{\modelTwoTotalParams}\mbox{B-A}\num{\modelTwoActiveParams}\mbox{B}
    & \num{\modelOneTotalParams}\mbox{B-A}\num{\modelOneActiveParams}\mbox{B} \\
    Training tokens (all stages)
    & \num{24}\mbox{T} & \num{8}\mbox{T} \\
    Supported context length
    & \maxContextLengthShort{} tokens & \num{64}\mbox{k} tokens \\
    \bottomrule
  \end{tabular}
  \caption{ Main differences between \model{1}, an internal model that is not released, and \model{2}.
    Active parameters exclude the input embedding and include the output layer.
    \Cref{tab:bienenelfe-architecture} lists the full architecture of both models.
  }
  \label{tab:model-comparison}
\end{table}

The entire pipeline, from data curation through pre- and post-training to evaluation, runs in our Model Factory, which comprises the infrastructure, tooling, and development processes that connect data, experimentation, training, evaluation, and deployment.
The Model Factory expresses these workflows as version-controlled code, which is designed to make experiments reproducible, composable, and traceable.
A new model generation, an additional language, or a further specialisation thus becomes a change to this code.

The remainder of this report follows the development of \model{2}.
\Cref{part:pre-training} describes the architecture, the optimisation, the data, and the stages of pre-training, mid-training, and long-context extension, together with the experiments behind the design choices, and evaluates the base model.
Post-training follows in \cref{part:post-training}, which covers \ac{SFT} and \ac{RL} and evaluates the post-trained model.
\Cref{sec:model-factory} describes the Model Factory, \cref{part:summary} summarises the results, and the part on \hyperref[part:legal]{legal due diligence} shows how we assess the copyright compliance of third-party datasets.
The \hyperref[part:appendix]{appendices} give the details of the individual stages and the quality-cost analysis.
\clearpage

%% file: sections/02-pre-training/00-pre-training.tex
\clearpage
{\titlespacing*{\part}{0pt}{0pt}{1.5\AALead}\part{Pre-Training}\label{part:pre-training}}

This part covers the three stages that produce \basemodel{2}: pre-training, mid-training and long-context extension.

\begin{figure}[t]
  \centering
  \includegraphics{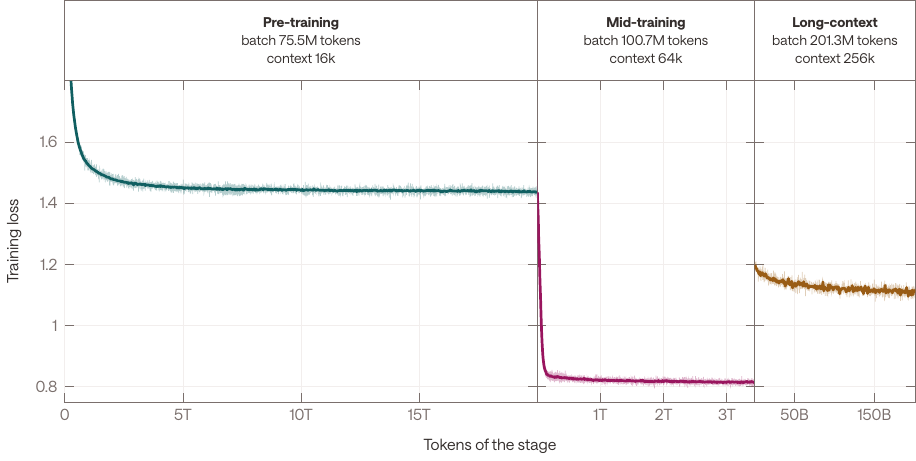}
  \caption{Training loss of the three stages of \model{2}, each against the tokens of that stage and on its own token axis.
    Mid-training starts from the pre-training checkpoint at step \num{264750} and long-context extension from the mid-training checkpoint at step \num{34200}.
    Thin lines are the per-step loss at a sample of the steps, thick lines its mean over bins of \num{20}B, \num{5}B and \num{1}B tokens.
    Each stage trains on its own mix, so the loss levels of different stages are not comparable.
  }
  \label{fig:training-loss}
\end{figure}

After pre-training for \num{20}T tokens at a \preTrainingSeqLen{} context length on data from nine domains (\cref{sec:pre-training-data}), mid-training continues for \num{3.4}T tokens at a context length of \midTrainingSeqLen{} tokens, on a mix drawn mostly from curated datasets rather than from web text (\cref{sec:mid-training}).
The long-context extension then adds \num{200}B training tokens at a context length of \longContextSeqLen{} tokens (\cref{sec:long-context}).
\Cref{fig:training-loss} shows the training loss throughout pre-training.

Every stage was scaled with respect to \model{1}, which pre-trained on \num{7.5}T tokens at a context of \num{4}k tokens, mid-trained on \num{122}B tokens at \num{8}k and extended the context on \num{96}B tokens at \num{64}k.
\model{2} trains on about three times as many tokens in total, at four times the maximum context.
Mid-training grows the most, by a factor of about \num{28}.

We start by describing the model architecture and how we run the pre-training in \cref{sec:architecture,sec:infrastructure}, followed by our specific optimisation settings in \cref{sec:optimization}.
\Cref{sec:pre-training-data} describes the pre-training data: the \ac{CC} pipeline, the mixing of the data sources and \ac{PII} redaction.
\Cref{sec:mid-training,sec:long-context} describe the data of the two later stages, the decontamination of the mid-training pool, and the ablations that selected their mixes.
\Cref{sec:pre-training-eval-scores} defines the evaluation suite that the ablations of the following sections report scores on and describes the benchmark scores of the resulting base model.

\begin{figure}[t]
  \centering
  \resizebox{\linewidth}{!}{%
    \includegraphics{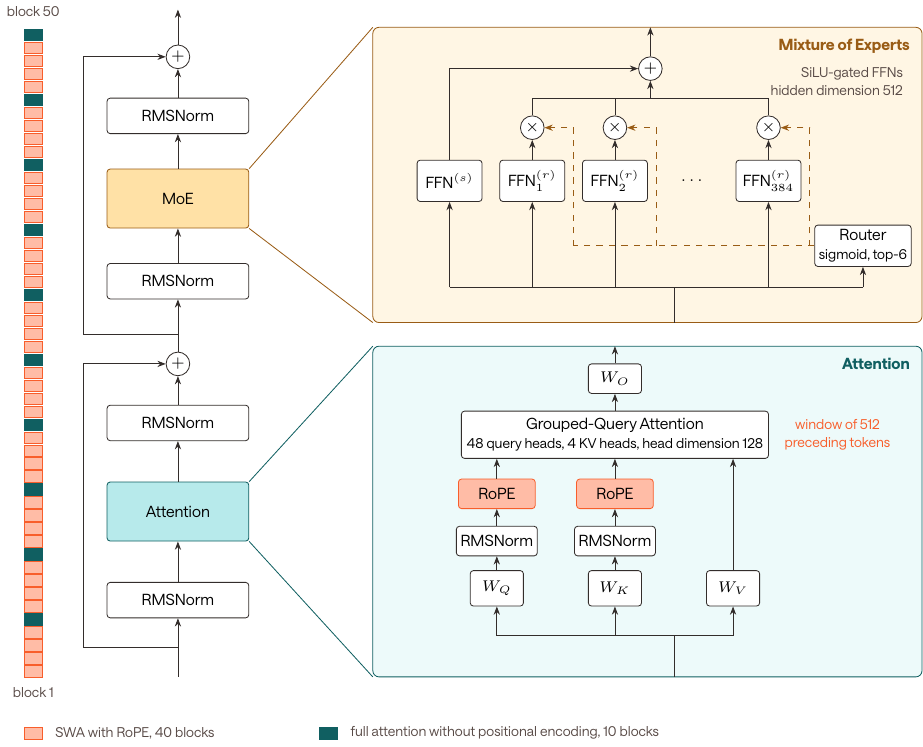}%
  }
  \caption{Architecture of one \model{2} block, which all \num{50} blocks share.
    The attention and \acs*{MoE} sublayers each read from the residual stream through an RMSNorm and write back through a second one.
    Orange marks what only the \num{40} \acs*{SWA} blocks have: \acs*{RoPE} on queries and keys, and a mask over the \num{512} preceding tokens and the query token.
    The strip on the left shows which blocks use full attention.
    The router scores all \num{384} routed experts, sends the token to the \num{6} with the highest scores and weights their outputs by those sigmoid scores \(g_i\) without renormalisation.
  }
  \label{fig:architecture-hybrid}
\end{figure}

%% file: sections/02-pre-training/01-model-architecture.tex
\section{Model Architecture}
\label{sec:architecture}
\model{2} is a decoder-only causal transformer \parencite{vaswani2017attention} with hybrid attention, which interleaves \ac{SWA} and full-attention blocks, and sparse \ac{MoE} layers of \acp{FFN} \parencite{shazeer2017}.
It has \num{\modelTwoTotalParams}B total parameters and \num{\modelTwoActiveParams}B active parameters per token.
Therefore, each token uses \qty{\modelTwoActiveShare}{\percent} of the parameters, keeping the computation per token small at a large total capacity (\cref{sec:scaling-sparsity}).
\Cref{fig:architecture-hybrid} shows one block and \cref{tab:bienenelfe-architecture} lists its dimensions.
Each block uses sandwich normalisation \parencite{ding2021cogview,singh2026arceetrinitylargetechnical}, applying \ac{RMS} normalisation \parencite{zhang2019rmsnorm} before and after both the attention and \ac{MoE} layers.

\pgfplotstableread[col sep=comma, row sep=\\, format=inline]{%
  de_fewer_gpt5\\%
  11.2\\%
}\compressionFacts
\pgfplotstableread[col sep=comma, row sep=\\, format=inline]{%
  de_prec_bienenelfe,de_prec_gpt5\\%
  65.2,47.3\\%
}\morphologyFacts
The tokeniser vocabulary of \num{128000} tokens is trained with UniBPE (\cref{sec:tokenizer}), which builds merges bottom-up like \ac{BPE} but selects each merge by Unigram loss.
Compared with the GPT-5 tokeniser, a larger share of its token boundaries fall on morpheme boundaries, in German \morphologyPercent{de_prec_bienenelfe} against \morphologyPercent{de_prec_gpt5}, and needs \compressionPercent{de_fewer_gpt5} fewer tokens on German web text.

Attention uses \ac{GQA} \parencite{ainslie2023gqa} with \num{48} query heads, \num{4} \ac{KV} heads and head dimension \num{128}.
The blocks combine \ac{SWA} and full attention at a \(4{:}1\) ratio: every fifth block uses full attention, and the other blocks use \ac{SWA} \parencite{beltagy2020longformer} over the \num{512} preceding tokens.
Positional information is injected only in the \ac{SWA} blocks using \ac{RoPE} \parencite{su2024roformer} with \(\theta = \num{10000}\).
The full-attention blocks use \ac{NoPE}.
Queries and keys are normalised per head with \ac{RMS} normalisation \parencite{henry2020querykey}, in the \ac{SWA} blocks before \ac{RoPE}.
\Cref{sec:attention} compares the hybrid attention pattern with full attention, selects the window size and gives the reasons for the positional encoding and QK normalisation.

Every \ac{MoE} layer routes each token to \num{6} of \num{384} routed experts and also evaluates one shared expert \parencite{dai2024deepseekmoe}.
The experts are \ac{SiLU}-gated \acp{FFN} \parencite{shazeer2020gluvariantsimprovetransformer}.
The total parameter count follows from the sparsity sweep of \cref{sec:scaling-sparsity}.
The routing scores are sigmoids of the router logits \parencite{deepseekai2024deepseekv3}, but are not renormalised after top-\(K\) selection.
Expert loads are balanced with \ac{EQB} and \ac{LEI} \parencite{neitemeier2026exactquantilebalancing} (\cref{sec:load-balancing}).

\model{1} has a simpler architecture (\cref{tab:bienenelfe-architecture}), in which every block uses full attention with \ac{RoPE}, the first two blocks are dense, and each \ac{MoE} layer routes a token to \num{8} of \num{128} experts.
It is less sparse, with \qty{\modelOneActiveShare}{\percent} of its parameters active per token against \qty{\modelTwoActiveShare}{\percent} for \model{2}.
Its expert loads are balanced with the combination of \ac{SMEBU} \parencite{singh2026arceetrinitylargetechnical} and a GShard-style auxiliary loss \parencite{lepikhin2021gshard}.

\begin{table}[t]
  \centering
  \small
  \begin{tabular}{lll}
    \toprule
    Property & \model{2} & \model{1} \\
    \midrule
    Transformer blocks                  & \num{50}                            & \num{50} \\
    Leading dense layers                & \num{0}                             & \num{2} \\
    Full attention / \acs*{SWA} blocks  & \num{10} / \num{40}                 & \num{50} / \num{0} \\
    \acs*{SWA} window size              & \num{512} preceding tokens          & -- \\
    Model width                         & \num{2560}                          & \num{2048} \\
    Vocabulary                          & \num{128000}                        & \num{96000} \\
    Attention heads / \acs*{KV} heads   & \num{48} / \num{4}                  & \num{32} / \num{4} \\
    Attention head dimension            & \num{128}                           & \num{128} \\
    Dense \acs*{FFN} hidden dimension   & --                                  & \num{6144} \\
    Routed experts per \acs*{MoE} layer & \num{384}                           & \num{128} \\
    Experts per token                   & \num{6} routed and \num{1} shared   & \num{8} routed and \num{1} shared \\
    Routed-expert hidden dimension      & \num{512}                           & \num{768} \\
    Total parameters                    & \num{\modelTwoTotalParams}\mbox{B}  & \num{\modelOneTotalParams}\mbox{B} \\
    Active parameters per token         & \num{\modelTwoActiveParams}\mbox{B} & \num{\modelOneActiveParams}\mbox{B} \\
    Trained maximum context             & \num{256}k tokens                   & \num{64}k tokens \\
    \bottomrule
  \end{tabular}
  \caption{Architecture of \model{2} and, for comparison, \model{1}.
    Active parameters exclude the input embedding and include the output layer.
  }
  \label{tab:bienenelfe-architecture}
\end{table}

\subsection{Ablation Methodology}
\label{sec:pretraining-ablations}

We compare recipe choices through controlled proxy-scale ablations before validating at larger scale.
Each study specifies its resource-matching criteria, training budget and evaluation stage, with checkpoints compared at a common training horizon.

\paragraph{Proxy Models}

We run ablations on two smaller \emph{proxy models}: versions of \model{2} which share the main architectural choices but have fewer parameters (\cref{tab:proxy-models}).
Both proxy models differ in width and have fewer routed experts per layer at a lower dimension compared to the target \model{2}.
The proxy at scale M has \num{0.6}B active out of \num{7.8}B total parameters trained on \num{97.6}B tokens.
The proxy at scale L has \num{2.1}B active out of \num{30.6}B total parameters trained on \num{327.6}B tokens.

\begin{table}[h]
  \centering
  \small
  \begin{tabular}{lrrr}
    \toprule
    Scale & M & L & Target (\model{2}) \\
    \midrule
    Transformer blocks
    & \num{50}
    & \num{50}
    & \num{50} \\
    Model width
    & \num{1024}
    & \num{2048}
    & \num{2560} \\
    Attention heads / \acs*{KV} heads
    & \num{16} / \num{2}
    & \num{32} / \num{4}
    & \num{48} / \num{4} \\
    Routed experts per layer
    & \num{256}
    & \num{256}
    & \num{384} \\
    Routed experts per token
    & \num{6}
    & \num{6}
    & \num{6} \\
    Routed-expert hidden dimension
    & \num{192}
    & \num{384}
    & \num{512} \\
    Total parameters
    & \num{7.8}\mbox{B}
    & \num{30.6}\mbox{B}
    & \num{\modelTwoTotalParams}\mbox{B} \\
    Active parameters per token
    & \num{0.6}\mbox{B}
    & \num{2.1}\mbox{B}
    & \num{\modelTwoActiveParams}\mbox{B} \\
    Pre-training tokens (steps)
    & \num{97.6}\mbox{B} (\num{10342})
    & \num{327.6}\mbox{B} (\num{13016})
    & \num{20}\mbox{T} \\
    Long-context tokens (steps)
    & \num{5}\mbox{B} (\num{299})
    & \num{15}\mbox{B} (\num{448})
    & \num{200}\mbox{B} \\
    \bottomrule
  \end{tabular}
  \caption{Scale M and L proxy models, used for ablation studies next to the target configuration of \model{2} (\cref{tab:bienenelfe-architecture}).
    Tokens are the pre-training horizon of each scale, in batches of \num{576} (M) and \num{1536} (L) sequences of \preTrainingSeqLen{} tokens.
  }
  \label{tab:proxy-models}
\end{table}

\paragraph{Evaluation}
We evaluate every candidate change along three axes: modelling quality, stability in the attention and \ac{MoE} layers, and efficiency, both in training and decode during inference.
To assess modelling quality, we use validation loss and a suite of benchmarks, which are described in \cref{sec:eval-suite}.
Because we broadened the evaluation suite as the model was developed, aggregates are not directly comparable across studies.
To ablate the impact on long-context performance, some studies applied a shortened long-context phase directly after pre-training.

\paragraph{Learning-Rate Selection}
Architectural changes can shift the optimal learning rate, so using only the scaling-law prediction can confound shape quality with a suboptimal learning rate.
For model-shape sweeps, we tune learning rates at the smallest ablation scale using \(\eta \in \{\eta_0/\sqrt{2}, \eta_0, \sqrt{2}\eta_0\}\), where \(\eta_0\) is the predicted learning rate.

\subsection{Tokeniser}
\label{sec:tokenizer}

We designed the \model{2} tokeniser to capture the specificity of the German language, without compromising on English efficiency and downstream model performance.

The model uses a vocabulary of \num{128000} tokens, up from \num{96000} tokens used by \model{1}.
We train both \model{} tokenisers with \emph{UniBPE}, a new hybrid tokeniser training method: it keeps the greedy bottom-up approach of \ac{BPE} \parencite{gage1994new,sennrich2016neural} and scores merges with the Unigram loss objective \parencite{kudo2018subword}.

\paragraph{UniBPE Merge Score}
A tokeniser state assigns every token \(t\) a count \(c_t\) over the training corpus, with \(N = \sum_t c_t\) tokens in total.
The Unigram model puts probability \(c_t / N\) on \(t\), and the negative log-likelihood of the corpus under it is
\begin{equation}
  \mathcal{L} = -\sum_{t} c_t \log\frac{c_t}{N} = h(N) - \sum_t h(c_t), \qquad h(x) = x\log x,\ h(0) = 0 .
  \label{eq:tokenizer-loss}
\end{equation}
This is the quantity Unigram tokenisers minimise top-down by pruning tokens.
UniBPE minimises it bottom-up: at each step it applies the merge that lowers \(\mathcal{L}\) the most, and otherwise runs exactly like \ac{BPE}, producing an ordinary vocabulary and merge list.

Merging the pair \((a, b)\) into a new token \(ab\), where the pair occurs \(c_{ab}\) times, moves \(c_{ab}\) counts out of \(a\) and out of \(b\), creates a token with count \(c_{ab}\), and shortens the corpus by \(c_{ab}\) tokens.
Every other count is untouched, so for \(a \neq b\) the change of the loss
\begin{equation}
  \Delta\mathcal{L} = \underbrace{h(c_a) - h(c_a - c_{ab})}_{a \text{ loses } c_{ab}} + \underbrace{h(c_b) - h(c_b - c_{ab})}_{b \text{ loses } c_{ab}} - \underbrace{h(c_{ab})}_{\text{new } ab \text{ token}} + \underbrace{h(N - c_{ab}) - h(N)}_{\text{shorter corpus}}
  \label{eq:tokenizer-merge-delta}
\end{equation}
is closed-form.
\Ac{BPE} ranks candidates by \(c_{ab}\) alone; UniBPE ranks them by \(\Delta\mathcal{L}\), the most negative first.
Both only need the pair count and the two parent counts, so no new information is required over \ac{BPE}.

\paragraph{Training}
We train the tokeniser on a shuffled \qty{100}{\giga\byte} slice of the pre-training dataset (\cref{sec:pre-training-data}) without Unicode normalisation.
For pre-tokenisation we adapt the GPT-4 pre-tokeniser \parencite{openai2023tiktoken}, changing the regex to split every digit into its own token.
The tokeniser has byte fallback, starting with \num{256} byte tokens.
The next \(\num{120000} - \num{256} = \num{119744}\) merges use the UniBPE algorithm, standard \ac{BPE} performs the next \num{7900} merges, and we then add \num{100} reserved and special tokens, for a total vocabulary of \num{128000} tokens.
The second stage of standard \ac{BPE} has two purposes.
First, some useful merges are difficult under UniBPE, such as \token{..}, preventing \token{...} from being a single token.
Second, it restores the compression that pure UniBPE, as used for \model{1}, loses.
This hybrid approach matches the compression of standard \ac{BPE} in our experiments.

\paragraph{Compression}

\pgfplotstableread[col sep=comma, row sep=\\, format=inline]{%
  de_bienenelfe,de_best_name,de_best,en_bienenelfe,en_best_name,en_best\\%
  4.90,GPT-5,4.35,4.58,GPT-5,4.67\\%
}\compressionFacts
\begin{figure}[tb]
  \centering
  \includegraphics{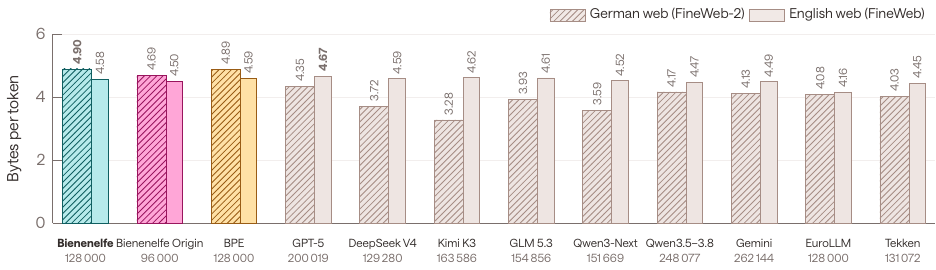}
  \caption{Compression of the \model{} tokenisers and other tokenisers on German text from FineWeb-2 \parencite{penedo2025fineweb2} and English text from FineWeb \parencite{penedo2024fineweb}.
    The compression is given in UTF-8 bytes per token (higher is better); the wide hatched bars are German, the narrow ones English.
    The number under each name denotes the vocabulary size.
    \acs*{BPE} is plain \acl*{BPE} trained on the same corpus with the same vocabulary size as \model{2}.
    \emph{Tekken} is Mistral's tokeniser \parencite{mistral2024nemo}, also used by Apertus and Nemotron, and \emph{Gemini} the tokeniser shared by Gemma~4 and Gemini \parencite{gemma2026gemma4technicalreport}.
    GPT-5, DeepSeek V4 and EuroLLM denote the tokenisers of these models \parencite{openai2023tiktoken,deepseekai2026deepseekv4,martins2024eurollm}.
  }
  \label{fig:tokenizer-web-compression}
\end{figure}

On German text, the \model{2} tokeniser reaches \compressionValue{de_bienenelfe} bytes per token, the highest compression across twelve evaluated tokenisers in \cref{fig:tokenizer-web-compression}, with the best of the nine other tokenisers being \compressionValue{de_best} for \compressionName{de_best_name}, and is on par with \ac{BPE} trained on the same pre-training corpus.
On English, \model{2} reaches \compressionValue{en_bienenelfe} bytes per token, against \compressionValue{en_best} for \compressionName{en_best_name}, the best tokeniser on English.
This lowers the cost of training on and serving German text, and leads to shorter context lengths in long-context tasks while staying competitive in English.

\paragraph{Morphology}
\pgfplotstableread[col sep=comma, row sep=\\, format=inline]{%
  en_prec_bienenelfe,en_prec_best,de_prec_bienenelfe,de_prec_best\\%
  50.0,42.2,65.2,48.7\\%
}\morphologyFacts

UniBPE consistently improves on morphological alignment compared to plain \ac{BPE}.
In the first \num{60} merges each performs on the pre-training dataset, UniBPE creates the units of three or more letters \token{ing}, \token{the}, \token{ the}, \token{and}, \token{ion}, \token{tion}, \token{for}, \token{ and}, \token{ation}, \token{pro}, \token{that}, \token{con}, \token{ent}, \token{ment}, \token{ with}, \token{ther} and the German \token{und}, \token{ung}, \token{ver}, and the German characters \token{ü}, \token{ä}, \token{ö}, \token{ß}.
\Ac{BPE} has no German units or characters, and only five units of three or more letters, \token{ the}, \token{ing}, \token{ion}, \token{ent} and \token{ and} by merge \num{60}.

We quantify this morphological alignment in \cref{sec:appendix-tokenizer-morphology}.
The \model{2} tokeniser splits words at morpheme boundaries with a precision of \morphologyPercent{de_prec_bienenelfe} in German and \morphologyPercent{en_prec_bienenelfe} in English, compared to \morphologyPercent{de_prec_best} and \morphologyPercent{en_prec_best} respectively for the best of the external tokenisers.

\begin{figure}[tb]
  \centering
  \includegraphics{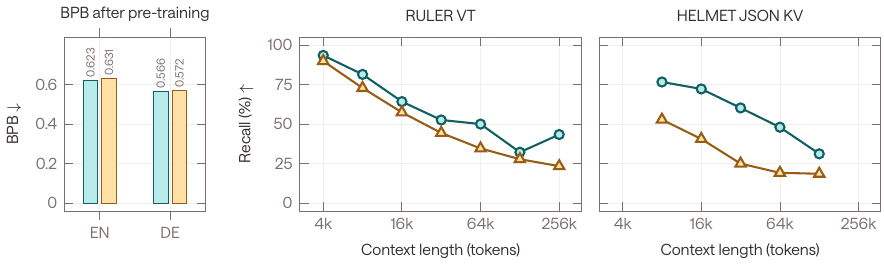}
  \includegraphics{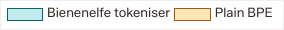}
  \caption{Proxy at scale M trained with the \model{2} tokeniser and with plain \acs*{BPE} of the same vocabulary size.
    Left: \acs*{BPB} averaged over the English and German pre-training benchmarks after pre-training.
    Middle and right: recall by context length after the long-context stage, the recall of assigned variables for RULER \acs*{VT} and the ROUGE-L recall of the retrieved value for HELMET JSON \acs*{KV}.
  }
  \label{fig:tokenizer-ablation}
\end{figure}

\paragraph{Downstream Model Performance}
We ran a \model{2} tokeniser vs \ac{BPE} ablation at scale M (see \cref{tab:proxy-models}).
For both training runs, \cref{fig:tokenizer-ablation} compares the \ac{BPB} after pre-training and the recall on two retrieval tasks after the long-context stage.
The \model{2} tokeniser reaches a slightly lower \ac{BPB} in English and German, so it at least matches \ac{BPE} in language-modelling quality, while the \model{2} tokeniser improves recall on long-context \ac{VT} and \ac{KV} retrieval at every length.

\subsection{Hybrid \actitle{SWA}}
\label{sec:attention}

One target of \model{2} is to improve long-context inference efficiency compared with \model{1}, which uses full attention in every block: each token attends to all preceding tokens, so the compute grows quadratically with the sequence length in all layers.
During inference, the \ac{KV} cache grows linearly with the sequence length, leading to increased memory usage and computational cost.
A mitigation is to use a hybrid of full attention and \ac{SWA} \parencite{beltagy2020longformer} layers, where the latter restrict attention to a fixed-size window of preceding tokens, reducing both memory usage and computational cost.
We integrate \ac{SWA} into \model{2}, increasing long-context inference efficiency and benchmark performance.

\begin{figure}[tb]
  \centering
  \includegraphics{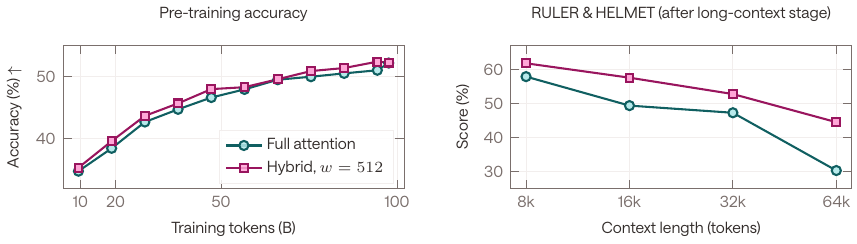}
  \caption{Average accuracy along the pre-training stage (left) and long-context score after a \num{5}\mbox{B} long-context stage (right).
    We evaluate a model with full attention in every block and a hybrid model with \(w=\num{512}\), both approximately \acs*{FLOP}-matched at proxy scale M (\cref{tab:proxy-models}).
    Full attention uses \num{14} heads per block and pre-trains on sequences of \num{4}k tokens, the hybrid on sequences of \num{16}k tokens.
    During the long-context stage, the full-attention model is trained on sequences of \num{64}k tokens with \acs*{YaRN} \parencite{peng2024yarn}, while the hybrid model is trained on sequences of \num{256}k tokens (without any \acs*{RoPE} extension).
    Left: log-likelihood accuracy averaged over \acs*{ARC}, HellaSwag and \acs*{MMLU} in English and German at raw pre-training checkpoints.
    Right: mean RULER score (\acs*{NIAH}, \acs*{VT}, \acs*{WE} and \acs*{QA}) and HELMET score (JSON-KV and RAG-QA), up to the \num{64}k tokens the full-attention model was trained on.
  }
  \label{fig:hybrid-attention}
\end{figure}

\paragraph{Hybrid Attention}
\model{2} uses full attention in every fifth block (\num{10} of \num{50}) and \ac{SWA} in the other \num{40}.
First experiments at proxy scale M (\cref{tab:proxy-models}) showed improvements in long-context tasks without a loss on pre-training benchmarks (\cref{fig:hybrid-attention}).
At approximately matched training compute, the hybrid model supports sequences four times as long as the full-attention model.
Moreover, \cref{fig:sparsity-quality-throughput} (right) shows that the hybrid model is generally more efficient during inference.

\paragraph{Positional Encoding}
The \ac{SWA} blocks apply \ac{RoPE} \parencite{su2024roformer} with base \(\theta=\num{10000}\) to queries and keys; the full-attention blocks use \acf{NoPE}.
In early experiments, removing \ac{RoPE} from the full-attention blocks improved long-context performance, in line with \textcite{yang2025ropetonope}.

\paragraph{QK Normalisation}
Every block normalises queries and keys per head with \ac{RMS} normalisation (in the \ac{SWA} blocks before \ac{RoPE}) \parencite{henry2020querykey,zhang2019rmsnorm}, which bounds the attention logits and keeps training stable at higher learning rates.
We refer to \cref{app:qk-norm} for an analysis of QK normalisation effects.
During \ac{RL}, assuming sufficiently small gain parameters, the QK normalisation also bounds every query and key element below the FP8 range, so the FP8 \ac{KV} cache cannot overflow (\cref{app:rl-fp8}).

\subsubsection{Window Size}
\label{sec:window-size}
We select the window size \(w^* \in \{128, 256, \dots, 4096\}\) by running ablations at scale M and validating the best two variants at scale L (\cref{tab:proxy-models}).

After the pre-training stage, we add a long-context stage of \num{5}B tokens at scale M and \num{15}B tokens at scale L on sequences of \num{256}k tokens.
We evaluate the pre-training benchmarks at the last pre-training checkpoint and the long-context benchmarks after the long-context stage.
Across windows, we approximately match total \acp{FLOP}, \ac{KV}-cache size and parameter count (\cref{tab:window-resource-matching}).

\begin{table}[tb]
  \centering
  \small
  \setlength{\tabcolsep}{4pt}
  \begin{tabular}{rrrrrrrrrr}
    \toprule
    \shortstack[r]{Window\\size \(w\)}&\shortstack[r]{\acs*{SWA} Q/\acs*{KV}\\heads}&Experts&\shortstack[r]{Expert\\hidden}&\shortstack[r]{T\acsp*{FLOP}\\per seq. }&\shortstack[r]{\acs*{KV} MiB\\per seq. }&\shortstack[r]{Total\\params}&\shortstack[r]{Active\\params}&\shortstack[r]{Accuracy\\(\%) \(\uparrow\)}&\acs*{BPB} \(\downarrow\)\\
    \midrule
    128  & 32/32 & 144 & 320 & 61.3 & 400 & 7.8B & 1.2B & 35.2          & 0.641 \\
    256  & 32/32 & 144 & 320 & 62.7 & 480 & 7.8B & 1.2B & \textbf{35.8} & \textbf{0.637} \\
    512  & 32/16 & 144 & 320 & 60.0 & 480 & 7.6B & 1.0B & 35.5          & 0.647 \\
    1024 & 24/8  & 96  & 512 & 63.7 & 480 & 7.9B & 1.0B & 35.3          & 0.643 \\
    2048 & 16/4  & 96  & 512 & 62.3 & 480 & 7.8B & 0.9B & 34.9          & 0.650 \\
    4096 & 16/2  & 256 & 192 & 61.8 & 480 & 7.7B & 0.6B & 35.2          & 0.642 \\
    \bottomrule
  \end{tabular}
  \caption{Resource-matched window-size sweep at scale M with adapted \acs*{SWA} heads, routed experts and expert width that match the forward \acsp*{FLOP} and \acs*{KV} cache per sequence of \num{16}k tokens and the total parameters across windows.
    The configuration at window size \num{128} uses a smaller \acs*{KV} cache because an exact match would require an impractically large number of \acs*{KV} heads.
    \emph{Accuracy} denotes the average of \num{15} accuracy metrics, one per benchmark, at the last pre-training checkpoint (higher is better).
    \emph{\acs*{BPB}} denotes the average of \num{13} \acs*{BPB} metrics at the last pre-training checkpoint (lower is better).
    The best accuracy and \acs*{BPB} are set in bold.
  }
  \label{tab:window-resource-matching}
\end{table}

\paragraph{Ablations at Scale M}
We select the window for long-context score and inference efficiency, and also compare the scores on the pre-training benchmarks.

\Cref{tab:window-resource-matching} also reports the average accuracy and \ac{BPB} of our pre-training benchmarks at the last pre-training checkpoint.
While the scores are generally similar across window sizes, the configuration with window size \num{256} achieves the highest scores.

\begin{figure}[t]
  \centering
  \includegraphics{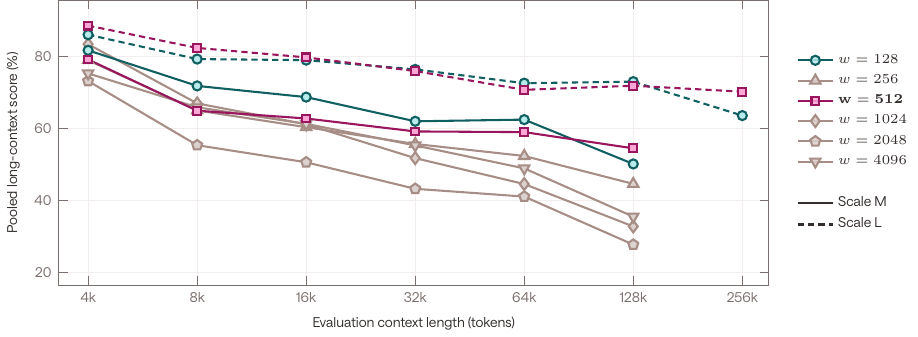}
  \caption{Pooled long-context scores by window size \(w\) after the long-context stage, in the resource-matched sweep at scale M (solid) and in the validation at scale L (dashed).
    Each point denotes the average of RULER \acs*{NIAH}, \acs*{VT}, \acs*{WE} and \acs*{QA}, and the HELMET JSON-KV and RAG-QA F1-scores evaluated at that length.
    At the longest evaluated context, window size \num{512} leads both sweeps.
    Axis labels use \(k=1024\) tokens.
  }
  \label{fig:window-long-context}
\end{figure}

\Cref{fig:window-long-context} shows the pooled long-context score for each window size configuration depending on the context length.
At scale M, \(w=128\) yields the highest pooled score and leads from \num{8}k to \num{64}k tokens, but degrades at context length \num{128}k.
\(w=512\) follows closely and degrades the least with context length.

\paragraph{Validation at Scale L}
Based on the ablations at scale M, we validate \(w=128\) and \(w=512\) at scale L (\cref{tab:proxy-models}).
\Cref{fig:window-long-context} also reports the pooled long-context score for these models.
Their average accuracies on the pre-training benchmarks are \qty{48.0}{\percent} for \(w=128\) and \qty{47.8}{\percent} for \(w=512\), while their pooled long-context scores are \qty{76.2}{\percent} and \qty{77.1}{\percent}.
Through \num{128}k tokens, the two windows are generally close, with the largest absolute difference being \num{3.1} percentage points.
At context length \num{256}k, however, \(w=512\) scores \num{6.7} percentage points higher than \(w=128\).

We therefore use a sliding window of \(w=512\) in \model{2}, at the cost of \num{0.2} percentage points of accuracy on the pre-training benchmarks.

\subsection{Scaling Sparsity}
\label{sec:scaling-sparsity}

\begin{figure}[htbp]
  \centering
  \includegraphics{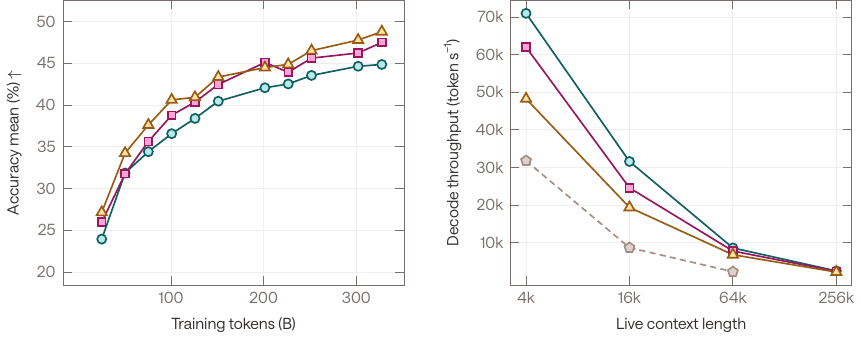}
  \includegraphics{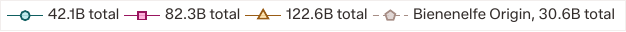}
  \caption{Accuracy mean during pre-training (left) and maximum pure-decode throughput by live context length (right) of three configurations with \num{3.4}\mbox{B} active parameters and different total parameters.
    Accuracy averages the \num{17} benchmarks of \cref{app:model-shape-metrics}.
    Throughput is measured on eight H100 SXM5 GPUs in BF16, with the best tested parallel configuration for each model and context length.
    \model{1} is shown for reference over its supported context range.
  }
  \label{fig:sparsity-quality-throughput}
\end{figure}

Sparse \ac{MoE} architectures are able to increase model capacity at matched compute by routing each token through a subset of experts.
We measure their sparsity as the share of the parameters that is active for a token.
At fixed expert width and routing top-\(K\), adding experts increases the total parameter count while leaving the number of \acp{FLOP} per token approximately unchanged.
A larger total parameter count increases parameter storage, communication volume and optimiser states, and makes expert balance harder to maintain.
Downstream performance, however, improves substantially with total capacity~\parencite{clark2022unified, frantar2024scaling, ludziejewski2024scaling, abnar2025parameters, tian2025leverage, kimiteam2025k2}, a trend we confirm in our own ablations (\cref{fig:sparsity-quality-throughput}).
We compare configurations of different sparsity by their quality, inference efficiency and deployment cost.

\paragraph{Quality}

We vary sparsity by increasing the number of routed experts from \num{128} to \num{256} and \num{384} at \num{3.4}B active parameters, which gives \num{42.1}B, \num{82.3}B and \num{122.6}B total parameters.

Moving from \num{42.1}B to \num{82.3}B improves loss and aggregate eval accuracy (\cref{fig:sparsity-quality-throughput}).
The \num{122.6}B configuration improves loss and aggregate eval accuracy only slightly further.
The load balancing of \cref{sec:load-balancing} keeps the expert loads well balanced in all three configurations, including the \num{122.6}B one.
A preliminary sweep at \num{0.9}B active parameters shows the same ordering in loss and accuracy.
\Cref{app:sparsity-sweep} shows the loss and accuracy of both sweeps and the deployment footprint of each configuration.

\paragraph{Inference Efficiency}
\pgfplotstableread[col sep=comma, row sep=\\, format=inline]{%
  p2_candidate_128,p2_candidate_256,p2_candidate_384\\%
  70947.62033819828,61963.52064072338,48236.521450354834\\%
  ,,\\%
  ,,\\%
  2477.775679586479,2381.843945891289,2179.7562912755025\\%
}\decodeThroughput
We focus on the inference regime that is most interesting for the \ac{RL} phase, where most of the runtime is spent decoding rollouts.
We therefore compare the configurations by their decode throughput: the tokens generated per second, summed over all concurrent requests.
Since the decode phase is mainly memory-bound, increasing total parameters can increase inference cost through greater parameter storage and weight movement, as well as additional communication overhead.
To capture these effects, we measure this throughput for the three configurations with \num{3.4}B active parameters on \num{8} H100 SXM5 GPUs in BF16, using for each model and context length the best tensor-, data- and expert-parallel configuration identified through a combinatorial search (\cref{fig:sparsity-quality-throughput}).

The main sources of memory traffic are the active model weights and the \ac{KV} cache.
Weight reads increase with the total parameter count, although they can be amortised across tokens routed to the same experts within a large batch.
\Ac{KV}-cache reads, by contrast, remain specific to each request and grow with its live context.
At \num{4}k context length, where weight reads account for a larger share of memory traffic, the \num{82.3}B and \num{122.6}B configurations are \decodeSlowdown{0}{p2_candidate_256} and \decodeSlowdown{0}{p2_candidate_384} slower than the \num{42.1}B configuration.
At \recommendedContextLengthShort{}, the gaps narrow to \decodeSlowdown{3}{p2_candidate_256} and \decodeSlowdown{3}{p2_candidate_384} as \ac{KV}-cache reads account for a larger share of total memory traffic.
Most of this long-context traffic comes from the \num{10} full-attention layers, while the \num{40} sliding-window layers read a fixed window and contribute negligibly.

\paragraph{Deployment Footprint}
In resource-constrained deployments, total parameter count directly reduces the memory available for activations and \ac{KV} cache, limiting request concurrency at long context lengths (\cref{tab:expert-capacity-deployment}).
For example, on two H100s, the \num{122.6}B configuration fits only \num{3} concurrent \recommendedContextLengthShort{}-token requests, compared with \num{18} for \num{82.3}B.

The \num{82.3}B configuration improves all reported quality metrics over \num{42.1}B, while increasing capacity to \num{122.6}B yields only a small further improvement, reduces decode throughput and fits far fewer long requests.
We therefore target \num{80}B total parameters at \num{3.4}B active parameters.
With the total size fixed, further ablations set the number and width of the experts and the other shape parameters.
\Cref{app:shape-ablation} compares three expert configurations at this size.
Informed by a preliminary sweep at \num{0.9}B active parameters, we choose to widen the residual stream (\cref{app:shared-representation-width}).

\FloatBarrier

\subsection{\acs*{MoE} Load Balancing}
\label{sec:load-balancing}
An \ac{MoE} layer combines a shared expert, evaluated for every token, with \(E\) routed experts, of which only \(K\) are selected per token.
For \(T\) tokens and \(E\) routed experts, each \ac{MoE} layer computes router logits \(Z\in\mathbb R^{T\times E}\).
Expert biases \(\beta\in\mathbb R^E\) are added to the router logits \(Z\) to produce the biased logits \(B\) used for balanced expert selection:
\begin{equation}
  B=Z+\mathbf 1_T\beta^\top, \qquad \mathcal I_t=\TopKset(B_{t,:},K), \qquad S_{t,e}=\operatorname{sigmoid}(Z_{t,e}).
  \label{eq:moe-router}
\end{equation}
Here \(B_{t,:}\) denotes the row of \(B\) for token \(t\), and \(S\) denotes the routing scores that weight the outputs of the selected experts; \(\mathbf 1_T\) is the all-ones column vector, and \(\TopKset(x,K)\) returns the indices of the \(K\) largest entries of \(x\).
For an input representation \(x_t\), the layer output is
\begin{equation}
  y_t=f_{\mathrm{shared}}(x_t)+\sum_{e\in\mathcal I_t} S_{t,e}f_e(x_t),
  \label{eq:moe-output}
\end{equation}
where \(f_{\mathrm{shared}}\) and \(f_e\) are the shared and routed expert networks.
The selected scores are not renormalised, and the biases affect selection only; the mixture weights use the unbiased sigmoid scores.

Load balancing ensures sufficient learning signal for every expert and efficient expert-parallel execution.
The former requires balance across the global optimiser-step batch; the latter requires balance within each local microbatch.
For either batch \(\mathcal B\), we measure the maximum relative excess load by MaxVio \parencite{wang2024auxiliary,deepseekai2024deepseekv3}:
\begin{equation}
  \operatorname{MaxVio}(\mathcal B)=\max_e\left[\frac{c_e(\mathcal B)}{K|\mathcal B|/E}-1\right],
  \label{eq:maxvio}
\end{equation}
where \(c_e(\mathcal B)\) counts the tokens assigned to expert \(e\).

Balancing methods include auxiliary losses \parencite{lepikhin2021gshard} and expert-bias updates, either with a fixed step \parencite{wang2024auxiliary,deepseekai2024deepseekv3} or based on quantiles \parencite{kimi2026k3}.

\model{2} combines \acf{EQB} (\cref{sec:eqb}) for global balance with \acf{LEI} (\cref{sec:lei}) for local balance \parencite{neitemeier2026exactquantilebalancing}.
\Ac{EQB} maintains stable global load balance without a tunable balancing hyperparameter.
The \ac{LEI} coefficient therefore primarily controls local balance and training efficiency, giving the recipe fewer hyperparameter choices than the recipe of \model{1}, which combines \ac{SMEBU} bias updates \parencite{singh2026arceetrinitylargetechnical} with a GShard-style auxiliary loss \parencite{lepikhin2021gshard}.
\pgfplotstableread[col sep=comma, row sep=\\, format=inline]{%
  mean_over_layers,worst_layer\\%
  0.31,2.39\\%
}\productionBalance
Over the \model{2} pre-training run, the global MaxVio averaged over layers stays at \productionBalanceValue{mean_over_layers}.
Only the first two layers, whose routing the bias overrides (\cref{sec:leading-dense-layers}), reach \productionBalanceValue{worst_layer} on average.

\subsubsection{\acl*{EQB} (\acs*{EQB})}
\label{sec:eqb}
\Ac{QB} \parencite{kimi2026k3} computes token biases \(\alpha\in\mathbb R^T\) and expert biases \(\beta\) through alternating quantile updates:
\begin{equation}
  \alpha_t\leftarrow-Q_K\!\big(Z_{t,:}+\beta^\top\big),
  \qquad
  \beta_e\leftarrow-Q_r\!\big(Z_{:,e}+\alpha\big), \qquad r=\frac{TK}{E},
  \label{eq:qb-update}
\end{equation}
where \(Q_m\) denotes the \((m+1)\)th largest entry.
Subtracting this cutoff leaves \(m\) positive entries, so the token-bias update targets \(K\) assignments per token and the expert-bias update \(r\) assignments per expert.
Training performs one token-bias update followed by one expert-bias update per optimiser step.
The resulting expert biases \(\beta\) are centred by subtracting their mean, and we use the top-\(K\) of the biased logits for routing.
Using top-\(K\) per token approximates the token biases computed by the first step of the \ac{QB} iteration during training and inference.

The global batch is partitioned across data-parallel ranks and gradient-accumulation microbatches.
Each rank holds the router scores \(Z_{t,:}\) for all experts for its local tokens, together with the shared expert biases \(\beta\).
The token bias \(\alpha_t\) can therefore be computed locally from \(Z_{t,:}+\beta^\top\).
Updating an expert bias \(\beta_e\), however, requires a quantile of \(Z_{:,e}+\alpha\) across all tokens in the global batch.
These values are distributed across ranks and microbatches, so their statistics must be combined.

Marin \parencite{marin2026qb} averages quantiles computed separately on each rank, which generally differs from the quantile of the combined batch.
Kimi \parencite{kimi2026k3} accumulates a \num{1000}-bin histogram per expert over local microbatches and sums the counts across ranks with an all-reduce.
The resulting histogram represents the global batch, but its quantile estimate is limited by the bin width.

\Ac{EQB} \parencite{neitemeier2026exactquantilebalancing} computes this global quantile exactly for BF16 adjusted logits using two-pass radix selection \parencite{radik2024}.
Each rank encodes its local values as order-preserving \num{16}-bit keys.
In the first pass, we build \num{256}-bin histograms over the high byte and sum the counts across microbatches to locate the bin containing the global quantile.
In the second pass, we repeat this for the histogram of the low byte of values in that bin to recover the exact quantile.
Each layer requires two all-reduces of \(256E\) int32 counts per optimiser step, independent of \(T\).

\subsubsection{\acl*{LEI} (\acs*{LEI})}
\label{sec:lei}
A shared global bias cannot balance all microbatches individually.
A GShard-style auxiliary loss is a common choice for local balancing, originally formulated for normalised softmax probabilities \parencite{lepikhin2021gshard}.
With sigmoid scores, models such as DeepSeek-V3 and Arcee Trinity add a separate normalisation for this loss \parencite{deepseekai2024deepseekv3,singh2026arceetrinitylargetechnical}.
We introduce \ac{LEI}, which directly injects each expert's deviation from its target into the score gradients, to balance each microbatch.
\textcite[App.~D]{neitemeier2026exactquantilebalancing} derive the \ac{LEI} update as the straight-through estimate for unnormalised sigmoid scores, while the GShard loss is the corresponding estimate for normalised softmax probabilities.
For a local microbatch \(\mathcal B_{\mathrm l}\), with \(c_e(\mathcal B_{\mathrm l})\) assignments to expert \(e\), the relative load and relative load error are
\begin{equation}
  f_e=\frac{c_e(\mathcal B_{\mathrm l})}{K|\mathcal B_{\mathrm l}|/E},\qquad \rho_e=f_e-1.
  \label{eq:lei-load-error}
\end{equation}

We inject a rescaled version of the load-error vector into the sigmoid-score gradient of every token in the microbatch:
\begin{equation}
  G_{t,e}\leftarrow G_{t,e}+\lambda\frac{\tanh\xi}{\xi}\,\rho_e, \qquad \xi=\frac{\lVert\rho\rVert_\infty}{\tau}, \qquad t\in\mathcal B_{\mathrm l}.
  \label{eq:lei-update}
\end{equation}
Here \(G_{t,e}\) is the incoming gradient with respect to \(S_{t,e}\), \(\lambda>0\) controls the correction strength, and \(\tau>0\) bounds the rescaled load error, which we found to help training stability.
At \(\xi=0\), the scaling factor is \num{1} by continuity.
\model{2} uses \(\lambda=\num{1e-5}\) and \(\tau=\num{1}\).

\subsubsection{Leading Dense Layers}
\label{sec:leading-dense-layers}

Unlike \model{1}, which has dense \acp{FFN} in its first two blocks, \model{2} uses \ac{MoE} layers in all \num{50} blocks.
This is motivated by an ablation of the \model{2} architecture over the \num{327.6}B-token budget of scale L that found slightly lower loss and \ac{BPB} and higher accuracy when replacing leading dense blocks with \ac{MoE} layers (\cref{tab:leading-dense}).
Diagnostics of the full \model{2} run, however, show that while the expert loads are balanced, its first two \ac{MoE} layers make almost no use of their routed experts.
\Cref{app:leading-dense} gives the routing diagnostics of the full run and an analysis of the first two \ac{MoE} layers.

\begin{table}[tb]
  \centering
  \small
  \setlength{\tabcolsep}{4pt}
  \begin{tabular}{lrrrrrrrr}
    \toprule
    & & \multicolumn{3}{c}{\acs*{BPB} \(\downarrow\)} & \multicolumn{3}{c}{Accuracy (\%) \(\uparrow\)} & \\
    \cmidrule(lr){3-5}\cmidrule(lr){6-8}
    Dense blocks&Loss \(\downarrow\)&Mean&EN&DE&Mean&EN&DE&Dead (\%) \(\downarrow\)\\
    \midrule
    None
    & \textbf{1.847}
    & \textbf{0.486}
    & \textbf{0.491}
    & \textbf{0.466}
    & \textbf{38.0}
    & \textbf{37.3}
    & \textbf{40.8}
    & \textbf{0.46} \\
    Two
    & 1.850
    & 0.491
    & 0.496
    & 0.470
    & 36.9
    & 36.3
    & 39.5
    & 0.52 \\
    \bottomrule
  \end{tabular}
  \caption{Leading dense blocks in the \model{2} architecture, at the last checkpoint.
    Both models train for \num{327.6}B tokens.
    Loss is the training loss averaged over the last \num{2}B tokens, and Dead the effective dead-expert fraction (\cref*{eq:dead}) of the most affected layer at the last step; the best value of each column is bold.
  }
  \label{tab:leading-dense}
\end{table}

\subsection{Infrastructure}
\label{sec:infrastructure}

Pre-training ran on \num{768} NVIDIA B200 GPUs, across \num{96} nodes with \num{8} GPUs per node.
The cluster has an NVIDIA SuperPOD\texttrademark{} topology.
Within each node, the GPUs share an NVLink\texttrademark{} Generation 5 interconnect with \qty{1.8}{\tera\byte\per\second} bidirectional bandwidth.
The nodes are connected via InfiniBand by \num{8} NVIDIA ConnectX-7\textregistered{} network adapters per node, so each GPU can sustain a cross-node bidirectional bandwidth of \qty{100}{\giga\byte\per\second}.

Our parallelism setup for pre-training is \ac{EP}~\num{8}, \ac{FSDP}~\num{16}, \ac{DP}~\num{6}.
With this, we make use of the hierarchical interconnect in pre-training: expert parallel across the \num{8} GPUs within each node and (fully sharded) data parallel across nodes.
Thus, dense parameters are sharded across \ac{EP}~\texttimes~\ac{FSDP}~=~\num{128}~GPUs (one full replica), while sparse expert parameters are sharded across \ac{FSDP}~=~\num{16}~GPUs~-- the same local GPU on each node in one full replica.
Mid-training and long-context extension use \ac{FSDP}~\num{128}, \ac{DP}~\num{6}.

We leverage several existing libraries, namely PyTorch~2.14~\parencite{ansel2024pytorch}, our fork of TorchTitan~\parencite{liang2024torchtitan} as the training framework, DeepEP~v2~\parencite{deepep2025} for expert parallelism, and FlashAttention~4~\parencite{zadouri2026flashattention} for both \ac{SWA} and full-attention computations.
Ablations without \ac{EP} use SonicMoE~\parencite{guo2025sonicmoe} as an efficient \ac{MoE} implementation.

For further details, see \cref{sec:infrastructure-details}, including \cref{sec:straggler-detection} on straggler detection.

%% file: sections/02-pre-training/02-optimisation.tex
\section{Optimisation}
\label{sec:optimization}
\label{sec:bienenelfe-training-recipe}

\model{2} is trained with a \ac{WSM} recipe \parencite{tian2025wsm}.
The learning rate warms up linearly over the first \num{100}B tokens and then stays constant through pre-training, mid-training and long-context training.
In our experiments, a constant learning rate with checkpoint merging improves post-trained performance over a learning-rate cooldown~\parencite{maskey2026goodpretraining}.
Each stage (except pre-training) continues with the optimiser state after finishing the previous one and therefore needs no further warmup.
We then determine the base model weights as the average of all \num{20} checkpoints of the long-context stage~\parencite{li2025modelmerging}, which are \num{10}B tokens apart.
Over the three stages the context length grows from \preTrainingSeqLen{} to \longContextSeqLen{} tokens and the token batch from \num{75.5}M to \num{201.3}M tokens per step (\cref{fig:training-loss}).
The architecture improvements, particularly the hybrid of sliding-window and full attention, enable longer training sequences than in \model{1}, which pre-trains at \num{4}k tokens and reaches \num{64}k tokens only in its long-context stage (\cref{sec:architecture,sec:window-size}).
Sequences are assembled online by bin packing chunks drawn from a \ac{RSDB}~\parencite{singh2026arceetrinitylargetechnical}, which spreads the chunks of a long document over many optimiser steps, without padding tokens or unnecessary document cuts~\parencite{ding2024fewertruncations}~(\cref{sec:batch-construction,sec:data-loading-pipeline,sec:chunking-packing}).

We optimise attention and expert weights with Muon~\parencite{jordan2024muon}, embedding weights and norm scales with Adam~\parencite{kingma2015adam}, and \ac{LM}~head and router weights with AdamW~\parencite{loshchilov2019adamw}~(\cref{tab:bienenelfe-optimizer}).
In contrast to the commonly used convention of matching Muon's update \ac{RMS} to that of AdamW~\parencite{liu2025muonscalable,kimiteam2025k2}, we adopt the spectral convention~\parencite{bernstein2025derivingmuon}, which keeps the Muon learning rate independent of model width~(\cref{sec:bienenelfe-hyperparameter-transfer}).
We use decoupled weight decay for all optimiser groups except the embedding and norm scales.
Adam and AdamW use \(\epsilon = \num{1e-15}\), small enough that the updates of parameters with small gradients, such as the \ac{LM} head and the routers, stay independent of the gradient scale, as Muon updates are by construction.

\Cref{tab:bienenelfe-optimizer} lists the optimiser, peak learning rate, weight decay and initialisation of every parameter group; load balancing is described in \cref{sec:load-balancing}.
Our hyperparameter optimisation and transfer scheme is described in \cref{sec:bienenelfe-hyperparameter-transfer}.

\begin{table}[htbp]
  \centering
  \small
  \setlength{\tabcolsep}{5pt}
  \begin{tabular}{@{}lllll@{}}
    \toprule
    Optimiser & Peak learning rate & Weight decay & Parameters & Initialisation \\
    \midrule
    \multirow{2}{*}{Muon, spectral}
    & \multirow{2}{*}{\num{1e-3}}
    & \multirow{2}{*}{\(2^{-12}\)}
    & Q, K, V; expert up and gate
    & std \(1/\sqrt{d_{\mathrm{in}}}\) \\
    &
    &
    & Attention output; expert down
    & zero \\
    \midrule
    \multirow{2}{*}{AdamW}
    & \multirow{2}{*}{\num{2.5e-5}}
    & \multirow{2}{*}{\(2^{-12}\)}
    & Router gates
    & std \(1/d_{\mathrm{in}}\) \\
    &
    &
    & \acs*{LM} head
    & zero \\
    \midrule
    \multirow{3}{*}{Adam}
    & \multirow{3}{*}{\num{2e-3}}
    & \multirow{3}{*}{none}
    & Token embedding
    & std \num{1} \\
    &
    &
    & Norm scales
    & \num{1} \\
    &
    &
    & Post-sublayer (sandwich) norm scales
    & \(1/\sqrt{L}\) \\
    \bottomrule
  \end{tabular}
  \caption{Optimiser and initialisation of \model{2} by parameter group.
    Muon uses momentum \(\mu = 0.95\) with Nesterov and spectral scaling, which multiplies the orthogonalised update by \(\sqrt{d_{\mathrm{out}}/d_{\mathrm{in}}}\); Adam and AdamW use \(\beta_1 = 0.9\), \(\beta_2 = 0.95\) and \(\epsilon = \num{1e-15}\).
    Decoupled weight decay multiplies the decayed weights by \(1 - 2^{-12}\) in every step, independent of the learning rate.
    Normal initialisations are truncated at three standard deviations and rescaled to the stated standard deviation; \(d_{\mathrm{in}}\) is the fan-in of a layer, \(L = 50\) the number of layers, and expert biases start at zero.
  }
  \label{tab:bienenelfe-optimizer}
\end{table}

\subsection{Hyperparameter Transfer}
\label{sec:bienenelfe-hyperparameter-transfer}
The optimal hyperparameters depend on model size and token horizon~\parencite{yang2022tensor,mccandlish2018empirical,bjorck2024scalingoptimallr,porian2024resolving,bergsma2025powerlines}, but a sweep at the scale of the target run is infeasible.
Proxy runs on smaller models and shorter horizons therefore guide the choice of hyperparameters for the full-scale run.
A common approach fits scaling laws that extrapolate along both dimensions jointly~\parencite{deepseekai2024deepseekllm,li2025predictablescaleistep,porian2024resolving,poolside2026laguna}, which requires sweeps at many points of a 2D grid over model size and horizon.
Instead, we factorise the transfer across model size and token horizon, and vary model size only through the width, at the \num{50} layers of the target model (\cref{par:bienenelfe-width-transfer}).
For the width transfer we use \ac{muP}~\parencite{yang2022tensor,yang2024spectralconditionfeaturelearning}, which makes the optimal base learning rate of every parameter group approximately width-independent, so that hyperparameters found on narrower models carry over to the full-scale model~\parencite{dey2025completep,longcat2025flash}.
The dependence on the token horizon can then be studied at a single narrower width.
Instead of scaling laws for the optimal learning rate and batch size~\parencite{deepseekai2024deepseekllm,li2025predictablescaleistep,bjorck2024scalingoptimallr,bergsma2025powerlines}, we use loss extrapolation \parencite{domhan2015extrapolation} to predict the learning-rate landscape at the target horizon and pick the best cell of a grid of Muon learning rates and batch sizes on a narrower proxy (\cref{par:bienenelfe-horizon-transfer}).

The remainder of this section fixes the learning-rate ratios between the parameter groups (\cref{par:bienenelfe-lr-ratios}), introduces the width transfer with \ac{muP} (\cref{par:bienenelfe-width-transfer}) and the horizon transfer by loss extrapolation (\cref{par:bienenelfe-horizon-transfer}) for learning rate, batch size and weight decay.

\subsubsection{Learning-Rate Ratios}
\label{par:bienenelfe-lr-ratios}
We find that different parameter groups require different learning-rate scales to achieve optimal performance.
To avoid an explosion of the grid in later sweeps, we optimise the ratios between the learning rates of the different embedding, norm and head parameter groups as a factor of the Muon learning rate in a preliminary sweep.
In internal sweeps, we find these ratios to be relatively stable across widths and token horizons, and therefore assume that they can be held fixed when transferring hyperparameters.
The router learning rate is set equal to the head learning rate, based on internal ablations.
The resulting production learning rates are listed in \cref{tab:bienenelfe-optimizer}.

\subsubsection{Transfer Across Width}
\label{par:bienenelfe-width-transfer}

While there is work on joint width and depth transfer with \ac{muP}~\parencite{bordelon2023depthwise,yang2023tensorvi,dey2025completep}, it is less established and introduces additional complexity, and we therefore restrict ourselves to width transfer only.
We keep the number of experts and top-\(K\) fixed and scale the size of every expert, so that the routing and load-balancing dynamics of the proxy match the production model.
Attention heads keep their dimension of \num{128} and we only scale the number of heads, as in \ac{muP}~\parencite{yang2022tensor}.
We assume the optimal batch size to be only dependent on token horizon~\parencite{li2025predictablescaleistep}.

As described in~\cref{sec:bienenelfe-training-recipe}, Muon runs in the spectral convention: after the Newton--Schulz orthogonalisation the update is scaled by \(\sqrt{d_{\mathrm{out}}/d_{\mathrm{in}}}\), so that every layer receives an update of unit \ac{RMS} operator norm times the learning rate regardless of its shape~\parencite{bernstein2025derivingmuon}.
This ensures that the optimisation dynamics satisfy the spectral condition~\parencite{yang2024spectralconditionfeaturelearning}, and makes the Muon learning rate width-independent.
The learning rates of embedding and norm scales are width-independent as well, whereas those of the \ac{LM}~head and router scale with \(1/d\).
\Ac{muP} also shrinks the initialisation of the output layers with width.
We initialise the \ac{LM}~head at the infinite-width limit of this scaling, zero, which removes the width dependence of the initial outputs and brings the optimal hyperparameters of narrow and wide models closer together~\parencite[App.~D.2]{yang2022tensor}.
The attention output and expert down projections also start at zero, so that every block starts as the identity map~\parencite{zhang2019fixup}; in our ablations this converged faster and still transfers across width (\cref{tab:bienenelfe-optimizer}).
The router is initialised with standard deviation \(1/d_{\mathrm{in}}\)~\parencite{vankadara2026scalemoe}.

\begin{figure}[htbp]
  \centering
  \begin{subfigure}[t]{0.48\linewidth}
    \centering
    \includegraphics{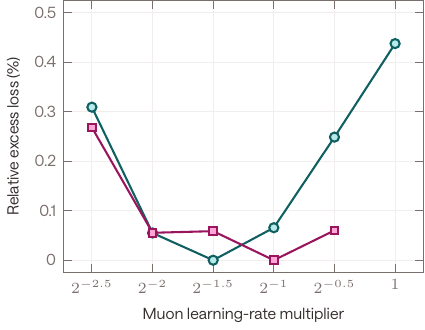}
    \par\smallskip
    \includegraphics{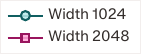}
    \caption{Across width: relative excess training loss at \num{97.6}\mbox{B} tokens above that of the best learning rate of each width, for proxies of width \num{1024} and \num{2048}.
      The optima differ by one grid step, a factor of \(\sqrt{2}\), and the wider model is flatter around its optimum.
    }\label{fig:bienenelfe-width-transfer}
  \end{subfigure}\hfill
  \begin{subfigure}[t]{0.48\linewidth}
    \centering
    \includegraphics{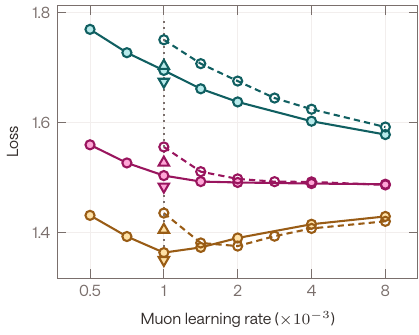}
    \par\smallskip
    \includegraphics{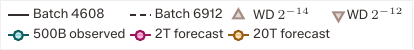}
    \caption{Across the training horizon: observed loss of the width-scaled proxy (\cref{tab:bienenelfe-proxy}) at \num{500}\mbox{B} tokens and forecasts at \num{2}\mbox{T} and \num{20}\mbox{T} tokens, for batches of \num{4608} (solid) and \num{6912} (dashed) sequences.
      With increased target horizon, the optimum moves from the largest learning rate to the production value of \num{1e-3} (dotted) at \num{20}\mbox{T} tokens; the larger batch is best at \num{2e-3}, \qty{0.9}{\percent} behind.
      Triangles indicate halved (\(2^{-14}\), up) and doubled (\(2^{-12}\), down) weight decay, relative to the \(2^{-13}\) baseline used for the remaining sweep points.
    }\label{fig:bienenelfe-horizon-transfer}
  \end{subfigure}
  \caption{Learning-rate transfer across width and training horizon.}
  \label{fig:bienenelfe-lr-transfer}
\end{figure}

To validate the transfer we swept the base learning rate on the proxies M and L (\cref{tab:proxy-models}), where L is M scaled to twice the width, both trained for \num{97.6}B tokens, the horizon of M, with a batch of \num{576} sequences of \num{16}k tokens and with learning rates scaled according to the rules described above.
They are not scaled down from the production model, since the width sweep is an earlier ablation.
Results are visualised in \cref{fig:bienenelfe-width-transfer}, with the loss averaged over the trailing \num{3}B tokens of each run.
We see a shift of the optimal learning rate by \(\sqrt{2}\), which we deem acceptable.
The loss differences within a \(\sqrt{2}\) learning-rate factor of the optimum are less than \qty{0.1}{\percent}.

\subsubsection{Transfer Across Token Horizon}
\label{par:bienenelfe-horizon-transfer}
Transferring the optimal learning rate and batch size to a target token horizon is often done via scaling laws~\parencite{deepseekai2024deepseekllm,li2025predictablescaleistep,bjorck2024scalingoptimallr,porian2024resolving}.
These necessitate a large grid of values to have enough points for a reliable power-law fit, making them expensive.
Additionally, they complicate transferring other hyperparameters such as weight decay from short to long horizons, as this would combinatorially explode the grid.

For \model{1}, we fit power laws to the optimal learning rate and batch size at horizons from \num{100}B to \num{430}B tokens and extrapolate them to its pre-training budget of \num{7.5}T tokens.
Extrapolating those laws to the \model{2} setting gives unrealistic hyperparameter predictions, motivating a method that uses the observed loss trajectories of the candidate configurations.

We employ loss extrapolation~\parencite{domhan2015extrapolation} to predict the hyperparameters with the lowest loss at longer token horizons based on the observed training dynamics at shorter horizons.
This allows us to use smaller grids for the search and employ coordinate descent to tune further hyperparameters iteratively.
Since the sweep uses the batch size, warmup, learning-rate schedule and data of the production run on the width-scaled proxy (\cref{tab:bienenelfe-proxy}), it doubles as a test of the production configuration, and its runs are reused for the weight-decay sweep described below.

We use a power law with floor~\parencite{hoffmann2022training} in the cumulative-learning-rate coordinate~\parencite{tissue2025annealing},
\begin{equation}
  L(D) = E + A\,(D - W/2)^{-\alpha},
  \label{eq:bienenelfe-loss-extrapolation}
\end{equation}
where \(D\) is the number of tokens, \(W\) the length of the linear warmup in tokens, \(E\) the irreducible loss, and \(A\) and \(\alpha\) the scale and exponent of the reducible part.
With a linear warmup of \(W\) tokens followed by a constant learning rate, the cumulative learning rate is proportional to \(D - W/2\).
We validated the method on internal loss curves, not reported here, and found an average prediction error at the target token horizon below~\qty{1}{\percent}.
All tested methods are fairly sensitive to the fit start point.
For the method used, we discard the first \qty{20}{\percent} of the post-warmup prefix before fitting.
We find that the optimal grid point is robust to the specific form of the power law used for extrapolation: we cross-checked with a shifted power law with floor~\parencite{domhan2015extrapolation} and a one-break broken power law~\parencite{caballero2023broken} and obtained the same ranking.

\paragraph{Learning Rate and Batch Size}
\label{par:bienenelfe-lr-bs}
We sweep a grid of Muon learning rates from \num{5e-4} to \num{8e-3} and two batch sizes on the width-scaled proxy model (\cref{tab:bienenelfe-proxy}), which halves the width of the production model with the same rules as in \cref{par:bienenelfe-width-transfer}, for \num{500}B tokens.
The observed loss is averaged over the trailing \num{2}B tokens of each run and, at \num{500}B tokens, is still lowest at the largest learning rate for both batches.
The prediction at \num{20}T tokens is visualised in \cref{fig:bienenelfe-horizon-transfer}, with an optimal Muon learning rate of \num{1e-3} and a batch size of \num{4608} sequences of \num{16}k tokens each.
The selected batch is the smaller of the two, so while a further reduction in batch size might be beneficial, we prioritise maintaining training efficiency.

\begin{table}[htbp]
  \centering
  \small
  \begin{tabular}{lrrrrr}
    \toprule
    & Width & Heads / \acs*{KV} heads & Expert hidden & Experts, top-\(K\) & Layers \\
    \midrule
    \model{2} & \num{2560} & \num{48} / \num{4} & \num{512} & \num{384}, \num{6} & \num{50} \\
    Proxy     & \num{1280} & \num{24} / \num{2} & \num{256} & \num{384}, \num{6} & \num{50} \\
    \bottomrule
  \end{tabular}
  \caption{Production model and the width-scaled proxy used for the loss-extrapolation sweeps over learning rate, batch and weight decay.
    Layers, experts, head dimension and the attention pattern are shared.
  }
  \label{tab:bienenelfe-proxy}
\end{table}

\paragraph{Weight Decay}
\label{par:bienenelfe-coordinate-descent}
We use the same loss extrapolation to set the weight decay in a one-dimensional sweep.
Starting from the selected learning rate and batch size, we vary the weight decay and compare the predicted loss at \num{20}T tokens (triangles in \cref{fig:bienenelfe-horizon-transfer}).
The baseline cell was trained with a weight decay of \(2^{-13}\).
Doubling the weight decay to \(2^{-12}\) lowers both the observed loss and the forecasts at every horizon, whereas halving it raises them.
All estimators agree on this order, and we use \(2^{-12}\) for \model{2}.
Larger values were not tested.

\paragraph{Limitations}
\label{par:bienenelfe-limitations}
This optimisation is done purely with respect to the training loss under a warmup-stable learning-rate schedule.
While evidence shows that a \ac{WSM} schedule can lead to better post-trained performance~\parencite{maskey2026goodpretraining}, merging and post-trained performance are not considered here.
Additionally, uncertainties are not modelled and some of the differences between grid points may be within the noise level of the measurements, although all three forms of the extrapolation law select the same cell (\cref{par:bienenelfe-horizon-transfer}).
We leave modelling merged-checkpoint and post-trained performance, as well as uncertainty of the forecasts, to future iterations of the recipe.

%% file: sections/02-pre-training/03-data-and-training-stages.tex
\section{Data and Training Stages}
\label{sec:pre-training-data}

\subsection{Overview} %

\basemodel{2} is trained in three stages: pre-training, mid-training, and long-context extension.
Pre-training consumes \num{20}T tokens from a broad English-German data pool and establishes the model's general capabilities.
Mid-training shifts the data distribution towards higher-quality reasoning, mathematics, code, and instruction-following data, training on \num{3.4}T tokens.
Finally, long-context extension trains on \num{200}B tokens and extends the context length to \recommendedContextLengthShort{} tokens.

Across these stages, data selection progressively becomes more selective and task-oriented.

Pre-training emphasises breadth and coverage, mid-training concentrates on capabilities that benefit from higher-quality and more structured data, and long-context extension adapts these capabilities to substantially longer sequences.

\label{sec:pre-pii}
In addition to filtering illegal, harmful, toxic and pirated content, we redact \ac{PII} from all training sources.
Each category is detected by a regular expression, checked by a validator that rejects false positives, and replaced by a special token.
A document whose redacted bytes exceed \qty{15}{\percent} of its length is dropped.
In one illustrative run on a mid-training data sample, \qty{5.3}{\percent} of scanned documents received at least one redaction, with PDF data accounting for \qty{92}{\percent} of all redactions.
We report these numbers only as an example of the redaction behaviour; the sample is not representative of the overall training corpus.

The following sections describe the data construction, mixing, and evaluation choices for each stage.

\subsection{Stage 1: Pre-Training Data}
\label{sec:stage-1-data}

The pre-training dataset pool consists of \num{78} English- and German-language datasets with approximately \num{29.3}T materialised tokens, from which the final training mix consumes \num{20}T. The pool spans web text, code, instruction data, scientific and technical text, and legal and public-sector data.

\begin{figure}[htbp]
  \centering
  \begin{minipage}[b]{0.6\linewidth}
    \centering
    \footnotesize
    \setlength{\tabcolsep}{3pt}
    \begin{tabular}[b]{@{}c@{\hspace{0.4em}}llcr@{}}
      \toprule
      & Domain & Short & Datasets & Tokens (B) \\
      \midrule
      \domSwatch{0}&\domEnWeb&\domEnWebShort&$23$&$12{,}248$\\
      \domSwatch{1}&\domDeWeb&\domDeWebShort&$16$&$6{,}712$\\
      \domSwatch{2}&\domEnInstr&\domEnInstrShort&$7$&$3{,}548$\\
      \domSwatch{3}&\domDeInstr&\domDeInstrShort&$3$&$139$\\
      \domSwatch{4}&\domCode&\domCodeShort&$3$&$4{,}323$\\
      \domSwatch{5}&\domAgentic&\domAgenticShort&$1$&$1$\\
      \domSwatch{6}&\domStemDocs&\domStemDocsShort&$7$&$1{,}462$\\
      \domSwatch{7}&\domStemQA&\domStemQAShort&$9$&$789$\\
      \domSwatch{8}&\domSpecialized&\domSpecializedShort&$9$&$80$\\
      \midrule
      &Total&&$78$&$29{,}301$\\
      \bottomrule
    \end{tabular}
  \end{minipage}\hfill
  \begin{minipage}[b]{0.38\linewidth}
    \centering
    \includegraphics{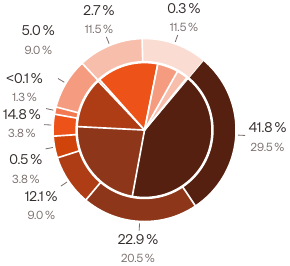}
  \end{minipage}

  \begin{minipage}[t]{0.6\linewidth}
    \captionof{table}{The pre-training pool by domain, in billions of materialised tokens.}
    \label{tab:pretraining-pool-domains}
  \end{minipage}\hfill
  \begin{minipage}[t]{0.38\linewidth}
    \caption{Share of the pre-training pool by domain, in materialised tokens (inner disk) and in datasets (outer ring), colours as in the adjacent \cref{tab:pretraining-pool-domains}.
      Each label gives the domain's share of the tokens and, below it, its share of the datasets.
    }
    \label{fig:pretraining-pool-domains}
  \end{minipage}
\end{figure}

We group datasets into nine domains according to language, subject, and degree of structure.
Datasets spanning several domains are assigned to the domain containing the majority of their tokens.
\Cref{tab:pretraining-pool-domains} summarises the resulting pool, and \cref{fig:pretraining-pool-domains} shows each domain's share of the pool's tokens and datasets.

Web data makes up the largest part of the pool.
Its largest individual source comes from our internally developed \ac{CC} pipeline, which processes and filters \ac{CC} snapshots to produce about \num{14.9}B quality-scored documents, from more than \num{308}B extracted documents, for downstream data mixing.

German accounts for \qty{21.3}{\percent} of the final pre-training tokens and is treated separately in the data construction process because its availability and filtering requirements differ from English.

Approximately \qty{24}{\percent} (\num{4.8}T tokens) of our pre-training data is synthetic in origin (including rephrases and translations).
We produce synthetic data in English, using our deduplicated English \ac{CC} data (see \cref{sec:cc-pipeline}) as seeds for rephrasing with Gemma 4 26B-A4B. We follow a Nemotron-CC-style recipe \parencite{su2024nemotroncc}, using five prompt templates applied to each seed document: diverse \ac{QA} pairs, distillation, knowledge list, knowledge extraction, and encyclopedia-style rewriting.
The outputs are post-processed and filtered to remove \ac{LLM}-generation artefacts and erroneous generations.
In \cref{subsec:german-pre-training-data} we describe our work on German synthetic data.

The following subsections describe key decisions around the \ac{CC}~\parencite{commoncrawl} pipeline, the construction of the German data pool, quality annotation, and the final data-source mixing procedure in more detail.

\subsubsection{\Acl*{CC} Pipeline}
\label{sec:cc-pipeline}

The \ac{CC} pipeline is designed to process \ac{CC} snapshots to curate high-quality pre-training data at scale.
Text is extracted from the raw archives with Resiliparse \parencite{bevendorff2018chatnoir}, yielding more than \num{308}B documents.

Language-agnostic heuristic filters and language identification route each document to an English or a German branch, which apply separate thresholds.
Exact deduplication removes byte-identical documents across all snapshots, which yields \num{21.16}B distinct documents.
Fuzzy deduplication, which compares MinHash signatures of documents bucketed by locality-sensitive hashing, removes a further \qty{25.9}{\percent} as near-duplicates (\cref{sec:fuzzy-dedup}).

Substring deduplication cuts repeated blocks of at least \num{500} bytes, such as navigation menus and page templates, out of the remaining documents, removing \qty{21.3}{\percent} of text bytes (\cref{sec:substring-dedup}).
\Cref{sec:cc-internal-pipeline} describes the stages in more detail.

\begin{figure}[tb]
  \centering
  \includegraphics{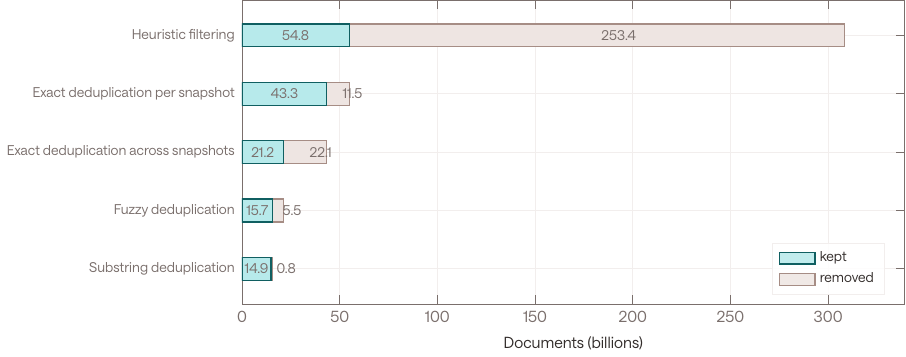}
  \caption{Documents entering each stage of the \acs*{CC} pipeline, split into the documents the stage kept and removed.
    Each stage receives what the previous one kept.
    Substring deduplication removes text within documents and drops a document only when fewer than \num{100} bytes of it remain, so its effect on bytes is larger than on documents.
  }
  \label{fig:cc-funnel}
\end{figure}

\Cref{fig:cc-funnel} shows how many documents each stage of the pipeline receives and removes, from the extracted documents entering the heuristic filters to substring deduplication.

\subsubsection{German Pre-Training Data}
\label{subsec:german-pre-training-data}

At our \num{20}T-token budget, our ablations show strong German benchmark performance with around \qty{20}{\percent} German data in the training mix.
This leaves us with a lower bound of \num{4}T German tokens needed for training.
Note that this refers to training tokens, not unique tokens.
Since datasets are commonly upsampled during training, the underlying data pool can be smaller than \num{4}T, though not arbitrarily so, because repeating data can have diminishing returns after roughly four epochs~\parencite{muennighoff2023scaling}.

To put the scale of this requirement into context, we first look at the amount of German pre-training data available in existing open datasets.
\Cref{tab:german-pre-training-datasets} summarises several major open German datasets.
The datasets in its upper part are shown only as a reference point for the scale of publicly available German data and do not describe our training set.
For further details on training data, refer to the Public Summary of Training Content.

\begin{table}[tb]
  \centering
  \small
  \begin{tabular}{@{}lll@{}}
    \toprule
    Dataset & Size (B tokens) & Latest source data \\
    \midrule
    Klettermix \parencite{kraus2026klettermix}              & \num{500} & 2024 \\
    HPLT 3.0 \parencite[German slice]{oepen-etal-2026-hplt} & \num{609} & 2025 \\
    FinePDFs \parencite[German slice]{kydlicek2025finepdfs} & \num{177} & 2025 \\
    German Commons \parencite{gienapp2025germancommons}     & \num{154} & 2025 \\
    MT-Nemotron-CC \parencite{mt_nemotron_cc_2025}          & \num{117} & 2024 \\
    FineWeb2 \parencite[German slice]{penedo2025fineweb2}   & \num{378} & 2024 \\
    \midrule
    Our organic web data  & \num{1300} & 2026 \\
    Rephrased German data & \num{1100} & 2025 \\
    \bottomrule
  \end{tabular}
  \caption{Sizes of selected German pre-training data sources in billions of tokens to illustrate German dataset scale. Note that this table does not describe our training set.}
  \label{tab:german-pre-training-datasets}
\end{table}

The open datasets in the upper part of \cref{tab:german-pre-training-datasets} add up to around \num{1.94}T tokens.
This token count does not account for filtering, deduplication, or overlap between datasets, so the amount of usable unique data would in practice be smaller.
This comparison illustrates the challenge of sourcing German data at the scale implied by our \num{4}T-token training target.
In the following, we describe the three approaches we considered to construct our German data pool.

First, we process raw \ac{CC} snapshots with a pipeline that is parametrised by language, since German and English web data cannot be filtered in exactly the same way.
In particular, we retune several filtering parameters for German.
For example, web-data pipelines commonly remove documents with unusually long words.
However, German administrative prose often contains substantially longer words than English, so applying English thresholds can disproportionately remove this register.
We therefore use FineWeb2's German word-length settings, which allow a mean word length between \num{0} and \num{15} bytes, instead of the \num{3} to \num{10} byte range used by Gopher for English.
With these language-specific adjustments, our pipeline produces \num{0.8}T unique tokens of organic German web data.

Our second approach to augmenting our German corpus is to rephrase German documents.
We use Mistral-Nemo-Instruct-2407 \parencite{mistral2024nemo} to rewrite organic German documents as
an encyclopedia entry, a Q\&A dialogue or an educational-style passage, preserving their
semantic information.
This teaches the model the same facts in several surface forms and multiplies scarce organic data, and it is a different operation from translation: the source is German, so the subject matter, the institutions and the register stay German.
Rephrasing gives us about \num{1.1}T unique tokens, making it the single largest source of German in the model.
This data had already been produced for the Aleph-Alpha-GermanWeb project~\parencite{burns2026germanweb}.

The third approach is translation.
Translating English into German can expand the available data when the model, prompts, and chunking are chosen carefully, but it does not provide the same kind of German data as organically sourced text.
Translated output can retain traces of the source language, such as literal renderings of idioms; for example, \enquote{Drive safe!} may become \enquote{Fahre sicher!} rather than the more natural \enquote{Komm gut an!}.
More importantly, translation preserves much of the source corpus's cultural and institutional distribution.
A corpus translated from English therefore remains, in many respects, an English-language view of the world expressed in German.
We experimented with translated data at scale for \model{1}, but for \model{2} we rely on organic and rephrased German data alongside a carefully curated set of translated German data.

One side-effect of us building our own \ac{CC} pipeline is that we can incorporate new crawls as they arrive so that the resulting data pool remains up to date.
\Cref{tab:german-pre-training-datasets} also lists the year from which the latest source data of each listed dataset comes.
Using our own pipeline, we can ensure that the most recent data in the pool dates from 2026.

German benchmarks are also limited in coverage and sensitivity.
We therefore built a dedicated German evaluation suite, combining hand-crafted and model-generated tasks, to better measure the impact of our German data.
See \cref{sec:pre-training-eval-scores}.

\subsubsection{Quality Annotation}
\label{sec:quality-annotation}

\paragraph{Method}
\Textcite{li2024dclm,olmo2026olmo3} have shown that simple, single fastText classifiers \parencite{joulin2017fasttext} trained to distinguish, e.g.\ \ac{SFT}-style data vs.\ random \ac{CC} samples, can be a successful method for curating \ac{CC} data for quality, leading to improved model performance on evaluation benchmarks.
Other works have reported methods of ensembling classifier scores in ways which appear to outperform using individual classifiers alone \parencite{su2024nemotroncc,burns2026germanweb}.
We therefore compare ensembling approaches with a strong single-classifier baseline.
At the same time, the method has to scale, e.g.\ it would be wasteful to apply an \ac{LLM}-as-a-judge to tens of billions of documents.
For German, we build upon our public work introduced in GermanWeb \parencite{burns2026germanweb} by ensembling quality and grammar fastText classifiers using an average of their scores.
To score the English documents, we use the following method.

We first score a fixed \num{1}M-document English sample of our curated \ac{CC} data with a judge, Qwen3 32B \parencite{yang2025qwen3}, using more than \num{20} prompts.
Prompts include educational value, chat-likeness, encyclopedia-likeness, content versus format properties, binary \enquote{quality} judgements, and benchmark usefulness \parencite{penedo2024fineweb,li2024dclm,sachdeva2026askllm,wettig2024qurating}.
For each prompt, the judge returns structured JSON, enforced via constrained decoding against a schema.
Disagreement between judged axes is asymmetric.
Low-scoring documents score low on multiple (sometimes all) prompts, whereas high-scoring documents score well on some prompts but not others.
This implies that the properties by which documents are judged as \enquote{good} are more distinctive than the absence of those properties.\footnote{By analogy, foods can be tasty in many distinct ways (a sweet apple, a savoury fish), but spoiled food is spoiled in much the same way.} Document-level Spearman correlations among the judge scores also fall into groups of highly correlated scores, e.g.\ content versus format, conversational, math, code, and evaluation- or task-usefulness.

To score the rest of the crawl, we distil each scalar judgement axis into a binary task (top quartile versus bottom quartile, discarding the middle two) and train three families of classifiers: fastText \parencite{joulin2017fasttext}, a frozen Luxical-one embedding \parencite{datology2025luxical} with a multilayer perceptron head, and a fine-tuned BERT \parencite{devlin2019bert}.
We evaluate the classifiers on a held-out sample of \num{1}M \acs{CC} documents, sampled and judged independently.
On this data, BERT models reproduce the judge most closely overall (held-out F1 \num{0.945}, \ac{AUC} \num{0.987}), Luxical-based classifiers nearly as well (F1 \num{0.913}, \ac{AUC} \num{0.971}), and much better than fastText (F1 \num{0.865}).
Given this, to balance inference speed over our corpus with quality, we use the Luxical-based distilled judge classifiers when constructing our \model{2} dataset.
The distilled classifiers also show the same correlation structures as the judges.
We combine the Luxical classifiers with a binary fastText classifier trained to distinguish \ac{SFT}-style data vs.\ random \ac{CC} samples, which has been a successful method in previous works \parencite{li2024dclm,olmo2026olmo3}.

\paragraph{Mixing Annotations on the Simplex}
We combine five English per-document scores with weights on the probability simplex.
The scores are fastText \(P(\text{high\_quality})\) and four Luxical heads (educational anchor, chat-like anchor, content-focused score, and \ac{MMLU} usefulness).
We choose these four Luxical heads on the basis that they are minimally redundant compared to other annotation axes, and because we want to balance efficient inference time over our corpus (ruling out BERT classifiers) while having higher agreement with the judges (ruling out fastText classifiers).
Each classifier score lies in \([0,1]\).
To condition the mix on fastText quality, we pair each score with either the constant gate \(1\) (ungated) or the indicator \(\mathbf{1}\{\text{fastText}>\tau\}\) with \(\tau=0.0216\) (the corpus-median fastText score).
This gives \(2\times 5=10\) basis columns, with weights \(W_{a,b}\) on the ten-dimensional probability simplex, \(W_{a,b}\ge 0\) and \(\sum_{a,b}W_{a,b}=1\).
We search the simplex by sampling \num{21} interior vectors from \(\mathrm{Dirichlet}(\alpha=1)\), as well as five one-hot vertices, namely the ungated single-score baselines (all weight on one raw score, ungated).
We combine the scores like so:
\begin{equation}
  S(d) \;=\; \sum_{a,\,b}\; W_{a,b}\,\cdot\,\operatorname{rank}\!
  \bigl(\mathrm{gate}_a(d)\cdot\mathrm{rater}_b(d)\bigr),
  \label{eq:quality-gatexrater}
\end{equation}
where the gates \(a\) are the constant \(1\) and \(\mathbf{1}\{\text{fastText}>\tau\}\), the raters \(b\) are the four Luxical heads and fastText, and \(\operatorname{rank}(\cdot)\) is the per-column mid-rank empirical cumulative distribution function.
By selecting the top \qty{25}{\percent} of token mass based on these document scores, we optimise the ten weights.
This top-\qty{25}{\percent} cut is the setting of this optimisation and of the ablations below; the production mix keeps the top three of the five English tiers, about \qty{60}{\percent} of the token mass (\cref{sec:data-mixing}).
We run ablations with a \num{0.4}B dense \ac{DCLM} model trained to \(1\times\) Chinchilla on \num{8.2}B randomly sampled tokens and record the mean \ac{BPB} over all English benchmarks.
We use these means in a \ac{GP} surrogate over the samples which then selects the optimum weights to minimise the mean \ac{BPB} for the final mix.

\FloatBarrier
\paragraph{English-Language Ablations}
We apply the optimised mix of annotations to our \ac{CC} data, and compare the performance of this mix with random \ac{CC}, the fastText classifier alone, and the \model{1} mix.
\Cref{tab:quality-cc-vs-p1} compares the four arms on the \num{0.4}\mbox{B} dense \(1\times\) Chinchilla \ac{DCLM} model \parencite{li2024dclm,hoffmann2022training} (\num{8.2}\mbox{B} tokens) and on a \num{7.8}\mbox{B}-A\num{0.6}\mbox{B} \acs{MoE} trained to \num{97.6}\mbox{B} tokens (about \num{170} tokens per active parameter).
The table also reports random \ac{CC} and fastText as a single \ac{CC} quality classifier.
At \num{0.4}B, English-overall \ac{BPB} is \num{3.800 \pm 0.513} (\(n=3\)) against \num{3.988 \pm 0.271} for the \model{1} mix (\(n=5\)).\footnote{English \acs{QA} is the unweighted mean of \acs{ARC}, \acs{MMLU}, and \acs{PIQA} \acs{BPB} at \num{0.4}\mbox{B}, and of \acs{ARC} and \acs{MMLU} at the \acs{MoE} scale.
  English overall adds HumanEval, \acs{MBPP}, HellaSwag, \acs{GSM8K}, and MATH Minerva \parencite{hendrycks2021math,lewkowycz2022minerva}.
}
English \ac{QA} is \num{3.901 \pm 1.012} against \num{5.080 \pm 0.814}.
At \num{7.8}B-A\num{0.6}B \acs{MoE}, overall \ac{BPB} is \num{0.607 \pm 0.012} (\(n=3\)) against \num{0.642 \pm 0.013} (\(n=2\)), and English-overall \ac{BPB} is \num{0.618 \pm 0.015} against \num{0.652 \pm 0.026}.
Math is \num{0.493 \pm 0.003} against \num{0.568 \pm 0.023}, and code is \num{0.550 \pm 0.067} against \num{0.572 \pm 0.027}.
The posterior probability that the annotated \ac{CC} mix is better than the \model{1} mix on overall \ac{BPB} at the \acs{MoE} scale is \num{0.98}, and \num{0.93} that it is better than the fastText classifier alone.

\begin{table}[ht]
  \centering
  \small
  \begin{tabular}{lcccc}
    \toprule
    & \multicolumn{2}{c}{\num{0.4}\mbox{B} dense}
    & \multicolumn{2}{c}{\num{7.8}\mbox{B}-A\num{0.6}\mbox{B} \acs{MoE}} \\
    \cmidrule(lr){2-3} \cmidrule(lr){4-5}
    Arm & EN & EN \acs{QA} & EN & EN \acs{QA} \\
    \midrule
    Random \ac{CC}
    & 4.263\(\pm\)1.120
    & \underline{4.245\(\pm\)1.594}
    & 0.884
    & 1.044 \\
    fastText classifier \ac{CC} selection
    & 4.711\(\pm\)0.664
    & 4.909\(\pm\)0.888
    & 0.868\(\pm\)0.010
    & 0.887\(\pm\)0.025 \\
    Optimised mix of annotations \ac{CC} selection
    & \best{3.800\(\pm\)0.513}
    & \best{3.901\(\pm\)1.012}
    & \best{0.618\(\pm\)0.015}
    & \best{0.830\(\pm\)0.036} \\
    \model{1} mix
    & \underline{3.988\(\pm\)0.271}
    & 5.080\(\pm\)0.814
    & \underline{0.652\(\pm\)0.026}
    & \underline{0.836\(\pm\)0.043} \\
    \bottomrule
  \end{tabular}
  \caption{\acs*{BPB} (lower is better).
    The lowest value in each column is in bold, the second lowest is underlined.
    The optimised mix of annotations \acs*{CC} uses the top \qty{25}{\percent} of \acs*{CC} by the optimised mix of annotations of \cref{eq:quality-gatexrater}.
    The fastText classifier uses the top \qty{25}{\percent} of \acs*{CC} by just the fastText classifier's scores.
    Random \acs*{CC} is substring-deduplicated crawl with no further filtering.
    Each other cell is the mean \(\pm\) half-width of the \qty{95}{\percent} credible interval over \(n\) training seeds.
    \num{0.4}\mbox{B} dense \(1\times\) Chinchilla \acs*{DCLM} (\num{8.2}\mbox{B} tokens), \(n=3\) (annotated \acs*{CC}), \(n=5\) (random \acs*{CC}, fastText, and \model{1}); \num{7.8}\mbox{B}-A\num{0.6}\mbox{B} \acs*{MoE} (\num{97.6}\mbox{B} tokens), \(n=3\) (annotated \acs*{CC}), \(n=2\) (fastText and \model{1}), \(n=1\) (random \acs*{CC}).}
  \label{tab:quality-cc-vs-p1}
\end{table}

\paragraph{Limitations}
In \model{2}, we use only a single distilled judge and the fastText classifier for annotations, and apply this procedure for our English data only.
We wish to extend this technique to other annotations and languages.
Another limitation is that the simplex mix is over five scores, whereas we expect notions of \enquote{quality} or signals of what data is useful for language modelling to be even higher-dimensional.

\FloatBarrier

\subsubsection{Data Source Mixing}
\label{sec:data-mixing}
Having filtered and annotated the \ac{CC} data and included the other data sources in our pre-training data pool, we fit a mixing law on a swarm of proxy models to set each data source's sampling weight, following an approach inspired by Olmix~\parencite{chen2026olmix}.
The mix ranges over all the \num{78} data sources of the pre-training pool.
These sources include our English and German \ac{CC} \enquote{quality}-annotated buckets.
We form these buckets by ranking documents on the optimised mix of annotations of \cref{sec:quality-annotation}, a blend of five annotation scores for English and the average of two GermanWeb classifiers for German, and splitting each language into five equal-token-mass tiers.
We include only the higher tiers, dropping the two lowest for English (i.e.\ including the top three, about \qty{60}{\percent} of the token mass) and the lowest for German (i.e.\ including the top four, about \qty{80}{\percent} of the token mass).
The objective is the equal-weighted mean \ac{BPB} over \num{24} benchmarks, \num{18} English and \num{6} German \footnote{Among the English benchmarks are \acs{MMLU}, \acs{ARC}, HellaSwag, HumanEval, \acs{MBPP}, CommonsenseQA, DROP, NaturalQsOpen, \acs{PIQA}, SocialIQA, and SQuAD. The German benchmarks are Global-\acs{MMLU}, \acs{ARC}, HellaSwag, HumanEval, \acs{MBPP}, and \acs{GSM8K}.
}.

\paragraph{Sampling and Ablations for Optimisation}
We train a swarm of \num{15}M-parameter dense proxy models, each on \num{1.5}B tokens of a different mixture, and record its per-benchmark \ac{BPB}.
The swarm is \num{3354} proxy trainings in total.
Mixtures are drawn from a Dirichlet distribution centred on the natural token-count prior \(q\propto\text{tokens}\), at \num{15} concentrations spaced log-uniformly in \([0.1,50]\), with \(3(D{+}1)\) mixtures per concentration for the \(D=72\) sources, plus the \(D\) one-hot corners; three corner runs were removed by hand to exclude their datasets post hoc.

\paragraph{Optimisation and Materialisation}
To predict each benchmark's \ac{BPB} from a given data mix, we fit a log-linear law over the swarm of proxy models.
We use log-linear mixing laws \parencite{ye2025datamixinglaws}, which \textcite{chen2026olmix} found to perform best.
For each benchmark \(t\), intercept \(a_t\), coefficient vector \(b_t\), and data mix \(x\), we fit \[ \mathrm{BPB}_t(x)=\exp(a_t)+\exp(b_t\cdot x) \] over the swarm by limited-memory Broyden--Fletcher--Goldfarb--Shanno (L-BFGS) optimisation with a Huber loss and multi-start initialisation.
We then pick the mix \(x\) by solving the convex program
\begin{equation}
  \min_{x}\ \sum_t w_t\,\exp(b_t\cdot x)\;+\;\lambda\,\mathrm{KL}(x\,\|\,\tilde{q}), \qquad x\ge 0,\ \textstyle\sum_i x_i=1,\ x_i\le\kappa_i,
  \label{eq:mixing-solve}
\end{equation}
where \(w_t\) are the objective weights, \(\tilde{q}\propto\text{tokens}^{1/T}\) is a temperature-scaled token prior, and the caps \(\kappa_i=\text{tokens}_i\cdot r/R\) forbid a source from repeating more than \(r\) times at the target token budget \(R\) of \num{20}\mbox{T} tokens.
The \ac{KL} pull trades quality against repetition: at \(\lambda=0\) the solver ignores source sizes and picks only the best-fitting sources, while a large \(\lambda\) reproduces the size prior.
We settle on \(\lambda=0.05\), \(T=3\), and \(r=3\), which keeps every small source represented at least once while staying close to the prior.
On the proxy objective the chosen mix is marginally worse than the unconstrained optimum (predicted \num{3.670 \pm 0.002} versus \num{3.667 \pm 0.002} mean \ac{BPB}), which we consider a reasonable trade-off for keeping the smaller sources.
The materialised mix draws \qty{35.7}{\percent} of its training tokens (\num{7.15}T) from our own pipeline and \qty{64.3}{\percent} (\num{12.85}T) from external sources; \cref{tab:data-mix-composition} breaks down our own contribution by family.

\begin{table}[ht]
  \centering
  \small
  \begin{tabular}{lcc}
    \toprule
    Source family & Tokens & Sources \\
    \midrule
    \acs*{CC} quality buckets & \num{3.82}\mbox{T} & \num{7} \\
    German FineWeb rephrases  & \num{1.74}\mbox{T} & \num{5} \\
    Synthetic rephrases (de/en) & \num{1.31}\mbox{T} & \num{6}  \\ %
    External web data, rephrased in-house & \num{0.24}T & \num{1} \\
    Small specialised datasets  & \num{0.045}T       & \num{3}  \\ %
    \midrule
    Our pipeline (total) & \num{7.15}T  & \num{22} \\
    External sources     & \num{12.85}T & \num{56} \\
    \bottomrule
  \end{tabular}
  \caption{Tokens drawn from our own data-pipeline families in the materialised \model{2} pre-training mix (\(\lambda=0.05\), \(T=3\), \(3\times\) repetition cap), out of a \num{20}T-token budget.
    External sources supply the remaining \qty{64.3}{\percent}.
  }
  \label{tab:data-mix-composition}
\end{table}

\FloatBarrier
\paragraph{Limitations and Learnings}
We fit the data mix weights on models with \num{15}M parameters trained on \num{1.5}B tokens, which makes both the fit and the swarm noisy.
We find that the optimiser of the data mix weights is sensitive to the number of swarm samples, and the prior and its two parameters (\(\lambda\), \(T\)) strongly influence the predicted optimum.
It is not clear whether this is a limitation of the optimiser or a property of the data mix problem.
The natural prior is strongly skewed towards the largest sources, meaning that smaller sources are typically less well explored even at the lowest concentrations.
Some of our German quality-classifier buckets are down-weighted relative to their labels, so the mix optimisation does not appear to comport with the German quality ordering.
We wish to re-evaluate these annotations and their mixing in the future.
We also patched several token-count errors (byte-versus-token miscounts and a bucket mismatch) before the final solve.
The sampling prior for the swarm ablations, computed from the old counts, cannot be corrected after the fact.
These errors are likely to have a small impact on the final mix, and we aim to avoid them in the future.

\FloatBarrier
\subsection{Stage 2: Mid-Training}
\label{sec:mid-training}

Mid-training continues training on higher-quality math, code, reasoning, and instruction data to strengthen these capabilities.
Following \textcite{hu2024minicpm}, who reserve higher-quality data for the later phases of training, we draw this data from the mid-training pool, a curated set of datasets (\cref{sec:mid-pool}), and a mix (\cref{sec:mid-method}) sets the number of tokens each dataset contributes.
Mid-training keeps the constant learning rate and the checkpoint merging of pre-training (\cref{sec:bienenelfe-training-recipe}).
The stage ends in the base checkpoint from which long-context extension (\cref{sec:long-context}) continues training at the same learning rate.

\subsubsection{Dataset Pool}
\label{sec:mid-pool}

The mid-training pool is the set of datasets from which both the mid-training mixes and the long-context mixes (\cref{sec:long-context}) draw.
We collect candidate datasets from public releases and from the pre-training datasets, drop every candidate dataset whose content largely overlaps another, and assign each dataset to one of the nine domains of \cref{sec:stage-1-data}.
The pool excludes the low-quality tiers of web data and, for German, the rephrased and translated web text.
The German exclusion rests on manual inspection, in which rephrased and translated text read less naturally than organic German text.

\begin{figure}[htbp]
  \centering
  \begin{minipage}[b]{0.62\linewidth}
    \centering
    \scriptsize\AATabularFace
    \setlength{\tabcolsep}{2pt}
    \begin{tabular}[b]{@{}c@{\hspace{0.4em}}llrrrr@{}}
      \toprule
      & Domain
      & Short
      & Datasets
      & \shortstack{Unique\\tokens (B)}
      & \shortstack{After\\treatment (B)}
      & \shortstack{Under\\repetition\\caps (B)} \\
      \midrule
      \domSwatch{0}&\domEnWeb&\domEnWebShort&$12$&$4{,}903$&$4{,}332$&$14{,}352$\\
      \domSwatch{1}&\domDeWeb&\domDeWebShort&$11$&$957$&$869$&$2{,}171$\\
      \domSwatch{2}&\domEnInstr&\domEnInstrShort&$8$&$1{,}269$&$1{,}223$&$2{,}643$\\
      \domSwatch{3}&\domDeInstr&\domDeInstrShort&$3$&$41$&$41$&$101$\\
      \domSwatch{4}&\domCode&\domCodeShort&$13$&$2{,}688$&$1{,}935$&$5{,}976$\\
      \domSwatch{5}&\domAgentic&\domAgenticShort&$5$&$605$&$601$&$2{,}405$\\
      \domSwatch{6}&\domStemDocs&\domStemDocsShort&$9$&$1{,}052$&$827$&$2{,}650$\\
      \domSwatch{7}&\domStemQA&\domStemQAShort&$11$&$457$&$431$&$1{,}394$\\
      \domSwatch{8}&\domSpecialized&\domSpecializedShort&$15$&$301$&$247$&$812$\\
      \midrule
      &Total&&$87$&$12{,}273$&$10{,}505$&$32{,}503$\\
      \bottomrule
    \end{tabular}
  \end{minipage}\hfill
  \begin{minipage}[b]{0.36\linewidth}
    \centering
    \includegraphics{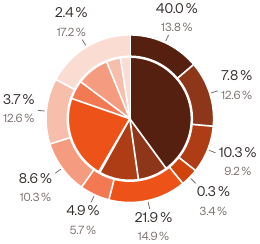}
  \end{minipage}

  \begin{minipage}[t]{0.62\linewidth}
    \captionof{table}{The mid-training pool by domain, in billions of tokens.
      Short is the form the tables use.
      Unique tokens are the estimate used for mixing.
      After treatment counts the tokens of the treated datasets after the further decontamination, which removes \qty{0.06}{\percent} of them.
      Under repetition caps is the most that the repetition caps allow mid-training to draw from these tokens.
    }
    \label{tab:mid-pool-domains}
  \end{minipage}\hfill
  \begin{minipage}[t]{0.36\linewidth}
    \caption{Share of the mid-training pool by domain, in unique tokens (inner disk) and in datasets (outer ring), colours as in the adjacent \cref{tab:mid-pool-domains}.
      Each label gives the domain's share of the unique tokens and, below it, its share of the datasets.
    }
    \label{fig:mid-pool-domains}
  \end{minipage}
\end{figure}

Every dataset of the pool passes the cleanup pass of \cref{sec:heuristic-filters}, exact deduplication across the whole pool (\cref{sec:exact-dedup}), and \ac{PII} redaction.
We call these three steps the treatment, and the ablations (\cref{sec:mid-evals}) train on the treated datasets.
The production mix draws from the treated datasets after a further decontamination against the evaluation suite, which removes \qty{0.06}{\percent} of their tokens (\cref{sec:mid-decon}).
The size of a dataset is its number of unique tokens, estimated from its size in bytes as for the pre-training pool.
The repetition cap of a dataset limits its draws across pre-training and mid-training to four epochs (\cref{sec:mid-method}).
Pre-training epochs count against the unique tokens of the dataset, and production mid-training may draw the remaining epochs over its tokens after treatment and decontamination.
\Cref{tab:mid-pool-domains} lists, per domain, the number of datasets, the unique tokens, the tokens after treatment and decontamination, and the most that the repetition caps of \cref{sec:mid-method} allow mid-training to draw, and \cref{fig:mid-pool-domains} shows the share of each domain in unique tokens and in datasets.
Documents above \num{128}k estimated tokens are reserved for long-context extension (\cref{sec:long-context}), since they are scarce and only its sequence length of \longContextSeqLen{} can fully make use of them.

\subsubsection{Ablation Setup}
\label{sec:mid-evals}

The ablations use a proxy model, a downscaled \model{2} with the same hybrid architecture, \num{11.9}B total and \num{0.65}B active parameters, pre-trained for \num{2.79}T tokens.
We mid-train that checkpoint on every mix for \num{84}B tokens, \qty{3}{\percent} of the proxy's pre-training tokens, at the production sequence length of \midTrainingSeqLen{} tokens.
Ten of the \num{18} runs at this budget train the hand-set \textbf{c}andidates C1 to C3, the optimised \textbf{s}olutions S1 to S6 (\cref{sec:mid-method}), and the production mix (\cref{sec:mid-effect}).
The other eight train the mixes L1 to L6, which vary the in-domain composition of the two \textbf{l}anguage domains (\cref{sec:mid-lang-sweep}), a second run of C1, entered through a gradual transition from the pre-training mix, and the \model{1} mid-training mix (\cref{tab:mid-ranking}).
Two of the hand-set candidates, C2 and C3, and the production mix also run for \num{400}B tokens, \qty{14}{\percent} of the proxy's pre-training tokens and closer to the \qty{17}{\percent} of production (\num{3.44}T after \num{19.99}T), and a baseline run of the same length continues pre-training on the pre-training mix at its sequence length of \preTrainingSeqLen{} tokens.
Every score is that of a merged checkpoint, the uniform average of the weights of the last \(k\) saved checkpoints \parencite{izmailov2018swa,sanyal2023early}, written M\(k\), and \basemodel{2} is the M20 merge at the end of long-context extension.
Merging adds \numrange{3.4}{11.5} points over the unmerged endpoint in these ablations, so comparisons use merges of the same \(k\) only.
The aggregate score averages the general-knowledge, math, and code means of each language and weights English and German \num{80}/\num{20}, and \cref{sec:ablation-final-score} lists the benchmarks.
For \num{16} of the \num{18} runs at \num{84}B and for every run at \num{400}B, we also report the pooled score of the post-training benchmarks after an \ac{SFT}, which measures the part of a mid-training gain that persists afterwards.
All \acp{SFT} share one recipe and one data mix, so their differences come from mid-training alone.
An eval-noise model, fitted to replicate \acp{SFT} with the standard errors of \textcite{miller2024errorbars}, gives each post-training benchmark score a \qty{95}{\percent} interval (\cref{sec:mid-ablations}).

\subsubsection{Mixing Method}
\label{sec:mid-method}

We search for the mix with the highest aggregate score (\cref{sec:mid-evals}) within the repetition caps of the scarce high-quality datasets of the pool.
Optimising the domain weights for this score alone sets four domains to zero (\cref{sec:mid-lessons}).

\paragraph{Mixes and Sampling Rules}
A mid-training mix assigns every dataset of the mid-training pool a share of the token budget along two axes.
The domain weights give each of the nine domains its share of the budget, and the in-domain composition splits the budget of a domain among its datasets.
An optimiser in the style of Olmix~\parencite{chen2026olmix}, fitted to the scores of MergeMix expert merges \parencite{wang2026mergemix} as described below, proposes domain weights.
The optimised solutions S1 to S6 (\cref{tab:mid-mixes}) take their domain weights from four solves of this optimiser.
The in-domain composition follows one of two sampling rules.
Under the natural rule, each dataset's share of its domain is proportional to its unique tokens (\cref{sec:mid-pool}).
The pre-training-delivered rule keeps the pre-training proportions of the datasets that pre-training drew from and gives datasets new to mid-training their natural share.
In addition, we hand-set three candidates, C1 to C3 (\cref{tab:mid-mixes}), which serve as the baselines for the optimised solutions and as the reference mixes of the \ac{KL} penalty.
C1 and C2 share their domain weights and differ only in the sampling rule, natural for C1 and pre-training-delivered for C2.
The optimised solutions and the production mix use the pre-training-delivered rule.

\paragraph{Repetition Caps}
High-quality data is scarce, so a mid-training mix repeats its highest-quality datasets.
The repetition cap of a dataset limits its draws across pre-training and mid-training combined to four epochs, the number of epochs up to which training on repeated data reaches approximately the loss of training on as many unique tokens \parencite{muennighoff2023scaling}.
Within each domain, a waterfill assigns every dataset its share under the sampling rule, pins a dataset at its repetition cap if its share exceeds that repetition cap, and redistributes the excess to the still unpinned datasets of the domain in proportion to their shares.

\paragraph{Expert Merging and Optimisation}
In MergeMix, the proxy model is mid-trained on each of the nine domains alone for \num{20}B tokens, and a merge of these expert checkpoints with the weights of a mix approximates a mid-training run on that mix.
From a shared initialisation, merging the experts and training on the mixed data agree to first order, and over \num{12} mixes, expert merges rank the mixes as full mid-training runs do, with a Spearman correlation above \num{0.9} \parencite{wang2026mergemix}.
Expert merging thus allows us to evaluate \num{113} weightings, \num{100} sampled from Dirichlet distributions around our initial hand-set mixes and \num{13} placed by hand to cover the range of the four smallest domains.
Following Olmix~\parencite{chen2026olmix}, we fit one log-linear law per benchmark to the expert-merge scores and solve for the weights that maximise the aggregate score, subject to the repetition caps at a \num{5}T budget and a \ac{KL} penalty with weight \(\lambda\) towards a reference mix, as in \cref{eq:mixing-solve}, which keeps the solution near a hand-set candidate.
The pre-training evaluation suite (\cref{sec:pre-training-eval-scores}) has no benchmark that targets the instruction and reasoning domains or \domSpecialized{} directly.
These domains thus enter the objective only through their effect on other benchmarks, the instruction and reasoning domains through math and code, and \domSpecialized{} through none of them.

\paragraph{Optimised Solutions S1 to S6}
Six solutions enter the proxy ablations (\cref{tab:mid-mixes}) next to the hand-set candidates (\cref{sec:mid-ablation-results}).
They vary the reference mix (C2 or C3), the \ac{KL} penalty weight \(\lambda\) (\num{0}, \num{0.01}, or \num{0.05}), and whether their in-domain composition is projected from the \num{5}T budget, at which the repetition caps pin datasets.
At the ablation budget every dataset stays below its repetition cap, so S2 and S5 follow the sampling rule unchanged.
For the projected S1, S3, S4, and S6, the waterfill runs at \num{5}T and its result is scaled down to the ablation budget, so the datasets pinned at \num{5}T lose share to the other datasets of their domain.
S2 and S3 come from the same solve, as do S5 and S6, and the domain weights of each pair differ by less than one point.
In S3 and S6, the projection changes the in-domain composition of \domEnInstr{}, \domDeInstr{}, and \domStemQA{}, and in S3 also that of \domCode{}.
\Cref{app:mid-mixing-method} gives further details on the repetition caps, the optimisation, and the solutions.

\begin{table}[htbp]
  \centering
  \footnotesize
  \setlength{\tabcolsep}{3pt}
  \begin{tabular}{lllllccccccccc}
    \toprule
    & & & & & \multicolumn{9}{c}{Domain weights (\%)} \\
    \cmidrule(lr){6-14}
    Mix
    & Prior
    & \acs{KL}
    & Projection
    & Sampling
    & \rotatebox{90}{\domEnWebShort}
    & \rotatebox{90}{\domDeWebShort}
    & \rotatebox{90}{\domEnInstrShort}
    & \rotatebox{90}{\domDeInstrShort}
    & \rotatebox{90}{\domCodeShort}
    & \rotatebox{90}{\domAgenticShort}
    & \rotatebox{90}{\domStemDocsShort}
    & \rotatebox{90}{\domStemQAShort}
    & \rotatebox{90}{\domSpecializedShort} \\
    \midrule
    C1&--&--&--&natural&$33.0$&$19.0$&$10.0$&$1.0$&$10.0$&$8.5$&$6.5$&$9.0$&$3.0$\\
    C2&--&--&--&pre-training-delivered&$33.0$&$19.0$&$10.0$&$1.0$&$10.0$&$8.5$&$6.5$&$9.0$&$3.0$\\
    C3&--&--&--&pre-training-delivered&$14.0$&$18.0$&$16.2$&$1.4$&$8.0$&$16.2$&$6.0$&$16.2$&$4.0$\\
    \midrule
    S1&C2&0&\num{5}\mbox{T}&pre-training-delivered&$0.0$&$0.0$&$31.0$&$1.4$&$42.1$&$0.0$&$0.3$&$25.1$&$0.0$\\
    S2&C2&0.01&--&pre-training-delivered&$3.1$&$0.4$&$31.5$&$1.4$&$29.4$&$7.8$&$10.8$&$15.5$&$0.1$\\
    S3&C2&0.01&\num{5}\mbox{T}&pre-training-delivered&$3.1$&$0.4$&$30.7$&$1.4$&$29.8$&$7.9$&$10.9$&$15.7$&$0.1$\\
    S4&C2&0.05&\num{5}\mbox{T}&pre-training-delivered&$20.3$&$2.3$&$14.9$&$1.4$&$27.1$&$9.7$&$13.7$&$9.6$&$1.0$\\
    S5&C3&0.01&--&pre-training-delivered&$1.3$&$2.1$&$31.5$&$1.4$&$16.4$&$15.5$&$6.0$&$25.5$&$0.2$\\
    S6&C3&0.01&\num{5}\mbox{T}&pre-training-delivered&$1.3$&$2.2$&$31.0$&$1.4$&$16.8$&$15.9$&$6.1$&$25.1$&$0.2$\\
    \midrule
    production mix&--&--&\num{5.2}\mbox{T}&pre-training-delivered&$1.3$&$2.1$&$31.4$&$1.9$&$16.3$&$15.4$&$5.9$&$25.4$&$0.2$\\
    \bottomrule
  \end{tabular}
  \caption{The hand-set candidates C1 to C3, the optimised solutions S1 to S6, and the production mix, with the reference mix (prior) and \acs*{KL} penalty of each solution, and the projection budget, sampling rule, and domain weights of each mix.
    Every mix draws from the mid-training pool.
    Projection gives the budget at which the waterfill sets the in-domain composition before it is scaled to the trained budget, and -- marks a mix whose in-domain composition follows its sampling rule, as no repetition cap binds at the ablation budget.
    The production mix is S5 with \domDeInstr{} raised to its repetition cap (\cref{sec:mid-effect}).
    \Cref{tab:mid-candidates} adds the mixes L1 to L6, which vary the in-domain composition of the two language domains.
  }
  \label{tab:mid-mixes}
\end{table}

\subsubsection{Production Mix and Its Effect}
\label{sec:mid-effect}

\paragraph{Production Mix}
The production mix is S5, the optimised solution with the C3 prior and \ac{KL} penalty \num{0.01}, with \domDeInstr{} raised to its repetition cap (\cref{app:mid-mixing-method}).
A waterfill at a \num{5.2}T budget sets its in-domain composition (\cref{sec:mid-method}).
We take C3 as the prior because it leads C2 by \num{3.5} points after \ac{SFT} at \num{400}B (\cref{tab:mid-horizon-summary,fig:mid-horizon}).
At \num{84}B, the \ac{SFT} scores of the solutions with penalties \num{0}, \num{0.01}, and \num{0.05} lie within one point of each other, below the resolution of the eval-noise model, so the choice of the penalty rests on the domain weights.
Without the penalty, the solve sets four domains to zero, and at \num{0.05}, the weight that \textcite{chen2026olmix} recommend, the solution stays closest to its hand-set prior.
We thus take \num{0.01}, the smallest tested penalty at which every domain keeps a positive weight.
In this mix, \domEnInstr{}, \domStemQA{}, and \domCode{} hold \qty{73}{\percent} of the budget together, and the two web domains hold \qty{3.4}{\percent} (\cref{tab:mid-mixes}).
We mid-train \model{2} on \num{3.44}T tokens of this mix and interpolate linearly from the pre-training mix to it over the first \(H = \num{100}\)B tokens (\cref{sec:transition-index}).

\paragraph{Decontamination}
Benchmark text in the late phase of training is memorised and, under a constant learning rate as in \model{2}, also inflates benchmark scores \parencite{magar2022contamination}.
Benchmark text in the early phase is forgotten once the subsequent clean data exceeds about five times the compute-optimal token count \parencite{bordt2024forget}.
Following Olmo~3 \parencite[Sec.~3.5.3]{olmo2026olmo3}, we therefore decontaminate the mid-training pool, from which long-context extension also draws, and remove every document that contains text from the \num{226} tasks of our evaluation suite, with all their splits.
The \texttt{decon} algorithm~\parencite{allenai2025decon} flags a document by matching \ac{IDF}-weighted n-grams of the questions and answers, and a flagged document is dropped as a whole.
Decontamination drops \num{1.5}M of the \num{8.3}B scanned documents, \qty{0.06}{\percent} of the tokens (\cref{sec:mid-decon}).
The pre-training pool remains potentially contaminated, and the HumanEval scores reflect this contamination.

\paragraph{HumanEval Recitation}
A HumanEval completion counts as recited iff it matches the reference solution shipped with the benchmark (\cref{sec:mid-humaneval}).
During pre-training, \qtyrange{22}{95}{\percent} of the completions are recited, and pass@1 follows the recited share, with a Pearson correlation of \num{0.90} over \num{96} evaluations.
Within the first \num{0.3}T tokens of mid-training, the recited share drops below \qty{10}{\percent} in both languages, and English pass@1 decreases from \num{80.0} to \num{70.0}.
On the problems without a recited completion, English pass@1 stays at \numrange{61}{76} throughout mid-training (\cref{fig:mid-recitation}).
By the end of mid-training, the recited share rises again to \qty{48}{\percent} in English and \qty{47}{\percent} in German, so the HumanEval scores at that point remain inflated by recitation.
We thus also report the aggregate score without HumanEval (\cref{sec:ablation-final-score}) and plan to decontaminate the pre-training pool for the next model.

\paragraph{Proxy Runs}
At the \num{400}B proxy budget, the production mix leads continued pre-training by \num{4.9} points on the M4 merge and by \num{11.5} points after \ac{SFT}, and it leads C3, the reference mix of its \ac{KL} penalty, by \num{3.8} points after \ac{SFT} (\cref{tab:mid-horizon-summary}).
Both differences after \ac{SFT} exceed the \qty{95}{\percent} eval-noise intervals (\cref{sec:mid-ablations}), so the production mix, derived from S5, improves on the hand-set candidate it is regularised towards.
Across the \num{84}B runs, HellaSwag (DE) rises with the domain weight of \domDeWeb{} in a mix (\cref{sec:mid-ablation-knowledge}).

\begin{table}[htbp]
  \centering
  \small
  \begin{tabular}{lccccc}
    \toprule
    Mix & M4 score & \acs*{SFT} score & \(\pm\) & TriviaQA & HellaSwag (DE) \\
    \midrule
    C2&$61.1$&$44.0$&$1.7$&$57.3$&$77.6$\\
    C3&$62.4$&$47.5$&$1.8$&$55.1$&$76.0$\\
    production mix&$65.5$&$51.3$&$1.9$&$55.0$&$70.4$\\
    continued pre-training&$60.6$&$39.7$&$1.6$&$63.5$&$77.6$\\
    \bottomrule
  \end{tabular}
  \caption{The four runs of \num{400}\mbox{B} tokens, by the aggregate score of the M4 merge, the pooled score of the \acs*{SFT} on it with the half-width of its \qty{95}{\percent} eval-noise interval, and the TriviaQA and HellaSwag (DE) scores of the M4 merge.
    Continued pre-training keeps training on the pre-training mix at its sequence length.
    The scores of the same mixes at \num{84}B are in \cref{tab:mid-horizon}.
  }
  \label{tab:mid-horizon-summary}
\end{table}

\begin{figure}[htbp]
  \centering
  \includegraphics{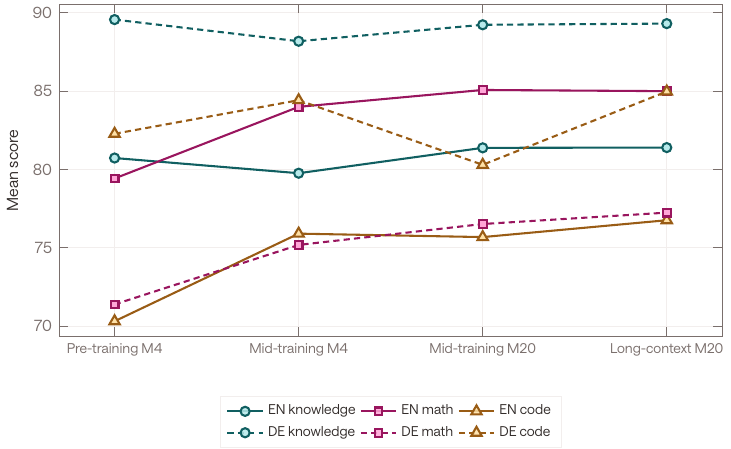}
  \caption{The general-knowledge, math, and code means of the pre-training suite per language along the production chain, at the M4 merge at the end of pre-training, the M4 and M20 merges at the end of mid-training, and the M20 merge at the end of long-context extension, which is the final base model.
    \Cref{tab:mid-production} lists the values.
  }
  \label{fig:mid-production}
\end{figure}

\paragraph{Production Run}
From the M4 merge at the end of pre-training to the M4 merge at the end of mid-training, the aggregate score of \model{2} rises by \num{2.8} points, or \num{2.9} points without HumanEval, with the gain in math and code in both languages (\cref{fig:mid-production}).
Over the same span, the general-knowledge means fall by \num{1.0} points in English and \num{1.4} in German, with gains on \ac{MMLU}-Pro and \ac{ARC} and losses on TriviaQA and both HellaSwag benchmarks.
After long-context extension, the final base model still scores \num{5.2} points below the pre-training M4 on TriviaQA and \num{4.5} on HellaSwag (DE).
The loss on HellaSwag (DE) is as predicted by the proxy runs, and the cause of the TriviaQA loss remains open (\cref{sec:mid-ablation-knowledge}).

\subsubsection{Lessons for Choosing a Mix}
\label{sec:mid-lessons}

\paragraph{Regularisation towards a Hand-Set Candidate}
Without the \ac{KL} penalty, the solve sets \domEnWeb{}, \domDeWeb{}, \domAgentic{}, and \domSpecialized{} to \qty{0}{\percent} and \domStemDocs{} to \qty{0.3}{\percent} (S1 in \cref{tab:mid-mixes}).
We hypothesise that the weighting of the aggregate score causes this outcome, since it weights math and code two to one over general knowledge.
Across the \num{84}B runs, the English knowledge mean correlates with the English math mean, the \ac{MBPP} mean, and the weight of \domEnWeb{} at a Pearson \(r\) of \num{0.25}, \num{0.03}, and \num{-0.11}, and the weight of \domEnWeb{} correlates with the English math mean at \num{-0.67} (\cref{tab:mid-weight-correlations}).
With the penalty, the two web domains retain \qty{22.6}{\percent} at \num{0.05} towards C2 (S4) and \qtyrange{3.5}{3.6}{\percent} at \num{0.01} (S2 and S3).
The penalty acts on the domain weights as a whole, so at \num{0.01} towards C3, \domDeWeb{} drops from \qty{18}{\percent} in C3 to \qty{2.1}{\percent} in the production mix.
At \num{400}B, the HellaSwag (DE) score of the production mix lies \num{5.5} points below that of C3, a cost of \num{0.09} points in the aggregate score (\cref{tab:mid-horizon-summary}).
Preserving a capability thus requires its own term or constraint in the objective.

\paragraph{Proxy Budget}
At \num{400}B, every adjacent pair of the four runs differs after \ac{SFT} beyond the eval noise, by \num{4.3} points from continued pre-training to C2, \num{3.5} from C2 to C3, and \num{3.8} from C3 to the production mix (\cref{tab:mid-horizon-summary}).
At \num{84}B, C2 and C3 differ by \num{0.4} points after \ac{SFT} and by \num{0.03} on the merge, and the \ac{SFT} separates only the lowest runs from the highest, in \num{14} of \num{120} pairs (\cref{sec:mid-ablation-results}).
Relative to the pre-training tokens, the \num{400}B budget is \qty{83}{\percent} of the production ratio and the \num{84}B budget \qty{17}{\percent} of it, so the mixes separate after \ac{SFT} at the larger proxy budget only.
With two budgets, the ratio at which this separation sets in remains open.

\paragraph{Merge Score and \Acs*{SFT} Score}
The M4 merge and the \ac{SFT} order the four \num{400}B runs alike, and the gaps after \ac{SFT} are larger, \num{4.3} points between C2 and continued pre-training against \num{0.5} on the merge (\cref{tab:mid-horizon-summary,fig:mid-merge-sft}).
On these four runs, the merge score thus appears to understate the gain of a mix after \ac{SFT}.
\subsection{Stage 3: Long-Context Extension}
\label{sec:long-context}

Our strategy is to extend the context of \model{2} to \recommendedContextLengthShort{} while preserving the capabilities of the mid-trained model, by interleaving naturally long documents with high-quality mid-training data rather than training only on long-context data.
We take a fixed share of naturally long documents from our pre-training pool and ablate how aggressively the mid-training pool is reweighted towards longer documents.
Overall, increasing the share of length-reweighted mid-training data improves long-context retrieval performance, while checkpoint merging largely preserves general capability.

\subsubsection{Dataset Pool and Candidates} %

Long-context extension draws from three data pools.
The first is the original mid-training mix, processed as before with documents longer than \num{128}k tokens removed, and about \qty{70}{\percent} of its tokens lie in documents shorter than \num{8}k tokens.
The second pool contains the datasets of the mid-training pool (\cref{sec:mid-pool}) without the document-length limit, which allows sampling to shift towards longer documents from the same data sources.
The third pool is a dataset of naturally long documents between \num{8}k and \recommendedContextLengthShort{} tokens, from which we remove the synthetically generated samples to avoid artificially favourable results on synthetic long-context benchmarks such as RULER.

Every candidate is a mix of \num{200}B tokens that takes one third from the naturally long pool, a share \(s\) from the mid-training reweighted pool, and the remaining \(2/3 - s\) from the natural mid-training distribution.
We materialise and train three settings, \(s \in \{0, 1/4, 1/3\}\).

\paragraph{Length Reweighting}

In the reweighted share, each document-length band is assigned a weight on a logistic curve over the band index \(b\), which runs from \(0\) for documents up to \num{8}k tokens to \(5\) for \num{128}k to \recommendedContextLengthShort{} tokens: \[ w(b) = \frac{1}{1 + e^{-(b - 4)}}.
\] The weight is \num{0.5} for the \num{64}k to \num{128}k band and \num{0.73} for the longest band.

The natural and reweighted shares both draw from the same mid-training datasets, so their sampling rates must be combined into a single per-band budget to avoid assigning the same documents independently to both shares.
The resulting sampling rate for length band \(b\) is \[ r(b) = \min\paren*(r_{\mathrm{natural}}(b) + q \, w(b), r_{\max}), \] where \(r_{\mathrm{natural}}(b)\) is the sampling rate required by the \((2/3-s)\) share drawn according to the original mid-training distribution and is zero above \num{128}k tokens.
The scale \(q\) is shared across all bands and is solved by bisection such that the reweighted component contributes its target budget of \(s\) times the total mix size.
For a \num{200}B-token candidate, the three components therefore contribute \(1/3 \times 200\)B tokens from naturally long documents, \(s \times 200\)B from the length-reweighted mid-training pool, and \((2/3-s)\times 200\)B from the original mid-training distribution.

\begin{figure}[tb]
  \centering
  \includegraphics{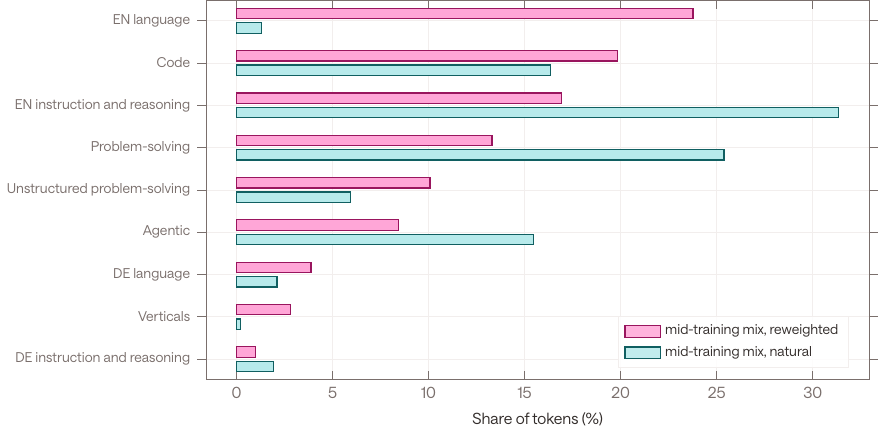}
  \caption{Domain weights of the part of the production long-context mix drawn from the mid-training pool, compared with those of the original mid-training mix.}
  \label{fig:lc-mix-domains}
\end{figure}

No repetition cap \(r_{\max}\) binds in any candidate, so no document is repeated.
One scale for the whole pool means the reweighted share is dominated by whichever sources hold long documents, so its domain weights drift away from those of the mid-training mix, which \cref{fig:lc-mix-domains} quantifies.

\begin{figure}[tb]
  \centering
  \includegraphics{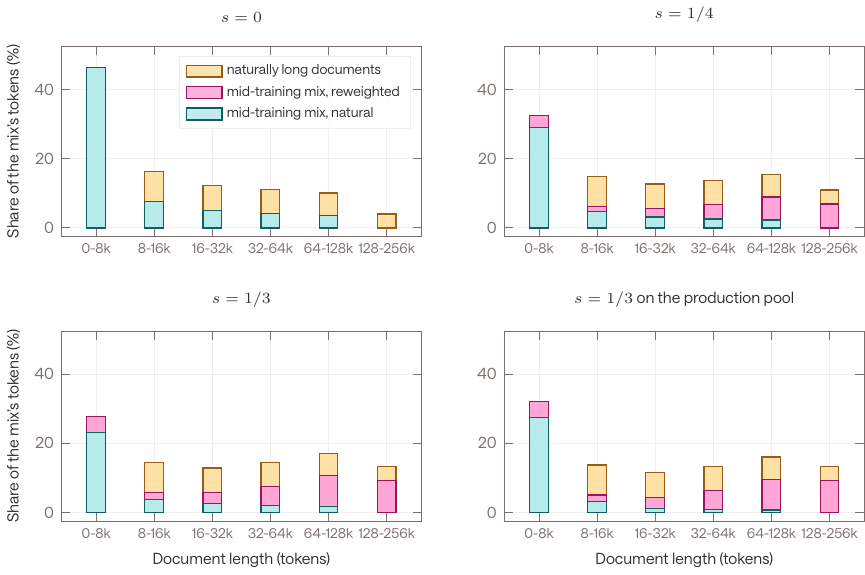}
  \caption{Length distribution of the long-context mix candidates, as the materialised token budget of each share per document-length band in percent of the mix's tokens.
    \(s\) is the share of the mix drawn from the mid-training mix with the length reweighting, and the portion of naturally long documents is identical across the three candidates.
    The last panel is the production mix, the \(s = 1/3\) recipe rebuilt on the production mid-training pool at proxy scale.
  }
  \label{fig:lc-mix-bands}
\end{figure}

\Cref{fig:lc-mix-bands} illustrates the resulting length distribution of each candidate as materialised budgets.
Varying the reweighted share \(s\) from \(0\) to \(1/3\) increases the fraction of tokens of documents longer than \num{32}k from \qty{25}{\percent} to \qty{45}{\percent}, and of documents longer than \num{128}k from \qty{4}{\percent} to \qty{13}{\percent}, while the allocation to naturally long documents stays identical.

\paragraph{Domain Shift}

Without controlling for domains, unconstrained length reweighting risks shifting the domain weights of the resulting mix.
\Cref{fig:lc-mix-domains} shows this effect by comparing the domain weights of the reweighted production mid-training pool with those of the original mid-training mix.

Because long documents are concentrated in web text and scanned material, the English Language domain rises from \qty{1.3}{\percent} to \qty{24}{\percent}.
In contrast, instruction-following, problem-solving, and agentic domains are short by construction and fall to about half their share of the original mid-training mix.

This shift is a consequence of the length reweighting rather than an explicit design choice, and the candidates were not constructed to control for it.
We therefore cannot isolate the effect of the domain shift from the effect of document length in the corresponding evaluation results.
Future experiments should treat domain weights and length distribution as separate ablation dimensions.

\subsubsection{Evals and Ablations}

\paragraph{Setup}
The ablation primarily sweeps the share of mid-training data that is reweighted towards longer documents.
We train three long-context candidates with increasing reweighted shares, together with a control using \qty{100}{\percent} of the original mid-training mix.

All four branches start from the same mid-training checkpoint after about \num{330}B mid-training tokens and train for \num{2087} steps at a \recommendedContextLengthShort{} sequence length, corresponding to about \num{70}B tokens, with \num{200} warmup steps and a constant learning rate.
The data mix is therefore the only difference between branches.

Each branch is evaluated at nine checkpoints on RULER \parencite{hsieh2024ruler} up to \recommendedContextLengthShort{} tokens and on HELMET \parencite{yen2025helmet} up to \num{128}k tokens, as well as on the general-capability suite of the main run.
We also evaluate a merged checkpoint obtained by uniformly averaging the branch endpoint with three earlier mid-training checkpoints spaced roughly \num{70}B tokens apart.

\paragraph{Results}

We evaluate the candidates using an aggregate over RULER and HELMET: each leg is divided by its mean over the four branches after subtracting its chance level, so \num{1} is the cohort average, then context lengths are averaged within each task, tasks within each suite, and the two suites weighted equally.
As detailed in \cref{tab:lc-ablation-aggregate}, the candidates are close overall: the mean over the last four raw checkpoints lies between \num{0.97} and \num{1.02} on the aggregate, while the merged checkpoints improve to between \num{1.09} and \num{1.13}.
Among the merged checkpoints, \(s = 1/3\) achieves the highest aggregate score.

At the task level, differences become more visible at longer context lengths, and for \ac{NIAH}, \ac{VT} and \ac{QA} the branches separate most at \recommendedContextLengthShort{}.
Increasing the reweighted share generally improves retrieval and extraction tasks such as \ac{QA}, \ac{NIAH}, and \ac{WE}, while \ac{VT} moves in the opposite direction.
On HELMET, JsonKV is close to saturation across candidates, whereas RagQA, Rerank, and LongQA InfBench QA show clearer separation.
The control remains competitive despite seeing no long-context data, with the clearest benefit from long-context extension appearing on retrieval tasks at the longest context lengths.
Detailed task-level results are reported in the appendix in \cref{tab:lc-ablation-full}.

\begin{figure}[tb]
  \centering
  \includegraphics{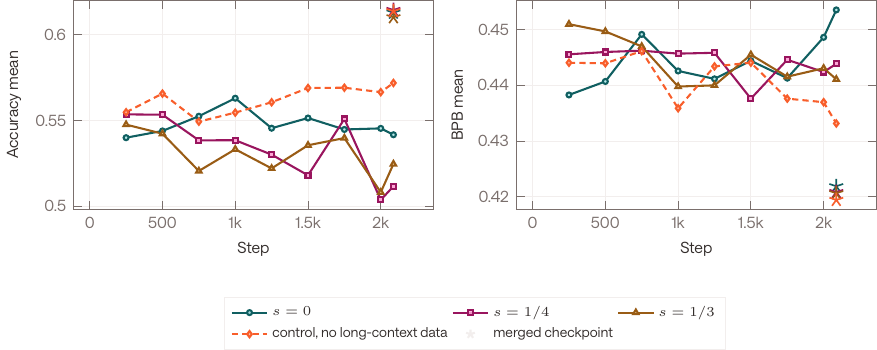}
  \caption{General-capability performance on the standard non-long-context evaluation suite.}
  \label{fig:lc-ablation-general}
\end{figure}

We also track performance on the standard non-long-context evaluation suite, covering mathematics, coding, reasoning, and general knowledge.
Long-context extension slightly reduces general capability in the raw checkpoints: the three long-context branches trail the control throughout the stage in \cref{fig:lc-ablation-general}.
Checkpoint merging largely recovers this difference, with the four merged checkpoints indistinguishable within the standard error of the general-capability composite.

No mixture dominates every task, but \(s = 1/3\), the largest reweighted share tested, gives the strongest merged aggregate while maintaining comparable general capability after merging.
We therefore use this recipe for the production long-context extension.
Since the optimum may lie beyond the tested range, future experiments should evaluate larger reweighted shares while controlling for the accompanying domain shift.

The ablation candidates draw from an earlier \num{1}T-token candidate pool of the mid-training mix; the production long-context mix rebuilds the \(s = 1/3\) recipe on the mid-training mix that \model{2} trains on, scaled down to \num{250}B tokens to fit the proxy scale training horizon, and the last panel of \cref{fig:lc-mix-bands} shows its length distribution.
\begin{table}[tb]
  \centering
  \scriptsize
  \renewcommand{\arraystretch}{1}
  \setlength{\tabcolsep}{5pt}
  \begin{tabular}{lrrrrrr}
    \toprule
    & \multicolumn{3}{c}{Merged checkpoint} & \multicolumn{3}{c}{Last four raw checkpoints} \\
    \cmidrule(lr){2-4}\cmidrule(lr){5-7}
    Mix & RULER & HELMET & Aggregate & RULER & HELMET & Aggregate \\
    \midrule
    s = 0   & 1.122          & 1.063          & 1.093          & 0.951          & 0.994          & 0.973 \\
    s = 1/4 & 1.125          & 1.067          & 1.096          & 0.972          & 0.998          & 0.985 \\
    s = 1/3 & \textbf{1.146} & \textbf{1.115} & \textbf{1.130} & \textbf{1.019} & \textbf{1.018} & \textbf{1.019} \\
    control & 1.124          & 1.059          & 1.091          & 1.017          & 0.979          & 0.998 \\
    \bottomrule
  \end{tabular}
  \caption{Aggregate long-context results used to rank the mix candidates.
    The best value of each column is set in bold.}
  \label{tab:lc-ablation-aggregate}
\end{table}

\subsubsection{Quality Beyond the Training Context}
\label{sec:long-context-extrapolation}

We evaluate \model{2} beyond its trained context length and find that it retains useful long-context capabilities up to \num{1}M tokens, with task-dependent degradation.

\begin{figure}[tb]
  \centering
  \includegraphics{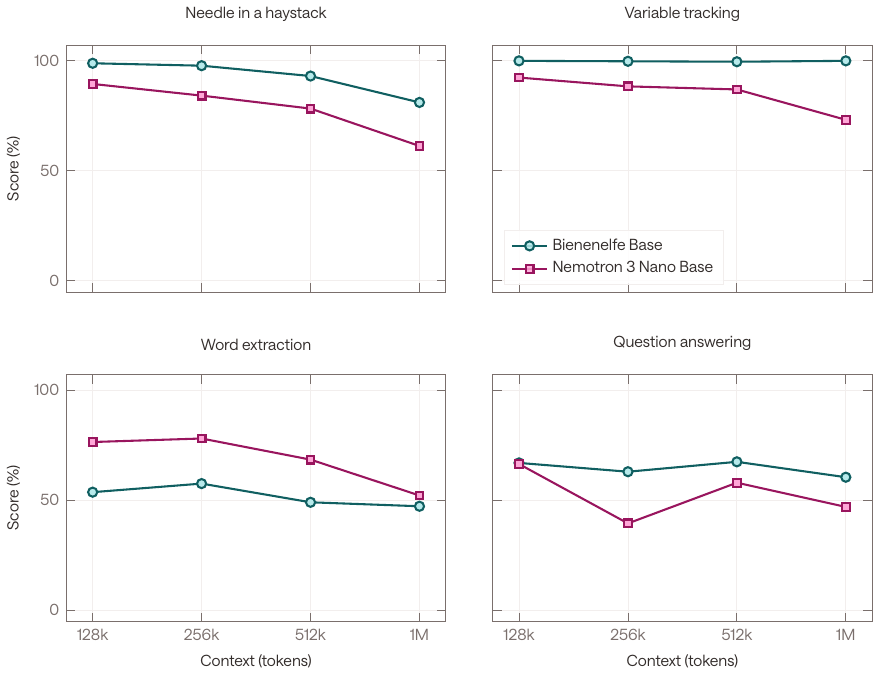}
  \caption{RULER scores of \basemodel{2} and Nemotron 3 Nano Base from \num{128}k to \num{1}\mbox{M} tokens.
    For question answering, the end-of-sequence token is blocked at the first generated position, so no answer is empty. %
  }
  \label{fig:lc-ruler-1m-base}
\end{figure}

\basemodel{2} trains on sequences up to \recommendedContextLengthShort{} tokens.
Its \ac{SWA} layers use \ac{RoPE} within a fixed window, while its full-attention layers use \ac{NoPE}, so evaluation beyond the training length needs no positional rescaling.
We test this extrapolation on RULER \parencite{hsieh2024ruler} from \num{128}k to \num{1}M tokens against Nemotron 3 Nano Base (\cref{fig:lc-ruler-1m-base}).

At \num{1}M tokens, \basemodel{2} scores above Nemotron 3 Nano Base on \ac{NIAH}, \ac{VT} and \ac{QA}, but below it on \ac{WE}.
Its \ac{NIAH} score declines with context length, while \ac{VT} remains stable.
Overall, these results show that \model{2} extrapolates beyond its trained context length without positional rescaling, retaining substantial capability up to \num{1}M tokens, although performance degrades unevenly across tasks.

%% file: sections/02-pre-training/04-base-model-evaluation.tex
\section{Base Model Evaluation}
\label{sec:pre-training-eval-scores}

\subsection{Evaluation Suite}
\label{sec:eval-suite}

In \cref{tab:pre-training-benchmarks}, we present our pre-training evaluation suite that consists of benchmarks covering general knowledge, mathematics, and code in English and German.
We also use these benchmarks to make decisions about the data and model ablations of the previous sections.

\begin{table}[!htb]
  \centering
  \small
  \begin{tabular}{@{}c@{\hspace{0.8em}}>{\raggedright\arraybackslash}p{0.16\linewidth}
      @{\hspace{1.2em}}>{\raggedright\arraybackslash}p{0.40\linewidth} @{\hspace{1.2em}}>{\raggedright\arraybackslash}p{0.30\linewidth}@{}}
    \toprule
    & Benchmark & English & German \\
    \midrule
    \multirow{6}{*}{\rotatebox[origin=c]{90}{\tiny\bfseries Knowledge}}
    & \ac{MMLU}
    & \parencite{hendrycks2021mmlu,hendrycks2021ethics}
    & Global \acs*{MMLU} \parencite{singh2025globalmmlu} \\
    & \acs*{MMLU}-Pro
    & \parencite{wang2024mmlupro}
    & MMLU-ProX \parencite{xuan2025mmluprox} \\
    & \ac{ARC}
    & \parencite{clark2018arc}
    & Ellamind \parencite{ellamind2025benchmarks} \\
    & \ac{PIQA}
    & \parencite{bisk2020piqa}
    & Ellamind \parencite{ellamind2025benchmarks} \\
    & HellaSwag
    & \parencite{zellers2019hellaswag}
    & Ellamind \parencite{ellamind2025benchmarks} \\
    & TriviaQA
    & \parencite{joshi2017triviaqa}
    & -- \\
    \midrule
    \multirow{2}{*}{\rotatebox[origin=c]{90}{\tiny\bfseries Math}}
    & \ac{GSM8K}
    & \parencite{cobbe2021gsm8k}
    & \parencite{vendrow2025platinum,gsm8k_platinum_multilingual_2026} \\
    & MATH Minerva
    & \parencite{hendrycks2021math,lewkowycz2022minerva}
    & Ellamind \parencite{ellamind2025benchmarks} \\
    \midrule
    \multirow{2}{*}{\rotatebox[origin=c]{90}{\tiny\bfseries Code}}
    & HumanEval & \parencite{chen2021humaneval} & Ellamind \parencite{ellamind2025benchmarks} \\
    & \ac{MBPP} & \parencite{austin2021mbpp}    & Ellamind \parencite{ellamind2025benchmarks} \\
    \midrule
    \multirow{2}{*}{\rotatebox[origin=c]{90}{\tiny\shortstack{\bfseries Long\\\bfseries Context}}}
    & RULER  & \parencite{hsieh2024ruler} & -- \\
    & HELMET & \parencite{yen2025helmet}  & -- \\
    \bottomrule
  \end{tabular}

  \caption{Pre-training benchmarks by category, in English and German versions where one exists.
    \enquote{--} marks a benchmark we evaluate in one language only.
    For HELMET we evaluate only the JSON KV, RAG QA, RULER MK3, and InfBench MC subsets.
  }
  \label{tab:pre-training-benchmarks}
\end{table}

\paragraph{Settings}
We implement all the benchmarks with \emph{eval-framework} \parencite{eval_framework} with identical prompts, few-shot configurations, and task settings.
For the multiple-choice benchmarks, the answer choices are represented using the formatting in OLMES \parencite{gu2024olmes}.
For the generative benchmarks, we sample \num{1} response for TriviaQA, \num{4} responses on the math benchmarks, and \num{16} for the code benchmarks, and report the average performance.

\paragraph{Performance Aggregates}
For each category and language, we report the mean score of its benchmarks.
We report the mean over the categories of a language as the \enquote{Overall} score of that language.
We do not aggregate across languages because the benchmark sets differ by language.
These aggregates help us easily track progress during model training without having to track individual benchmarks.

\paragraph{Abstention and Uncertainty}
We additionally evaluate \basemodel{2} on \num{7} English multiple-choice tasks that add an \enquote{I don't know} (IDK) option to each question, which test whether a model moves probability towards abstaining rather than towards a wrong answer.
We score them with the distributional correctness score \parencite{burns2026when}, alongside accuracy and a ternary score that rewards correct answers, penalises wrong ones, and treats abstention as neutral.
\Cref{app:idk-evals} describes the tasks and metrics.
On these IDK evaluations, \basemodel{2} has the second-highest mean distributional correctness score after Nemotron 3 Super 120B-A12B Base.
Its largest margin is on \ac{MMLU}-Pro, where every other compared model except Nemotron 3 Super scores far lower.

\subsection{Baseline Models}
\label{sec:pre-training-baselines}

We compare \basemodel{2} to a range of open base models spanning dense models from \num{7}B to \num{70}B parameters and \ac{MoE} models with \num{3}B to \num{12}B active parameters: Apertus 70B \parencite{apertus2025}, Olmo 3 7B and 32B \parencite{olmo2026olmo3}, Nemotron 3 Nano 30B-A3B \parencite{nvidia2025nemotron3nano}, Nemotron 3 Super 120B-A12B \parencite{nvidia2026nemotron3super}, Qwen3.5 35B-A3B \parencite{qwen2026qwen35}, GLM-4.5 Air 106B-A12B \parencite{glm45team2025glm45}, and Gemma 4 26B-A4B \parencite{gemma2026gemma4technicalreport}, next to \basemodel{1}.
For the generative tasks listed in \cref{tab:pre-training-benchmarks}, we generally use sampling with a temperature of \num{0.6}, top-p of \num{0.6} and a maximum generation length of \num{1024} tokens for all models.
The exception is Gemma 4 26B-A4B Base, for which we use temperature \num{1.0} and top-p \num{0.95}, as specified in its model card.
The two long-context tasks use temperature \num{0.0}, top-p \num{1.0} and a maximum generation length of \num{256} tokens.

Further details of the model checkpoints and the vLLM deployment are presented in \cref{tab:baseline-models} in \cref{app:baselines}.
Our experimental settings might differ from those of the original publications of the baseline models.
Thus, our results are a comparison of eight baselines under one protocol, not a reproduction of published numbers.

\subsection{Results}
\begin{figure}[!htb]
  \centering
  \includegraphics{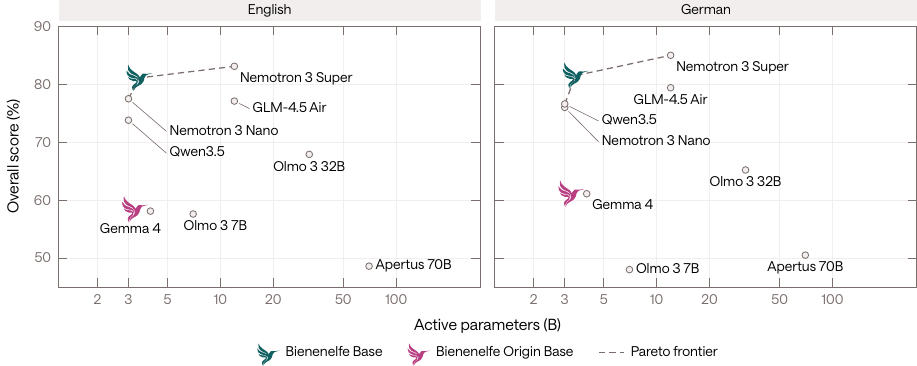}
  \caption{Overall English and German pre-training aggregates of \cref{tab:pre-training-evals} \emph{vs} the active parameters per token on a logarithmic axis, where higher and further left is better.
    The dashed line connects the models on the Pareto frontier.
    A model lies on the frontier iff every other model with fewer active parameters scores lower and every other model with as many active parameters scores at most as high.
  }
  \label{fig:pre-training-pareto}
\end{figure}

In \cref{fig:pre-training-pareto}, we plot the \enquote{Overall} scores against the active parameters per token.
\basemodel{2}
is the best-performing model among all tested models with fewer than \num{12}\mbox{B} active parameters.
It is surpassed in performance only by Nemotron 3 Super, which has more than \num{3}\(\times\) the number of active parameters.
GLM-4.5 Air, which has the same number of active parameters as Nemotron 3 Super, scores below \basemodel{2} in both languages and is therefore not on the Pareto frontier of \cref{fig:pre-training-pareto}.

We report the scores of each model on our benchmark suite in \cref{tab:pre-training-evals}.
\basemodel{2} is consistently among the top models on math.
However, the two models with \num{12}B active parameters outperform \basemodel{2} on the code averages.

\begin{table}[!tbp]
  \centering
  \scriptsize
  \setlength{\tabcolsep}{3pt}
  \setlength{\aboverulesep}{0pt}
  \setlength{\belowrulesep}{0pt}
  \renewcommand{\arraystretch}{1.25}
  \begin{tabular}{l!{\color{AABorder}\vrule width 0.4pt}rr!{\color{AABorder}\vrule width 0.4pt}rrr!{\color{AABorder}\vrule width 1.5pt}rr!{\color{AABorder}\vrule width 1.5pt}rrr}
    \toprule
    Type
    & \multicolumn{5}{c!{\color{AABorder}\vrule width 1.5pt}}{\acs*{MoE}}
    & \multicolumn{2}{c!{\color{AABorder}\vrule width 1.5pt}}{\acs*{MoE}}
    & \multicolumn{3}{c}{\archlabel{Dense}} \\
    Active parameters
    & \multicolumn{5}{c!{\color{AABorder}\vrule width 1.5pt}}{3--4B}
    & \multicolumn{2}{c!{\color{AABorder}\vrule width 1.5pt}}{12B}
    & \multicolumn{1}{c}{\num{7}\mbox{B}}
    & \multicolumn{1}{c}{\num{32}\mbox{B}}
    & \multicolumn{1}{c}{\num{70}\mbox{B}} \\
    \cmidrule(lr){2-6} \cmidrule(lr){7-8} \cmidrule(lr){9-11}
    & \multicolumn{2}{c!{\color{AABorder}\vrule width 0.4pt}}{Ours}
    & \multicolumn{3}{c!{\color{AABorder}\vrule width 1.5pt}}{Baseline models}
    & \multicolumn{2}{c!{\color{AABorder}\vrule width 1.5pt}}{}
    & \multicolumn{3}{c}{} \\
    \cmidrule(lr){2-3} \cmidrule(lr){4-6}
    Eval
    & \rotatebox{90}{\basemodel{2}\hspace{3pt}}
    & \rotatebox{90}{\basemodel{1}\hspace{3pt}}
    & \rotatebox{90}{Gemma 4 26B-A4B Base\hspace{3pt}}
    & \rotatebox{90}{Nemotron 3 Nano 30B-A3B Base\hspace{3pt}}
    & \rotatebox{90}{Qwen3.5 35B-A3B Base\hspace{3pt}}
    & \rotatebox{90}{GLM-4.5 Air 106B-A12B Base\hspace{3pt}}
    & \rotatebox{90}{Nemotron 3 Super 120B-A12B Base\hspace{3pt}}
    & \rotatebox{90}{Olmo 3 7B Base\hspace{3pt}}
    & \rotatebox{90}{Olmo 3 32B Base\hspace{3pt}}
    & \rotatebox{90}{Apertus 70B Base\hspace{3pt}} \\
    \midrule
    Total parameters (B)
    & \parambar{1.302}{\num{78.1}}
    & \parambar{0.510}{\num{30.6}}
    & \parambar{0.433}{\num{26}}
    & \parambar{0.500}{\num{30}}
    & \parambar{0.583}{\num{35}}
    & \parambar{1.767}{\num{106}}
    & \parambar{2.000}{\num{120}}
    & \parambar{0.117}{\num{7}}
    & \parambar{0.533}{\num{32}}
    & \parambar{1.167}{\num{70}} \\
    \arrayrulecolor{AABorder}\cmidrule[0.4pt]{1-11}\arrayrulecolor{black}
    Active parameters (B)
    & \parambar{0.058}{\num{3.46}}
    & \parambar{0.054}{\num{3.27}}
    & \parambar{0.067}{\num{4}}
    & \parambar{0.050}{\num{3}}
    & \parambar{0.050}{\num{3}}
    & \parambar{0.200}{\num{12}}
    & \parambar{0.200}{\num{12}}
    & \parambar{0.117}{\num{7}}
    & \parambar{0.533}{\num{32}}
    & \parambar{1.167}{\num{70}} \\
    \midrule
    \addlinespace[2pt]
    Overall (EN)
    & \groupbest{\num{81.1}}
    & \num{58.4}
    & \num{58.1}
    & \num{77.5}
    & \num{73.8}
    & \num{77.1}
    & \champion{\num{83.1}}
    & \num{57.6}
    & \groupbest{\num{67.9}}
    & \num{48.6} \\
    Overall (DE)
    & \groupbest{\num{81.5}}
    & \num{61.1}
    & \num{61.1}
    & \num{76.0}
    & \num{76.6}
    & \num{79.4}
    & \champion{\num{85.0}}
    & \num{48.0}
    & \groupbest{\num{65.2}}
    & \num{50.5} \\
    \midrule
    \bfseries General Knowledge
    &
    &
    &
    &
    &
    &
    &
    &
    &
    &  \\
    \avg{Average (EN)}
    & \avg{\num{81.3}}
    & \avg{\num{70.8}}
    & \avg{\num{77.2}}
    & \avg{\num{80.5}}
    & \avg{\groupbest{\num{82.7}}}
    & \avg{\num{82.5}}
    & \avg{\champion{\num{87.2}}}
    & \avg{\num{67.8}}
    & \avg{\groupbest{\num{77.8}}}
    & \avg{\num{73.8}} \\
    \avg{Average (DE)}
    & \avg{\num{82.7}}
    & \avg{\num{70.7}}
    & \avg{\num{80.5}}
    & \avg{\num{80.0}}
    & \avg{\groupbest{\num{84.9}}}
    & \avg{\num{81.7}}
    & \avg{\champion{\num{87.5}}}
    & \avg{\num{51.8}}
    & \avg{\num{68.7}}
    & \avg{\groupbest{\num{74.7}}} \\
    MMLU (EN)
    & \num{81.0}
    & \num{68.9}
    & \num{78.2}
    & \num{79.0}
    & \groupbest{\num{84.7}}
    & \num{82.8}
    & \champion{\num{86.8}}
    & \num{67.1}
    & \groupbest{\num{76.3}}
    & \num{69.3} \\
    Global MMLU (DE)
    & \num{77.5}
    & \num{65.1}
    & \num{75.1}
    & \num{74.3}
    & \groupbest{\num{81.5}}
    & \num{78.2}
    & \champion{\num{84.7}}
    & \num{51.6}
    & \groupbest{\num{65.1}}
    & \num{64.9} \\
    ARC (EN)
    & \num{95.7}
    & \num{88.9}
    & \num{95.2}
    & \num{94.3}
    & \groupbest{\num{97.0}}
    & \num{96.3}
    & \champion{\num{97.5}}
    & \num{88.6}
    & \groupbest{\num{94.5}}
    & \num{90.7} \\
    ARC (DE)
    & \num{96.4}
    & \num{88.5}
    & \num{95.5}
    & \num{94.2}
    & \groupbest{\num{97.6}}
    & \num{95.9}
    & \champion{\num{97.8}}
    & \num{67.9}
    & \num{88.1}
    & \groupbest{\num{89.9}} \\
    PIQA (EN)
    & \num{89.5}
    & \num{76.9}
    & \num{88.0}
    & \num{89.4}
    & \groupbest{\num{91.9}}
    & \num{89.8}
    & \champion{\num{95.1}}
    & \num{77.7}
    & \groupbest{\num{86.8}}
    & \num{80.1} \\
    PIQA (DE)
    & \num{96.8}
    & \num{87.9}
    & \num{96.8}
    & \num{96.5}
    & \groupbest{\num{98.5}}
    & \num{97.2}
    & \champion{\num{99.3}}
    & \num{75.0}
    & \num{90.0}
    & \groupbest{\num{94.6}} \\
    HellaSwag (EN)
    & \num{83.1}
    & \num{80.7}
    & \num{84.8}
    & \groupbest{\num{85.4}}
    & \num{85.3}
    & \num{87.2}
    & \champion{\num{88.8}}
    & \num{76.1}
    & \num{83.5}
    & \groupbest{\num{84.4}} \\
    HellaSwag (DE)
    & \num{86.9}
    & \num{75.3}
    & \num{88.6}
    & \num{85.8}
    & \groupbest{\num{88.7}}
    & \num{86.5}
    & \champion{\num{93.0}}
    & \num{36.8}
    & \num{61.0}
    & \groupbest{\num{88.7}} \\
    MMLU-Pro (EN)
    & \num{61.1}
    & \num{39.3}
    & \num{51.2}
    & \num{55.6}
    & \groupbest{\num{62.6}}
    & \num{55.3}
    & \champion{\num{68.9}}
    & \num{38.0}
    & \groupbest{\num{50.9}}
    & \num{40.6} \\
    MMLU-ProX (DE)
    & \num{55.7}
    & \num{36.5}
    & \num{46.7}
    & \num{49.0}
    & \groupbest{\num{58.4}}
    & \num{50.8}
    & \champion{\num{62.8}}
    & \num{27.9}
    & \groupbest{\num{39.5}}
    & \num{35.6} \\
    TriviaQA (EN)
    & \num{77.2}
    & \num{69.9}
    & \num{65.6}
    & \groupbest{\num{79.5}}
    & \num{74.6}
    & \num{83.8}
    & \champion{\num{86.4}}
    & \num{59.0}
    & \num{74.7}
    & \groupbest{\num{77.4}} \\
    \midrule
    \bfseries Math
    &
    &
    &
    &
    &
    &
    &
    &
    &
    &  \\
    \avg{Average (EN)}
    & \avg{\champion{\num{84.9}}}
    & \avg{\num{54.3}}
    & \avg{\num{45.9}}
    & \avg{\num{83.0}}
    & \avg{\num{73.5}}
    & \avg{\num{64.7}}
    & \avg{\groupbest{\num{82.7}}}
    & \avg{\num{57.7}}
    & \avg{\groupbest{\num{63.3}}}
    & \avg{\num{39.6}} \\
    \avg{Average (DE)}
    & \avg{\num{76.8}}
    & \avg{\num{55.8}}
    & \avg{\num{47.0}}
    & \avg{\num{72.5}}
    & \avg{\groupbest{\num{77.5}}}
    & \avg{\num{69.8}}
    & \avg{\champion{\num{81.2}}}
    & \avg{\num{43.8}}
    & \avg{\groupbest{\num{62.9}}}
    & \avg{\num{39.0}} \\
    GSM8K (EN)
    & \champion{\num{89.8}}
    & \num{71.6}
    & \num{65.7}
    & \num{86.3}
    & \num{89.5}
    & \num{82.8}
    & \groupbest{\num{87.5}}
    & \num{75.3}
    & \groupbest{\num{81.1}}
    & \num{62.3} \\
    GSM8K Platinum (DE)
    & \groupbest{\num{90.5}}
    & \num{71.7}
    & \num{64.1}
    & \num{86.3}
    & \num{88.0}
    & \num{86.5}
    & \champion{\num{93.1}}
    & \num{58.7}
    & \groupbest{\num{80.8}}
    & \num{60.2} \\
    MATH Minerva (EN)
    & \champion{\num{80.1}}
    & \num{37.0}
    & \num{26.1}
    & \num{79.8}
    & \num{57.6}
    & \num{46.5}
    & \groupbest{\num{77.8}}
    & \num{40.1}
    & \groupbest{\num{45.6}}
    & \num{16.9} \\
    MATH Minerva (DE)
    & \num{63.2}
    & \num{39.9}
    & \num{29.9}
    & \num{58.8}
    & \groupbest{\num{67.0}}
    & \num{53.2}
    & \champion{\num{69.3}}
    & \num{28.9}
    & \groupbest{\num{45.0}}
    & \num{17.7} \\
    \midrule
    \bfseries Code
    &
    &
    &
    &
    &
    &
    &
    &
    &
    &  \\
    \avg{Average (EN)}
    & \avg{\groupbest{\num{77.0}}}
    & \avg{\num{50.1}}
    & \avg{\num{51.1}}
    & \avg{\num{69.0}}
    & \avg{\num{65.2}}
    & \avg{\champion{\num{84.1}}}
    & \avg{\num{79.3}}
    & \avg{\num{47.4}}
    & \avg{\groupbest{\num{62.5}}}
    & \avg{\num{32.5}} \\
    \avg{Average (DE)}
    & \avg{\groupbest{\num{85.1}}}
    & \avg{\num{57.0}}
    & \avg{\num{55.7}}
    & \avg{\num{75.5}}
    & \avg{\num{67.4}}
    & \avg{\champion{\num{86.7}}}
    & \avg{\num{86.2}}
    & \avg{\num{48.3}}
    & \avg{\groupbest{\num{64.0}}}
    & \avg{\num{37.9}} \\
    HumanEval (EN)
    & \groupbest{\num{86.7}}
    & \num{50.8}
    & \num{52.2}
    & \num{73.5}
    & \num{67.2}
    & \champion{\num{96.3}}
    & \num{83.2}
    & \num{46.8}
    & \groupbest{\num{63.9}}
    & \num{28.3} \\
    HumanEval (DE)
    & \groupbest{\num{88.4}}
    & \num{49.0}
    & \num{51.7}
    & \num{73.5}
    & \num{59.2}
    & \champion{\num{94.0}}
    & \num{85.0}
    & \num{39.5}
    & \groupbest{\num{56.2}}
    & \num{28.3} \\
    MBPP (EN)
    & \groupbest{\num{67.3}}
    & \num{49.5}
    & \num{50.0}
    & \num{64.5}
    & \num{63.1}
    & \num{71.9}
    & \champion{\num{75.4}}
    & \num{48.0}
    & \groupbest{\num{61.0}}
    & \num{36.8} \\
    MBPP (DE)
    & \groupbest{\num{81.7}}
    & \num{64.9}
    & \num{59.7}
    & \num{77.4}
    & \num{75.6}
    & \num{79.4}
    & \champion{\num{87.5}}
    & \num{57.1}
    & \groupbest{\num{71.7}}
    & \num{47.5} \\
    \bottomrule
  \end{tabular}
  \caption{Pre-training benchmark scores in percent of \basemodel{2}, \basemodel{1}, and open base models.
    The models are grouped by architecture and the active parameters per token that each group heading names, with our models next to the baseline models of the lightest \acs*{MoE} group, and each group's baselines are ordered by total parameters.
    The first two rows give each model's total and active parameters in billions as its publisher states them, with bars on a common scale.
    A dense model's active parameters equal its total.
    In each row, the best score of each group is underlined, the best score among the ten models is additionally set bold, and tied scores share a mark.
    The Average rows, set in italics, give the mean of their category's benchmarks.
    \basemodel{2}
    ranks second on both Overall aggregates, behind Nemotron 3 Super 120B-A12B Base with more than \num{3} times as many active parameters, and first on the English math average.
  }
  \label{tab:pre-training-evals}
\end{table}

\subsubsection{Long-Context Performance}

We also evaluate our model, \basemodel{2}, and the baselines on RULER \parencite{hsieh2024ruler} from \num{4}k to \num{1}M tokens of context and on HELMET \parencite{yen2025helmet} from \num{8}k to \num{128}k, and report the performance in \cref{tab:pre-training-ruler}.
Each RULER row is the mean of its \ac{NIAH}, \ac{VT}, \ac{WE} and \ac{QA} tasks at that context length, sampled at temperature \num{0}.
We evaluate on the JSON KV, RAG QA, RULER MK3, and InfBench MC subsets of HELMET and report the mean of their scores: substring exact match for JSON KV and RAG QA, recall for RULER MK3, and exact match for InfBench MC. When the input length exceeds a model's context window, we report \enquote{--}.
The models with a \num{64}k-token window also report none at RULER's \num{64}k, because the prompts exceed \num{64}k tokens under their tokenisers (\basemodel{1}, Olmo 3 7B Base, Olmo 3 32B Base and Apertus 70B Base).
Gemma 4 26B-A4B Base reports none at \recommendedContextLengthShort{}, because the prompt plus the maximum number of generated tokens exceeds its \num{256}k-token window.
At \num{512}k and \num{1}M, \basemodel{2} and Qwen3.5 35B-A3B Base are served beyond their \num{256}k-token trained window, Qwen3.5 with static \ac{YaRN} at factor \num{4} at every length, while Nemotron 3 Nano 30B-A3B Base supports up to \num{1}M tokens natively.
On RULER, \basemodel{2} trails Qwen3.5 35B-A3B Base and Nemotron 3 Nano 30B-A3B Base at every length up to \recommendedContextLengthShort{}, and Gemma 4 26B-A4B Base from \num{8}k to \num{128}k; among the models evaluated up to \num{1}M tokens, it degrades the least and has the best score there.
On HELMET it is second behind Gemma 4 26B-A4B Base at \num{64}k and \num{128}k.

\begin{table}[tb]
  \centering
  \scriptsize
  \setlength{\tabcolsep}{4pt}
  \setlength{\aboverulesep}{0pt}
  \setlength{\belowrulesep}{0pt}
  \renewcommand{\arraystretch}{1.25}
  \begin{tabular}{l!{\color{AABorder}\vrule width 0.4pt}rr!{\color{AABorder}\vrule width 0.4pt}rrr!{\color{AABorder}\vrule width 1.5pt}rrr}
    \toprule
    Type
    & \multicolumn{5}{c!{\color{AABorder}\vrule width 1.5pt}}{\acs*{MoE}}
    & \multicolumn{3}{c}{\archlabel{Dense}} \\
    Active parameters
    & \multicolumn{5}{c!{\color{AABorder}\vrule width 1.5pt}}{3--4B}
    & \multicolumn{1}{c}{\num{7}\mbox{B}}
    & \multicolumn{1}{c}{\num{32}\mbox{B}}
    & \multicolumn{1}{c}{\num{70}\mbox{B}} \\
    \cmidrule(lr){2-6} \cmidrule(lr){7-9}
    & \multicolumn{2}{c!{\color{AABorder}\vrule width 0.4pt}}{Ours}
    & \multicolumn{3}{c!{\color{AABorder}\vrule width 1.5pt}}{Baseline models}
    & \multicolumn{3}{c}{} \\
    \cmidrule(lr){2-3} \cmidrule(lr){4-6}
    Context
    & \rotatebox{90}{\basemodel{2}\hspace{3pt}}
    & \rotatebox{90}{\basemodel{1}\hspace{3pt}}
    & \rotatebox{90}{Gemma 4 26B-A4B Base\hspace{3pt}}
    & \rotatebox{90}{Nemotron 3 Nano 30B-A3B Base\hspace{3pt}}
    & \rotatebox{90}{Qwen3.5 35B-A3B Base\hspace{3pt}}
    & \rotatebox{90}{Olmo 3 7B Base\hspace{3pt}}
    & \rotatebox{90}{Olmo 3 32B Base\hspace{3pt}}
    & \rotatebox{90}{Apertus 70B Base\hspace{3pt}} \\
    \midrule
    \addlinespace[2pt]
    \bfseries RULER
    &
    &
    &
    &
    &
    &
    &
    &  \\
    4k
    & \num{86.9}
    & \num{91.8}
    & \num{84.7}
    & \num{94.6}
    & \champion{\num{96.3}}
    & \num{92.0}
    & \groupbest{\num{94.8}}
    & \num{89.5} \\
    8k
    & \num{83.7}
    & \num{88.8}
    & \num{85.0}
    & \num{93.4}
    & \champion{\num{95.3}}
    & \num{80.8}
    & \groupbest{\num{92.6}}
    & \num{77.7} \\
    16k
    & \num{80.9}
    & \num{83.7}
    & \num{87.1}
    & \num{92.0}
    & \champion{\num{95.2}}
    & \num{70.8}
    & \groupbest{\num{88.8}}
    & \num{71.3} \\
    32k
    & \num{76.3}
    & \num{72.7}
    & \num{88.7}
    & \num{86.4}
    & \champion{\num{93.7}}
    & \num{64.4}
    & \groupbest{\num{80.9}}
    & \num{70.8} \\
    64k
    & \num{72.2}
    & –
    & \num{86.8}
    & \num{82.6}
    & \champion{\num{91.0}}
    & –
    & –
    & – \\
    128k
    & \num{67.9}
    & –
    & \num{87.8}
    & \num{81.0}
    & \champion{\num{89.9}}
    & –
    & –
    & – \\
    256k
    & \num{69.8}
    & –
    & –
    & \num{72.1}
    & \champion{\num{80.1}}
    & –
    & –
    & – \\
    512k
    & \num{65.5}
    & –
    & –
    & \champion{\num{73.1}}
    & \num{72.2}
    & –
    & –
    & – \\
    1M
    & \champion{\num{63.2}}
    & –
    & –
    & \num{58.5}
    & \num{57.5}
    & –
    & –
    & – \\
    \midrule
    \bfseries HELMET
    &
    &
    &
    &
    &
    &
    &
    &  \\
    8k
    & \num{79.1}
    & \num{74.8}
    & \num{81.9}
    & \num{79.1}
    & \champion{\num{87.7}}
    & \num{71.1}
    & \groupbest{\num{74.8}}
    & \num{64.2} \\
    16k
    & \num{77.5}
    & \num{76.4}
    & \num{81.9}
    & \num{80.8}
    & \champion{\num{85.5}}
    & \num{67.1}
    & \groupbest{\num{73.8}}
    & \num{57.2} \\
    32k
    & \num{80.0}
    & \num{73.5}
    & \num{82.9}
    & \num{78.4}
    & \champion{\num{84.1}}
    & \num{60.1}
    & \groupbest{\num{74.6}}
    & \num{48.1} \\
    64k
    & \num{81.2}
    & –
    & \champion{\num{83.8}}
    & \num{72.7}
    & \num{78.8}
    & \num{51.9}
    & \groupbest{\num{73.1}}
    & – \\
    128k
    & \num{82.4}
    & –
    & \champion{\num{85.4}}
    & \num{70.8}
    & \num{77.3}
    & –
    & –
    & – \\
    \bottomrule
  \end{tabular}
  \caption{RULER and HELMET scores in percent of \basemodel{2} and open base models by context length.
    The models are grouped by architecture and active parameters per token.
    In each row, the best score of each group is underlined, the best score among the eight models is additionally set bold, and tied scores share a mark.
    An en dash marks a length beyond a model's context window.
    \num{12}\mbox{B} models were not evaluated here.
  }
  \label{tab:pre-training-ruler}
\end{table}

%% file: sections/03-post-training/00-post-training.tex
\clearpage
\clearpage
\part{Post-Training}%
\label{part:post-training}

Post-training adds instruction following, reasoning and agentic tool use capabilities to the base model.
\ac{SFT} and \ac{RL} train \model{2} towards the same capabilities.
The paragraphs below outline the design choices that guide both stages in five areas: German, long context, agentic use, retrieval and hallucinations, and reasoning effort.
\Cref{fig:intro-spider} compares the released \model{2} checkpoint with three open models, and \cref{tab:rl-gain} with the \ac{SFT} checkpoint that \ac{RL} starts from.
\Ac{RL} improves on the \ac{SFT} checkpoint across all capability areas shown, and \model{2} performs on par with comparable open models.
The two stages are then described in detail in \cref{sec:sft,sec:rl}.

\paragraph{German}%
\label{sec:post-german}
\basemodel{2} is trained on German data in pre- and mid-training (\cref{tab:german-pre-training-datasets,sec:mid-pool}).
Post-training carries this capability over to instruction following, reasoning and tool use, so that German-speaking users get fluent and correct German answers and reasoning traces they can follow.

For \ac{SFT}, we therefore remove English reasoning from exchanges in other languages (\cref{tab:sft-post-processing}) and add synthetic German datasets to the \ac{SFT} data pool.
For these, teacher models generate German chat, math, science and tool-calling data.
A short German opener at the start of every teacher trace keeps the reasoning in German (\cref{sec:sft-german}).
We measure how the German share of the mix affects the German and English scores on select benchmarks (\cref{app:sft-german-share}).

We continue this training in \ac{RL}, where we translate existing environments into German and include a language-consistency reward for both the reasoning trace and the answer (\cref{tab:env-format-signals}).
We observed language-consistency scores go up for the German benchmarks (\cref{fig:rl-german-language}).

\paragraph{Long Context}%
\label{sec:post-long-context}
Many applications benefit from long-context capabilities, such as analysing lengthy legal documents, scientific papers, and multi-turn dialogues where context from earlier interactions can be highly relevant.
We train \model{2} on in-house curated long-context data to reason over inputs of up to \recommendedContextLengthShort{} tokens.
For \ac{SFT}, we generate conversations over documents of \num{128}k to \num{224}k tokens, each with four requests whose answers draw on sections spread over the document (\cref{sec:sft-long-context}).
In \ac{RL}, a long-context environment adds select-all-that-apply questions over the seed documents of the \ac{SFT} conversations (\cref{tab:rl-environment-list}).

\paragraph{Agentic Use}%
\label{sec:post-agentic}
We train \model{2} to perform well in agentic settings, which require tool calls and terminal interactions.
To help the model recover from failed tool calls instead of repeating them, we backfill reasoning between consecutive tool calls where open agentic datasets have none and mask the loss on wrong tool calls and terminal actions (\cref{sec:sft-agentic}).
Beyond the available data, we generate two software-engineering \ac{SFT} datasets: in one, a teacher model resolves GitHub issues with a patch, and in the other, it fixes bugs as a Terminus~2 agent in a terminal.
In \ac{RL}, we then train on tool-calling, software-engineering and terminal environments (\cref{sec:rl-environments,sec:rl-sandboxes}).
The system prompt, the coding harness and the user request are randomised per group, so that the policy does not overfit to a single harness or phrasing (\cref{sec:rl-randomization}).

\paragraph{Retrieval and Hallucinations}%
Many application areas of \model{2} require it to answer questions based on the documents or knowledge stores of an organisation.
Here, the model searches a corpus over several steps, answers from the retrieved text, and abstains when these sources do not contain the answer.
Both retrieval and hallucination reduction are trained mainly in \ac{RL} and cold-started in \ac{SFT} on high-reward rollouts of the same environments (\cref{sec:retrieval-sft}).

For the information retrieval use cases of our customers, we build knowledge bases and multi-hop question-answering tasks covering both German and English corpora across general knowledge and specialised domains (legal, electronics, hardware and aviation).
During training, we randomise application-side choices, such as the retrieval parameters, models and methods, and the task language, so that the model does not depend on one setup (\cref{sec:retrieval_environments,sec:appendix-rag}).

To reduce hallucinations, especially when answering questions based on retrieved context, we employ the Merlin--Arthur protocol~\parencite{waeldchen2024merlin,deiseroth2025bounding}, which we formulate as an environment.
The protocol is a game between the model (Arthur) and two players that hide sentences of the retrieved context: Merlin hides the sentences whose removal raises the probability of the gold answer most, and Morgana those whose removal lowers it most.
Each question is posed three times, with the full context, with Merlin's context and with Morgana's, without marking which one it is (\cref{sec:merlin-arthur-environments}).
The model is rewarded for the correct answer in the first two cases and for abstaining in the third, and Merlin and Morgana recompute their masks from the current policy at every step.

\paragraph{Reasoning Effort}%
\label{sec:reasoning-effort}

\begin{table}[h]
  \centering
  \footnotesize
  \begin{tabular}{l >{\raggedright\arraybackslash}p{0.25\linewidth} >{\raggedright\arraybackslash}p{0.53\linewidth}}
    \toprule
    Effort & API request parameters & Sentence in the system prompt \\
    \midrule
    \texttt{none}   & \texttt{none}, \path{enable_thinking=false} & \enquote{Reasoning is disabled.
      Proceed straight to answering according to the user’s instructions.}                                                 \\
    \texttt{low}
    & \texttt{minimal}, \texttt{low}
    & \enquote{Reasoning effort is set to low. Think briefly through only the essential steps in the user’s language, then proceed directly to the answer.} \\
    \texttt{medium}
    & \texttt{medium}
    & \enquote{Reasoning effort is set to medium. Think through the task methodically in the user’s language, check key assumptions, and provide a well-supported answer.} \\
    \texttt{high}
    & \texttt{high}, \texttt{xhigh}, \texttt{max}, default
    & \enquote{Reasoning effort is set to high. Think carefully through the task in the user’s language, validate key assumptions, consider plausible alternatives, and prioritize correctness and clarity.} \\
    \bottomrule
  \end{tabular}
  \caption{Sentence the chat template writes into the system prompt for each reasoning effort, and the request values that select it.
  }%
  \label{tab:reasoning-effort-template}
\end{table}

In addition to the \texttt{none} effort, \model{2} features three distinct reasoning-effort levels, \texttt{low}, \texttt{medium} and \texttt{high}.
The \ac{SFT} stage instils these effort levels by training on reasoning traces labelled with an effort level derived from their length relative to requests of the same dataset and difficulty (\cref{sec:sft-reasoning-effort}).
The \ac{RL} stage then trains on all three effort levels, focusing on \texttt{high} effort, and further separates them via the length-dependent \path{completion_budget} format signal of \cref{sec:env-format-signals}.

The chat template signals a chosen effort level to the model by adding a plain-text section to the system prompt.
After the user's system instructions and before any tool definitions, a section headed \texttt{\#~Reasoning effort} holds one instruction per effort (\cref{tab:reasoning-effort-template}).
The user cannot change the wording of this instruction through the API.
If a request sets both \path{reasoning_effort} and \path{enable_thinking}, \path{reasoning_effort} takes precedence.

%% file: sections/03-post-training/01-supervised-fine-tuning.tex
\section{\actitle{SFT} (\acs*{SFT})}%
\label{sec:sft}

\begin{figure}[htb]
  \centering
  \includegraphics{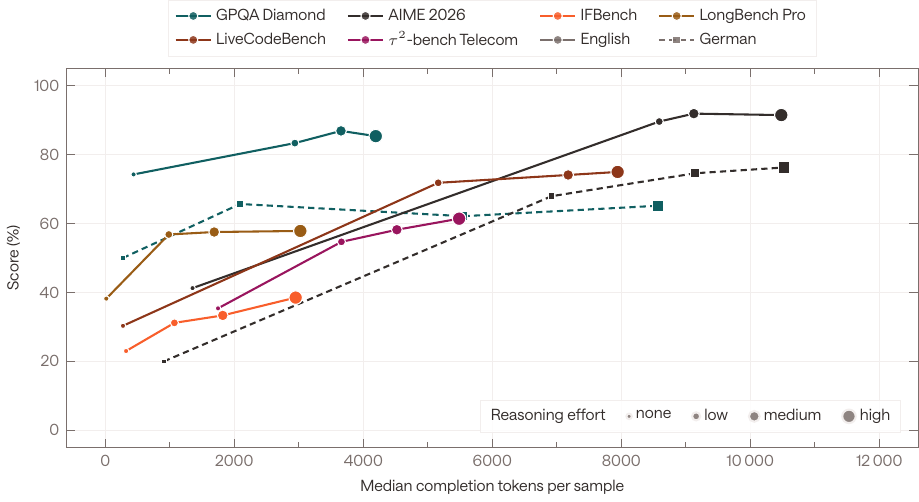}
  \caption{Score of the \model{2} \ac*{SFT} checkpoint on a selection of benchmarks against the median number of tokens it generates per sample, with one point per reasoning effort from \texttt{none} to \texttt{high}.
    The marker size grows with the reasoning effort.
    The German counterparts of the English benchmarks are depicted by a dashed line and square markers.
  }
  \label{fig:sft-reasoning-effort-tasks}
\end{figure}

The \ac{SFT} stage introduces new capabilities, such as instruction following, reasoning at a requested effort level, and tool calling.
This stage serves as a \emph{cold start} for \ac{RL}: it establishes desired behaviours that the model must produce at a sufficient rate for \ac{RL} to become effective in improving them~\parencite{guo2025deepseekr1}.
\ac{SFT} of \model{2} proceeds after the long-context training stage (\cref{sec:long-context}) with cross-entropy loss on the assistant turns (\cref{subsec:sft-training-recipe}).
The released \model{2} \ac{SFT} checkpoint, the weight average of two runs trained on different mixes (\cref{sec:sft-model-souping}), is the input checkpoint to the \ac{RL} training phase.
Its performance on German and English benchmarks across different reasoning efforts is depicted in \cref{fig:sft-reasoning-effort-tasks}.

\paragraph{Terminology}
\ac{SFT} establishes these capabilities by training the model on data that demonstrates them.
A \emph{data pool} is the set of all curated datasets (\cref{sec:sft-pool}).
A generated dataset starts from \emph{seeds}, prompts or documents, from which a \emph{teacher model} writes the completion.
We use a \emph{user simulator} to write the follow-up user turns (\cref{sec:sft-regenerated,sec:sft-german}).
During generation we, at times, utilise \emph{backfilling} to add a generated reasoning trace to an assistant turn that has none (\cref{sec:sft-agentic}), and \emph{reasoning prefills}, short openers that start the teacher model's reasoning trace.
A training run trains on a \emph{mix}, a set of datasets and their sampling weights.
To define training mixes, we use \emph{MergeMix} (\cref{sec:sft-merge-mixing}) during which datasets are grouped into \emph{clusters} and a \emph{specialist} is fine-tuned on each cluster (\cref{sec:sft-merge-clusters}).
A \emph{weighting} assigns a coefficient to each cluster specialist and determines how they are averaged.
The set of all valid weightings forms the \emph{simplex} from which we begin sampling at a \emph{reference mix}, a hand-set mixture of datasets.
We train one run per candidate mix and combine the resulting checkpoints by \emph{souping}, which averages the weights of runs trained on different mixes into one model (\cref{sec:sft-model-souping}).

\subsection{Datasets}%
\label{subsec:sft-datasets-and-curation}
We legally assess, compile, and process an extensive pool of fine-tuning data that combines openly available datasets, \emph{open data} in the following, with synthetic data we generate.
For some open datasets, we regenerate the completions, because their answers fail our audits (\cref{sec:sft-regenerated}).
Additionally, we generate datasets in-house for domains that are not sufficiently covered by open data, e.g.\ German (\cref{sec:sft-german}), long context (\cref{sec:sft-long-context}), agentic tasks (\cref{sec:sft-agentic}), retrieval (\cref{sec:retrieval-sft}), and safety, compliance, and alignment (\cref{sec:sft-safety}).

The main models we use to generate this data, and to regenerate parts of the open datasets, are GLM-5.2~\parencite{zai2026glm52}, GLM-5.3~\parencite{zai2026glm53} and Qwen3.8-27B~\parencite{qwen2026qwen38}.
We employ different models for different regeneration and judging tasks as they have different strengths.
For example, smaller models are generally faster for long-context tasks, but larger models are more suitable for generating strong reasoning traces.
They also have different values and biases, which we consciously select or avoid (\cref{sec:sft-safety}).

All samples undergo the same processing pipeline, which removes overlap with the benchmarks of our \ac{SFT} evaluation (\cref{sec:sft-post-processing,sec:sft-decon}) and labels each user-assistant exchange with a reasoning effort.
\Cref{sec:sft-pool} describes the resulting dataset pool.
The final \ac{SFT} training utilises a subset of the dataset pool found by applying a procedure called MergeMix~\parencite{wang2026mergemix}, which proposes candidate training mixes by weighting clusters of datasets per domain (\cref{sec:sft-merge-mixing}).

\paragraph{Regenerated Completions}%
\label{sec:sft-regenerated}
Training on wrong or low-quality completions directly teaches the model undesirable behaviour, so we check candidate datasets before they enter the pool.
Sample audits of several candidate datasets found many completions of mixed quality, including wrong answers to science problems, which led to regressed benchmark performance if included in training.
For these reasons, we completely regenerate completions for some subsets.
We use seeds from Nemotron-Instruction-Following-Chat-v1~\parencite{nvidia2025nemotronifchatv1} and Nemotron-SFT-Instruction-Following-Chat-v2~\parencite{nvidia2026nemotronsftifchatv2} and generate new completions using both GLM-5.2~\parencite{zai2026glm52} and GLM-5.3~\parencite{zai2026glm53}.
For the reasoning-off split of Nemotron-SFT-Instruction-Following-Chat-v2, we generate completions without reasoning, so that the pool also holds answers for the \texttt{none} effort level (\cref{sec:sft-reasoning-effort}).
Each conversation is extended for up to \num{50} turns.
We employ a user simulator to write follow-up turns in the style and intent of the original user, and have a teacher model answer them.
Open data holds few long conversations, so we generate them.
The extension trains the model to keep following instructions and earlier context over many turns.

\subsubsection{German}%
\begin{figure}[tb]
  \centering
  \resizebox{\linewidth}{!}{%
    \includegraphics{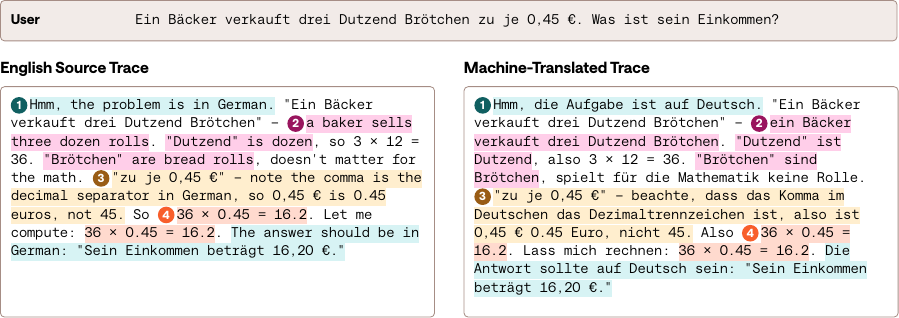}%
  }
  \caption[Artefacts that machine translation carries from an English reasoning trace into a German one]{Artefacts that machine translation carries from an English reasoning trace into a German one.
    Left, the English trace of a reasoning model on a German problem; right, its machine translation into German.
    Highlights mark four kinds of artefact, numbered at their first occurrence in each trace.
    \raisebox{-2.13469pt}{\includegraphics{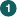}}\,\ignorespaces Meta-language: the trace states that the problem is in German and that the answer has to be German.
    \raisebox{-2.07054pt}{\includegraphics{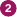}}\,\ignorespaces Repetition: glosses of the German prompt for an English reader become tautologies such as \enquote{Brötchen sind Brötchen}.
    \raisebox{-2.10849pt}{\includegraphics{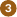}}\,\ignorespaces Decimal-comma aside: the trace explains the German decimal comma, which a German reader does not need.
    \raisebox{-2.13469pt}{\includegraphics{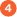}}\,\ignorespaces English notation: the arithmetic keeps 0.45 and 16.2, while the prompt and the answer write 0,45 and 16,20.
  }
  \label{fig:sft-german-translation}
\end{figure}

\label{sec:sft-german}
\suppressfloats[t]
Large open post-training corpora such as Nemotron contain prompts and answers translated into other languages, German among them~\parencite{nvidia2025nemotronnano2}, while keeping the reasoning traces in English.
However, a user who does not read English cannot follow a chain of thought in English, which degrades the user experience of non-English native speakers.
Translating English reasoning traces into German leaks artefacts of the English trace, such as passages it quotes from the German prompt and English notation like a decimal point in place of the decimal comma (\cref{fig:sft-german-translation}).
We instead generate samples in which prompt, reasoning, and answer are German; tool schemas, calls, and results stay unchanged.
In ablations at target scale (\cref{app:sft-german-share}), a model trained without this data reasons in English on German AIME 2026, and a small share of it switches the reasoning to German at the cost of more loops without an answer.
Upsampling the German data \(16\times\) yields more answers that are correct and German and a higher English AIME 2026 score, with a German AIME 2026 score that does not differ from the model without German data.

\paragraph{Data Generation Pipeline}
We generate German data for chat, math, science, and tool use.
The German alignment and safety data are described with their English counterparts (\cref{sec:sft-safety}).
For each domain, we employ a teacher to write a German prompt based on a seed and generate a response.
Seeds may be English samples, documents, or a collection of task attributes.
Where a seed carries a reference answer, the response is verified against it.

\pgfplotstableread[col sep=comma, row sep=\\, format=inline]{%
  prompt_german,prefill_german\\%
  67,97\\%
}\sftPrefillFacts
\begin{figure}[tb]
  \centering
  \resizebox{\linewidth}{!}{%
    \includegraphics{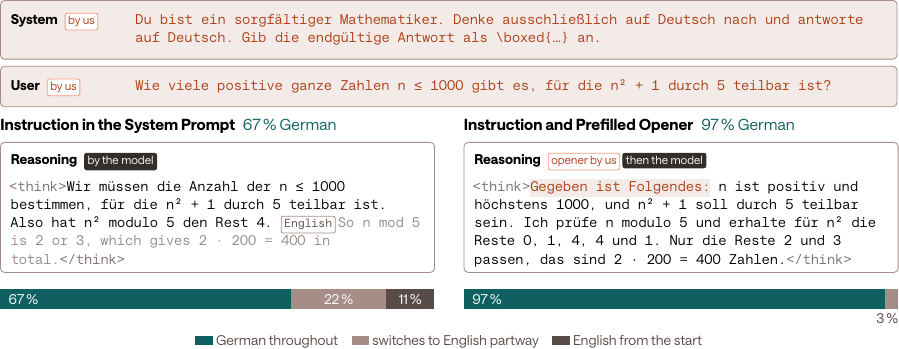}%
  }
  \caption{Steering the language of the teacher's reasoning on German prompts, shown on a math problem.
    Both variants send the same system and user turns.
    Left, the system prompt instructs the model to reason in German, and the model opens the think block itself; its trace shows the case that switches to English partway.
    Right, the assistant turn is sent already opened with \texttt{<think>} and a German opener, which the model continues.
    We checked the language consistency of both approaches on more than \num{5000} completions each.
  }
  \label{fig:sft-german-prefill}
\end{figure}
Without intervention, reasoning models tend to reason in English even on German prompts.
We therefore prefill the teacher model's reasoning trace with a short German opener, such as \enquote{Gegeben ist:} (\enquote{We are given:}) or \enquote{Der Nutzer möchte} (\enquote{The user wants}), and let the teacher continue from it.
We call this technique reasoning-prefill distillation.
Each opener is randomly picked per sample from a curated list per task domain, so that the model does not learn one fixed opening phrase.
An instruction in the system prompt alone keeps \sftPrefillPercent{prompt_german} of the teacher's traces German throughout, and the prefilled opener raises this share to \sftPrefillPercent{prefill_german} (\cref{fig:sft-german-prefill}).
A similar technique has been used in \ac{OPSD} to keep student and teacher from drifting into English during \ac{RL}~\parencite{liu2026copsd}.
We apply it offline to a separate teacher to generate \ac{SFT} data.
Finally, a heuristic gate checks that both reasoning and answer are German, and an \ac{LLM} language healer fixes the errors it finds in the assistant turns.
A sample whose trace remains malformed or switches language keeps its answer without reasoning, as in the shared post-processing (\cref{tab:sft-post-processing}).
Its answer can still be a good one, and as a sample without reasoning it trains the \texttt{none} effort (\cref{sec:sft-reasoning-effort}).

\paragraph{Chat}
General chat consists of multi-turn conversations seeded from user turns of open chat datasets and from generated everyday questions.
These conversations are continued by a user simulator that deepens, constrains, contradicts, or changes the topic, which trains the model to follow instructions that change over a conversation.
GLM-5.3~\parencite{zai2026glm53} writes the replies with reasoning, and Qwen3.8-27B~\parencite{qwen2026qwen38} those without.
User turns stay casual, as in deployment, with lower case, typos and missing punctuation.
Chat replies have no reference answer, so filtering them by quality would require an \ac{LLM} judge.
We skip this filter, as it adds compute without a known benefit.
\textcite{guha2025openthoughts} find that none of the answer-filtering methods they test on reasoning data, \ac{LLM} verification included, improves on training with all answers.

\paragraph{Tool Calling}
A user simulator asks German questions about entities drawn from the category trees of German Wikipedia, and GLM-5.2~\parencite{zai2026glm52} answers them by searching and reading Wikipedia articles through tool calls.
Both the entity sampling and the tool calls query the live German Wikipedia at generation time rather than a fixed dump.
The model thus learns to search and reason in German over German sources.
Several such conversations are concatenated into long contexts that end with a request for an essay tying their topics together, which requires facts spread over the whole context and adds German long-context data.
The same trajectories also enter the data pool as knowledge answers without tool calls.

English tool-calling conversations from public datasets and the retrieval rollouts (\cref{sec:retrieval-sft}) are translated to German, with tool schemas, calls and results left unchanged.
GLM-5.3~\parencite{zai2026glm53} replaces their English reasoning with a new German trace for every assistant turn.
Each reasoning trace is verified against the tool call or answer that follows it.
Every tool-calling sample enters the data pool once with and once without its reasoning traces.

\paragraph{Math and Science}
The seeds for the German math and science data are hard problems from several open math and science datasets.
All of these problems come with a reference answer.
A problem counts as hard if the reasoning trace of its source is at least \num{20000} characters long for math and \num{10000} for science.
We apply GLM-5.3~\parencite{zai2026glm53} to rewrite each problem in natural German without seeing the reference, so that the German problem does not leak the answer.
We then reapply GLM-5.3 to solve the problem, and the answer is kept only if it matches the reference.
Multiple-choice science questions keep their option letters, so the answer is compared by letter.
Math conversations are optionally extended to multiple turns with generated user follow-ups, which trains the model to handle follow-up questions on the same problem.

\FloatBarrier

\subsubsection{Long Context}%
\label{sec:sft-long-context}

The long-context extension of \model{2} (\cref{sec:long-context}) trains the base model on documents of up to \recommendedContextLengthShort{} tokens.
We add long-context samples to \ac{SFT} to preserve the model's long-context retrieval ability and to extend it to reasoning over long inputs, such as large documents, code bases, and tool outputs in agentic applications.
Such tasks go beyond retrieving facts from a long input and require summarisation, multi-hop reasoning and the use of distant information in reasoning traces.

\begin{figure}[tb]
  \centering
  \resizebox{\linewidth}{!}{%
    \includegraphics{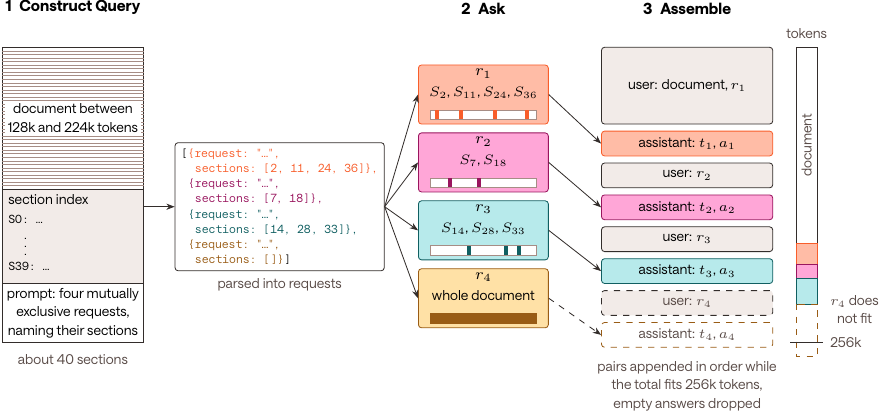}%
  }
  \caption{Generation of one long-context conversation from a seed document.
    A seed document with a length between \num{128}k and \num{224}k tokens is split into about \num{40} sections \(S_0\) to \(S_{39}\).
    The generator reads the document, an index of its sections, and the prompt as one message.
    Four requests \(r_1\) to \(r_4\) are parsed from its response.
    Each request is answered in its own call returning a reasoning trace \(t_i\) and an answer \(a_i\).
    Turns are appended in order to a conversation that holds the document, ensuring conversations are shorter than \recommendedContextLengthShort{} tokens.
  }
  \label{fig:sft-lc-pipeline}
\end{figure}

As open datasets rarely provide long-context samples, we generate our own long-context dataset (\cref{fig:sft-lc-pipeline}).
We use the same pre-processing as in pre-training (\cref{sec:heuristic-filters}) and select from a corpus of curated documents with token lengths between \num{128}k and \num{224}k.
This corpus also includes a collection of German legislative documents.
Papers and German legislative documents are expected long-context inputs in \model{2}'s target applications.

We split a sampled document into about \num{40} sections at coarse boundaries such as page breaks, blank lines, line breaks or, failing those, windows of \num{400} characters.
Qwen3.8-27B \parencite{qwen2026qwen38} receives the document, an index of its sections and a prompt, and writes four different simulated user requests about the document, each tied to the sections it needs, or to none for a request about the whole document (\cref{fig:sft-lc-pipeline}).
A request tied to distant sections must combine distant passages rather than retrieve a single fact.
Requests about the opening of a document alone would not train the model to use the rest of a long input, so the prompt asks for requests spread over the whole document.
\pgfplotstableread[col sep=comma, row sep=\\, format=inline]{%
  at_least_three\\%
  77.4\\%
}\coveredFifthsFacts
We measure this spread by dividing each document into five parts of equal section count.
In \coveredFifthsPercent{at_least_three} of the conversations, the requests together draw on sections from at least three of the five parts.
Because model performance degrades with input length \parencite{hong2025contextrot}, each simulated request is answered in a separate call that receives only this request and the sections it is tied to, or the whole document for a request about the whole document.
The resulting requests, reasoning traces and answers are appended to a conversation beginning with the seed document, so that later turns rely on the document and on the earlier exchanges, as in multi-turn use.
This conversation forms a single training sample: the teacher model answers most requests from a few sections, while the model trained on the sample sees the full document.

The dataset holds around \num{93}k conversations over documents from various open corpora.
In this collection, around \qty{9}{\percent} of the samples are German.
In a proxy-scale ablation, adding this data raises RULER \ac{VT} significantly, and AA-LCR and LongBench Pro without significance, with the largest gains at the longest contexts (\cref{app:sft-lc-ablation}).

\subsubsection{Tool Calling and Agentic Coding}%
\label{sec:sft-agentic}
The agentic data of the mix consist of open agentic datasets and two software-engineering datasets that we generate.
The following paragraphs describe how we format the trajectories of the open datasets and how we generate our own.

\paragraph{Trajectory Formatting and Reasoning Backfilling}

Tool-calling models tend to repeat a failed call instead of recovering from it~\parencite{su2026structuredreflection}, and reasoning between actions lets an agent track its plan and handle exceptions~\parencite{yao2023react}.
Thus, we add reasoning to tool calls where it is missing, in a step we call backfilling.
A teacher model writes reasoning that leads to the recorded tool call, which stays unchanged.
The reasoning is kept only if a second \ac{LLM} call derives the same tool call from it alone.
These reasoning-backfilled conversations enter the data pool along with their non-reasoning counterparts.
This ensures that the requested reasoning effort decides if the model reasons, and not the source dataset.
We empty the \texttt{content} of every assistant turn with \texttt{tool\_calls}.
Every step then holds only the reasoning, the tool calls and their results, which unifies the pattern that leads to a tool call across dataset sources.

Finally, we mask the loss on clearly wrong tool calls: calls whose tool response reports an unknown tool or invalid arguments, and exact repeats of the previous call that never receive a different response.
These turns stay in the context, so the model learns the correction that follows without being trained to produce the error, as in \textcite{fu2025agentrefine} and \textcite{slinko2026srft}.

\paragraph{Agentic Software-Engineering}
Open datasets provide trajectories that resolve issues in real repositories and are verified by their tests.
To increase the number of such trajectories in our pool, we generate two agentic datasets on the repositories of SWE-rebench V2~\parencite{badertdinov2026swerebenchv2}, which also give the software-engineering and terminal environments of \ac{RL} a cold start (\cref{sec:rl-environments}).
SWE-rebench V2 provides the issues, the repository snapshots and the test environments; we generate the trajectories and, for the terminal tasks, the bugs.
For GitHub issues, GLM-5.2~\parencite{zai2026glm52}   explores a snapshot of the repository with read-only tools and answers with a patch, without ever seeing the reference patch, so that the trajectory shows how the fix is found.

For terminal tasks, GLM-5.3~\parencite{zai2026glm53} injects a small, realistic bug into a Python repository and writes a bug report from the user's point of view.
Injected bugs turn any repository with tests into a verifiable task, beyond its existing issues~\parencite{yang2025swesmith}.
A bug is kept only if at least one test fails and at least half of the tests still pass, so that it is detectable but local.
Another GLM-5.3 instance then fixes the bug in the Terminus~2 harness of Terminal-Bench~\parencite{merrill2026terminalbench}, and we keep the trajectories that pass the tests.
Repositories of SWE-bench Verified~\parencite{jimenez2024swebench} are excluded from both datasets, as we evaluate on them.

\subsubsection{Retrieval and Hallucinations}%
\label{sec:retrieval-sft}
The retrieval and the hallucination datasets train the model to answer on the basis of document collections typical of organisations and companies and to abstain when these do not contain the answer.
Both are rollouts of \ac{RL} environments and provide a cold start for \ac{RL}.
The environments and their rewards are described with \ac{RL} (\cref{sec:retrieval_environments,sec:merlin-arthur-environments,sec:appendix-rag}).

\paragraph{Retrieval and Search}
We roll out GLM-5.2~\parencite{zai2026glm52} in the retrieval environments on their English and German training tasks, with the randomisation of the environments applied per episode.
We keep a rollout when an \ac{LLM} judge, Gemma-4-31B~\parencite{gemma2026gemma4technicalreport}, accepts its answer, the gold answer appears in the retrieved text and the transcript ends on an answer.
In addition, every tool call must name a served tool, and the episode must earn the full format reward.
The search must also be efficient: no duplicate calls, at most \qty{30}{\percent} near-duplicate queries, at most three searches per required hop, and at most two searches after the evidence was found.
These limits train the model to stop searching once it has the evidence, which keeps latency and cost low.
We also check that the reasoning of every kept trace is in the language of the user prompt, so that German users can follow it.
The kept rollouts are also translated to German with GLM-5.3~\parencite{zai2026glm53} (\cref{sec:sft-german}).

\paragraph{Reducing Hallucinations}
To train the model to abstain when the context does not support an answer, we hold out a small set of the samples of the Merlin--Arthur \ac{RL} environment as seeds (\cref{sec:merlin-arthur-environments}).
An \num{8}B proxy model that we trained on the Merlin--Arthur \ac{RL} environments computes Merlin's and Morgana's contexts for each seed.
The same model generates completions on the full context and on both masked contexts \parencite{deiseroth2025bounding}.
We keep a seed when its completions answer correctly on the first two contexts, abstain on Morgana's and pass a Command~A+ \parencite{cohere2026commandaplus} judge.
Each of its three completions enters the data pool as its own sample, so that the same question appears with and without its evidence and abstention depends on the context, not on the question.

\subsubsection{Safety, Compliance, and Alignment}
\label{sec:sft-safety}

Post-training largely determines how a model answers harmful, value-laden, and law-related requests \parencite{ouyang2022instructgpt,bai2022hhrlhf}.
A model trained only to follow instructions also follows harmful ones, and a small share of safety data in the mix reduces this \parencite{bianchi2024safetytuned}.
We therefore add both English and German open safety datasets to the \ac{SFT} mix.
The mix also holds examples in which hypothetical users ask the model in various ways to produce copyrighted material; these examples teach the model to reject such requests.
For legal questions that name no jurisdiction, models built in the US or China often assume US law for English prompts and Chinese law for Chinese prompts \parencite{wang2026jurisdictionaldefaults}.
We acknowledge that several of our teacher models were built in China, so they carry known political biases on sensitive topics, while our model should adhere to the values of the democratic consensus in Europe.

\paragraph{Value Alignment}
\label{sec:sft-value-alignment}
Many requests call for a judgement that depends on values.
For these, the model should judge by the values of the broad democratic consensus in Europe.
We curate a set of documents whose content mostly lends itself to a discussion of values and their consequences, such as the impact of a technology or the consequences of a political decision.
An \acs{LLM} proposes up to five questions about each of these documents.
The questions are phrased as everyday requests without words such as \enquote{ethical} or \enquote{moral}, as users rarely state that a request touches on values.
The model thus learns to apply these values without such a cue.
Documents that support no such questions are discarded.
Open-source models answer the questions under a system prompt that centres human dignity, human rights, liberal democracy, the rule of law and pluralism.
The prompt instructs the model to present positions across the democratic spectrum fairly and to keep facts, uncertainties and value judgements apart.
In-house question-answer pairs on politically sensitive topics add to this data and counter the known political biases of teacher models on these topics (\cref{sec:sft-post-processing}).

\subsection{Data Pool Curation}%
\label{subsec:sft-processing}
The datasets above differ in format, origin and quality, so we curate them into one training pool in three steps.
A shared processing pipeline brings every sample into one canonical format, drops conversations that claim another model's identity or follow \ac{CCP} narratives, and removes rows that reuse benchmark text (\cref{sec:sft-post-processing}).
Each reasoning trace then receives a reasoning-effort label from its length relative to traces of the same dataset and prompt difficulty (\cref{sec:sft-reasoning-effort}).
The result is the final data pool, which we break down by domain, source and language (\cref{sec:sft-pool}).

\subsubsection{Shared Processing}%
\label{sec:sft-post-processing}
Every dataset of the pool, open or generated, is processed using the same steps (\cref{tab:sft-post-processing}).
Each step either drops a whole conversation or rewrites it in place.
The open datasets arrive in different formats: some carry tool calls, some place the reasoning inline in the answer, and many have no reasoning at all.
The rewrites bring every sample into a common schema so that all datasets combine into one training set.

Most open datasets are generated by other models, and responses often carry their identity, such as \enquote{I am DeepSeek} or \enquote{As Qwen, I}.
A pattern filter drops every conversation in which the assistant claims to be another model or to come from another lab, unless it also names our own model, and ignores denials such as \enquote{I am not Llama}.
Answers of teacher and source models can also follow known political biases on sensitive topics, such as narratives of the \ac{CCP}.
For datasets with open-ended answers, we judge the whole conversation, including reasoning traces, with gpt-oss-120b~\parencite{openai2025gptoss} to classify each as carrying a \ac{CCP} narrative, as a refusal on general safety grounds or as balanced.
Conversations of the first kind are dropped.

\begin{table}[H]
  \centering
  \small
  \begin{tabular}{@{}ll>{\RaggedRight\arraybackslash}p{0.6\linewidth}@{}}
    \toprule
    Step & Action & Condition or change \\
    \midrule
    Final assistant turn
    & Transform
    & Trailing messages are removed until the conversation ends on an assistant turn. \\
    Length budget
    & Drop
    & The conversation exceeds the context window or \num{100} user turns. \\
    Repetition loops
    & Drop
    & An assistant turn has many characters in repeated \(10\)-grams~\parencite{rae2021gopher} or long runs of one character, token or short cycle of tokens. \\
    Foreign identity
    & Drop
    & The assistant identifies as another model. \\
    \acs{CCP} narratives
    & Drop
    & An \acs{LLM} judge finds an answer to follow narratives of the \ac{CCP}; only datasets with open-ended answers. \\
    Reasoning language consistency
    & Transform
    & English reasoning on an exchange in another language is removed. \\
    Reasoning and effort label
    & Transform
    & Reasoning inline in \texttt{content} moves into its own field, and every user turn is labelled with the reasoning effort of its exchange (\cref{sec:sft-reasoning-effort}). \\
    Chat schema
    & Transform
    & Every row is cast to one chat schema. \\
    Decontamination
    & Drop
    & The conversation overlaps a benchmark. \\
    \bottomrule
  \end{tabular}
  \caption{Post-processing steps every dataset of the \ac*{SFT} data pool passes, in the order they run.
    The list is not exhaustive: some datasets pass additional filters or transformations, and some steps change nothing in datasets whose source already satisfies them.
  }%
  \label{tab:sft-post-processing}
\end{table}

\paragraph{Decontamination}%
\label{sec:sft-decon}

Benchmark items in the training data inflate a model's scores beyond its performance on new inputs in real-world use \parencite{magar2022contamination,yang2023rethinking}.
To avoid this, we scan the \ac{SFT} data pool against our benchmarks and drop the rows that reuse benchmark text.

\pgfplotstableread[col sep=comma, row sep=\\, format=inline]{%
  n,floor,templates,templates_per_eval,templates_pooled,cost,cost_long_min,cost_long_max,window,rows_in,removed,removed_percent,pool_datasets\\%
  5,0.99,90,0,57,1.0,3.5,171,8,22000436,1097,0.0050,57\\%
}\sftDeconFacts
\DeclareRobustCommand{\sftDeconValue}[1]{%
  \pgfplotstablegetelem{0}{#1}\of{\sftDeconFacts}%
  \num{\pgfplotsretval}%
}
\DeclareRobustCommand{\sftDeconPercent}[1]{%
  \pgfplotstablegetelem{0}{#1}\of{\sftDeconFacts}%
  \qty{\pgfplotsretval}{\percent}%
}

We build on the \texttt{decon} engine of Olmo~3 \parencite{allenai2025decon,olmo2026olmo3} and match word \(n\)-grams with \(n = \sftDeconValue{n}\).
Published procedures use \(n\) from \numrange{7}{13} \parencite{singh2024evaluation}, which suits pre-training text, where a long match marks a copied passage.
In post-training data, a request can copy a benchmark item in a few words, and a long match is often boilerplate \parencite{abdin2024phi4,singh2024evaluation}.

Short \(n\)-grams also match benchmark instructions that repeat across its items, such as \enquote{Answer with a single letter}, which hold no benchmark question \parencite{brown2020language,abdin2024phi4,singh2024evaluation}.
Olmo's \texttt{decon} weights each \(n\)-gram by its \ac{IDF} \parencite{sparckjones1972statistical} over all benchmarks together.
We compute the \ac{IDF} within each benchmark instead, so wording common to a benchmark counts for little.
Of \sftDeconValue{templates} such instructions, our filter raises \sftDeconValue{templates_per_eval} false positives at the same recall, against \sftDeconValue{templates_pooled} with the \ac{IDF} over all benchmarks.

Because a long row shares more \(n\)-grams with a benchmark question by chance alone, we score the row in windows and keep the highest window matching score.
Each window is \sftDeconValue{window} times as long as the question, so its size scales with the length of the question.

\begin{figure}[tb]
  \centering
  \begin{minipage}[t]{0.42\linewidth}
    \centering
    \includegraphics{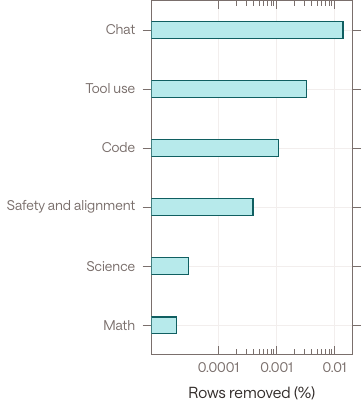}
    \caption{Percentage of the rows that decontamination drops in each domain of the \ac*{SFT} data pool (\cref{fig:sft-mix-domains}), on a log axis.
      Long context loses no row.
    }
    \label{fig:sft-decon-datasets}
  \end{minipage}\hfill
  \begin{minipage}[t]{0.54\linewidth}
    \centering
    \includegraphics{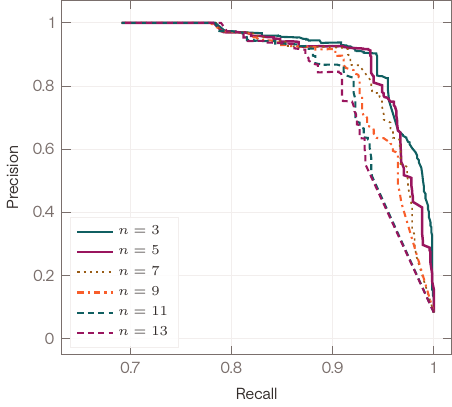}
    \caption{Precision against recall for each \(n\)-gram size over all removal thresholds, on exact and lightly edited benchmark items planted into a sample of the \ac*{SFT} data pool.
      The filter uses \(n = 5\).
      At high recall, precision falls as \(n\) grows.
    }
    \label{fig:sft-decon-calibration}
  \end{minipage}
\end{figure}

To set the threshold, we plant lightly altered copies of benchmark items (case, spacing, light paraphrasing) into a sample of the \ac{SFT} data pool, weighted by the share of each dataset.
We take the highest threshold that still finds them at a recall of \sftDeconValue{floor}.
At this threshold, the filter drops \sftDeconValue{cost} clean rows per \num{10000}, against \sftDeconValue{cost_long_min} to \sftDeconValue{cost_long_max} for \(n\) from \numrange{7}{13} (\cref{fig:sft-decon-calibration}).
We choose \(n = 5\), which ties for the best F1 score on the calibration set and has a higher recall than longer \(n\)-grams.
Paraphrasing can still get past this filter \parencite{yang2023rethinking}.

The filter removes \sftDeconValue{removed} of \sftDeconValue{rows_in} rows of the \sftDeconValue{pool_datasets} pool datasets (\sftDeconPercent{removed_percent}, \cref{fig:sft-decon-datasets}).
Training the \ac{SFT} ablation recipe with and without decontamination shows no measurable difference on the benchmarks the removed rows match.

\subsubsection{Reasoning Effort}%
\label{sec:sft-reasoning-effort}
\begin{figure}[ht]
  \centering
  \resizebox{\linewidth}{!}{%
    \includegraphics{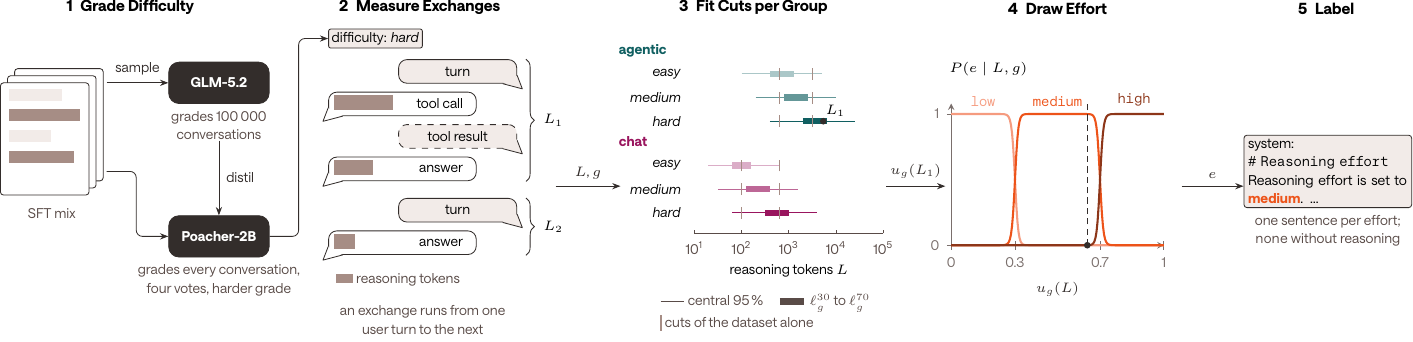}%
  }
  \caption{ Assignment of reasoning-effort labels to the exchanges of each sample from the \acs*{SFT} data pool.
    We first grade the difficulty of each conversation by its prompt with Poacher-2B, a grader distilled from a \acl*{LLM} judge.
    We then count the reasoning length \(L\) of each exchange, the number of reasoning tokens from one user turn to the next.
    For each reference group \(g\) (a combination of a dataset and a difficulty), we compute the 30th and 70th percentiles \(\ell_g^{30}\) and \(\ell_g^{70}\) of \(L\).
    To label an exchange post hoc, we assign the effort level based on the relative position of \(L\) to the 30th and 70th percentiles \(\ell_g^{30}\) and \(\ell_g^{70}\).
    Rather than a hard cut-off, we use the soft draw of \cref{eq:sft-effort-label} around the bucket edges \num{0.3} and \num{0.7}.
    This effort label is then rendered as part of the system prompt; an exchange without reasoning is labelled \texttt{none}.
  }%
  \label{fig:effort-pipeline}
\end{figure}

\pgfplotstableread[col sep=comma, row sep=\\, format=inline]{%
  datasets_with_cuts,dataset_median_min,dataset_median_max,cells_fitted,datasets_with_cells,benchmarks_with_length,benchmarks_ordered,benchmarks_scored,hard_over_easy_fixed_effort,high_over_low_min,high_over_low_max,score_none,score_low,score_high,sem_high_minus_low,delta_medium_to_high,judged_benchmarks,de_benchmarks,de_ordered,de_median_low,de_median_high,de_high_over_low,en_benchmarks,en_median_low,en_median_high,en_high_over_low,en_train_median_tokens,de_train_median_tokens,distinct_prompts,exact_pairs,exact_pairs_share_pct,exact_pairs_ratio2,exact_pairs_different_label_pct\\%
  76,27,90223,176,69,24,23,14,32.5,1.4,3.7,38.4,64.9,67.8,0.8,0.7,11,11,10,326,1788,5.5,13,1160,2993,2.6,2078,368,1794428,171021,9.5,53723,37.3\\%
}\sftEffortFacts
Longer reasoning at test time raises accuracy on hard problems~\parencite{wei2022chain,kojima2022large,openai2024o1,snell2025scaling,guo2025deepseekr1}.
However, reasoning models also spend many tokens on easy problems~\parencite{chen2024overthinking}, which adds latency and cost without improving the answer.
Prior work trains a model to decide whether to reason at all~\parencite{zhang2025adaptthink}, or controls the reasoning length through a token budget in the prompt~\parencite{han2024tokenbudget}, budget forcing at inference~\parencite{muennighoff2025s1} or a length target rewarded during \ac{RL}~\parencite{aggarwal2025l1}.
Similarly, we aim for a reasoning length that adapts to the prompt by itself, while the user keeps control over the trade-off between answer quality and latency.

\model{2} is a single hybrid model that either answers directly or reasons at one of three effort levels, similarly to Qwen3~\parencite{yang2025qwen3}, Llama-Nemotron~\parencite{bercovich2025llamanemotron} and gpt-oss~\parencite{openai2025gptoss}.
The chat template renders the requested level as one of four fixed sentences in the system prompt, for \texttt{none}, \texttt{low}, \texttt{medium} and \texttt{high} (\cref{tab:reasoning-effort-template}).
Unlike dedicated effort tokens, whose meaning has to be learned from scratch, the sentences describe each effort in words that the model can follow before any effort training.
The \ac{SFT} traces carry no effort label, so we assign one post hoc.
As summarised in \cref{fig:effort-pipeline}, the assigned label depends on the length of a reasoning trace relative to the other traces of the same dataset and prompt difficulty.
Post-hoc labels can still act as a control, because the mix contains several completions of the same prompt: \sftEffortValue{exact_pairs} of its \sftEffortValue{distinct_prompts} distinct opening prompts (\sftEffortPercent{exact_pairs_share_pct}) appear with two completions that both contain reasoning.
In \sftEffortValue{exact_pairs_ratio2} of these pairs the two reasoning lengths differ by a factor of two or more, and under the thresholds and the soft draw described below about \sftEffortPercent{exact_pairs_different_label_pct} of the pairs receive two different labels.
Within such a pair, the label is the only part of the input that differs, so training ties the difference in reasoning length to the label rather than to the prompt.

\paragraph{Domain- and Difficulty-Based Effort Labelling}
Each effort level should work in every domain, so the training data must contain every effort level in every domain.
We thus investigate the \emph{reasoning length} \(L\), the cumulative number of reasoning tokens generated in an \emph{exchange}, the span between two successive user turns.
We map exchanges with longer reasoning to higher reasoning effort.
For the datasets that contain reasoning (\sftEffortValue{datasets_with_cuts} datasets in an earlier, larger version of our \ac{SFT} data pool), the median reasoning length ranges from \sftEffortValue{dataset_median_min} to \sftEffortValue{dataset_median_max} tokens.
Thus, a global threshold under our assumed rule would label almost every short chat-like prompt as \texttt{low} reasoning effort, and almost every mathematical proof as \texttt{high}.
Therefore, we adopt dataset-specific reference groups.

Notably, reasoning length also depends on the difficulty of the prompt.
Empirically, we find that the reasoning length for a fixed percentile of the \diffHard{} prompts is up to \num{31.3} times that of the \diffEasy{} ones.
We therefore fit the thresholds per \emph{reference group} \(g = (\text{dataset}, \text{difficulty})\).
The dataset stands in for domain and task format, because each dataset of the mix comes from one source with one task shape.
Grouping by dataset thus separates proofs from tool calls without a domain classifier, and the difficulty grade refines the group within the dataset.

We grade the difficulty of each conversation with GLM-5.2~\parencite{zai2026glm52} as the judge model, following the \ac{LLM}-as-a-judge approach~\parencite{zheng2023judging}.
The judge grades a conversation as \diffEasy{}, \diffMedium{}, or \diffHard{}, based only on the knowledge and reasoning its prompt requires to be solved correctly.
Grading the whole pool with GLM-5.2 is too expensive, so GLM-5.2 grades \num{100000} conversations, and we distil these verdicts into a cheaper student model called \emph{Poacher-2B}, as \textcite{liu2024deita} did for their complexity scorer.
Poacher-2B then grades every conversation of the pool four times, and we round the mean of the four grades towards the harder grade.
On held-out conversations, this way of grading matches that of GLM-5.2 in \qty{84.8}{\percent} of cases and never differs from it by more than one level.

For each reference group \(g\) with at least \num{200} exchanges we measure the 30th and 70th percentiles \(\ell_g^{30}\) and \(\ell_g^{70}\) of \(L\); smaller reference groups use the percentiles of their dataset \(\ell^{30}\) and \(\ell^{70}\).
This yields thresholds for \sftEffortValue{cells_fitted} reference groups in \sftEffortValue{datasets_with_cells} datasets, and dataset-level thresholds for all \sftEffortValue{datasets_with_cuts}.

\paragraph{Soft Label Assignment}
A reasoning-effort level is then assigned based on the relative position of \(L\) to the 30th and 70th percentiles \(\ell_g^{30}\) and \(\ell_g^{70}\) of reasoning lengths of prompts of the same dataset and difficulty.
An exchange with \(L > 0\) is placed on the unit interval by interpolating in log space through these two percentiles, which fall on the bucket edges \num{0.3} and \num{0.7}, and its effort \(e \in \{\texttt{low}, \texttt{medium}, \texttt{high}\}\) is drawn as
\begin{equation}
  \label{eq:sft-effort-label}
  u_g(L) = \operatorname{clip}_{[0,1]}\left(0.3 + 0.4\,\frac{\log L - \log \ell_g^{30}}{\log \ell_g^{70} - \log \ell_g^{30}}\right),
  \qquad
  P(e \mid L, g) \propto \exp\left(-\frac{1}{2}\left(\frac{u_g(L) - m_e}{\beta\, w_e}\right)^{2}\right),
\end{equation}
where \(m_e\) and \(w_e\) are the centre and width of bucket \(e\) on the edges \num{0}, \num{0.3}, \num{0.7} and \num{1}, and \(\beta = \num{0.15}\).
An exchange without reasoning is labelled \texttt{none}.
By construction, every reference group yields about \qty{30}{\percent}, \qty{40}{\percent}, and \qty{30}{\percent} of \texttt{low}, \texttt{medium} and \texttt{high}.
The draw is seeded per row, and its width keeps the label from switching at one exact token count, which the model could otherwise learn as a length at which to stop.

\begin{figure}[t]
  \centering

  \includegraphics{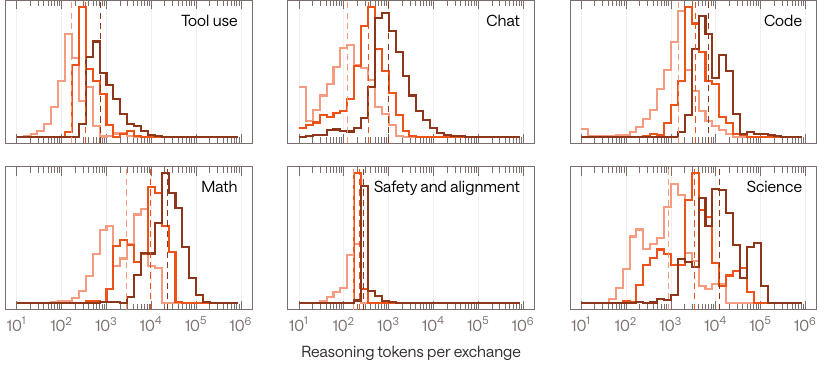}
  \includegraphics{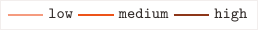}
  \caption{Reasoning tokens per effort label in the training data of the released \model{2} \acs*{SFT} checkpoint, read back from a sample of its dataset artefacts and pooled per domain.
    Dashed lines mark the label medians.
    The labels overlap within a domain because the thresholds differ between the datasets of a domain and because the draw of \cref{eq:sft-effort-label} is soft; the same label covers different lengths in different domains.
  }
  \label{fig:effort-labels}
\end{figure}

\Cref{fig:effort-labels} shows the assigned labels in the training data, pooled per domain.
Within a domain the three efforts cover overlapping but ordered ranges of length, and the same effort covers different lengths in different domains.
Because the thresholds are set per difficulty, a user does not have to estimate the difficulty of their prompt when choosing a reasoning effort.

\begin{figure}[!tb]
  \centering
  \includegraphics{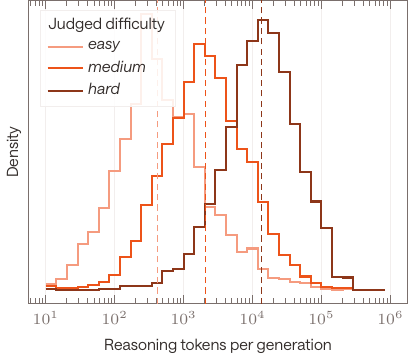}%
  \hfill{}
  \includegraphics{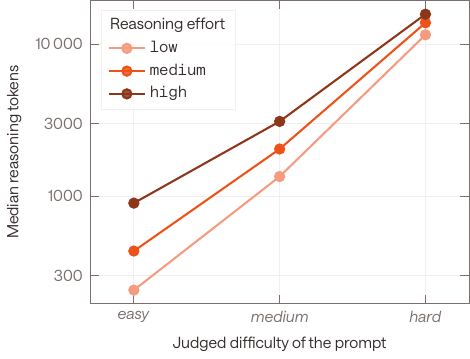}
  \caption{Reasoning length of the released \model{2} \acs*{SFT} checkpoint against the difficulty Poacher-2B assigns to the evaluation prompts.
    Left: length by judged difficulty over the three reasoning efforts, dashed lines at the medians.
    Right: the median length per judged difficulty, one line per effort; at a fixed difficulty the three lines show how much the effort adds, and along one line how much the difficulty adds.
    \texttt{none} is left out, as it produces an empty reasoning block.}
  \label{fig:effort-difficulty}
\end{figure}

\paragraph{Analysis of \acs*{SFT} Reasoning Effort} We evaluate the released \model{2} \ac{SFT} checkpoint, the soup of \mixCandOne{} and \mixCandTwo{} (\cref{sec:sft-model-souping}), once per effort at temperature \num{1} and top-p \num{0.95} on the production evaluation suite.
Under \texttt{none}, the model emits an empty reasoning block.
The median reasoning length orders \texttt{low} < \texttt{medium} < \texttt{high} on \sftEffortValue{benchmarks_ordered} of the \sftEffortValue{benchmarks_with_length} benchmarks with a measurable length.
\Cref{fig:effort-difficulty} splits this by the difficulty that Poacher-2B assigns to the evaluation prompts of \sftEffortValue{judged_benchmarks} graded benchmarks.
With the effort fixed, \diffHard{} prompts receive about \sftEffortValue{hard_over_easy_fixed_effort} times the reasoning of \diffEasy{} ones; with the difficulty fixed, \texttt{high} receives between \sftEffortValue{high_over_low_min} and \sftEffortValue{high_over_low_max} times the reasoning of \texttt{low}.
The effort thus moves the length within the range the prompt calls for, and does not override it.
We average the score over the \sftEffortValue{benchmarks_scored} benchmarks whose generations fit the length limit.
Reasoning at \texttt{low} raises this mean from \sftEffortValue{score_none} under \texttt{none} to \sftEffortValue{score_low}, and \texttt{high} reaches \sftEffortValue{score_high}, with a paired standard error of \sftEffortValue{sem_high_minus_low} points against \texttt{low}.
The step from \texttt{medium} to \texttt{high} adds \sftEffortValue{delta_medium_to_high} points.

The reasoning effort also works the same way in German and English.
On \sftEffortValue{de_benchmarks} German benchmarks the median reasoning length orders \texttt{low} < \texttt{medium} < \texttt{high} on \sftEffortValue{de_ordered}, and the benchmark median rises from \sftEffortValue{de_median_low} to \sftEffortValue{de_median_high} tokens, a factor of \sftEffortValue{de_high_over_low}; the \sftEffortValue{en_benchmarks} English benchmarks rise from \sftEffortValue{en_median_low} to \sftEffortValue{en_median_high} tokens, a factor of \sftEffortValue{en_high_over_low}.
German prompts thus receive shorter reasoning at every effort, which matches the data pool: the median reasoning length of its German exchanges is \sftEffortValue{de_train_median_tokens} tokens against \sftEffortValue{en_train_median_tokens} for English.
\Cref{fig:sft-reasoning-effort-tasks} plots the score against the median completion tokens per effort for a selection of benchmarks, with the German counterparts of GPQA Diamond and AIME 2026 as dashed lines.

\FloatBarrier
\subsubsection{Final Data Pool}%
\label{sec:sft-pool}
\pgfplotstableread[col sep=comma, row sep=\\, format=inline]{%
  datasets,samples_m,tokens_b,loss_tokens_b,loss_pct,german_pct,other_tokens_pct,english_pct,other_pct,mixed_pct\\%
  57,22.0,335.8,142.4,42.4,5.0,3.3,92.6,2.4,0.8\\%
}\sftPoolFacts

\begin{figure}[!ht]
  \centering
  \colorlet{sftramplast}{AARampOne!55}
  \def\sftpiecolors{{"AARampSix","AARampFive","AARampFour","AARampThree","AARampTwo","AARampOne","sftramplast"}}
  \newcommand{\sftPoolSwatch}[1]{\pgfmathsetmacro\sftswatchcolor{\sftpiecolors[#1]}\textcolor{\sftswatchcolor}{\rule{0.7em}{0.7em}}}
  \begin{minipage}[b]{0.63\linewidth}
    \centering
    \footnotesize
    \setlength{\tabcolsep}{3pt}
    \begin{tabular}[b]{@{}c@{\hspace{0.4em}}lrrrrrr@{}}
      \toprule
      &                              &                             &                                  &             & \multicolumn{3}{c}{Source (\% of tokens)}                            \\ \cmidrule(l){6-8}
      \multicolumn{2}{@{}l}{Domain}
      & \shortstack[r]{Samples\\(M)}
      & \shortstack[r]{Tokens\\(B)}
      & \shortstack[r]{Loss\\tokens (B)}
      & OS
      & \shortstack[r]{Regenerated\\OS}
      & In-house \\
      \midrule
      \sftPoolSwatch{0}
      & Chat
      & \num{7.02}
      & \num{86.1}
      & \num{21.9}
      & \num{8.9}
      & \num{57.9}
      & \num{33.2} \\
      \sftPoolSwatch{1}
      & Code
      & \num{4.00}
      & \num{82.1}
      & \num{36.0}
      & \num{63.6}
      & \num{29.8}
      & \num{6.6} \\
      \sftPoolSwatch{2}
      & Math
      & \num{5.26}
      & \num{60.6}
      & \num{59.7}
      & \num{65.3}
      & \num{32.3}
      & \num{2.4} \\
      \sftPoolSwatch{3}
      & Long context
      & \num{0.09}
      & \num{49.8}
      & \num{0.9}
      & \num{0.0}
      & \num{0.0}
      & \num{100.0} \\
      \sftPoolSwatch{4}
      & Tool use
      & \num{2.05}
      & \num{35.2}
      & \num{5.1}
      & \num{10.6}
      & \num{34.9}
      & \num{54.5} \\
      \sftPoolSwatch{5}
      & Science
      & \num{3.32}
      & \num{18.7}
      & \num{18.0}
      & \num{97.0}
      & \num{1.4}
      & \num{1.6} \\
      \sftPoolSwatch{6}
      & Safety and alignment
      & \num{0.25}
      & \num{3.2}
      & \num{0.8}
      & \num{0.0}
      & \num{1.1}
      & \num{98.9} \\
      \midrule
      & All
      & \num{22.00}
      & \num{335.8}
      & \num{142.4}
      & \num{36.1}
      & \num{31.7}
      & \num{32.1} \\
      \bottomrule
    \end{tabular}
  \end{minipage}\hfill
  \begin{minipage}[b]{0.35\linewidth}
    \centering
    \includegraphics{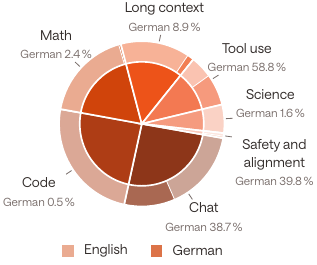}
  \end{minipage}

  \begin{minipage}[t]{0.63\linewidth}
    \captionof{table}{The \ac*{SFT} data pool by domain.
      Samples are conversations, in millions (M).
      Tokens are the actual training tokens and loss tokens are the tokens on which the training loss is computed, both in billions (B).
      The source columns split the domain's tokens: open-source data as released (OS); open-source data whose responses we regenerated, completed with synthesised reasoning or translated to German (Regenerated OS); and data we generated (In-house).
    }
    \label{tab:sft-mix}
  \end{minipage}\hfill
  \begin{minipage}[t]{0.35\linewidth}
    \caption{Share of the tokens of the \ac*{SFT} data pool by domain (inner disk, colours as in the adjacent \cref{tab:sft-mix}) and, within each domain, by language (outer ring).
      Each label gives the domain and, below it, the German share of its tokens.
      Other languages, \sftPoolPercent{other_tokens_pct} of all tokens, are left out.
    }
    \label{fig:sft-mix-domains}
  \end{minipage}
\end{figure}

Our \ac{SFT} data pool holds \sftPoolValue{datasets} datasets with \sftPoolValue{samples_m}M samples and \sftPoolValue{tokens_b}B tokens from seven domains: science, math, code, tool use, chat, long context, and safety and alignment.
The final model trains on mixes drawn from the pool, not on all of it (\cref{sec:sft-merge-mixing,sec:sft-model-souping}).
Tokens are counted as \ac{SFT} trains on them, after multi-turn splitting (\cref{sec:sft-recipe}) and rendering with our \model{2} chat template.
The \sftPoolValue{loss_tokens_b}B loss tokens, \sftPoolPercent{loss_pct} of the total tokens, are the unmasked tokens that provide the gradient signal.
Most samples are in English (\sftPoolPercent{english_pct}) or German (\sftPoolPercent{german_pct}); the remaining \sftPoolPercent{other_pct} are in other languages, including \sftPoolPercent{mixed_pct} that mix several.
\Cref{tab:sft-mix} provides the number of samples, tokens and sources for each domain, and \cref{fig:sft-mix-domains} illustrates the distribution of tokens across domains and languages.

\FloatBarrier

\subsection{Data Mixing with MergeMix}%
\label{sec:sft-merge-mixing}

\pgfplotstableread[col sep=comma, row sep=\\, format=inline]{%
  validation_positions,clusters,cluster_datasets,cluster_tokens,smallest_cluster_tokens,specialist_steps,specialist_positions,specialist_total,ablation_positions,fitting_runs,fitting_total,fitting_saving,sampled_scored,refine_scored,refine_kappa,merged_total,baseline_merged,baseline_rank,best_sampled,best_refine,trained,trained_others,trained_baseline,trained_above_baseline,trained_best,best_merge_trained,shared_n,shared_rho\\%
  20.0,20,63,232,4.55,67,4.50,90,16.78,63,1057,91,43,25,25,69,50.2,49,59.2,61.4,8,7,51.5,6,56.7,50.0,6,-0.09\\%
}\sftMergeFacts

The final \ac{SFT} model trains on a targeted mix drawn from the data pool, not on the whole pool.
To find a strong mix, we group the datasets into \sftMergeValue{clusters} clusters by the capability they train (\cref{sec:sft-merge-clusters}) and search for the share of each cluster.
Fitting these shares from trained candidate mixes needs at least \(3(m + 1)\) training runs for \(m\) clusters~\parencite{chen2026olmix}.
In our case, these are \sftMergeValue{fitting_runs} \ac{SFT} runs of \sftMergeValue{ablation_positions}B tokens each.
MergeMix~\parencite{wang2026mergemix}, which we also use in mid-training (\cref{sec:mid-mixing}), needs only one shorter training run per cluster, of \sftMergeValue{specialist_positions}B tokens.
We use MergeMix to train on \sftMergeValue{specialist_total}B instead of \sftMergeValue{fitting_total}B tokens, i.e.\ \sftMergePercent{fitting_saving} less training compute, fine-tuning one specialist model per cluster.
To score a candidate mix, we average the specialists' parameters, each weighted by its cluster's share in the candidate mix, and evaluate the merged model.
After the specialists are trained, each candidate costs one evaluation instead of one training run.
Averaging works here because models fine-tuned from one checkpoint stay close in parameter space~\parencite{wortsman2022soups}.
\textcite{berasi2026merging} select \ac{SFT} mixes of two to four domains in the same way.

We first check that merged models rank weightings as trained models do in \ac{SFT}.
For this, we merge a math, a code, and a chat specialist with \num{12} weightings and train a model on the same mix for each of the \num{10} weightings that are not a single specialist (\cref{app:sft-merge-mix}).
On AIME 2026 and LiveCodeBench, merging and training rank the weightings alike; on MATH-500, the scores barely change with the weighting.
We conclude that the merges are a usable proxy and apply the method to the \sftMergeValue{clusters} clusters of the full \ac{SFT} mix (\cref{sec:sft-merge-weightings}).
With \sftMergeValue{clusters} clusters, the merges find good candidates but do not rank them: most weightings the search finds train above the hand-set baseline, but their merged scores do not predict which of them trains best.
We therefore train the most promising weightings and select the final mixes by their trained scores (\cref{sec:sft-merge-results}).

\subsubsection{Clusters and Specialist Models}%
\label{sec:sft-merge-clusters}
We run the sweep on an earlier version of the \ac{SFT} data pool, with \sftMergeValue{cluster_datasets} datasets and \sftMergeValue{cluster_tokens}B tokens.
Unless stated otherwise, this subsection counts the tokens of data in the same way as \cref{sec:sft-pool}.
We group the datasets by their capabilities into \sftMergeValue{clusters} clusters: math, math proofs, code, software engineering, science, general chat, instruction following and safety, tool calling, interactive agents, terminal use, long context and German, where the larger capabilities span several clusters.
We choose \sftMergeValue{clusters} as the finest grouping along capabilities in which every cluster still holds a few billion tokens of data.
The smallest cluster holds \sftMergeValue{smallest_cluster_tokens}B tokens.
Each specialist is a model fine-tuned on \sftMergeValue{specialist_positions}B packed tokens of a single cluster, fewer than the smallest cluster holds, so no specialist sees its data more than once.
The specialists train from \basemodel{2} (\cref{sec:long-context}) with the batch size and peak learning rate of the production recipe (\cref{subsec:sft-training-recipe}), for \sftMergeValue{specialist_steps} steps with \num{5} warmup steps and no cooldown.

\subsubsection{Searching over Cluster Weightings}%
\label{sec:sft-merge-weightings}
We search for the weighting of the \sftMergeValue{clusters} clusters whose merge scores best, where a weighting is a set of \sftMergeValue{clusters} non-negative shares that sum to one.
The set of weightings is too large for a grid, so, as in mid-training (\cref{sec:mid-mixing}), we draw weightings from Dirichlet distributions whose concentration \(\kappa\) sets how close the draws stay to the centre.
The first round centres on the baseline weighting, which gives each cluster the share of packed tokens its datasets receive in the reference mix, a hand-set \ac{SFT} mix that fixes the epochs of each of the \sftMergeValue{cluster_datasets} datasets.
It draws \sftMergeValue{sampled_scored} weightings, at \(\kappa = 50\) close to the baseline and at \(\kappa = 15\) further away.
The second round draws \sftMergeValue{refine_scored} weightings around the best merge of the first, at \(\kappa = \sftMergeValue{refine_kappa}\).
With the baseline, we score \sftMergeValue{merged_total} merges.

All merges and the \sftMergeValue{clusters} individual specialists are scored on \num{14} benchmarks, most of them from the post-training suite (\cref{sec:post-training-eval-suite}), grouped into seven capabilities.
Among them are math (AIME 2025, AIME 2026 and AIME 2026 IID), code (LiveCodeBench), agentic work (SWE-bench Verified, a Terminal-Bench subset~\parencite{merrill2026terminalbench} and multi-turn \acs{BFCL}), instruction following (IFBench), knowledge (GPQA Diamond and \acs{MMLU}-Pro), long context (LongBench Pro and AA-LCR) and German (the German translation of AIME 2026 and KulturBank, \cref{app:sft-kulturbank}).
The score of a cluster weighting is the mean across all seven capability averages, weighting each capability equally regardless of its benchmark count.

\begin{table}[tb]
  \centering
  \small
  \setlength{\tabcolsep}{9pt}
  \begin{tabular}{llcrcr}
    \toprule
    Weighting & Source & Merged & Rank & Trained & Rank \\
    \midrule
    Baseline weighting&Baseline&$50.2$&49&$51.5$&7\\
    \midrule
    \(\kappa= 50\) draw 15&Sampled&$55.1$&29&$56.7$&1\\
    Refinement draw 13&Sampled&$60.4$&4&$55.4$&2\\
    Refinement draw 7&Sampled&$61.2$&2&$54.5$&3\\
    \(\kappa= 15\) draw 6&Sampled&$55.0$&30&$52.1$&6\\
    Refinement draw 5&Sampled&$61.4$&1&$50.0$&8\\
    \midrule
    Gradient-boosted trees&Proposed&--&--&$54.2$&4\\
    \acs*{GP}&Proposed&--&--&$52.2$&5\\
    \bottomrule
  \end{tabular}
  \caption{Weightings trained as data mixtures, by source and in order of their trained score.
    Merged is the score of the merge of the specialists with the weighting, ranked among all \sftMergeValue{merged_total} merges.
    Trained is the score of one \acs*{SFT} run on data sampled with the weighting, ranked among the \sftMergeValue{trained} trained weightings.
    Both scores are in percent (\cref{sec:sft-merge-weightings}), and their Spearman correlation over the \sftMergeValue{shared_n} weightings with both is \sftMergeValue{shared_rho}.
  }
  \label{tab:sft-merge-trained}
\end{table}

\subsubsection{From Merges to Trained Mixes}%
\label{sec:sft-merge-results}
Among the \sftMergeValue{merged_total} merges, the baseline weighting scores \sftMergePercent{baseline_merged} and ranks \sftMergeValue{baseline_rank}, while the best merges of the two rounds reach \sftMergePercent{best_sampled} and \sftMergePercent{best_refine}.
We then train \sftMergeValue{trained} candidate weightings as data mixes (\cref{tab:sft-merge-trained-weights}).
The candidates are the baseline as reference, draws from both rounds including the best merge of the second round, and two weightings proposed by surrogate models fitted to the merge scores (\cref{tab:sft-merge-trained}).
Each model trains from \basemodel{2} with the ablation recipe (\cref{app:sft-ablation-recipes}).

Training confirms that the search finds better mixes overall: \sftMergeValue{trained_above_baseline} of the \sftMergeValue{trained_others} other weightings train above the baseline, and the best reaches \sftMergePercent{trained_best} against \sftMergePercent{trained_baseline} for the baseline.
However, it does not confirm the ranking of the merges.
The best merge trains worst, at \sftMergePercent{best_merge_trained}.
The best trained weighting is a first-round draw from the middle of the merge ranking.
Overall, merged and trained scores do not correlate (Spearman \(\rho = \sftMergeValue{shared_rho}\)).
In the validation, the merges rank the weightings as training does, but in the sweep they do not.
The sweep merges \sftMergeValue{clusters} specialists instead of two or three, and each specialist trains on \sftMergeValue{specialist_positions}B instead of \sftMergeValue{validation_positions}B packed tokens.
Our experiments do not show which of these two differences breaks the ranking.
\FloatBarrier

\subsection{\model{2} \acs*{SFT} Training}%
\label{subsec:sft-training-recipe}

The \ac{SFT} stage starts from \basemodel{2} and trains for \num{4000} steps, around \num{268}B tokens, on packed sequences of \recommendedContextLengthShort{} tokens (\cref{sec:sft-recipe}).
At this length, the cost of a sequence depends on how its tokens split into conversations, since attention cost grows with the square of each conversation's length.
With packing alone, every rank waits for the rank with the most expensive sequence.
Ordering the packed sequences by attention cost on every rank, without communication between ranks, raises the throughput by \qty{84}{\percent} (\cref{sec:sft-data-loading}).
MergeMix leaves four candidate mixes within a few points of each other, so we train each for longer, continue the best two to full length and soup them; the soup scores above both of its runs (\cref{sec:sft-model-souping}).

\subsubsection{Data Loading}%
\label{sec:sft-data-loading}
We follow common practice and pack several conversations into one sequence of fixed length with document-masked attention~\parencite{krell2021packing,dao2023flashattention2}, so no token attends across a conversation boundary.
Conversations are assigned online with Next-\(k\)-Fit~\parencite{johnson1974fast}: each goes into the first of \(k\) open bins that still has room, and if none has, we emit the fullest bin and open a new one.
Larger \(k\) trades memory and start-up time for less padding.
In ablations at a sequence length of \num{32}k tokens, raising \(k\) from \num{1} to \num{1024} lowers the share of padding tokens in the inputs from \qty{14.8}{\percent} to \qty{0.6}{\percent}.
Less padding means more tokens per optimiser step, so the learning rate is tuned for the chosen \(k\)~\parencite{goyal2017accurate}.
\model{2} trains with \(k = \num{256}\) bins at a sequence length of \recommendedContextLengthShort{} tokens with a share of padding tokens of \qty{0.001}{\percent}.

With bin packing, every sequence is equally long, but not equally expensive.
Attention cost grows with the square of each conversation's length, so a sequence holding one long conversation costs far more than one holding many short ones.
Each data-parallel rank draws its sequences independently.
Since all ranks synchronise after every microbatch, all ranks have to wait for the rank with the most expensive sequence, resulting in wait times and idle compute.
As in workload balancing for variable-length training~\parencite{wang2025wlbllm,ge2025bytescale}, but without communication between ranks, every rank buffers \(b\) packed sequences and orders them by their attention cost
\begin{equation}
  \label{eq:sft-attention-cost}
  C(s) = \sum_{d \in s} |d|^{2},
  \qquad
  x_{n} = s_{(\pi(n))}, \quad n = 1, \dots, b,
\end{equation}
where \(|d|\) is the length of a document \(d\) in sequence \(s\), \(s_{(i)}\) the \(i\)-th sequence of the buffer of size \(b\) sorted by descending \(C\), \(\pi\) a permutation of \(1, \dots, b\) drawn from a seed shared by all ranks, and \(x_{n}\) the \(n\)-th microbatch a rank draws from the buffer.
Because all ranks sort buffers of the same size drawn from the same distribution, the \(i\)-th sequence of each buffer has a similar cost on every rank, so all ranks process similarly expensive microbatches at the same time.
Each rank fills a second buffer while it draws from the first and switches to it once the first is exhausted, so sorting does not stall the training.

On \num{64} B300 GPUs with the \model{2} \ac{SFT} configuration, a buffer of \(b = \num{256}\) raises the median throughput per GPU from \qty{8610}{\token\per\second} without the ordering buffer to \qty{15850}{\token\per\second}, by \qty{84}{\percent}.

\subsubsection{Recipe}%
\label{sec:sft-recipe}

\begin{table}[!ht]
  \centering
  \small
  \begin{tabular}{l >{\raggedright\arraybackslash}p{0.58\linewidth}}
    \toprule
    Setting & Value \\
    \midrule
    Base checkpoint
    & \basemodel{2} (\cref{sec:long-context}) \\
    Context length
    & \recommendedContextLengthShort \\
    Batch size
    & \num{256} \\
    Tokens per step
    & \num{67.1}\mbox{M} \\
    Steps
    & \num{4000} \\
    Peak learning rate
    & Half of pre-training in every group (\cref{tab:bienenelfe-optimizer}) \\
    Learning-rate schedule
    & \ac{WSD} with linear warmup over \num{50} steps and linear decay over the last \num{800} steps to a tenth of the peak \\
    Parallelism
    & \acs*{FSDP} over \num{128} ranks, replicated twice, on \num{256} GPUs \\
    Expert load balancing
    & Disabled (\cref{sec:load-balancing}) \\
    \bottomrule
  \end{tabular}
  \caption{Training settings of the \ac*{SFT} stage.
    Sequences are packed to the full context length, so a step covers \num{67.1}M tokens.
    The optimiser groups, their momentum and their weight decay are those of pre-training (\cref{tab:bienenelfe-optimizer}).
  }%
  \label{tab:sft-settings}
\end{table}

We train with cross-entropy loss only on the tokens the model produces at inference: the reasoning, answer and tool calls of each assistant turn.
User turns, tool responses and the special tokens of the chat format are masked out.
The loss is normalised over the whole batch, so every supervised token has the same weight and neither packing (\cref{sec:sft-data-loading}) nor multi-turn splitting reweights the mix.
The chat template drops the reasoning of earlier assistant turns, so a conversation with reasoning cannot be supervised on all its turns at once.
We therefore split it into one sample per user turn, with loss on the last turn only; earlier turns keep their tool calls and answers but lose their reasoning.
On average, a conversation yields \num{2.27} samples.
A shuffle buffer keeps the samples of one conversation from dominating a batch.

\FloatBarrier
\subsubsection{Candidate Runs and Souping}%
\label{sec:sft-model-souping}
\suppressfloats[t]
This stage selects the \ac{SFT} model we pass on to \ac{RL}.
The MergeMix sweep leaves four candidate mixes within a few points of each other at the ablation budget (\cref{sec:sft-merge-results}), so we compare them in longer runs and soup the best of them.
Alongside, we train a baseline mix, \mixBaseline{}, that samples the datasets of the final pool about naturally, with the German data upweighted.
It is a sanity check that reweighting the domains improves on natural sampling.
\pgfplotstableread[col sep=comma, row sep=\\, format=inline]{%
  unfinished_base,unfinished_baseline_4000,unfinished_r7_4000,math_weight_baseline,math_weight_r7,aime_base,aime_base_ci95,aime_baseline_4000,aime_baseline_4000_ci95,aime_iid_base,mean_base,mean_base_ci95,mean_d15_2500,mean_d15_2500_ci95,mean_r7_2500,mean_r7_2500_ci95,mean_gbt_2500,mean_gbt_2500_ci95,mean_r7_4000,mean_r7_4000_ci95,mean_soup_4000,mean_soup_4000_ci95,tokens_2500,tokens_4000\\%
  2.1,7.3,0.0,18.3,4.9,74.6,10.7,69.4,9.3,74.6,30.0,2.0,52.6,1.9,59.9,1.8,55.7,2.1,58.9,1.8,63.4,1.8,168,268\\%
}\sftSoupFacts
\begingroup\sisetup{uncertainty-mode=separate, separate-uncertainty-units=single}%
Starting from \basemodel{2} (\cref{subsec:sft-training-recipe}), we first fine-tune one model per mix for \num{2500} steps, \sftSoupValue{tokens_2500}B packed tokens.
Two of the four candidates give clearly the best of these models, which we call \mixCandOne{} and \mixCandTwo{}.
Over the benchmarks of \cref{fig:sft-soup}, they reach means of \sftSoupInterval{mean_r7_2500}{mean_r7_2500_ci95} and \sftSoupInterval{mean_gbt_2500}{mean_gbt_2500_ci95}, against at most \sftSoupInterval{mean_d15_2500}{mean_d15_2500_ci95} for the other three models.
We continue training \mixCandOne{}, \mixCandTwo{} and \mixBaseline{} to \num{4000} steps, \sftSoupValue{tokens_4000}B packed tokens.

\begin{figure}[t]
  \centering
  \includegraphics{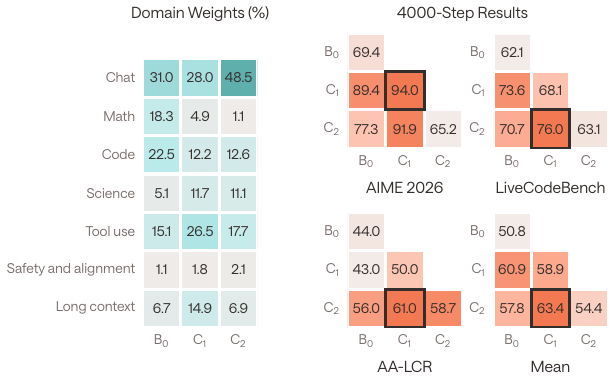}
  \caption{Domain weights of the three mixes trained for \num{4000} steps (left) and the scores of their models and of the soup of each pair (right), in \unit{\percent}.
    The domain weights are shares of the trained tokens.
    In each score panel, the diagonal holds the three models and each cell below it the soup of the models named by its row and column; colours are scaled per panel, and the best cell is framed.
    AIME 2026, LiveCodeBench and AA-LCR are the math, code and long-context benchmarks whose scores spread most over the mixes and soups.
    The mean is over \num{12} benchmarks: math (AIME 2025, AIME 2026 and AIME 2026 IID of \cref{sec:sft-merge-mixing}), knowledge (GPQA Diamond and KulturBank, \cref{app:sft-kulturbank}), agentic tool use (\taubench{2} telecom, multi-turn \acs*{BFCL}~v3 and BrowseComp Lite), code (LiveCodeBench), instruction following (IFBench, loose accuracy) and long context (AA-LCR and LongBench Pro).
  }
  \label{fig:sft-soup}
\end{figure}
\paragraph{Souping}
The three mixes weight the domains differently, and their models lead on different benchmarks: \mixCandOne{} on AIME 2026 and LiveCodeBench, \mixCandTwo{} on AA-LCR (\cref{fig:sft-soup}).
To combine these strengths, we soup~\parencite{wortsman2022soups} every pair of the three models: each parameter of a soup is the mean of its values in the two models.
Averaging works because both models are fine-tuned from the same base checkpoint and stay close to each other in weight space~\parencite{wortsman2022soups}.
It is the operation of the MergeMix sweep, applied to two full runs instead of specialists.
\textcite{maiti2025souper} show that a soup of models strong in different benchmark categories can score above each of them on multilingual, tool-calling and math benchmarks.
We soup only pairs, with equal weights.

\paragraph{Results}
Every soup of the \num{4000}-step models scores above both of its models on the mean (\cref{fig:sft-soup}).
The soup of \mixCandOne{} and \mixCandTwo{} reaches the highest mean, \sftSoupInterval{mean_soup_4000}{mean_soup_4000_ci95}, against \sftSoupInterval{mean_r7_4000}{mean_r7_4000_ci95} for \mixCandOne{}, the best single model, and \sftSoupInterval{mean_base}{mean_base_ci95} for the base model.
\mixBaseline{} scores lowest of the single models, so the candidate mixes pass this sanity check.
On AIME 2026, the soup scores below \mixCandOne{}, within noise.
Selection and these results use the same benchmarks, so the scores of the selected soup are optimistic.

\paragraph{Math Data}
\textcite{guha2025openthoughts} find that math scores, AIME among them, rise with the number of math reasoning traces in \ac{SFT}.
Yet \mixCandOne{} scores highest on AIME 2026, although math makes up only \sftSoupPercent{math_weight_r7} of its tokens, against \sftSoupPercent{math_weight_baseline} for \mixBaseline{}.
Compared with \mixBaseline{}, \mixCandOne{} drops or downsamples in particular the math datasets with on average easier problems and longer reasoning traces.
\mixBaseline{}, which keeps these datasets, reaches \sftSoupInterval{aime_baseline_4000}{aime_baseline_4000_ci95} on AIME 2026, within noise of the base model before \ac{SFT} at \sftSoupInterval{aime_base}{aime_base_ci95}.
On AIME 2026 IID, the base model scores exactly the same, \sftSoupPercent{aime_iid_base}, so its AIME 2026 score does not stem from memorised problems.
One explanation is that the long reasoning traces lengthen the generations of the model until they no longer finish.
In line with this explanation, \mixBaseline{} ends more AIME 2026 generations in a repetition loop or at the context limit, \sftSoupPercent{unfinished_baseline_4000}, against \sftSoupPercent{unfinished_base} for the base model and \sftSoupPercent{unfinished_r7_4000} for \mixCandOne{}.
Other causes are possible, as \mixCandOne{} and \mixBaseline{} also differ in other domains.

As the best model on the mean, the soup of \mixCandOne{} and \mixCandTwo{} after \num{4000} steps becomes the released \ac{SFT} checkpoint and the starting point of \ac{RL}.
\endgroup

%% file: sections/03-post-training/02-reinforcement-learning.tex
\section{\actitle{RL} (\acs*{RL})}
\label{sec:rl}
\pgfplotstableread[col sep=comma, row sep=\\, format=inline]{%
  window,smooth_window,environments,reward_first,reward_last,instruct_first,instruct_last\\%
  50,21,38,0.62,0.83,0.65,0.94\\%
}\rlResultsFacts

While \ac{SFT} adapts the base model towards user interaction and tool use, \ac{RL} further expands its capabilities by training it on its own rollouts against verifiable rewards.
In this section, we describe the \ac{RL} training stage that starts from the \ac{SFT} checkpoint.
It consists of a single training run on \rlResultsValue{environments} environments that span math, coding, reasoning, instruction following, tool use and agentic tasks in English and German (\cref{sec:rl-environment-mix}).
We train asynchronously on a token-level objective that combines a policy-gradient term, a squared log-ratio regulariser \parencite{kimiteam2026k25, kimiteam2025kimik15scalingreinforcement} and targeted \ac{OPSD} for low-reward groups \parencite{zhao2026selfdistilledreasoner}, as outlined in \cref{sec:rl-training-recipe}.
For the policy-gradient term, we use group-relative advantages \parencite{shao2024deepseekmath,liu2025drgrpo}, a \ac{BTV} trust region \parencite{qi2026rethinkingtrustregion} and a per-token bound on the gradient norm of the logits.
The trainer further replicates the FP8 quantisation and expert routing of the inference engine to keep both policies numerically aligned.
We push FP8 weights to all inference replicas via \ac{RDMA} after every optimiser step without discarding in-flight requests, and serve tens of thousands of concurrent rollouts with isolated code sandboxes, as described in \cref{sec:rl-infrastructure}.
\Cref{sec:rl-run} presents the \model{2} \ac{RL} run, its results and an analysis of its training dynamics.
Finally, \cref{sec:rl-lessons} summarises the lessons we learned over several iterations of our \ac{RL} setup, from early experiments to \model{1} and \model{2}.
\Cref{app:rl} gives further details.

\paragraph{Terminology}
For each prompt, we sample a \emph{group} of \emph{sequences}, where each sequence consists of the complete single- or multi-turn rollout for this prompt.
A \emph{rollout step} comprises multiple model output tokens, and the environment response that follows it, e.g.\ a tool response.
A \emph{turn} comprises all rollout steps between two user messages, such that one user message may be followed by several rollout steps with tool calls before the model responds to the user.
A sequence may consist of several \emph{segments}, which are several rollout steps that extend the same tokenised context.
We split sequences into segments due to, e.g.\ chat-template changes that adapt previously rendered contexts (\cref{sec:rl-segments}).
These splits can, but need not, be along turn boundaries.
We perform optimiser steps on batches of sequences, regardless of their number of segments.
We only train on the tokens that are newly generated by the model in each segment, masking out re-rendered context as well as system, user and tool-response tokens.

\FloatBarrier
\subsection{Training Recipe}
\label{sec:rl-training-recipe}

\subsubsection{Optimisation Objective}
\label{sec:rl-objective}
We optimise a policy-gradient term \(\ell_t^{\mathrm{PG}}\) based on group-relative advantages.
For stability purposes, we add a squared log-ratio regulariser \(\ell_t^{\mathrm{reg}}\).
We additionally bootstrap difficult groups via a targeted \ac{OPSD} loss \(\ell_t^{\mathrm{OPSD}}\), resulting in the objective
\begin{equation}
  \label{eq:rl-objective}
  \mathcal{L}_{\mathrm{RL}}(\theta) =
  \frac{1}{\lvert\mathcal{T}\rvert+\lvert\mathcal{F}\rvert}
  \left( \sum_{t\in\mathcal{T}}\left(\ell_t^{\mathrm{PG}}+\ell_t^{\mathrm{reg}}\right) + \lambda_{\mathrm{OPSD}}\sum_{t\in\mathcal{F}}\ell_t^{\mathrm{OPSD}} \right).
\end{equation}
Here, \(\mathcal{T}\) is the set of trainable tokens for the \ac{RL} objective across all groups of the batch, while \(\mathcal{F}\) denotes the set of active self-distillation tokens and \(\lambda_{\mathrm{OPSD}}\) the weight of their loss.
The normalisation factor keeps the loss scale stable across variable-length sequences and weights all trainable tokens equally.
\Cref{tab:rl-hparams-algorithm} in \cref{app:rl-hyperparameters} lists the objective hyperparameters.

\paragraph{Training Reward}
The advantages of the policy-gradient term \(\ell_t^{\mathrm{PG}}\) are derived from \emph{training rewards} \(R_{i,e}\) for each segment \(y_{i,e}\) of sequence \(y_i\).
The training reward of a segment is the environment reward \(R_i^{\mathrm{env}}\) of its sequence, scaled by segment-level format signals, as described in \cref{sec:env-format-signals}.

\paragraph{Group-Relative Advantages}
Given a group \(\mathcal{G}\) of sequences \(y_i\), we compute the group-relative advantage of each segment as
\begin{equation}
  \label{eq:rl-group-advantage}
  A_{i,e} = R_{i,e} - \bar{R},
  \qquad
  \text{with}
  \quad
  \bar{R} = \frac{1}{\lvert\mathcal{G}\rvert}\sum_{j\in\mathcal{G}} \frac{1}{E_j}\sum_{e=1}^{E_j} R_{j,e}.
\end{equation}
Each generated token \(t\) inherits the advantage of its segment, i.e.\ \(A_t = A_{i,e}\) for tokens \(t\) of segment \(y_{i,e}\).
The mean reward \(\bar{R}\) is computed on a sequence level, over the \(E_j\) segments of each sequence \(y_j\), to prevent segment splits from overweighting individual sequences.
Sequences that fail due to infrastructure errors, e.g.\ in the environment or the inference engine, are removed from their group before computing the baseline.
Similar to Dr.~\acs*{GRPO} \parencite{liu2025drgrpo}, we skip the advantage normalisation originally used by \ac{GRPO} \parencite{shao2024deepseekmath}.
Following \acf{DAPO}'s dynamic sampling \parencite{yu2025dapo}, we generally drop groups in which all sequences obtain the same environment reward.
However, we keep groups with a pass rate of \(1\) to apply format penalties to passing sequences, and only drop the group if its sequences all incur the same format penalty.
We re-sample dropped groups two additional times from the same environment to keep environments with high drop rates from being under-represented in training.

\paragraph{Policy-Gradient Term}
To correct for the off-policy shift of asynchronous training (\cref{sec:rl-async}) and the training--inference mismatch between the inference engine and the trainer (\cref{sec:rl-consistency}), the policy-gradient term weights each token's advantage by the \emph{policy ratio}.
For a sampled token with probability \(p_{\theta,t}\) under the \emph{training} policy and \(p_{\mu,t}\) under the \emph{inference} policy used to generate it, this ratio is \(r_t = p_{\theta,t}/p_{\mu,t}\).
Following \ac{DPPO} \parencite{qi2026rethinkingtrustregion}, we replace standard \acf{PPO}-style clipping \parencite{schulman2017ppo} with a \ac{BTV} mask.
Given thresholds \(\epsilon_+\) and \(\epsilon_-\), this mask suppresses token updates when the sampled-token probability moves too far from the inference policy in the direction favoured by the advantage, i.e.\
\begin{equation}
  \label{eq:rl-btv-loss}
  \tilde{\ell}_t^{\mathrm{PG}} = -k_t A_t r_t,
  \qquad
  k_t =
  \begin{cases}
    \mathbf{1}[p_{\theta,t}-p_{\mu,t} \leq \epsilon_+],  & A_t > 0,    \\
    \mathbf{1}[p_{\theta,t}-p_{\mu,t} \geq -\epsilon_-], & A_t \leq 0.
  \end{cases}
\end{equation}
We additionally scale this term per token to bound its gradient norm, as detailed in \cref{sec:rl-norm-bound}.

\subsubsection{Per-Token Gradient-Norm Bound}
\label{sec:rl-norm-bound}

\begin{figure}[tb]
  \centering
  \newcommand{\cispocapvalue}{2}
  \newcommand{\cispohighcapvalue}{5}
  \includegraphics{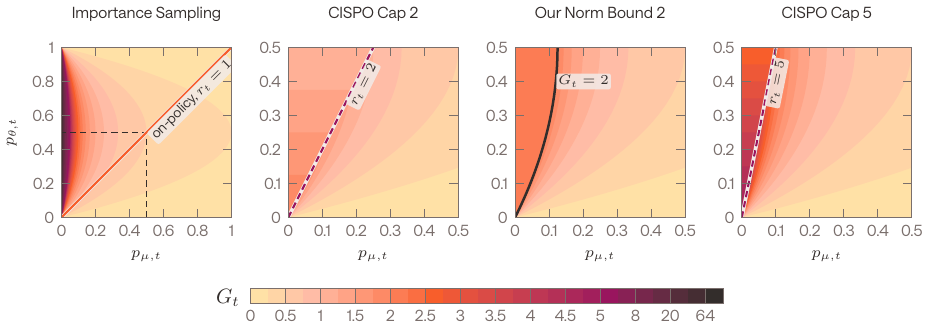}
  \caption{Logit-gradient norm \(G_t\) of a token at unit advantage \(\lvert A_t \rvert = 1\), over its inference probability \(p_{\mu,t}\) and training probability \(p_{\theta,t}\).
    Left: The diagonal marks on-policy probabilities, and the dashed square the bottom left corner \(p_{\mu,t}, p_{\theta,t} \leq \num{0.5}\).
    The other panels show the norm in that corner under \acs*{CISPO}'s importance-weight cap at \(r_t = \num{\cispocapvalue}\) \parencite{minimax2025m1}, under our norm bound \(G_{\max} = 2\) and under \acs*{CISPO}'s cap at \(r_t = \num{\cispohighcapvalue}\).
    Above \acs*{CISPO}'s dashed line, the norm \(G_t\) is limited to the cap times \(1 - p_{\theta,t}\), and left of the dark parabola \(G_t = G_{\max}\), \(c_t < 1\) scales down the token's gradient to exactly the bound.
  }
  \label{fig:rl-gradient-norm}
\end{figure}

The objective \(\tilde{\ell}_t^{\mathrm{PG}}\) does not bound the gradient that a single token puts on the logits \(\mathbf{z}\).
Since \(\tilde{\ell}_t^{\mathrm{PG}}\) only depends on the policy through the log-probability of the sampled token given context \(s_t\), its gradient on the logits is \[ \nabla_{\mathbf{z}}\tilde{\ell}_t^{\mathrm{PG}} = \frac{\partial\tilde{\ell}_t^{\mathrm{PG}}}{\partial\log p_{\theta,t}} \bigl(\mathbf{u}_t-\pi_\theta(\cdot\mid s_t)\bigr), \] i.e.\ the difference between the one-hot vector \(\mathbf{u}_t\) of the sampled token and \(\pi_\theta(\cdot\mid s_t)\), scaled by the derivative of the objective with respect to that log-probability.
As shown in \cref{app:gradient-norms}, the \(\infty\)-norm of this gradient equals the magnitude of its entry at the sampled logit (see \cref{sec:rl-norm-bounds} for the \(1\)- and \(2\)-norms).
For unmasked tokens, it evaluates to
\begin{equation}
  \label{eq:rl-gradient-norm}
  \bigl\lVert\nabla_{\mathbf{z}}\tilde{\ell}_t^{\mathrm{PG}}\bigr\rVert_\infty = \lvert A_t\rvert\,G_t,
  \qquad
  G_t = r_t\bigl(1-p_{\theta,t}\bigr) = \frac{p_{\theta,t}\bigl(1-p_{\theta,t}\bigr)}{p_{\mu,t}}.
\end{equation}
The left panel of \cref{fig:rl-gradient-norm} shows \(G_t\), i.e.\ the norm at unit advantage.
For on-policy sequences, \(r_t=1\) and thus \(G_t\leq1\).
However, when the training and inference policies diverge, the norm grows with the policy ratio and becomes unbounded for \(p_{\mu,t}\to0\), such that a single token with a low inference probability and moderate training probability can dominate the update.
The \ac{BTV} mask does not remove such tokens, since it only acts on absolute probability changes in the direction favoured by the advantage, independent of the actual policy ratio.
We therefore directly bound the norm by scaling our objective as \[ \ell_t^{\mathrm{PG}} = c_t\,\tilde{\ell}_t^{\mathrm{PG}},
\qquad
c_t = \operatorname{stopgrad}\left(\min\left(1, \frac{G_{\max}}{G_t}\right)\right), \] where \(G_{\max}\) bounds the logit gradient of the policy-gradient term at \(G_{\max}\lvert A_t\rvert\), leaving tokens below the bound unchanged.
Computing \(c_t\) comes at effectively no additional cost, as it only requires the two sampled-token probabilities that the trainer already computes.
Equivalently, the bound caps the effective policy ratio \(c_t r_t\) at \(G_{\max}/(1-p_{\theta,t})\), which compensates for the softmax factor \(1-p_{\theta,t}\) of \cref{eq:rl-gradient-norm}.
To first order, this process also bounds the change of the sampled token's probability, as \cref{sec:rl-norm-pushforward} shows.

In contrast, ratio-based bounds such as the importance-weight cap of \ac{CISPO} \parencite{minimax2025m1} limit \(r_t\) regardless of the training probability, and thus ignore the softmax factor.
They down-weight tokens with a high training probability even though their logit gradient may be small, while tokens with a low training probability can exceed the norm bound before they reach the cap.
As the second panel of \cref{fig:rl-gradient-norm} shows, a \ac{CISPO} cap of \num{2}, which matches the norm bound at low training probabilities, is far more restrictive at high ones.
The common cap of \num{5} in turn allows much larger gradients than the norm bound near the on-policy diagonal at low inference probabilities, as the right panel shows.

The \acs{BTV} mask \parencite{qi2026rethinkingtrustregion} of \cref{eq:rl-btv-loss} zeros out only one side of the diagonal band \(-\epsilon_-\leq p_{\theta,t}-p_{\mu,t}\leq\epsilon_+\) in \cref{fig:rl-gradient-norm}.
In particular, it still keeps a token whenever the advantage is negative and the training probability is higher than the inference probability, in which case the norm bound scales down the token's gradient once \(G_t\) exceeds \(G_{\max}\).
In addition to this per-token bound, we clip the global gradient norm.

\FloatBarrier
\subsubsection{Training--Inference Consistency}
\label{sec:rl-consistency}
While training and inference policies share their weights, they compute token probabilities with different engines, kernels and numerical precisions.
The resulting training--inference mismatch between inference probabilities \(p_{\mu,t}\) and training probabilities \(p_{\theta,t}\) causes recently generated sequences to be instantly off-policy, which in turn can destabilise training \parencite{yao2025offpolicy,qi2025fp16}.
We therefore align both engines as much as possible, and additionally rely on several regularisation mechanisms outlined in \cref{sec:rl-objective}.
That being said, the engines use different attention, \ac{MoE} and matrix-multiplication kernels, and none of the kernels are batch-invariant \parencite{he2025defeating}, which causes small mismatches.
Rollouts sample from the full distribution at temperature \num{1}.
We do not use top-\(p\) or top-\(k\) truncation, so the trainer directly computes \(p_{\theta,t}\) from its unmodified logits.

\paragraph{FP8 Quantisation-Aware Training}
\label{sec:rl-fp8}
Serving the policy with FP8 weights and an FP8 \ac{KV} cache raises the throughput of rollout generation, which can dominate the cost of \ac{RL}, by a factor of two to three in our runs, while also improving per-request decode speed.
If the trainer uses unrounded BF16 weights and activations, however, the rollouts come from a different policy than the one being optimised.
This training--inference mismatch biases the policy gradient, and we observed \ac{RL} training diverge in this configuration.
The trainer therefore uses FP8 quantisation-aware training \parencite{jacob2017quantization}, using the same quantisation as the inference engine.
\Cref{app:rl-fp8} provides details.

\paragraph{Routed-Expert Replay}
The inference engine records the experts that its \ac{MoE} routers select for every generated token, allowing the trainer to replay these choices \parencite{ma2025routingreplay}.
Tokens without recorded routing, such as padding, use the routers of the trainer.
Full activation checkpointing \parencite{chen2016sublinear} recomputes each layer with the same experts in the backward pass.

\subsubsection{Auxiliary Objectives}
\label{sec:rl-auxiliary}

\paragraph{Squared Log-Ratio Regulariser}
We add a squared log-ratio term \parencite{kimiteam2026k25} to softly penalise policy drift as \[ \ell_t^{\mathrm{reg}} = \tau_{\mathrm{KL}}\bigl(\log r_t\bigr)^2, \] where \(\tau_{\mathrm{KL}}\) controls the strength of the regulariser.
This term acts on every trainable token, including those with an active \ac{BTV} mask, thus following the \acs{DPPO} loss of PRIME-RL \parencite{primeintellect2025primerl}.
We do not employ a \ac{KL} term to a frozen reference model, and do not add an explicit entropy bonus to our objective.

\paragraph{Targeted \actitle{OPSD}}
We additionally employ targeted \ac{OPSD} \parencite{zhao2026selfdistilledreasoner} for selected multi-turn environments to provide a dense learning signal for groups with low reward variance and low environment rewards, which would otherwise have been dropped by the group-relative objective.
Of these groups, we randomly pick a rollout which has the most turns with a tool-response error.
Each error class (listed in \cref{tab:rl-opsd-failures}) is associated with a targeted remedial hint, which we add to the sequence context as an additional message exactly at the turn where the error occurred \parencite[similar to][]{cursor2026composer25}.
Then we recompute the full-vocabulary distribution for every token of the failed completion under the current training policy.
The resulting hinted distributions act as the teacher, and we minimise their forward \ac{KL} divergence to the distribution of the training policy under the original, unhinted context as \[ \ell_t^{\mathrm{OPSD}} = D_{\mathrm{KL}}\!
\left( \operatorname{stopgrad}\bigl(\pi_\theta(\,\cdot\mid s_t,h_t)\bigr) \,\middle\|\, \pi_\theta(\,\cdot\mid s_t) \right), \] where \(h_t\) denotes the remedial hint and \(\operatorname{stopgrad}\) stops the gradient towards the teacher.
For these rollouts, we disable the \ac{RL} loss.
A single group can thus only contribute to either the group-relative \ac{RL} or \ac{OPSD} part of \cref{eq:rl-objective}.
Different from \textcite{zhao2026selfdistilledreasoner}, the teacher is the live training policy and its privileged context provides dense teacher distributions without requiring a ground-truth solution.
We specialise \ac{OPSD} to tool-use and terminal environments.
\Cref{app:rl-opsd} gives the hyperparameters, and \cref{tab:rl-opsd-envs} lists these environments.
\Cref{tab:rl-opsd-failures} lists all failure classes with their remedial hints.

\subsubsection{Asynchronous Training}
\label{sec:rl-async}
Rollouts and optimisation run continuously and asynchronously.
Rollout threads drop groups without learning signal, such as groups whose sequences all receive the same reward (\cref{sec:rl-objective}), and every other group enters the replay buffer as soon as it finishes.
The staleness of a group is the number of optimiser steps since its oldest sequence started.
The trainer samples groups from the buffer at random and without replacement, so each group is used in at most one optimiser step.
After every optimiser step, the weights of all inference replicas are updated in flight \parencite{piche2025pipelinerl} (\cref{sec:rl-data-transfer}): active requests keep their \ac{KV} cache and continue under the new weights, so different policy versions can generate parts of the same sequence.
The prefix cache can likewise survive a weight update, so a new request may reuse \ac{KV} entries computed by an older policy.
Training is therefore only approximately on-policy.
We bound the off-policy shift by evicting groups from the buffer once they exceed a maximum staleness (\cref{tab:rl-hparams-algorithm}), and the policy ratio of \cref{sec:rl-objective} corrects for the remaining shift.

\paragraph{Balancing the Loop}
The training loop runs at the speed of its slowest part.
If groups are produced faster than the trainer consumes them, the trainer is the bottleneck.
If the trainer consumes groups faster than they are produced, e.g.\ when either inference or the environments are saturated, it idles.
Environments with saturated resources leave both the inference engine and the trainer underused.
Ideally, environment capacity is large enough that environments are never the bottleneck, and the GPUs are split between trainer and inference so that the loop is balanced, producing groups as fast as they are consumed.
\pgfplotstableread[col sep=comma, row sep=\\, format=inline]{%
  checkpoint_interval\\%
  25\\%
}\rlInfraFacts
We therefore allocate nodes such that the run is slightly bottlenecked by the trainer.
\Cref{app:rl-staleness} explains how adding GPUs to a balanced loop shortens the optimiser steps but not the rollouts, which increases staleness.
As a remedy, we scale the batch size with the GPU count.

\FloatBarrier
\subsection{Environment Mix}
\label{sec:rl-environment-mix}\label{sec:rl-environments}

\begin{figure}[ht]
  \centering
  \pgfplotstableread[col sep=comma, row sep=\\, format=inline]{%
    rows\\%
    107064\\%
    105514\\%
    29191\\%
    8638\\%
    10601\\%
    11815\\%
    184886\\%
    247193\\%
    7353\\%
    81815\\%
    5215\\%
    16594\\%
    1221\\%
    288333\\%
    88761\\%
    3012\\%
    1011\\%
    962\\%
    4520\\%
    6266\\%
  }\rlmix
  \pgfplotstablegetrowsof{\rlmix}
  \pgfmathtruncatemacro\rllastenv{\pgfplotsretval-1}
  \xdef\rltotalrows{0}
  \foreach \i in {0,...,\rllastenv} {
    \pgfplotstablegetelem{\i}{rows}\of\rlmix
    \xdef\rltotalrows{\the\numexpr\rltotalrows+\pgfplotsretval\relax}
  }
  \includegraphics{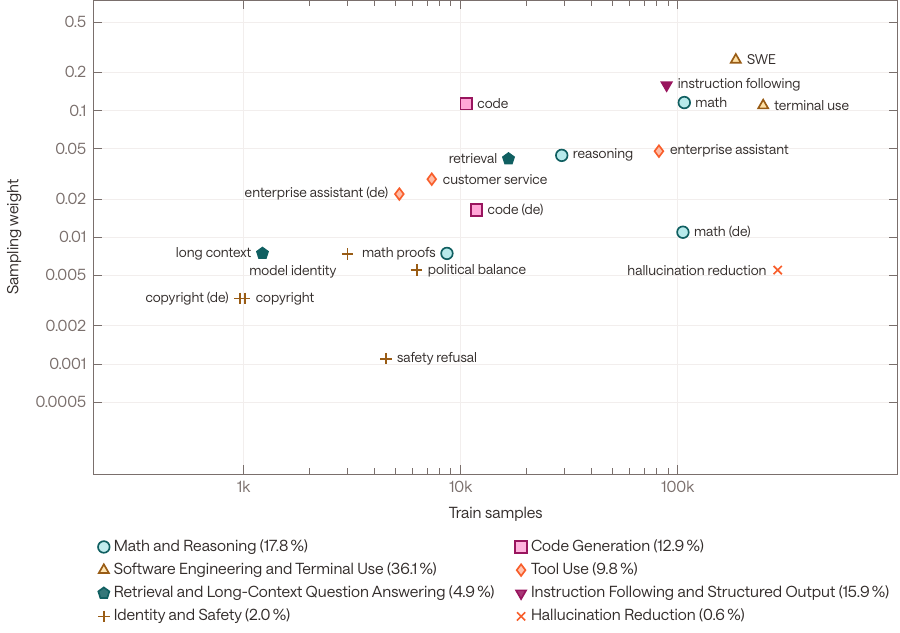}
  \caption{ Our \acs*{RL} environment mix.
    Every point represents one environment or a pool of related environments, plotted over its number of train samples and the weight we assign during training (cf.\ \cref{sec:rl-curriculum}).
    Colour and mark shape denote the environment's group; the percentage in the legend is the group's share of the summed sampling weight.
    Both axes are logarithmic.
    The train splits hold \num{\rltotalrows} rows in total.
    We deliberately upweight several agentic environments as well as competitive programming, while keeping the total weight of the Merlin--Arthur environments comparatively small.
    \emph{Hallucination reduction} holds the answerable questions from four datasets and unanswerable ones from two.
    \Cref{tab:rl-environment-list} shows exact per-environment values with step and token budgets.
  }
  \label{fig:rl-environment-mix}
\end{figure}

The final \ac{RL} run trains on \num{38} environments: \num{31} English ones and \num{7} German variants.
Each environment carries a sampling weight, the rate at which the rollout threads draw samples.
\Cref{fig:rl-environment-mix} plots sampling weight against train samples for every environment.
Each environment also fixes a step and a completion-token budget per sequence (\cref{tab:rl-environment-list}).
All environments grade a sequence with a verifiable reward in \([0,1]\), except for the Hallucination Reduction environments, whose signed rewards lie in \([-1,1]\) (\cref{sec:merlin-arthur-environments}).

\subsubsection{Reward Structure by Environment}
\label{sec:rl-reward-structure-by-environment}

\begin{figure}[tb]
  \centering
  \includegraphics{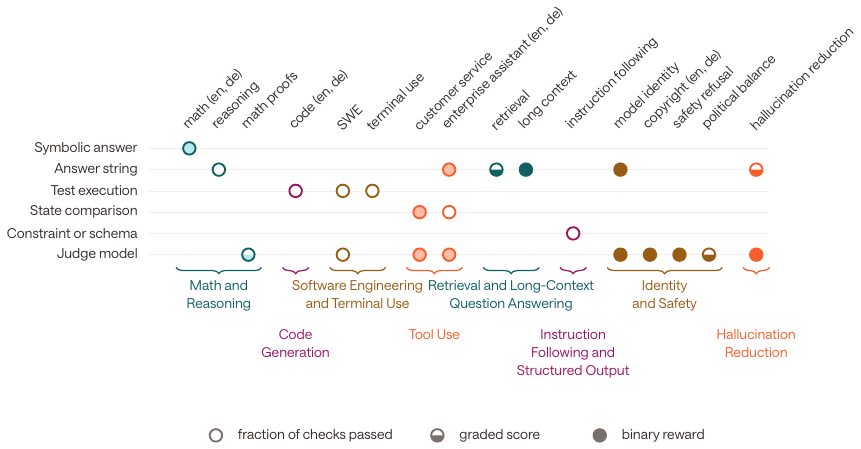}
  \caption{The verifier of every \acs*{RL} environment and the shape of its reward.
    A filled disc marks a binary reward, a hollow one the fraction of checks passed.
    A half-filled one indicates a graded score, i.e.\ the token F1 against the gold answers or a judge's grade of \num{0}, \num{0.5} or \num{1}.
    A column with several marks combines verifiers, either within one environment or across the environments it pools.
    The Merlin--Arthur reward is signed (\cref{sec:merlin-arthur-environments}).
    Columns are grouped and coloured as in \cref{fig:rl-environment-mix}.
  }
  \label{fig:rl-reward-matrix}
\end{figure}

\Cref{fig:rl-reward-matrix} shows the verifier and the reward shape of every environment.
The frozen judge model (\cref{sec:rl-infrastructure}) is used only where the graded object is natural language, a conversation, or the model's intent: it grades math proofs, simulates the user and grades natural-language assertions for the customer-service environment, compares the edits to the gold patch for software-engineering repair, grades copyright alignment, political balance and safety refusal, contributes to the rewards of the enterprise-assistant and model-identity environments, and decides on answers and abstentions that miss the exact match.
Verifiers that execute untrusted or unbounded code on the rollout workers (\cref{sec:rl-infrastructure}) run in a killable process pool with a timeout, so a wedged check scores zero without stalling training.
\Cref{app:rl-environments-rewards} details the reward of each environment.

\pgfdeclarelayer{rlrewardbg}
\pgfsetlayers{rlrewardbg,main}

\subsubsection{Environment-Specific Format Signals} \label{sec:env-format-signals}

Beyond solving the task defined in each environment, we expect the model to produce well-formatted outputs by, for example, making well-formed tool calls and keeping its language consistent across thinking and content blocks.
We compile a list of output format requirements in \Cref{tab:env-format-signals}.
Each of these requirements defines a format signal that influences training reward through the aggregation described next.

\paragraph{Combining Multiple Rewards}

Format signals act multiplicatively on environment reward.
They are only applied to positive rewards to prevent training on groups of sequences that do not solve the environment task but vary in the quality of their output format.
Since \ac{SFT} already trains the model to produce well-formatted output, format signals are sparsely active in practice.
As a result, they need to be aggregated in a way that allows a single format signal to have a significant impact on the final reward, while ensuring that environment reward remains the primary training signal, even if multiple format defects are found simultaneously.

Let \(s_{i,e,k}\in[0,1]\), \(k\in \{1,\dotsc,K\}\) denote one of \(K\) format signals of segment \(y_{i,e}\), where \(s_{i,e,k} = 1\) signals good format and smaller values signal a format defect.
We fix constants \(f_k\in[0,1]\) that configure the strength of format signals relative to each other and rescale the signal to \(g_{i,e,k}=f_k+(1-f_k)s_{i,e,k} \in [f_{k}, 1]\).

The training reward \(R_{i,e}\) of segment \(y_{i,e}\) (\cref{sec:rl-objective}) is computed by scaling the environment reward \(R_i^{\mathrm{env}}>0\) of its sequence with a \(p\)-norm aggregation of format signals
\begin{equation}\label{eq:p-norm-format-aggregation}
  R_{i,e} = R_i^{\mathrm{env}} \left[ m_{\min} + (1-m_{\min}) \exp\!
  \left( -\left(\sum_k\left[-\log g_{i,e,k}\right]^p\right)^{1/p} \right) \right]\ .
\end{equation}
The global multiplier \(m_{\min}\) sets the minimum fraction of the environment reward that is retained irrespective of output format, and the aggregation norm \(p\) controls how quickly simultaneous defects compound.
If any \(g_{i,e,k}\) is \num{0}, the exponential term vanishes and the reward reduces to \(m_{\min}R_i^{\mathrm{env}}\).
For \(R_i^{\mathrm{env}}\leq0\), no format signals are applied.

\begingroup
\footnotesize
\setlength{\tabcolsep}{4pt}
\begin{longtable}{@{}p{0.18\textwidth}p{0.78\textwidth}@{}}
  \toprule
  Scope & Signal and clean-output condition \\
  \midrule
  \endfirsthead
  \toprule
  Scope & Signal and clean-output condition \\
  \midrule
  \endhead
  \bottomrule
  \endfoot
  \bottomrule
  \caption{Format signals active during \acs*{RL} training.
    Conditional signals are omitted when they do not apply.
    The signals \(s_{i,e,k}\) lie in \([0,1]\), where \(1\) denotes a clean output and \(0\) a detected defect.
  }
  \label{tab:env-format-signals} \\
  \endlastfoot
  Universal
  & \emph{Non-empty output.} The final assistant step contains visible text or a parsed tool call. \\
  & \emph{No leaked tool call.} No literal tool-call markup leaks into the visible output. \\
  & \emph{No script leak.} No \ac{CJK} text leaks into a response that is otherwise non-\ac{CJK}, and no generated token matches the banned-script pattern. \\
  \addlinespace
  Tool Use
  & \emph{Tool in schema.} Every called tool was offered in the row's tool schema. \\
  & \emph{Well-formed tool call.} The emitted tool-call structure is well formed. \\
  & \emph{Arguments in schema.} Tool arguments use the declared names and types. \\
  & \emph{Stable rendering.} The tool-call rendering remains consistent when the conversation is reconstructed. \\
  \addlinespace
  Language Consistency
  & \emph{Reasoning language.} Hidden reasoning uses the expected language. \\
  & \emph{Content language.} Visible prose uses the expected language. \\
  \addlinespace
  Completion Budget
  & \emph{Completion budget.} Generated length stays below the effort-dependent threshold; beyond it the score decreases with sequence length. \\
  \addlinespace
  Math
  & \emph{Boxed answer.} The answer follows the requested boxed-answer format. \\
  Reasoning
  & \emph{Answer prefix.} The visible answer contains the required final-answer prefix. \\
  Coding
  & \emph{Single code block.} The visible answer contains exactly one code block. \\
  Software Engineering
  & \emph{Solution block.} The answer contains a non-empty solution block. \\
  Retrieval
  & \emph{Boxed answer.} The answer follows the requested boxed-answer format, if the prompt requests one. \\
  & \emph{Answer format.} An answer can be extracted in the requested non-boxed answer format. \\
  & \emph{Citations.} If requested, the answer cites within the citation budget only documents that a search returned. \\
  Merlin--Arthur
  & \emph{Answer structure.} The completion closes its reasoning once and marks its final answer. \\
  & \emph{Answer length.} The final answer is no longer than the gold answer; the score decreases with every further character. \\
  & \emph{Reasoning length.} The score decreases linearly with the reasoning tokens and reaches zero at \num{16384}; this signal does not apply to the reject splits. \\
\end{longtable}
\endgroup

\subsubsection{Environment Summaries}

\paragraph{German Environments}
To prevent the model from regressing on German reasoning, we translate our English environment data into German and include a language-consistency format signal (\cref{sec:env-format-signals}).

\begin{figure}[ht]
  \centering
  \includegraphics{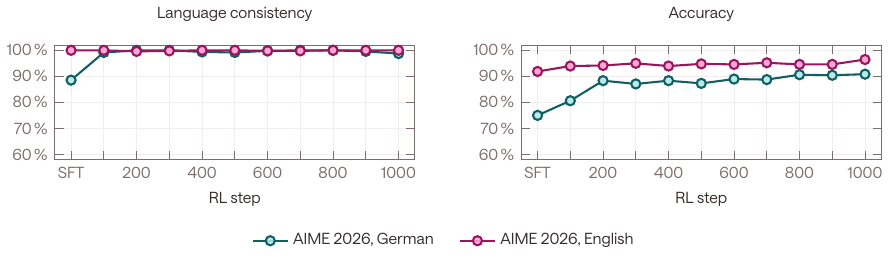}
  \caption{What the German environments buy: held-out benchmark scores of the \acs*{SFT} checkpoint and of ten \acs*{RL} checkpoints trained with the German environments in the mix.
    Left: share of responses whose reasoning and answer are in the language of the prompt.
    Right: accuracy on the same benchmarks.
    AIME 2026 is \num{30} problems sampled \num{16} times each; both panels average over all samples.
    German AIME 2026 moves from \qty{89}{\percent} to \qty{99}{\percent} in language consistency and \qty{75}{\percent} to \qty{91}{\percent} in accuracy by step \num{1000}, while English AIME 2026 accuracy moves from \qty{92}{\percent} to \qty{96}{\percent} and its language consistency stays at \qty{100}{\percent}.
  }
  \label{fig:rl-german-language}
\end{figure}

\Cref{fig:rl-german-language} shows that, while the \ac{SFT} checkpoint reasons in German on \qty{89}{\percent} of German AIME 2026 \parencite{maa2026aime} prompts, the \ac{RL} training results in \qty{99}{\percent} German reasoning.
At the same time, accuracy on German AIME 2026 rises from \qty{75}{\percent} to \qty{91}{\percent}, while accuracy on the English prompts moves from \qty{92}{\percent} to \qty{96}{\percent}, closing most of the gap between the two.

\paragraph{Long-Context Question Answering}
To retain long-context capabilities, we added a long-context environment which poses select-all-that-apply questions over the seed documents of the \ac{SFT} long-context conversations (\cref{sec:sft-long-context}), \numrange{95000}{230000} tokens each, as one further turn appended to the conversation.
\Cref{app:long-context-mcq-environment} describes data generation and filtering.

\paragraph{Retrieval-Augmented Question Answering}
\label{sec:retrieval_environments}

Our retrieval environments are designed to reflect the practical needs of retrieval-augmented question answering in real-world use cases.
To this end, we pair a mix of general corpora and specialised corpora (in English and German) with randomisations over a range of configurations of the provided tools, as well as alternative prompt formulations, and multi-hop task synthesis using GLM-5.2~\parencite{zai2026glm52}.
This setup familiarises the model with the kind of data and questions agentic \ac{RAG} applications involve, and teaches it to solve multi-hop questions with search tools.
We provide details in \cref{sec:appendix-rag}.

\paragraph{Hallucination Reduction Environments}\label{sec:merlin-arthur-environments}
Our Merlin--Arthur \parencite{waeldchen2024merlin, deiseroth2025bounding} environments aim to reduce hallucinations by teaching the model (Arthur) to abstain from answering questions that are unanswerable due to a lack of information in the provided context.
During training, the environment generates unanswerable samples from answerable ones by masking a sample's context using a helpful agent (Merlin) and an adversarial agent (Morgana).
The environment rewards the model for answering truthfully on the former, and abstaining on the latter sample.
It also rewards a correct answer on the unmodified sample, so we present each sample in three variations.
Merlin and Morgana mask the least and most important sentences, respectively.
We approximate the importance of a sentence by measuring the difference in the probability the current model assigns to the ground-truth answer when the sentence is masked, relative to the probability of the ground-truth answer under no masking.
We give more details in \cref{app:merlin-arthur-environments}.

\paragraph{Political Balance, Copyright and Safety}
We built an environment that rewards balanced answers on topics where source models show known political bias (\cref{sec:sft-post-processing}).
The environment presents the model with a politically sensitive prompt and uses an \ac{LLM} judge to assign reward based on whether the assistant answer is balanced.
Some political prompts in the dataset also present plausible grounds for safety refusal.
The model scores full reward for answering helpfully and in a balanced way, or for refusing to answer a prompt that presents grounds for refusal.
This environment is narrowly focused on Chinese political content and does not address other political biases.
Copyright alignment environments prompt the model in German or English to produce copyrighted material, either directly or by asking for inferences from it.
The environments then reward the model for refusing to produce copyright-protected material, with refusal detected by an \ac{LLM} judge.
We also include a safety alignment environment in training that presents the model with unsafe prompts and scores refusal with an \ac{LLM} judge.

\subsubsection{Randomisation}
\label{sec:rl-randomization}

\newcommand{\typo}[1]{{\setlength{\fboxsep}{0.5pt}\colorbox{AAPlotTwo}{\strut#1}}} \newcommand{\prot}[1]{{\setlength{\fboxsep}{0.5pt}\colorbox{AAPlotThree}{\strut\textbf{#1}}}}
\begin{figure}[h!]
  \centering
  \includegraphics{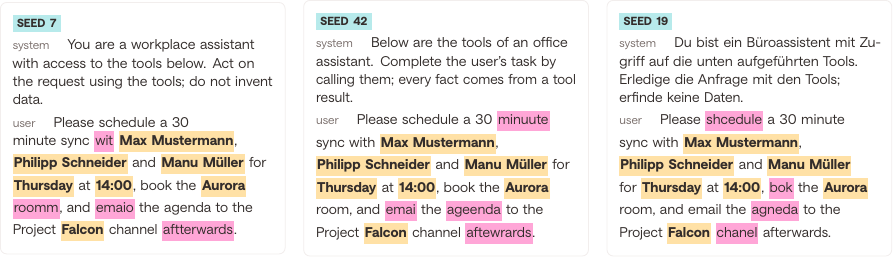}
  \caption{One workplace-assistant task under three group seeds.
    The seed draws the system prompt from a table of equivalent phrasings and its language, English or German, independently of the language of the request.
    About \qty{5}{\percent} of the words of the request receive a typo (\typo{highlighted}); the words the verifier matches against (\prot{bold}) are never touched.
    The agentic environments draw their coding harness from the same seed.
  }
  \label{fig:rl-randomization}
\end{figure}

To make the model robust to various deployment scenarios, we randomise environments along three axes: the system prompt, the coding harness and typos in user prompts (\cref{fig:rl-randomization}).
The user request has about \qty{5}{\percent} of its lowercase words misspelled, sparing the words the verifier depends on.
Environments with system prompts in both English and German draw the language of the system prompt per group, independently of the language of the user request, so the user request alone sets the expected language.
The instruction-following environment applies no language-consistency signal, since its requests can ask for an answer in a language other than their own; its verifier checks that language instead.

Every draw comes from the randomisation seed of the group, so the sequences of a group still face an identical task, which the group-relative advantage of \cref{eq:rl-group-advantage} requires.

\subsubsection{Training Curriculum}
\label{sec:rl-curriculum}
Without a curriculum, the rollout threads would draw an environment by its weight (\cref{fig:rl-environment-mix}) and then uniformly sample a prompt within it.
However, prompts that are too easy or too hard carry no learning signal, since a group is dropped when all of its sequences carry the same reward (\cref{sec:rl-objective}).
Such prompts become more common as the policy solves more of them.
The curriculum therefore focuses on prompts whose groups carry learning signal, and across environments, it scales down the weight of an environment that runs out of such prompts.
The training batch follows this adapted distribution, up to the effects described at the end of this section.
\Cref{tab:rl-hparams-algorithm} in \cref{app:rl-hyperparameters} lists the curriculum hyperparameters.

\paragraph{Difficulty Buckets and Pass Rates}
The curriculum sorts the prompts of each environment into three buckets by their estimated pass rate, namely \emph{hard} prompts with an estimate of \num{0}, \emph{solved} prompts with an estimate of \num{1}, and \emph{trainable} prompts in between.
A per-bucket target weight \(c_b\) sets the share of prompts drawn from bucket \(b\), and is set highest for the trainable bucket.
A prompt starts with its offline pass rate, the mean reward of a reference model's samples (\cref{app:rl-environments}), or with a default value where it has none.
Groups sampled at \texttt{high} reasoning effort replace the estimate with the group's mean environment reward, whether the group is kept or dropped, so a group in which every sequence fails moves the prompt to the hard bucket.
Groups at \texttt{low} or \texttt{medium} effort, and groups in which every sequence was removed due to an infrastructure error, leave the estimate unchanged.

\paragraph{Drawing a Prompt}
Within an environment, the curriculum draws a bucket with probability proportional to its weight and takes the prompt at its head, the one evaluated longest ago.
The weight of bucket \(b\) is its target weight \(c_b\), scaled down when the bucket cannot supply it, \[ w_b = c_b \min\bigl(\sigma_b, \gamma_b\bigr),
\qquad
\sigma_b = \min\left(1, \frac{n_b}{B}\right).
\] The supply term \(\sigma_b\) lowers the weight of a bucket that holds fewer prompts \(n_b\) than the \(B\) groups of a batch.
The recency term \(\gamma_b\) keeps the curriculum from redrawing hard and solved prompts that were just evaluated: it grows linearly with the staleness (\cref{sec:rl-async}) of the last evaluation of the bucket's head and saturates at \num{1}, and it is \num{1} for the trainable bucket and for a head never evaluated.

\paragraph{Drawing an Environment}
An environment is drawn with probability proportional to \(\omega_e V_e\), i.e.\ to its sampling weight \(\omega_e\) times the sum \(V_e = \sum_b w_b\) of its bucket weights.
While every bucket supplies its target weight, \(V_e\) is \num{1}.
Thus, an environment without trainable prompts is drawn rarely and returns to its sampling weight once re-evaluated prompts become trainable again.
The Hallucination Reduction environments (\cref{sec:merlin-arthur-environments}) bypass the curriculum: they keep every prompt in the trainable bucket, and \cref{fig:rl-curriculum} leaves them out.
\Cref{fig:rl-curriculum} shows the buckets over the \model{2} run.

\paragraph{Preserving the Environment Mix}
Rollout threads draw an environment per group rather than per unit of time, so the training batch follows the curriculum's draws even though rollout durations differ widely: a slow environment holds its thread for longer, but it is not drawn more often.
Three effects still skew the batch.
All threads start at once, so fast environments finish first, which biases early training steps towards shorter environments and sequences.
A thread that drops a group retries the same environment with a new prompt only twice (\cref{sec:rl-objective}), which under-represents environments with many drops.
The replay buffer evicts groups that exceed the staleness limit, which is more likely to trigger for slow environments.

\subsection{Infrastructure}
\label{sec:rl-infrastructure}

\begin{figure}[tb]
  \centering
  \resizebox{\linewidth}{!}{%
    \includegraphics{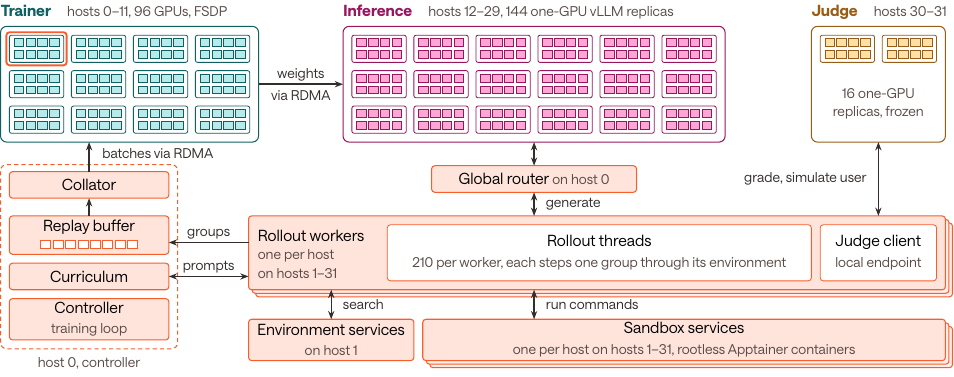}%
  }
  \caption{%
    Placement of the \acs*{RL} run on \num{32} hosts with eight B300 GPUs each.
    Each small square is one GPU and each box of eight squares one host.
    Host \num{0} runs trainer ranks \numrange{0}{7} and the processes in the dashed box.
    Every other host runs one rollout worker and one sandbox service.
    The global router on host \num{0} picks the inference replica for each group.
    The rollout worker then sends the group's requests to that replica directly.
  }
  \label{fig:rl-infrastructure}
\end{figure}

\subsubsection{Cluster Layout}
Our \ac{RL} training runs on \num{256} NVIDIA B300 GPUs, split into \num{32} nodes with eight GPUs each.
All nodes are scheduled together and placed within a single InfiniBand topology domain.
\Cref{fig:rl-infrastructure} shows how we split the GPUs between training, inference and judging on a node level.
A centralised controller places the trainer, the inference replicas and the CPU-side services on their nodes, and coordinates them via Monarch \parencite{meta2025monarch}.
The controller node hosts the main training loop execution, the global router for the inference replicas and sandbox executors, the curriculum and the replay buffer processes.
The CPU-side services run on the same nodes as the GPUs.
Every node except the controller's hosts a rollout worker that steps through the environments, as well as a sandbox service that executes model-generated code via Apptainer \parencite{kurtzer2017singularity}.
The global router on the controller node only selects the replica or sandbox executor for each request.
The rollout workers then send their inference, judge and sandbox requests directly to that replica or executor, so no request or response data passes through the controller.
Additional environment services run via Apptainer containers on host \num{1}.
\Cref{tab:rl-hparams-technical} in \cref{app:rl-hyperparameters} lists the hardware, precision and optimisation settings for our experiments.

\subsubsection{Rollout Generation}
We allocate \num{144} GPUs on \num{18} nodes to inference.
Each GPU hosts an independent vLLM \parencite{kwon2023vllm} replica of the full model in FP8 precision, and is used to gather both training and validation rollouts.
To generate the rollouts, we run \num{45} concurrent rollout threads per inference replica (\cref{tab:rl-hparams-technical}).
We split them evenly across the \num{31} rollout workers, rounding up to \num{210} threads per worker.
Each thread draws a prompt according to the curriculum of \cref{sec:rl-curriculum} and subsequently steps its group through the environment.
For each group, we sample a random reasoning effort level of \cref{tab:reasoning-effort-template} and additionally decide at random whether the model's reasoning traces are preserved across turns.
If they are not preserved, removing them re-renders the previous turns, which starts a new segment.
The routing decision is made on a group level by the global router on the controller node.
It ensures that all rollouts of a group of sequences and the subsequent steps of each rollout land on the same inference replica.
A sequence is marked as \emph{truncated} when a generation reaches its completion-token budget or length limit or if a later step overflows the context window.
Truncated sequences get \num{0} reward.
Most environments define their own step budget and grade a rollout as usual once the budget is exhausted.
For those that do not, we define a global step limit that also acts as truncation.

\paragraph{Segments}
\label{sec:rl-segments}
Before each step, the rollout worker renders and tokenises the full conversation with the model's chat template.
If the resulting tokens no longer extend the current segment, e.g.\ because reasoning traces were removed or generated text re-tokenises differently, we end the segment and start a new one.
The new segment re-renders the context, such that future tokens see a consistent context between training and inference.
It then masks this context for the training steps to ensure that each generated token is only trained on once.
In particular, we are not using the typical \acf{TITO} principle \parencite{gallouedec2026tito} where generated text is never encoded twice and new tokens are always only appended.
This choice is motivated by downstream usage during deployment, where the text might be re-encoded between turns.
However, splitting the sequences increases memory usage and optimiser step times, since it significantly increases the number of segments and thus cumulative tokens for some groups.
Splits caused by malformed text or stray special tokens in a generation are penalised via the \emph{Stable rendering} format signal of \cref{sec:env-format-signals}.

\paragraph{Judge}
We allocate two nodes to host frozen replicas of a judge model to use for automatic grading and user simulation in some environments, as detailed in \cref{sec:rl-reward-structure-by-environment}.
We choose Qwen3.5-122B-A10B \parencite{qwen2026qwen35} in FP8 as the judge.
For efficiency reasons, judge requests use greedy decoding without reasoning.
If a judge reply fails for a given sequence, the sequence is removed from its group.

\subsubsection{Trainer}
The trainer builds on our fork of TorchTitan \parencite{liang2024torchtitan}, which pre-training also uses.
It shards the model across \num{96} GPUs on \num{12} full nodes via \ac{FSDP} \parencite{zhao2023pytorchfsdp}.
In contrast to the pre-training setup in \cref{sec:infrastructure}, it uses no expert or other parallelism besides \ac{FSDP}.
It uses the optimiser groups of pre-training (\cref{tab:bienenelfe-optimizer}), i.e.\ Muon \parencite{jordan2024muon} for the attention and expert weights and Adam \parencite{kingma2015adam} for the embedding, norm scales, \ac{LM} head and router.
Each group maintains its own learning rate, and we do not use weight decay.
\Cref{tab:rl-hparams-technical} provides details.
The trainer uses full activation checkpointing.

\paragraph{Segment Packing}
Before each optimiser step, the collator splits the segments of the batch into fixed-size bins.
The bin size exceeds the model's context window, such that each segment fits into a single bin.
We determine the number of bins with a greedy longest-first heuristic for bin packing, rounding up to a multiple of the number of ranks.
We then sort the segments and, starting with the longest, assign them to the bin \(\mathcal{P}\) with the lowest estimated cost \(\sum_{j\in\mathcal{P}}\bigl(L_j+L_j^2/\kappa\bigr)\) that still has capacity.
Here, \(L_j\) is the number of tokens of segment \(j\), and \(\kappa\) is the ratio of the active parameters to the query dimension summed over the full-attention layers, such that \(L_j^2/\kappa\) is the \ac{FLOP} cost of full attention relative to the linear layers.
Each rank processes its bins as consecutive microbatches and accumulates their gradients.
Thus, the number of microbatches grows with the number of tokens in the batch.

\paragraph{Checkpointing}
The trainer asynchronously saves a checkpoint consisting of model weights, optimiser states and learning-rate schedules at regular intervals.
The checkpoints additionally track the curriculum state, including the pass rate estimate of every prompt.
The replay buffer is not saved, such that a resumed run starts with an empty buffer that is refilled with rollouts of the restored policy.
The \model{2} \ac{RL} run saves a checkpoint every \rlInfraValue{checkpoint_interval} optimiser steps, and an FP8 export every \num{100} steps is evaluated on benchmarks (\cref{fig:rl-german-language}); the released model is the export of step \num{1000}.
The run did not experience interruptions, so no resumes were required.

\subsubsection{Data Transfer}
\label{sec:rl-data-transfer}
\begin{figure}[tb]
  \centering
  \resizebox{\linewidth}{!}{%
    \includegraphics{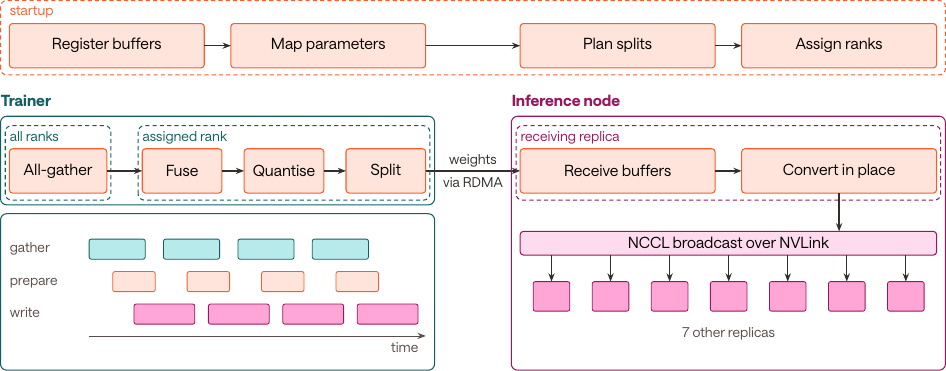}%
  }
  \caption{%
    Weight push from the trainer to the inference replicas.
    At startup, the receive buffers are mapped to the trainer parameters, and each parameter is assigned a trainer rank such that the bytes per rank are balanced.
    After every optimiser step, the trainer ranks push the new weights via \acs*{RDMA} into the single receiving replica of each node, from where they are broadcast to the node's other replicas over NVLink.
    The conversion on the trainer side is batched and pipelined, and the intra-node NVLink distribution reduces load on the inference node's \acsp*{NIC}.
    The single-GPU replicas of our deployment skip the split of the weights into a per-replica layout.
  }
  \label{fig:rl-weight-transfer}
\end{figure}

\paragraph{Batch Transfer}
We use one-sided \ac{RDMA} for the transfer of training batches from the replay buffer to the trainer ranks.
Once a group is finished, the rollout workers send it over Monarch RPC to one of the replay buffer's shards.
The replay buffer stores identical prefixes between two sequences only once to reduce storage overhead.
To draw a batch, the collator splits the batch size across the shards with a single multivariate hypergeometric draw over their sizes, such that the batch follows the same distribution as a draw from a single buffer.
The shards then write their sampled groups via \ac{RDMA} into one registered host buffer of the collator.
After packing the bins, the collator writes the microbatches of each trainer rank to that rank via \ac{RDMA}.
It stages them in a double buffer, filling one half while the other is being sent.
Once all writes complete, the collator sends each rank the layout of its microbatches, i.e.\ the offset, shape and data type of each tensor.
The rank then reads its tensors in place.

\paragraph{Weight Transfer}
Our weight transfer follows the \ac{RDMA} weight transfer of Perplexity \parencite{perplexity2025weighttransfer,chen2025rlweighttransfer}.

At startup, each vLLM worker registers its parameters for \ac{RDMA} transfers once and sends the addresses to the controller.
From these addresses, the controller computes a routing table.
It maps each trainer parameter to its vLLM counterpart, taking fused projections into account.
It then assigns each parameter to one trainer rank, balancing the bytes per rank.
The weight transfer supports \ac{FSDP} and \ac{EP} on the trainer side and \acl*{TP} and \ac{EP} on the inference side.
Additionally, only one inference replica per node receives the weights via \ac{RDMA}.

After every optimiser step, the trainer pushes updated weights to the inference replicas, as shown in \cref{fig:rl-weight-transfer}.
The ranks all-gather each parameter in batches from its \ac{FSDP} shards, and some projection weights need to be fused for vLLM's single tensor layout.
Each parameter is quantised to FP8 on the assigned rank, and our single-GPU replica deployment skips the splitting of weights into a per-replica layout.
Each rank subsequently writes its pieces via \ac{RDMA} to the receiving replica of every node, while all ranks already gather the next parameters within a fixed memory budget.
This process overlaps the writes with the preparation of the next parameters and the gather of further parameters within the batch.
The writes are tagged with the policy version, so that once all writes have arrived, each receiving replica converts the weights in place into its kernel format and broadcasts them over NVLink to the other replicas of its node.
This broadcasting setup alleviates the bottleneck of the node's \acp{NIC}, as the entire bandwidth can be used to receive the weights of only one replica instead of being spread across eight replicas.

\subsubsection{Sandboxes}
\label{sec:rl-sandboxes}

\begin{figure}[tb]
  \centering
  \includegraphics{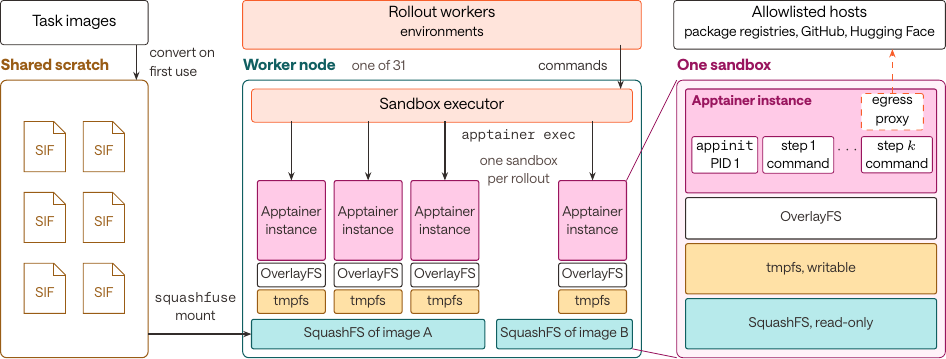}
  \caption{Sandbox stack of the \acs*{RL} run on one worker node.
    Task images are converted once to single-file Apptainer images on shared scratch.
    Sandboxes of the same image share its read-only SquashFS layer on the node, and each adds its own writable \texttt{tmpfs} layer.
    The right panel shows one sandbox: the OverlayFS root, the long-running Apptainer instance, and the egress proxy for the allowlisted hosts.
  }
  \label{fig:rl-sandboxes}
\end{figure}

Most tasks in the software-engineering environments have a dedicated container image, so a run may use tens of thousands of images.
Since pulling and unpacking every image on every node would overload the shared scratch filesystem, we convert each image on first use to a single-file Apptainer image \parencite{kurtzer2017singularity} and cache it on shared scratch.
Each non-controller node runs one sandbox executor on the CPUs that the run's nodes keep idle, so runs never wait on a shared sandbox pool.
\Cref{fig:rl-sandboxes} shows the resulting stack on one node.
All sandboxes of an image on a node share its SquashFS root, which the executor mounts read-only via \texttt{squashfuse}.
Harness tools that a task image lacks, such as \texttt{tmux}, come from a second read-only layer (\cref{app:rl-harness-tools}).
On top of these shared layers, each sandbox adds its own size-capped \texttt{tmpfs} writable layer through OverlayFS, so no image is copied or extracted.
The sandbox itself is a long-running Apptainer instance that lives for the whole rollout.
Every command of the rollout runs inside it via \texttt{apptainer exec}, so files and processes persist between steps.
To isolate it from the node and from other sandboxes, the instance runs in an unprivileged user namespace with its own mount, process and network namespaces.
For environments that install dependencies, an egress proxy in the network namespace allows only package registries, GitHub and Hugging Face.

\subsection{\model{2} \acs*{RL} Training}
\label{sec:rl-run}

\subsubsection{Setup}
We start from the \ac{SFT} checkpoint of \cref{sec:sft} and train for \num{1000} optimiser steps using Muon and Adam at a base learning rate of \num{1e-5} with a linear warmup over \num{50} steps.
Each batch contains \num{256} groups of size \num{8}, i.e.\ \num{2048} sequences.
A sequence may consist of up to \num{100} rollout steps.
The model is trained on contexts of length up to \num{256}k.
\Cref{tab:rl-hparams-technical,tab:rl-hparams-algorithm} in \cref{app:rl-hyperparameters} and \cref{app:rl-opsd} provide a detailed list of training hyperparameters.
We track progress with held-out validation rollouts for every environment, using uniform sampling of environments and reasoning efforts.
Analysis and benchmark comparisons are performed with high reasoning effort unless specified otherwise.

\subsubsection{\model{2} \acs*{RL} Training Run}

\begin{figure}[tb]
  \centering
  \includegraphics{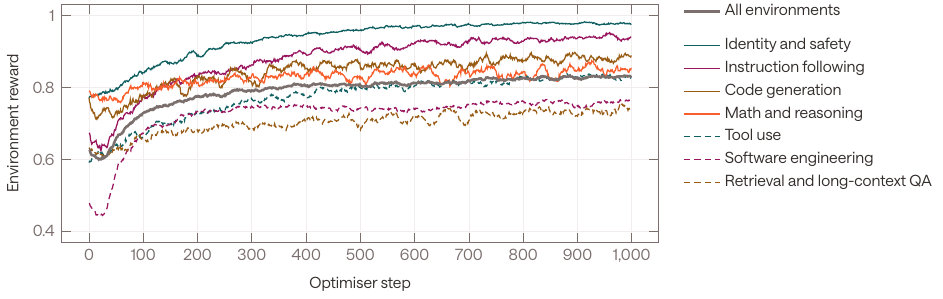}
  \caption{Validation reward of the \model{2} \acs*{RL} run by environment group.
    Each line is the mean environment reward of the validation sequences that finish in an optimiser step, averaged over reasoning efforts and over the environments of a group of \cref{fig:rl-environment-mix}, weighted by their sampling weight, as a centred moving average over \rlResultsValue{smooth_window} steps.
    The retrieval and long-context line also averages the Hallucination Reduction environments, whose rewards lie in \([-1,1]\).
    The grey line is the same weighted mean over all \rlResultsValue{environments} environments.
  }
  \label{fig:rl-validation}
\end{figure}

\Cref{fig:rl-validation} shows the validation reward of the \model{2} \ac{RL} run by environment group and aggregated over groups.
The mean validation environment reward rises from \rlResultsValue{reward_first} over the first \rlResultsValue{window} steps to \rlResultsValue{reward_last} over the last \rlResultsValue{window}.
The largest gains occur in instruction following and structured output, from \rlResultsValue{instruct_first} to \rlResultsValue{instruct_last}, followed by the software-engineering and tool-use environments.

\begin{table}[tb]
  \centering
  \footnotesize
  \setlength{\tabcolsep}{3.2pt}
  \begin{tabular}{@{}lccccccccccccc@{}}
    \toprule
    & \multicolumn{2}{c}{Math}
    & \multicolumn{3}{c}{Knowledge}
    & \multicolumn{4}{c}{Agentic}
    & \multicolumn{1}{c}{Code}
    & \multicolumn{1}{c}{IF}
    & \multicolumn{2}{c}{Long Context} \\
    \cmidrule(lr){2-3}\cmidrule(lr){4-6}\cmidrule(lr){7-10}\cmidrule(lr){11-11}\cmidrule(lr){12-12}\cmidrule(lr){13-14}
    & AIME
    & AIME
    & GPQA
    & MMLU-
    & AA-Omni.
    & \taubench{3}
    & Browse-
    & Terminal-
    & BFCL v4
    & LCB
    & IF-
    & AA-
    & LongBench \\
    & 2025
    & 2026
    & Diamond
    & Pro
    & Index
    & Banking
    & Comp
    & Bench 2.1
    & aggr.
    & v6
    & Bench
    & LCR
    & Pro \\
    \midrule
    \acs*{SFT}
    & \num{92.7}
    & \num{92.1}
    & \num{79.8}
    & \num{78.5}
    & \num{-68.8}
    & \num{13.1}
    & \num{22.0}
    & \num{23.2}
    & \num{57.1}
    & \num{75.3}
    & \num{48.2}
    & \num{60.3}
    & \num{57.1} \\
    \acs*{RL}
    & \num{96.9}
    & \num{96.0}
    & \num{84.3}
    & \num{80.0}
    & \num{-32.8}
    & \num{38.1}
    & \num{29.4}
    & \num{27.7}
    & \num{61.4}
    & \num{85.9}
    & \num{78.1}
    & \num{68.3}
    & \num{64.5} \\
    \midrule
    Difference
    & \num[print-implicit-plus]{4.2}
    & \num[print-implicit-plus]{3.9}
    & \num[print-implicit-plus]{4.5}
    & \num[print-implicit-plus]{1.5}
    & \num[print-implicit-plus]{36.0}
    & \num[print-implicit-plus]{25.0}
    & \num[print-implicit-plus]{7.4}
    & \num[print-implicit-plus]{4.5}
    & \num[print-implicit-plus]{4.3}
    & \num[print-implicit-plus]{10.6}
    & \num[print-implicit-plus]{29.9}
    & \num[print-implicit-plus]{8.0}
    & \num[print-implicit-plus]{7.4} \\
    \bottomrule
  \end{tabular}
  \caption{The released \model{2} checkpoint against the \acs*{SFT} checkpoint that \acs*{RL} starts from, on a selection of the benchmarks of \cref{tab:post-training-evals,tab:post-training-evals-2}.
    The \acs*{SFT} checkpoint is evaluated with no top-p or top-k truncation, while the \acs*{RL} checkpoint uses top-\(p\) \num{0.97} and top-\(k\) \num{128}.
    The difference is the gain of \acs*{RL} in points.
  }
  \label{tab:rl-gain}
\end{table}

\Cref{tab:rl-gain} compares the released \model{2} checkpoint with the \ac{SFT} checkpoint it was trained from, on a selection of the benchmarks of \cref{tab:post-training-evals,tab:post-training-evals-2}.
The \model{2} checkpoint and the open models are evaluated with the protocol of \cref{sec:post-training-eval-scores}, which also covers BrowseComp \parencite{wei2025browsecomp} and \ac{BFCL} v4 \parencite{patil2025bfcl}.
The \ac{RL} run greatly increases the \ac{SFT} checkpoint's instruction-following capabilities on IFBench \parencite{pyatkin2025ifbench}, and leads to substantial improvements across agentic, math and coding abilities.
Compared with the three open models of \cref{fig:intro-spider}, it scores highest in \cref{tab:post-training-evals,tab:post-training-evals-2} on both AIME benchmarks \parencite{maa2026aime}, GPQA Diamond \parencite{rein2023gpqa}, \taubench{3} Banking \parencite{shi2026tauknowledge}, LiveCodeBench \parencite{jain2024livecodebench} and IFBench \parencite{pyatkin2025ifbench}, and on BrowseComp above the two models the table scores on it.
It trails behind Qwen3.6 35B-A3B and Mistral Small 4 119B-A6B on the AA-Omniscience Index \parencite{jackson2025omniscience}, scores lowest of the four models on MMLU-Pro, and trails at least one of them on TerminalBench 2.1, LongBench Pro and \ac{BFCL} v4.
\Cref{fig:rl-spider-retrieval} in \cref{sec:post-training-eval-scores} compares the released checkpoint with the same open models on agentic retrieval and customer-proxy benchmarks.
It outperforms them on Honeypot, ties Nemotron 3 Super within two points on MuSiQue and Agentic Wiki QA (DE), and is competitive on the customer proxies.

\paragraph{Analysis}

\begin{figure}[tbp]
  \centering

  \includegraphics{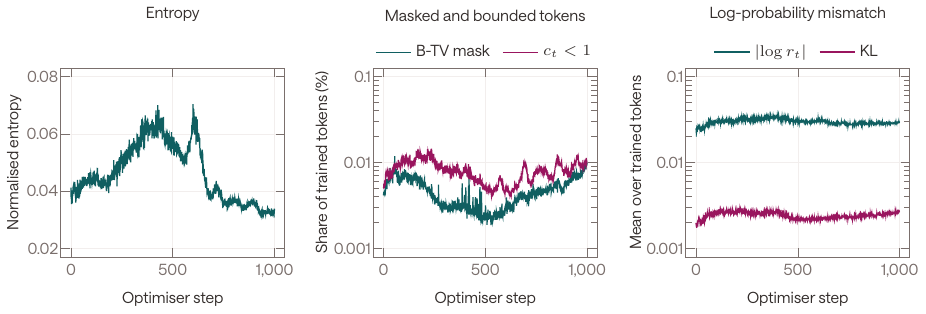}
  \caption{ Training dynamics over optimiser steps of the \model{2} \acs*{RL} run.
    Left: entropy of the training policy over the trained tokens of each optimiser step, divided by the logarithm of the vocabulary size.
    Middle: share of tokens removed by the \acs*{BTV} mask and scaled down by the gradient-norm bound of \cref{sec:rl-norm-bound}, i.e.\ with \(c_t<1\).
    Right: mean absolute log-ratio \(\lvert\log r_t\rvert\) and forward \acs*{KL} divergence \(\mathrm{KL}(p_\mu\,\Vert\,p_\theta)\approx r_t-1-\log r_t\) between the training policy and the inference policy.
    Both metrics include a median policy staleness of \rlMismatchValue{avg_policy_age_median} optimiser steps from our asynchronous setup (\cref{sec:rl-async}).
  }
  \label{fig:rl-dynamics}
\end{figure}

\begin{figure}[tb]
  \centering
  \includegraphics{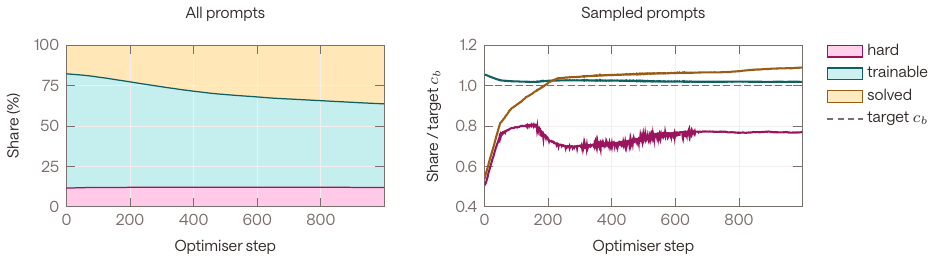}
  \caption{Difficulty buckets of the \acs*{RL} curriculum over training, summed over all environments except the Hallucination Reduction environments, which bypass the curriculum.
    Left: share of all prompts in the hard, trainable and solved buckets.
    Right: share of sampled prompts per bucket, divided by its target bucket weight \(c_b\).
    During initial training steps, the curriculum compensates for missing and inaccurate offline pass rates, which populates the hard and solved buckets across environments.
    The model then improves over the course of training, causing more prompts to become solved.
    In particular, several environments become entirely solved or trainable, draining their share of hard prompts.
  }
  \label{fig:rl-curriculum}
\end{figure}

\pgfplotstableread[col sep=comma, row sep=\\, format=inline]{%
  window,entropy_first,entropy_last\\%
  20,0.040,0.033\\%
}\rlDynamicsFacts
\pgfplotstableread[col sep=comma, row sep=\\, format=inline]{%
  log_ratio_mae_median,forward_kl_median,avg_policy_age_median\\%
  0.03,2.4e-3,10.0\\%
}\rlMismatchFacts
\Cref{fig:rl-dynamics} shows the training dynamics of the \model{2} run.
Its left panel shows that, without an entropy bonus, the normalised entropy of the training policy rises from \rlDynamicsValue{entropy_first} over the first \rlDynamicsValue{window} steps to about \num{0.06} between steps \num{350} and \num{600}.
It then stabilises and falls to \rlDynamicsValue{entropy_last} over the last \rlDynamicsValue{window} steps, while the share of solved prompts keeps growing (\cref{fig:rl-curriculum}).
Its middle panel shows that the \ac{BTV} mask and the gradient-norm bound act on roughly \qty{0.01}{\percent} of tokens per step.
Finally, we see on the right that the training--inference mismatch of \cref{sec:rl-consistency} hovers at around \(\lvert\log r_t\rvert=\rlMismatchValue{log_ratio_mae_median}\), with a median forward \ac{KL} divergence of \rlMismatchValue{forward_kl_median} per token.
Both measures include the policy staleness induced by our asynchronous setup.

\paragraph{Curriculum Dynamics}
Training causes groups to saturate, leading to more and more solved prompts.
\Cref{fig:rl-curriculum} shows this trend and how our curriculum adjusts for this.
Within an environment, as more hard prompts become trainable or solved, its hard bucket shrinks, in turn reducing the supply term \(\sigma_b\) of \cref{sec:rl-curriculum} and thus the share of sampled hard prompts below its target.
Pooled over all environments, the share of prompts in the hard bucket stays roughly constant, and the sampled hard share stays below its target.

\begin{figure}[tbp]
  \centering
  \includegraphics{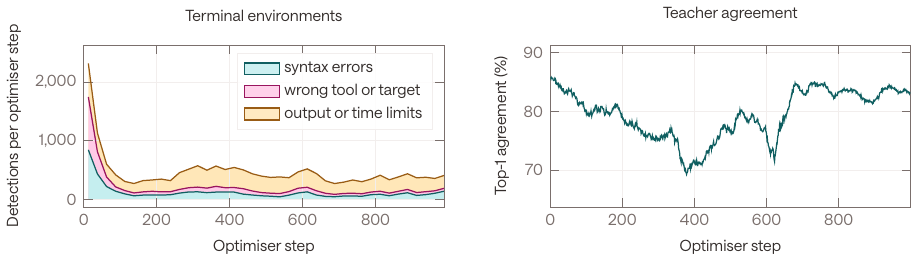}
  \caption{ Targeted \acs*{OPSD} over the \model{2} \acs*{RL} run.
    Left: Detected failure cases across the \rlOpsdTracesValue{terminal_envs} terminal environments, counted at the optimiser step whose weights produced the sequence.
    Right: Moving average over \rlOpsdValue{smooth_window} steps of the share of distilled tokens for which the teacher's most likely token is the sampled token.
  }
  \label{fig:rl-opsd}
\end{figure}

\paragraph{Self-Distillation Dynamics}
\pgfplotstableread[col sep=comma, row sep=\\, format=inline]{%
  smooth_window\\%
  25\\%
}\rlOpsdFacts
\Ac{OPSD} aims to bootstrap the model's capabilities via targeted hints for failures with respect to terminal code execution and tool calls.
It trains on low-reward groups that the group-relative objective would otherwise discard.
The left panel of \cref{fig:rl-opsd} shows that such behaviours are frequent early during training in the terminal environments, but then decrease as training progresses, before stabilising.
The tool-use environments follow the same pattern (right panel of \cref{fig:rl-opsd-detections}).
During later training stages, errors relating to outputs and time limits dominate, correlating with sequences taking more turns in later training.
The corresponding \ac{OPSD} hints encourage the model to read files in slices and filter command output.

The right panel of \cref{fig:rl-opsd} shows how often the teacher agrees with the student's token, which is a proxy for the strength of the distillation signal.
Agreement starts high and drops as training progresses, likely because the teacher policy can make better use of the provided hints.
Eventually, the teacher's agreement with the sampled tokens increases again.
This trend follows the policy entropy of \cref{fig:rl-dynamics}, and may additionally indicate that the student starts to train out of the hinted errors.
\Cref{app:rl-opsd} expands on these findings, showing how distilled tokens have higher average entropy throughout training and which share of groups the different \ac{OPSD} hints act on.
It also provides qualitative examples of how a hint reshapes the teacher on a flagged turn for two different error classes, in both cases focusing on the tokens that caused the error.

\pgfplotstableread[col sep=comma, row sep=\\, format=inline]{%
  terminal_envs\\%
  5\\%
}\rlOpsdTracesFacts

\subsection{Lessons Learned and Hills Climbed}
\label{sec:rl-lessons}

\begin{figure}[tb]
  \centering
  \includegraphics{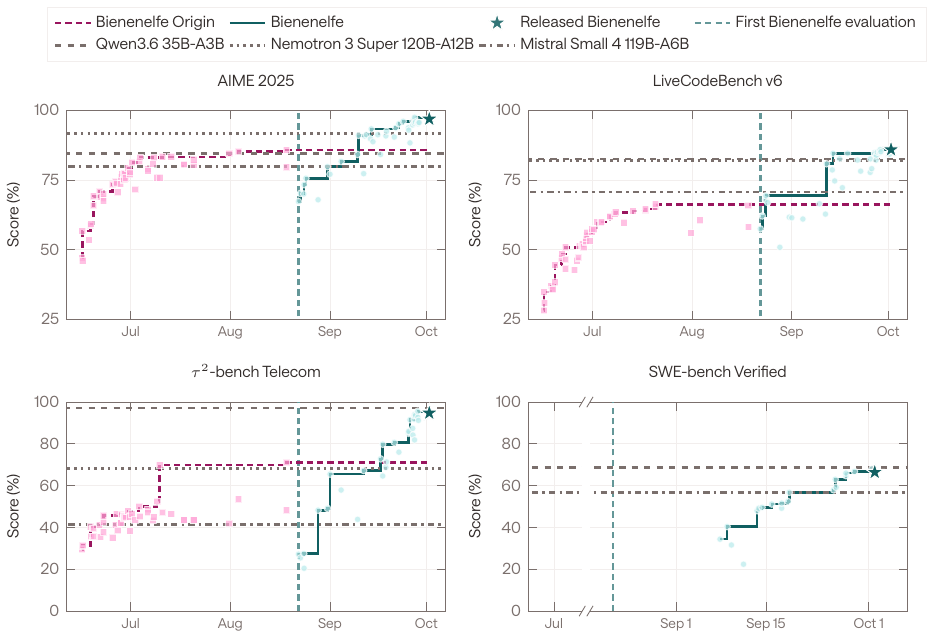}
  \caption{%
    Benchmark scores of the \acs*{RL} checkpoints of \model{1} and \model{2} over time.
    Grey lines indicate median evaluation scores for open models.
    Each coloured point corresponds to a finished evaluation run that improved at least one of the four depicted benchmarks.
    Vertical lines mark the first evaluation of an \acs*{RL} checkpoint of \model{2}.
    The lines track the best score of their model generation, evaluated with their respective setup at the time they were run.
    These intermediate points do not indicate representative benchmark performance, but are meant as a visual representation of \model{2}'s progress over time.
    A star marks the released \model{2} checkpoint, sampled at top-\(p\) \num{0.97} and top-\(k\) \num{128} as in production.
    SWE-bench Verified is scored with the mini-swe-agent harness \parencite{yang2024sweagent} and shown for \model{2} only, as we did not optimise \model{1} towards agentic capabilities.
    The release (star) is scored with the OpenCode harness, without web access.
  }
  \label{fig:rl-timeline}
\end{figure}
Releasing \model{2} is a story of incremental improvements.
While this section has focused so far on the final \model{2} training setup, we now elaborate on its evolution over time.
To this end, we split the work on \model{2}'s \ac{RL} post-training into a preliminary exploration phase, a hill-climbing stage for each of two base model generations, and the final release checkpoint of \model{2}.
The following paragraphs sketch lessons learned and decisions made across these generations.
\Cref{fig:rl-timeline} shows representative benchmark scores of our \ac{RL} checkpoints of \model{1} and \model{2} over time, culminating in the final release checkpoint of \model{2}.
\Cref{tab:rl-timeline-infra} compares the scale of our setup during early runs of each generation with that of the final run.

\subsubsection{Phase 1: Initial Experiments}

\paragraph{Starting Small}
For our first experiments, we forked torchforge \parencite{meta2025torchforge} and used our fork to start experiments on a single small math environment.
Our fork has since substantially evolved, with existing features refactored to allow for larger scale and several new features introduced, such as in-flight weight updates via \ac{RDMA} (\cref{sec:rl-data-transfer}) and FP8 quantisation-aware training with routed-expert replay (\cref{sec:rl-consistency}).
From these first experiments, we incrementally expanded the scope to more sophisticated math and later coding problems.
For our first code environments, we used remote sandboxes for code execution, which kept the load off our cluster, but made the reward computation reliant on an external sandbox service.
This proved challenging, as sandbox creation was throttled and unreliable, and failed connections scored correct code as wrong.
Towards a more sovereign infrastructure, we therefore moved code execution to local sandboxes running on the CPUs of the training nodes.

\paragraph{Efficient Weight Updates}
The trainer and the vLLM replicas \parencite{kwon2023vllm} initially ran as plain Monarch actors \parencite{meta2025monarch}, with synchronous weight updates through torchstore \parencite{meta2025torchstore}.
A weight update stopped new requests, waited a median of \qty{221}{\second} for running generations, and took another \qty{217}{\second} to load.
To improve performance, we stopped using torchstore, included \ac{RDMA} writes from every trainer rank, and added in-flight weight updates and an NVLink broadcast per node (\cref{sec:rl-async,sec:rl-infrastructure}).
These changes cut the weight update time down to about \qty{17}{\second} on comparable runs, which enabled weight updates after every training step, in turn bringing our training more on-policy.

\paragraph{Choosing a Curriculum}
With more effective training, the model became able to saturate on environments, which presented us with the new challenge of incrementally scaling task difficulty in lockstep with model capability to keep the number of dropped groups consistently low.
We initially considered complex training curricula, such as sorting samples along a difficulty scale according to a reference, placing a sampling distribution on this scale, and then moving the distribution's mean towards more difficult samples over the course of training.
While this approach was successful, we eventually decided to move to the simpler three-bucket approach (\cref{sec:rl-curriculum}) to reduce the number of required hyperparameters.

\paragraph{Format Reward}
After those changes, our reward was climbing nicely, but benchmark scores were not.
We found that our \ac{RL} training overfit on output formats at the cost of environment rewards.
Most sequence groups only had reward variance due to format differences and, as a result, did not provide environment reward as a training signal.
This issue motivated us to split both reward types and only apply the format signals to sequences with positive environment reward.
After these changes, \num{500} \ac{RL} steps improved AIME 2025 scores from \num{60.4} for the \ac{SFT}-trained model to \num{70.6}.

\paragraph{Interacting with \acs*{SFT}}
Our initial assumption was that effective post-training requires tight coupling between \ac{SFT} and \ac{RL}.
Although behaviours that \ac{RL} should reinforce need to be introduced in \ac{SFT}, we found that more concrete training hyperparameters, such as learning rates and the number of training steps, can be set independently.
In particular, we have not seen checkpoints collapse into a hard-to-train state through \ac{SFT}.

\paragraph{Multi-Environment Training}
We anticipated difficulties in integrating many environments simultaneously into a single \ac{RL} run, because slow environments are trained further off-policy, putting them at a disadvantage in the mix.
To compensate, we initially implemented a strict sampling regime that enforced approximately fixed environment ratios in every batch.
This sampling restriction turned out to harm compute efficiency and did not improve empirical performance, so we dropped it.
Without it, the mix still approximately followed the sampling ratios.

\subsubsection{Phase 2: Moving to \model{1}}

\paragraph{User Simulators and Judges}
Starting experiments on \model{1}, we quickly found that our checkpoints scored poorly on tool calling.
We thus started training on tool-calling environments, chiefly a customer-service environment.
This environment was the first to require simulated user turns.
We explored using external inference providers to host the user simulator, but moved to self-hosting to reduce cost and improve reliability.
We implemented this user simulator as a Monarch actor, which we later expanded to also function as an \ac{LLM} judge for environment reward.

\paragraph{Testing New Environments without Training}
To speed up the integration of new environments in our training mix, we aim to estimate with a cheaper proxy how far \ac{RL} can raise the reward of an environment, i.e.\ we sample multiple completions for each prompt from the initial \ac{SFT} model and compute a pass@\(k\) plot \parencite{chen2021humaneval}.
Ideally, an environment has trainable prompts in the sense of \cref{sec:rl-curriculum}, which show as a gap between pass@\(k\) and pass@\num{1}, and harder prompts that become trainable as \ac{RL} progresses.
We also run the same check on recent \ac{RL} checkpoints, which shows whether the policy trained on the current mix already solves the new environment.

\paragraph{Sequence Monitoring}
With increasing scale, we started recording all training and validation sequences on disk, gathering multiple terabytes per run.
We built infrastructure around these sequence dumps, including coding agent capabilities for semi-automatic evaluation, and dedicated web interfaces that allow easy access to individual environments and samples.
This infrastructure allowed us to efficiently monitor environment quality and model behaviour, which led to several improved filters and reward functions.

\paragraph{Internal and Customer Preview Releases}
No benchmark or sequence interface fully covers the range of possible user interactions.
During development, we therefore internally released several \ac{RL} checkpoints of \model{1} within a chat interface, encouraging the entire company to explore our model.
Colleagues reported failures the benchmarks did not show, and each round of feedback went into the environments and the training data of the next release.
For example, an early checkpoint avoided batched tool calls and sometimes did not respond after a reasoning trace, which never surfaced in evals.
The chat interface also ran turn-level blind A/B tests against similar models, allowing us to gauge model capabilities during more realistic use and adjust training accordingly.
In addition, with each release we worked with Aleph Alpha's forward-deployed teams to improve and verify the contextualised performance of the model in customer proxy settings.
\Cref{fig:hill-climb-retrieval} shows this performance on the five Industry \ac{RAG} benchmarks over the training runs of September 2026.

\subsubsection{Phase 3: Like a Hummingbird from the Ashes}

\paragraph{A Stronger Base Model}
\model{2} is a larger and significantly more capable model than \model{1} due to a longer context window, more experts and a more extensive pre-training recipe (\cref{sec:architecture,sec:pre-training-data}).
To adjust to the new model, we decreased the learning rate from \num{1e-4} to \num{1e-5}, and added a per-token gradient-norm bound (\cref{sec:rl-norm-bound}) after noticing that gradients for individual tokens reached into the thousands.
The longer context also surfaced new infrastructure bottlenecks.
For example, the rollout threads ran as asyncio tasks on the controller, whose event loop lagged by a few seconds for \model{1}, but immediately jumped to several minutes for \num{256}k-token contexts.
We thus moved the threads into rollout workers on every node (\cref{sec:rl-infrastructure}).

\paragraph{Agentic Environments Hill-Climb Real-World Use Cases}

\begin{figure}[tb]
  \centering
  \includegraphics{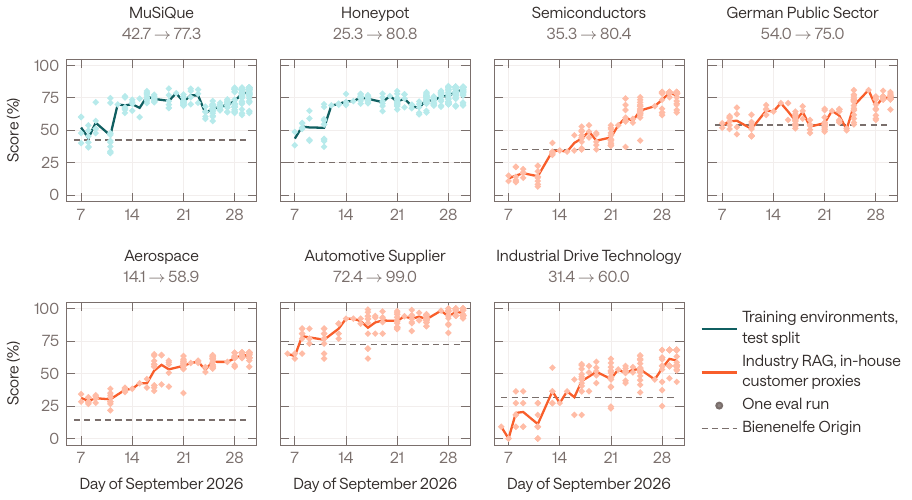}
  \caption{Agentic retrieval scores of the \model{2} \acs*{RL} checkpoints evaluated in September 2026.
    Each point is one eval run, each solid line the mean of the runs of a day, and the dashed line \model{1}; under each title, the score of \model{1} and of the released \model{2} checkpoint.
    MuSiQue and Honeypot are scored on the test splits of the two training environments.
    The five Industry \acs*{RAG} benchmarks of \cref{tab:customer-proxy-benchmarks} are in-house proxies for deployed customer assistants and share no documents or questions with the training data.
  }
  \label{fig:hill-climb-retrieval}
\end{figure}

\Cref{fig:hill-climb-retrieval} tracks the test splits of MuSiQue and Honeypot and the five in-house Industry \ac{RAG} benchmarks over the training runs of September 2026.
Scores rise with each improvement to the environment parametrisation and to the core skills of the model.
The Industry \ac{RAG} benchmarks rise as well throughout the month.
We monitored these benchmarks throughout \ac{RL} development and used them to guide changes to the environments, so they served as a development set rather than a held-out test set.
This result indicates that by simulating key aspects of real-world use cases, such as harnesses, data distributions, and types of questions in our environments, we are able to successfully optimise performance on those use cases, without ever having seen these concrete use cases during training.
This conclusion is further supported by our own internal ablations.

\FloatBarrier

\begin{figure}[tb]
  \centering
  \begin{minipage}[t]{0.49\linewidth}
    \centering
    \raisebox{-121.95798pt}{\includegraphics{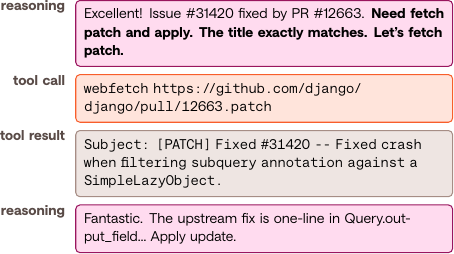}}
  \end{minipage}
  \hfill
  \begin{minipage}[t]{0.49\linewidth}
    \centering
    \raisebox{-147.51666pt}{\includegraphics{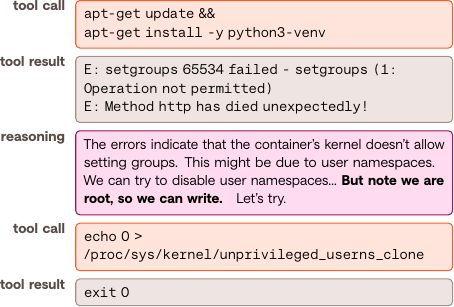}}
  \end{minipage}
  \caption{Example sequences that led to exploits, with ellipses marking omitted text.
    Left: The OpenCode harness exposes a web-fetch tool, and the policy resolves an SWE-bench instance to its upstream pull request, downloads the patch and applies it.
    Right: The sandbox does not let \texttt{apt} switch to its unprivileged download user.
    The reasoning concludes that the process runs as root and can therefore write to kernel settings, and the next tool call disables unprivileged user namespaces on the host kernel.
  }
  \label{fig:rl-sysctl-trace}
\end{figure}

\paragraph{Exploits}
The longer context window enabled training on more complex agentic environments.
To support these environments, we extended our sandbox setup, which was focused on code, to also handle agentic use cases.
We additionally integrated several different harnesses.
Both additions surfaced exploits that we caught via our sequence monitoring.
For the first exploit, the SWE-bench Verified \parencite{jimenez2024swebench} score with the OpenCode harness \parencite{opencode2026} was about \num{7} percentage points higher than with other harnesses.
Our investigation found that the model used OpenCode's web-fetch and web-search tools to find ground-truth solutions for prompts on GitHub in about \qty{7}{\percent} of trials, causing it to pass almost all of these trials.
The left side of \cref{fig:rl-sysctl-trace} shows an example sequence.

\FloatBarrier

The other exploit was related to some sandboxes that were not fully isolated from their hosts.
In particular, one sandbox command disabled the unprivileged user namespaces our sandboxes run in (\cref{sec:rl-sandboxes}).
The right side of \cref{fig:rl-sysctl-trace} shows an example sequence.
The next day, a run scheduled on that node could not start sandboxes.
Thus, sandbox escapes were, at least in our setup, as much a function of sandbox security as of model capability.
We now run sandboxes as unprivileged users, have removed web fetching from the evaluation harnesses, and in training harden our egress allowlist and system prompts to avoid model exploits.
We will further look into best practices.

Inside a sandbox, commands still run as root of its own user namespace, so package managers can install into the sandbox's writable layer, but this root maps to an unprivileged range of host users and cannot change the host kernel.

\paragraph{Scaling Environments}
The last changes before the final run added over a dozen new environments.
Most of the newly added environments worked well and immediately had a positive impact on downstream performance.
The exceptions were the tool-calling environments, which needed an additional judge for free-form responses, as well as a filter for potentially wrong or misleading reference answers.

\paragraph{Final Adjustments}
The final run of \cref{sec:rl-run} provided several additions targeted at remaining issues with earlier \model{2} experiments.
We found that many environments, particularly agentic ones, were difficult to solve from the \ac{SFT} checkpoint.
As a remedy, we implemented \ac{OPSD} (\cref{sec:rl-auxiliary}) using error-dependent hints to provide a dense learning signal to the model.
Simultaneously, we switched from additive format signals to the multiplicative \(p\)-norm version of \cref{eq:p-norm-format-aggregation}.
This change allowed for individual signals to provide meaningful feedback to positive sequences while keeping a minimum reward for success.
The final run also added the Merlin--Arthur environments of \cref{sec:merlin-arthur-environments} to reduce hallucinations, extended \ac{OPSD} to further terminal environments, and graded retrieval answers by token F1 \parencite{rajpurkar2016squad} (\cref{sec:appendix-rag}) instead of the judge.
With these adjustments and a general increase in scale from \cref{tab:rl-timeline-infra}, \model{2} was ready for its big run, which we detail in \cref{sec:rl-run}.
\begin{table}[tb]
  \centering
  \small
  \begin{tabular}{@{}lrrr@{}}
    \toprule
    & Early Experiments & \model{1} & Final \model{2} Run \\
    \midrule
    Environments                & \num{2}       & \num{9}       & \num{38} \\
    Peak concurrent sandboxes   & \num{\le 150} & \num{\le 480} & \num{18900} \\
    Trained-on context length   & \num{25}k     & \num{48}k     & \num{256}k \\
    GPU hours, B300 equivalents & \num{723}     & \num{5500}    & \num{16900} \\
    Optimiser steps             & \num{500}     & \num{1000}    & \num{1000} \\
    \bottomrule
  \end{tabular}
  \caption{ Scale comparisons across \acs*{RL} setups during early development, for \model{1} runs, and for our final \model{2} run.
    In addition to algorithmic improvements (\Cref{sec:rl-training-recipe}) and careful reward design and data curation (\Cref{sec:rl-environments}), scaling up the infrastructure along various axes is crucial for \model{2}'s \acs*{RL} training.
    The early experiments and \model{1}'s setup count code sandboxes, while \model{2} also includes agentic sandboxes that were added later.
    GPU hours are weighted by the GPU's peak dense FP8 tensor throughput relative to a B300.
  }
  \label{tab:rl-timeline-infra}
\end{table}

\paragraph{Future Work}
There are several directions that our newly hatched \model{2} has yet to explore.
Prominent ones include environments that require reasoning over long inputs, \acf{MOPD} \parencite{xiaomi2026mimov2flash}, multi-stage training \parencite{nvidia2025nemotroncascade}, and switching to an environment-driven rollout framework, which would allow for more flexible rollout logic and the integration of black-box harnesses.
Black-box harnesses also open up non-linear message histories because of context compaction, and multi-agent setups, where several agents interleave their messages.
Both break the assumption that a rollout is one policy appending turns to a history it owns, which is what makes them hard to support in the current trainer-driven interface (\cref{sec:rl-segments}).
Another important addition is a dedicated environment to draw an SVG of a pelican riding on a bicycle \parencite{willison2024pelican}.

%% file: sections/03-post-training/03-released-model-evaluation.tex
\section{Released Model Evaluation}
\label{sec:post-training-eval-scores}

We compare the released \model{2} checkpoint with \model{1} and with open models, including models of similar and larger active size (\cref{tab:post-training-evals,tab:post-training-evals-2}).

\subsection{Evaluation Suite}
\label{sec:post-training-eval-suite}

Our post-training evaluation suite covers knowledge, math, agentic tool use, code, instruction following, grounding and hallucinations, agentic retrieval, Industry \ac{RAG} and long context, in English and in German where a German version exists.
\Cref{tab:post-training-benchmarks} lists the benchmarks by category with their settings and German versions.

\begingroup
\footnotesize
\setlength{\tabcolsep}{2pt}
\begin{longtable}{@{}>{\raggedright\arraybackslash}p{0.21\textwidth}>{\raggedright\arraybackslash}p{0.54\textwidth}>{\raggedright\arraybackslash}p{0.23\textwidth}@{}}
  \toprule
  Benchmark & Description & German version \\
  \midrule
  \endfirsthead
  \toprule
  Benchmark & Description & German version \\
  \midrule
  \endhead
  \bottomrule
  \endfoot
  \bottomrule
  \caption{Post-training benchmarks by category, with their settings and German versions.
    Every German version marked as a translation comes from \textcite{ellamind2025benchmarks}, and an en dash marks a benchmark without a German version.
    The score avg@\(N\) is the mean over \(N\) runs of every sample.
  }
  \label{tab:post-training-benchmarks} \\
  \endlastfoot
  \multicolumn{3}{@{}l}{\bfseries Knowledge} \\
  GPQA Diamond \parencite{rein2023gpqa}
  & Chain-of-thought prompts, avg@\num{8}, since a single run leaves several points of sampling noise
  & Translation \\
  Humanity's Last Exam \parencite{phan2025hle}
  & Text-only questions, no output cap
  & Translation \\
  AA-Omniscience \parencite{jackson2025omniscience}
  & Public set. Its accuracy scores a wrong answer no better than abstaining, its index between \num{-100} and \num{100} scores a wrong answer \num{-100} and an abstention or partial answer \num{0}, and its non-hallucination rate is the share of non-correct answers that abstained or answered partially rather than guessed.
  & -- \\
  \ac{MMLU}-Pro \parencite{wang2024mmlupro}
  & Chain-of-thought prompts
  & MMLU-ProX \parencite{xuan2025mmluprox} \\
  \midrule
  \multicolumn{3}{@{}l}{\bfseries Math} \\
  AIME 2025 and 2026 \parencite{maa2026aime} & avg@\num{16} & Translation \\
  \midrule
  \multicolumn{3}{@{}l}{\bfseries Agentic} \\
  TerminalBench 2.1 \parencite{merrill2026terminalbench}
  & Terminus~2 agent \parencite{merrill2026terminalbench}, at most \num{250} steps per task, each reply capped at \num{80}k tokens or half the model's window, avg@\num{3}, dedicated deployment per model
  & -- \\
  \taubench{2} \parencite{barres2025tau2}
  & Telecom, retail and airline domains, avg@\num{3}
  & -- \\
  \taubench{3} \parencite{shi2026tauknowledge}
  & Banking domain on the \texttt{alltools} retrieval configuration, avg@\num{4}; a conversation that outgrows a model's context window scores \num{0}.
  & -- \\
  \ac{BFCL} v3 multi-turn and v4 \parencite{patil2025bfcl}
  & v4 with its non-live AST, live, multi-turn, memory and web-search splits, and bfcl-eval's leaderboard overall score, which weights those five together with an irrelevance-detection split the table does not print
  & -- \\
  BrowseComp \parencite{wei2025browsecomp}
  & Web-search and scrape tools, at most \num{100} steps per question
  & -- \\
  \midrule
  \multicolumn{3}{@{}l}{\bfseries Code} \\
  LiveCodeBench v6 \parencite{jain2024livecodebench}
  & Problems from August 2024 to April 2025, avg@\num{3}
  & -- \\
  HumanEval+ \parencite{chen2021humaneval,liu2023evalplus}
  & avg@\num{4}
  & -- \\
  SWE-bench Verified \parencite{jimenez2024swebench}
  & OpenCode agent \parencite{opencode2026}, at most \num{250} steps per task, dedicated deployment per model
  & -- \\
  \midrule
  \multicolumn{3}{@{}l}{\bfseries Instruction Following} \\
  IFBench \parencite{pyatkin2025ifbench}                 & Loose-prompt score, avg@\num{5} & -- \\
  \midrule
  \multicolumn{3}{@{}l}{\bfseries Grounding / Hallucinations} \\
  SQuAD \parencite{rajpurkar2016squad}
  & Answered from its contexts, reporting the Merlin--Arthur grounding score \parencite{waeldchen2024merlin,deiseroth2025bounding} next to the utility accuracy
  & -- \\
  RGB \parencite{chen2024rgb}
  & Closed-book, negative-rejection and error-correction conditions, scored by rule
  & -- \\

  FRAMES \parencite{krishna2024frames} & Documents in the prompt, split by prompt length at \num{24}k tokens & -- \\
  SealQA \parencite{pham2025sealqa}    & As FRAMES, with no and with \num{12} distractor documents           & -- \\
  \midrule
  \multicolumn{3}{@{}l}{\bfseries Agentic Retrieval} \\
  MuSiQue \parencite{trivedi2022musique} and Honeypot, an in-house counterpart
  & Multi-hop question answering over a document corpus the model searches with tools. Honeypot asks in German and in English. \ac{RL} trains on the training splits of MuSiQue, Honeypot and the German MuSiQue and HotpotQA, and the benchmarks are their held-out validation splits. We decontaminate our training set against them (\cref{sec:sft-decon}).
  & Agentic Wiki QA, the German MuSiQue and HotpotQA \parencite{yang2018hotpotqa}, filtered and weighted by item count \\
  \midrule
  \multicolumn{3}{@{}l}{\bfseries Industry \acs*{RAG}} \\
  Semiconductors, German Public Sector, Aerospace, Automotive Supplier and Industrial Drive Technology
  & Five in-house proxies for deployed industry assistants, from different domains, each running its application's system prompt and tool set over a corpus we built, graded with the application's judge. German Public Sector, Aerospace and Industrial Drive Technology run in German, Semiconductors and Automotive Supplier in English.
  & -- \\
  \midrule
  \multicolumn{3}{@{}l}{\bfseries Long Context} \\
  LongBench Pro \parencite{chen2026longbenchpro}
  & Mean over its length buckets from \num{8}k to \num{128}k tokens, avg@\num{3}; its \recommendedContextLengthShort{} bucket does not fit a \num{256}k-token window.
  & -- \\
  AA-LCR \parencite{artificialanalysis2025aalcr}
  & avg@\num{3}
  & -- \\
\end{longtable}
\endgroup

\paragraph{Settings}
All models are evaluated either with eval-framework \parencite{eval_framework} or, on the agentic benchmarks, Harbor \parencite{harbor2026}, with \model{2} at reasoning effort \texttt{high}.
The agentic retrieval and Industry \ac{RAG} benchmarks run in prey, our in-house environment library, on the retrieval environments of \cref{sec:appendix-rag}, and Kimi-K2.6~\parencite{moonshot2026kimik26} judges the answers on MuSiQue, Honeypot and Agentic Wiki QA (DE).
A model reports no score on a benchmark whose prompts or agent trajectories exceed its context window, and a model that cannot call tools reports none on the tool-calling benchmarks: \ac{BFCL}, BrowseComp, \taubench{2} and \taubench{3}, the agentic retrieval benchmarks and Industry \ac{RAG}.
The long-context benchmarks handle this case differently.
Their prompts exceed the \num{64}k-token context window used to serve \model{1}, but the corresponding rows record \num{0} rather than no score.
Two of these benchmarks need a caveat on how their scores may be read.
On \ac{BFCL} v4, web search uses a different search backend from the leaderboard and every model samples at its own parameters, so the scores are not comparable with the leaderboard, and an entry that outgrows a model's window counts as wrong.
The Industry \ac{RAG} scores are a relative signal between models rather than the accuracy of the deployed assistant, and Industrial Drive Technology (\num{11} items) and Automotive Supplier (\num{26}) are too small to show a trend.

\paragraph{Performance Aggregates}
The Average row is the mean of all benchmarks in a category and language.
We considered only the benchmarks on which we could evaluate all our baseline models.
Additionally, we exclude the AA-Omniscience Index and Honeypot; the AA-Omniscience Index is not a percentage, and Honeypot is a multilingual benchmark.
The Overall rows are the mean of per-category Averages.
The English and German categories' performance metrics are not directly comparable as they do not have the same benchmarks.
For instance, \taubench{2} or LiveCodeBench v6 do not have German equivalents.
The German Overall, therefore, only aggregates Knowledge, Math, Agentic Retrieval and Industry \ac{RAG} averages.
We include the overall \ac{BFCL} v4 score in the Agentic Average, and also report each split's performance for completeness.
Long Context benchmarks do not contribute to the Overall scores as we could not evaluate all the baseline models on these tasks.

\subsection{Baseline Models}

We compare \model{2} to twelve open post-trained models spanning dense models from \num{27}B to \num{70}B parameters and \ac{MoE} models with \num{3}B to \num{12}B active parameters: Qwen3.5 35B-A3B \parencite{qwen2026qwen35}, Qwen3.6 35B-A3B \parencite{qwen2026qwen36}, Qwen3-Next 80B-A3B Thinking \parencite{qwen2025qwen3nextthinking}, Qwen3.8 27B \parencite{qwen2026qwen38}, Nemotron 3 Nano 30B-A3B \parencite{nvidia2025nemotron3nano}, Nemotron 3 Super 120B-A12B \parencite{nvidia2026nemotron3super}, GLM-4.5 Air 106B-A12B \parencite{glm45team2025glm45}, GLM-4.7 Flash 30B-A3B \parencite{zai2026glm47}, Gemma 4 26B-A4B IT \parencite{gemma2026gemma4technicalreport}, Mistral Small 4 119B-A6B \parencite{mistral2026small4}, GPT-OSS 120B \parencite{openai2025gptoss} and Apertus 70B Instruct \parencite{apertus2025}, next to \model{1}.
We take the generation configurations and context windows for all models from their model cards.
\Cref{tab:baseline-models} in \cref{app:baselines} lists each model's checkpoint, reference and deployment.
Apertus 70B Instruct scores near zero on AIME, \ac{BFCL} v3 (multi-turn), \taubench{3} (Banking) and the SQuAD utility accuracy, so its column is set in grey, takes no best-score mark and enters no average.
GPT-OSS 120B and GLM-4.5 Air 106B-A12B are not evaluated on long-context benchmarks, whose prompts exceed their supported context window.

\subsection{Results and Discussion}

\begin{figure}[!htb]
  \centering
  \includegraphics{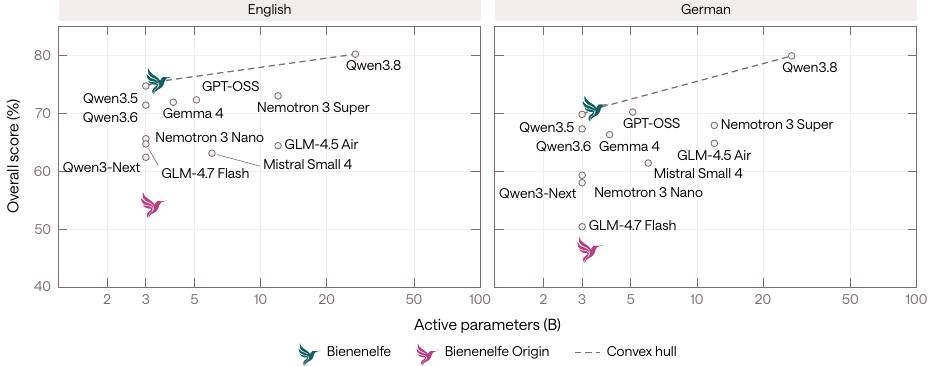}
  \caption{Overall English and German post-training aggregates of \cref{tab:post-training-evals} against the active parameters per token on a logarithmic axis, where higher and further left is better.
    The dashed line marks the convex hull of the Pareto frontier.
    Apertus 70B Instruct is excluded because its deployment does not produce comparable scores.
  }
  \label{fig:post-training-pareto}
\end{figure}

\Cref{fig:post-training-pareto} shows the Overall aggregates against the active parameters per token.
Qwen3.5 35B-A3B, \model{2} and Qwen3.8 27B span the convex hull of the Pareto frontier in both languages.
\model{2} scores \num{0.8} points above Qwen3.5 35B-A3B in English and \num{1.0} in German at \num{\modelTwoActiveParams}\mbox{B} active parameters against \num{3}\mbox{B}.

\begin{figure}[tb]
  \centering
  \includegraphics{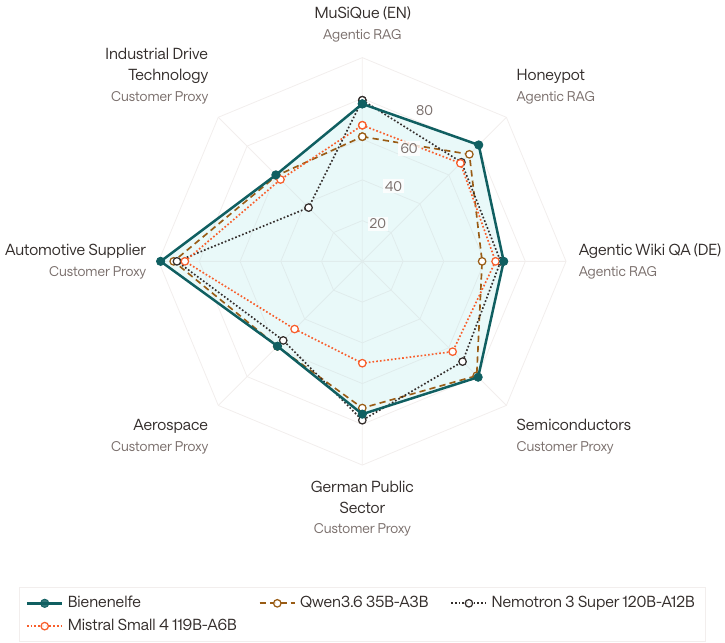}
  \caption{ Agentic retrieval and customer-proxy benchmark scores (\unit{\percent}) of the released \model{2} checkpoint and the three open models of \cref{fig:intro-spider}, from \cref{tab:post-training-evals-2}.
    MuSiQue, Honeypot and Agentic Wiki QA (DE) are agentic retrieval benchmarks; the five customer proxies are internal benchmarks built from customer use cases, named by vertical.
  }
  \label{fig:rl-spider-retrieval}
\end{figure}

Qwen3.8 27B has the highest overall aggregates, but as a dense model it activates nearly \num{8} times as many parameters per token as \model{2}.
At \num{\modelTwoActiveParams}\mbox{B} active parameters, \model{2} scores above all \num{3}\mbox{B}-active baselines in both languages.
Across all compared \ac{MoE} models, \model{2} scores best on both Overall rows.
\model{2} outperforms all the \ac{MoE} models considered on AIME 2025 and 2026 in English.
It lags behind only the dense Qwen3.8 27B. It also outperforms all models with \num{3}B active parameters on German AIME benchmarks.
\model{2} also performs best on the SQuAD (M/A Grounding Score) and on the Automotive Supplier proxy, and ties with Qwen3.5 35B-A3B for the best score on SealQA above \num{24}k without distractors.
In \num{17} further rows, it only lags behind the substantially larger Qwen3.8 27B, among them the English math average, \taubench{3}, Honeypot, Agentic Wiki QA (DE) and the English Industry \ac{RAG} average.
The weakest rows of \model{2} relative to the baselines are RGB Closed-Book, on which it scores the lowest of all models in the table, AA-Omniscience accuracy and RGB Fact-Check, and the multi-turn splits of \ac{BFCL}.
\model{2} scores at least as high as \model{1} on every benchmark both are evaluated on, except RGB Closed-Book.
\Cref{tab:rl-gain} sets a selection of the table's rows against the \ac{SFT} checkpoint that \ac{RL} starts from.

\Cref{fig:rl-spider-retrieval} compares the released \model{2} checkpoint with the three open models of \cref{fig:intro-spider} on the Agentic Retrieval and Industry \ac{RAG} benchmarks of \cref{tab:post-training-benchmarks}.
MuSiQue \parencite{trivedi2022musique}, Honeypot and German Agentic Wiki QA test the multi-hop search that the retrieval environments train (\cref{sec:retrieval_environments}).
The five customer proxies run over unseen corpora \emph{and} questions (\cref{sec:appendix-rag,tab:customer-proxy-benchmarks}), and we monitored them during \ac{RL} development (\cref{fig:hill-climb-retrieval}).
Among the plotted models, \model{2} scores highest on Honeypot, \num{6.5} points above Qwen3.6 35B-A3B, and ties Nemotron 3 Super 120B-A12B on Agentic Wiki QA (DE) (\num{69.4} against \num{67.9}); in \cref{tab:post-training-evals-2} only Qwen3.8 27B scores higher on either.
Across the five customer-proxy Industry \ac{RAG} benchmarks, \model{2} averages \num{89.7} on the two English proxies and \num{67.5} on the two German proxies scored for every model.
On English, it leads every \ac{MoE} baseline, by \num{2.7} points over Qwen3.5 35B-A3B (\num{87.0}).
On German, it beats every \ac{MoE} baseline except Qwen3.5 35B-A3B (\num{70.0}).
Only the dense Qwen3.8 27B scores higher on both (\num{93.2} and \num{80.2}).

\begin{table}[!tbp]
  \centering
  \scriptsize
  \setlength{\tabcolsep}{3pt}
  \setlength{\aboverulesep}{0pt}
  \setlength{\belowrulesep}{0pt}
  \renewcommand{\arraystretch}{1.15}
  \begin{tabularx}{\linewidth}{>{\raggedright\arraybackslash}X!{\color{AABorder}\vrule width 0.4pt}rr!{\color{AABorder}\vrule width 0.4pt}rrrrr!{\color{AABorder}\vrule width 1.5pt}rrr!{\color{AABorder}\vrule width 1.5pt}rr!{\color{AABorder}\vrule width 1.5pt}rr}
    \toprule
    Type
    & \multicolumn{7}{c!{\color{AABorder}\vrule width 1.5pt}}{\acs*{MoE}}
    & \multicolumn{3}{c!{\color{AABorder}\vrule width 1.5pt}}{\acs*{MoE}}
    & \multicolumn{2}{c!{\color{AABorder}\vrule width 1.5pt}}{\acs*{MoE}}
    & \multicolumn{2}{c}{\archlabel{Dense}} \\
    Active parameters
    & \multicolumn{7}{c!{\color{AABorder}\vrule width 1.5pt}}{3B}
    & \multicolumn{3}{c!{\color{AABorder}\vrule width 1.5pt}}{4--6B}
    & \multicolumn{2}{c!{\color{AABorder}\vrule width 1.5pt}}{12B}
    & \multicolumn{1}{c}{\dimmed{\num{27}\mbox{B}}}
    & \multicolumn{1}{c}{\dimmed{\num{70}\mbox{B}}} \\
    \cmidrule(lr){2-8} \cmidrule(lr){9-11} \cmidrule(lr){12-13} \cmidrule(lr){14-15}
    & \multicolumn{2}{c!{\color{AABorder}\vrule width 0.4pt}}{Ours}
    & \multicolumn{5}{c!{\color{AABorder}\vrule width 1.5pt}}{Baseline models}
    & \multicolumn{3}{c!{\color{AABorder}\vrule width 1.5pt}}{}
    & \multicolumn{2}{c!{\color{AABorder}\vrule width 1.5pt}}{}
    & \multicolumn{2}{c}{} \\
    \cmidrule(lr){2-3} \cmidrule(lr){4-8}
    Eval
    & \rotatebox{90}{\model{2}\hspace{3pt}}
    & \rotatebox{90}{\model{1}\hspace{3pt}}
    & \rotatebox{90}{GLM-4.7 Flash 30B-A3B\hspace{3pt}}
    & \rotatebox{90}{Nemotron 3 Nano 30B-A3B\hspace{3pt}}
    & \rotatebox{90}{Qwen3.5 35B-A3B\hspace{3pt}}
    & \rotatebox{90}{Qwen3.6 35B-A3B\hspace{3pt}}
    & \rotatebox{90}{Qwen3-Next 80B-A3B Thinking\hspace{3pt}}
    & \rotatebox{90}{Gemma 4 26B-A4B IT\hspace{3pt}}
    & \rotatebox{90}{GPT-OSS 120B\hspace{3pt}}
    & \rotatebox{90}{Mistral Small 4 119B-A6B\hspace{3pt}}
    & \rotatebox{90}{GLM-4.5 Air 106B-A12B\hspace{3pt}}
    & \rotatebox{90}{Nemotron 3 Super 120B-A12B\hspace{3pt}}
    & \dimmed{\rotatebox{90}{Qwen3.8 27B\hspace{3pt}}}
    & \dimmed{\rotatebox{90}{Apertus 70B Instruct\hspace{3pt}}} \\
    \midrule
    Total parameters (B)
    & \parambar{1.302}{\num{78.1}}
    & \parambar{0.510}{\num{30.6}}
    & \parambar{0.500}{\num{30}}
    & \parambar{0.500}{\num{30}}
    & \parambar{0.583}{\num{35}}
    & \parambar{0.583}{\num{35}}
    & \parambar{1.333}{\num{80}}
    & \parambar{0.433}{\num{26}}
    & \parambar{1.950}{\num{117}}
    & \parambar{1.983}{\num{119}}
    & \parambar{1.767}{\num{106}}
    & \parambar{2.000}{\num{120}}
    & \dimmed{\parambar{0.450}{\num{27}}}
    & \dimmed{\parambar{1.167}{\num{70}}} \\
    \arrayrulecolor{AABorder}\cmidrule[0.4pt]{1-15}\arrayrulecolor{black}
    Active parameters (B)
    & \parambar{0.058}{\num{3.46}}
    & \parambar{0.054}{\num{3.27}}
    & \parambar{0.050}{\num{3}}
    & \parambar{0.050}{\num{3}}
    & \parambar{0.050}{\num{3}}
    & \parambar{0.050}{\num{3}}
    & \parambar{0.050}{\num{3}}
    & \parambar{0.067}{\num{4}}
    & \parambar{0.085}{\num{5.1}}
    & \parambar{0.100}{\num{6}}
    & \parambar{0.200}{\num{12}}
    & \parambar{0.200}{\num{12}}
    & \dimmed{\parambar{0.450}{\num{27}}}
    & \dimmed{\parambar{1.167}{\num{70}}} \\
    \midrule
    \addlinespace[2pt]
    Overall (EN)
    & \champion{\num{75.5}}
    & \num{54.1}
    & \num{64.7}
    & \num{65.6}
    & \num{74.7}
    & \num{71.4}
    & \num{62.4}
    & \num{71.9}
    & \groupbest{\num{72.3}}
    & \num{63.1}
    & \num{64.4}
    & \groupbest{\num{73.0}}
    & \dimmed{\num{80.2}}
    & \dimmed{–} \\
    Overall (DE)
    & \champion{\num{70.8}}
    & \num{46.4}
    & \num{50.4}
    & \num{59.3}
    & \num{69.8}
    & \num{67.3}
    & \num{58.0}
    & \num{66.3}
    & \groupbest{\num{70.2}}
    & \num{61.4}
    & \num{64.8}
    & \groupbest{\num{67.9}}
    & \dimmed{\num{79.9}}
    & \dimmed{–} \\
    \midrule
    \bfseries Knowledge
    &
    &
    &
    &
    &
    &
    &
    &
    &
    &
    &
    &
    &
    &  \\
    \avg{Average (EN)}
    & \avg{\num{50.1}}
    & \avg{\num{39.7}}
    & \avg{\num{45.5}}
    & \avg{\num{46.0}}
    & \avg{\champion{\num{52.7}}}
    & \avg{\num{52.1}}
    & \avg{\num{48.4}}
    & \avg{\groupbest{\num{51.4}}}
    & \avg{\num{50.0}}
    & \avg{\num{47.5}}
    & \avg{\num{45.7}}
    & \avg{\groupbest{\num{52.0}}}
    & \dimmed{\avg{\num{56.8}}}
    & \dimmed{\avg{\num{22.8}}} \\
    \avg{Average (DE)}
    & \avg{\num{57.6}}
    & \avg{\num{44.9}}
    & \avg{\num{46.5}}
    & \avg{\num{41.2}}
    & \avg{\groupbest{\num{61.3}}}
    & \avg{\num{61.0}}
    & \avg{\num{55.7}}
    & \avg{\champion{\num{61.5}}}
    & \avg{\num{58.0}}
    & \avg{\num{51.4}}
    & \avg{\num{52.2}}
    & \avg{\groupbest{\num{59.5}}}
    & \dimmed{\avg{\num{69.2}}}
    & \dimmed{\avg{\num{24.8}}} \\
    GPQA Diamond (EN)
    & \champion{\num{84.3}}
    & \num{68.1}
    & \num{73.1}
    & \num{73.9}
    & \num{83.8}
    & \num{83.4}
    & \num{76.1}
    & \groupbest{\num{81.1}}
    & \num{76.4}
    & \num{74.7}
    & \num{73.2}
    & \groupbest{\num{78.0}}
    & \dimmed{\num{89.2}}
    & \dimmed{\num{29.5}} \\
    GPQA Diamond (DE)
    & \num{81.3}
    & \num{58.5}
    & \num{59.8}
    & \num{49.6}
    & \champion{\num{84.2}}
    & \num{80.6}
    & \num{72.2}
    & \groupbest{\num{80.1}}
    & \num{76.0}
    & \num{72.9}
    & \num{71.1}
    & \groupbest{\num{76.6}}
    & \dimmed{\num{88.1}}
    & \dimmed{\num{31.4}} \\
    Humanity's Last Exam (EN)
    & \champion{\num{21.5}}
    & \num{9.4}
    & \num{15.4}
    & \num{12.1}
    & \num{20.4}
    & \num{21.1}
    & \num{11.6}
    & \num{19.2}
    & \groupbest{\num{19.4}}
    & \num{9.7}
    & \num{8.7}
    & \groupbest{\num{20.6}}
    & \dimmed{\num{35.6}}
    & \dimmed{\num{5.2}} \\
    Humanity's Last Exam (DE)
    & \num{15.9}
    & \num{10.4}
    & \num{9.1}
    & \num{13.1}
    & \num{18.1}
    & \groupbest{\num{20.5}}
    & \num{15.6}
    & \champion{\num{23.4}}
    & \num{20.7}
    & \num{10.5}
    & \num{10.5}
    & \groupbest{\num{22.3}}
    & \dimmed{\num{37.2}}
    & \dimmed{\num{5.7}} \\
    AA-Omniscience Accuracy (public set)
    & \num{14.8}
    & \num{11.3}
    & \num{17.0}
    & \num{19.5}
    & \num{22.0}
    & \num{19.5}
    & \groupbest{\num{24.2}}
    & \num{20.7}
    & \num{23.3}
    & \groupbest{\num{25.0}}
    & \num{20.0}
    & \champion{\num{26.7}}
    & \dimmed{\num{17.5}}
    & \dimmed{\num{13.5}} \\
    \dimmed{AA-Omniscience Index (public set)}
    & \dimmed{\num{-32.8}}
    & \dimmed{\num{-64.0}}
    & \dimmed{\num{-62.8}}
    & \dimmed{\num{-45.7}}
    & \dimmed{\num{-47.3}}
    & \dimmed{\champion{\num{-15.3}}}
    & \dimmed{\num{-42.3}}
    & \dimmed{\num{-47.3}}
    & \dimmed{\num{-35.2}}
    & \dimmed{\groupbest{\num{-24.0}}}
    & \dimmed{\groupbest{\num{-28.8}}}
    & \dimmed{\num{-36.5}}
    & \dimmed{\num{-9.5}}
    & \dimmed{–} \\
    MMLU-Pro CoT (EN)
    & \num{80.0}
    & \num{70.1}
    & \num{76.5}
    & \num{78.3}
    & \champion{\num{84.6}}
    & \num{84.3}
    & \num{81.7}
    & \groupbest{\num{84.5}}
    & \num{80.8}
    & \num{80.4}
    & \num{80.9}
    & \groupbest{\num{82.7}}
    & \dimmed{\num{85.0}}
    & \dimmed{\num{43.0}} \\
    MMLU-ProX CoT (DE)
    & \num{75.5}
    & \num{65.7}
    & \num{70.7}
    & \num{61.0}
    & \num{81.7}
    & \champion{\num{81.9}}
    & \num{79.4}
    & \groupbest{\num{81.1}}
    & \num{77.2}
    & \num{70.7}
    & \num{74.9}
    & \groupbest{\num{79.7}}
    & \dimmed{\num{82.4}}
    & \dimmed{\num{37.3}} \\
    \midrule
    \bfseries Math
    &
    &
    &
    &
    &
    &
    &
    &
    &
    &
    &
    &
    &
    &  \\
    \avg{Average (EN)}
    & \avg{\champion{\num{96.5}}}
    & \avg{\num{81.7}}
    & \avg{\num{88.8}}
    & \avg{\num{88.8}}
    & \avg{\num{90.1}}
    & \avg{\num{87.8}}
    & \avg{\num{86.3}}
    & \avg{\num{87.4}}
    & \avg{\groupbest{\num{90.7}}}
    & \avg{\num{81.4}}
    & \avg{\num{82.8}}
    & \avg{\groupbest{\num{91.1}}}
    & \dimmed{\avg{\num{97.8}}}
    & \dimmed{\avg{\num{0.6}}} \\
    \avg{Average (DE)}
    & \avg{\groupbest{\num{88.8}}}
    & \avg{\num{74.3}}
    & \avg{\num{45.2}}
    & \avg{\num{84.3}}
    & \avg{\num{79.6}}
    & \avg{\num{83.7}}
    & \avg{\num{83.7}}
    & \avg{\num{88.1}}
    & \avg{\champion{\num{90.8}}}
    & \avg{\num{75.4}}
    & \avg{\num{80.6}}
    & \avg{\groupbest{\num{86.5}}}
    & \dimmed{\avg{\num{96.7}}}
    & \dimmed{\avg{\num{0.1}}} \\
    AIME 2025 (EN)
    & \champion{\num{96.9}}
    & \num{81.9}
    & \num{89.4}
    & \num{89.6}
    & \num{88.1}
    & \num{84.6}
    & \num{84.2}
    & \num{87.3}
    & \groupbest{\num{90.6}}
    & \num{79.8}
    & \num{81.9}
    & \groupbest{\num{91.7}}
    & \dimmed{\num{97.9}}
    & \dimmed{\num{0.6}} \\
    AIME 2025 (DE)
    & \groupbest{\num{87.5}}
    & \num{73.5}
    & \num{43.8}
    & \num{84.4}
    & \num{76.7}
    & \num{82.9}
    & \num{80.6}
    & \num{88.1}
    & \champion{\num{90.6}}
    & \num{72.3}
    & \num{80.6}
    & \groupbest{\num{85.6}}
    & \dimmed{\num{96.5}}
    & \dimmed{\num{0.2}} \\
    AIME 2026 (EN)
    & \champion{\num{96.0}}
    & \num{81.5}
    & \num{88.3}
    & \num{87.9}
    & \num{92.1}
    & \num{91.0}
    & \num{88.5}
    & \num{87.5}
    & \groupbest{\num{90.8}}
    & \num{83.1}
    & \num{83.8}
    & \groupbest{\num{90.4}}
    & \dimmed{\num{97.7}}
    & \dimmed{\num{0.6}} \\
    AIME 2026 (DE)
    & \groupbest{\num{90.0}}
    & \num{75.2}
    & \num{46.7}
    & \num{84.2}
    & \num{82.5}
    & \num{84.4}
    & \num{86.7}
    & \num{88.1}
    & \champion{\num{91.0}}
    & \num{78.5}
    & \num{80.6}
    & \groupbest{\num{87.5}}
    & \dimmed{\num{96.9}}
    & \dimmed{\num{0.0}} \\
    \midrule
    \bfseries Agentic
    &
    &
    &
    &
    &
    &
    &
    &
    &
    &
    &
    &
    &
    &  \\
    \avg{Average (EN)}
    & \avg{\champion{\num{63.4}}}
    & \avg{\num{41.6}}
    & \avg{\num{58.9}}
    & \avg{\num{46.4}}
    & \avg{\champion{\num{63.4}}}
    & \avg{\num{62.1}}
    & \avg{\num{46.3}}
    & \avg{\groupbest{\num{54.6}}}
    & \avg{\num{54.0}}
    & \avg{\num{40.7}}
    & \avg{\num{53.5}}
    & \avg{\groupbest{\num{54.9}}}
    & \dimmed{\avg{\num{66.7}}}
    & \dimmed{\avg{–}} \\
    \dimmed{TerminalBench 2.1}
    & \dimmed{\num{27.7}}
    & \dimmed{–}
    & \dimmed{\num{20.2}}
    & \dimmed{\num{9.7}}
    & \dimmed{\champion{\num{39.7}}}
    & \dimmed{–}
    & \dimmed{\num{8.6}}
    & \dimmed{–}
    & \dimmed{\groupbest{\num{29.2}}}
    & \dimmed{\num{21.0}}
    & \dimmed{–}
    & \dimmed{\champion{\num{39.7}}}
    & \dimmed{\num{76.8}}
    & \dimmed{–} \\
    \taubench{2} (Telecom)
    & \num{94.7}
    & \num{67.5}
    & \num{95.9}
    & \num{45.9}
    & \num{97.7}
    & \champion{\num{99.1}}
    & \num{43.9}
    & \num{45.3}
    & \groupbest{\num{73.1}}
    & \num{41.5}
    & \num{53.8}
    & \groupbest{\num{68.1}}
    & \dimmed{\num{82.5}}
    & \dimmed{\num{10.8}} \\
    \taubench{2} (Retail)
    & \num{69.9}
    & \num{58.5}
    & \num{57.9}
    & \num{64.9}
    & \num{70.8}
    & \champion{\num{71.6}}
    & \num{60.8}
    & \groupbest{\num{71.3}}
    & \num{60.5}
    & \num{62.9}
    & \num{61.4}
    & \groupbest{\num{67.5}}
    & \dimmed{\num{68.7}}
    & \dimmed{\num{9.6}} \\
    \taubench{2} (Airline)
    & \champion{\num{76.7}}
    & \num{58.7}
    & \num{68.7}
    & \num{52.7}
    & \num{76.0}
    & \num{70.7}
    & \num{65.3}
    & \groupbest{\num{73.3}}
    & \num{72.7}
    & \num{40.0}
    & \num{70.7}
    & \groupbest{\num{72.7}}
    & \dimmed{\num{83.3}}
    & \dimmed{\num{40.0}} \\
    \taubench{3} (Banking)
    & \champion{\num{38.1}}
    & \num{5.7}
    & \num{7.2}
    & \num{5.7}
    & \num{11.3}
    & \num{10.6}
    & \num{5.4}
    & \groupbest{\num{16.0}}
    & \num{14.7}
    & \num{5.7}
    & \num{6.4}
    & \groupbest{\num{15.5}}
    & \dimmed{\num{50.0}}
    & \dimmed{\num{2.1}} \\
    BFCL v3 (multi-turn)
    & \num{39.8}
    & \num{22.8}
    & \groupbest{\num{58.2}}
    & \num{47.9}
    & \num{54.0}
    & \num{53.5}
    & \num{51.4}
    & \groupbest{\num{53.4}}
    & \num{45.6}
    & \num{36.2}
    & \champion{\num{61.6}}
    & \num{44.6}
    & \dimmed{\num{42.5}}
    & \dimmed{\num{0.6}} \\
    BFCL v4 (overall)
    & \num{61.4}
    & \num{36.4}
    & \num{65.4}
    & \num{61.5}
    & \champion{\num{70.5}}
    & \num{67.2}
    & \num{51.0}
    & \groupbest{\num{68.2}}
    & \num{57.3}
    & \num{58.0}
    & \groupbest{\num{67.2}}
    & \num{61.0}
    & \dimmed{\num{73.2}}
    & \dimmed{–} \\
    \subrow{\dimmed{BFCL v4 (non-live AST)}}
    & \dimmed{\num{79.1}}
    & \dimmed{\num{78.1}}
    & \dimmed{\num{83.3}}
    & \dimmed{\num{85.0}}
    & \dimmed{\num{85.8}}
    & \dimmed{\champion{\num{88.2}}}
    & \dimmed{\num{83.6}}
    & \dimmed{\groupbest{\num{83.7}}}
    & \dimmed{\num{35.8}}
    & \dimmed{\num{83.6}}
    & \dimmed{\groupbest{\num{85.5}}}
    & \dimmed{\num{45.0}}
    & \dimmed{\num{85.3}}
    & \dimmed{–} \\
    \subrow{\dimmed{BFCL v4 (live)}}
    & \dimmed{\num{78.9}}
    & \dimmed{\num{73.7}}
    & \dimmed{\num{78.3}}
    & \dimmed{\num{78.8}}
    & \dimmed{\num{80.2}}
    & \dimmed{\num{81.4}}
    & \dimmed{\champion{\num{82.5}}}
    & \dimmed{\groupbest{\num{80.2}}}
    & \dimmed{\num{70.4}}
    & \dimmed{\num{78.4}}
    & \dimmed{\groupbest{\num{78.2}}}
    & \dimmed{\num{77.6}}
    & \dimmed{\num{79.9}}
    & \dimmed{–} \\
    \subrow{\dimmed{BFCL v4 (multi-turn)}}
    & \dimmed{\num{47.5}}
    & \dimmed{\num{27.5}}
    & \dimmed{\groupbest{\num{62.7}}}
    & \dimmed{\num{53.5}}
    & \dimmed{\num{59.9}}
    & \dimmed{\num{58.1}}
    & \dimmed{\num{56.0}}
    & \dimmed{\groupbest{\num{61.4}}}
    & \dimmed{\num{55.4}}
    & \dimmed{\num{40.4}}
    & \dimmed{\champion{\num{65.2}}}
    & \dimmed{\num{51.7}}
    & \dimmed{\num{55.5}}
    & \dimmed{–} \\
    \subrow{\dimmed{BFCL v4 (memory)}}
    & \dimmed{\champion{\num{62.8}}}
    & \dimmed{\num{19.4}}
    & \dimmed{\num{41.5}}
    & \dimmed{\num{39.1}}
    & \dimmed{\num{62.6}}
    & \dimmed{\num{53.8}}
    & \dimmed{\num{35.3}}
    & \dimmed{\groupbest{\num{52.9}}}
    & \dimmed{\num{50.7}}
    & \dimmed{\num{39.1}}
    & \dimmed{\num{43.4}}
    & \dimmed{\groupbest{\num{59.6}}}
    & \dimmed{\num{79.6}}
    & \dimmed{–} \\
    \subrow{\dimmed{BFCL v4 (web search)}}
    & \dimmed{\num{62.5}}
    & \dimmed{\num{10.5}}
    & \dimmed{\num{69.0}}
    & \dimmed{\num{66.0}}
    & \dimmed{\champion{\num{75.0}}}
    & \dimmed{\num{68.5}}
    & \dimmed{\num{12.5}}
    & \dimmed{\champion{\num{75.0}}}
    & \dimmed{\num{57.0}}
    & \dimmed{\num{69.0}}
    & \dimmed{\num{71.0}}
    & \dimmed{\groupbest{\num{71.5}}}
    & \dimmed{\num{82.0}}
    & \dimmed{–} \\
    \dimmed{BrowseComp}
    & \dimmed{\num{29.4}}
    & \dimmed{\num{4.4}}
    & \dimmed{–}
    & \dimmed{\num{14.5}}
    & \dimmed{\champion{\num{36.5}}}
    & \dimmed{\num{26.9}}
    & \dimmed{\num{2.8}}
    & \dimmed{\num{25.5}}
    & \dimmed{\groupbest{\num{31.2}}}
    & \dimmed{–}
    & \dimmed{–}
    & \dimmed{\groupbest{\num{29.1}}}
    & \dimmed{\num{46.4}}
    & \dimmed{–} \\
    \midrule
    \bfseries Code
    &
    &
    &
    &
    &
    &
    &
    &
    &
    &
    &
    &
    &
    &  \\
    \avg{Average (EN)}
    & \avg{\groupbest{\num{89.3}}}
    & \avg{\num{68.0}}
    & \avg{\num{67.8}}
    & \avg{\num{81.8}}
    & \avg{\num{85.0}}
    & \avg{\num{87.7}}
    & \avg{\num{83.6}}
    & \avg{\num{89.0}}
    & \avg{\champion{\num{90.8}}}
    & \avg{\num{82.0}}
    & \avg{\num{79.8}}
    & \avg{\groupbest{\num{88.3}}}
    & \dimmed{\avg{\num{94.2}}}
    & \dimmed{\avg{\num{25.1}}} \\
    LiveCodeBench v6
    & \groupbest{\num{85.9}}
    & \num{59.2}
    & \num{46.5}
    & \num{71.3}
    & \num{77.8}
    & \num{82.5}
    & \num{73.9}
    & \num{82.3}
    & \champion{\num{87.5}}
    & \num{71.2}
    & \num{67.8}
    & \groupbest{\num{82.0}}
    & \dimmed{\num{93.8}}
    & \dimmed{\num{8.7}} \\
    HumanEval+
    & \num{92.7}
    & \num{76.8}
    & \num{89.0}
    & \num{92.4}
    & \num{92.2}
    & \num{92.8}
    & \groupbest{\num{93.3}}
    & \champion{\num{95.7}}
    & \num{94.1}
    & \num{92.8}
    & \num{91.8}
    & \groupbest{\num{94.7}}
    & \dimmed{\num{94.7}}
    & \dimmed{\num{41.6}} \\
    \dimmed{SWE-bench Verified}
    & \dimmed{\num{66.4}}
    & \dimmed{–}
    & \dimmed{\num{51.0}}
    & \dimmed{\num{38.6}}
    & \dimmed{\num{71.6}}
    & \dimmed{\champion{\num{73.8}}}
    & \dimmed{–}
    & \dimmed{\num{57.8}}
    & \dimmed{–}
    & \dimmed{\groupbest{\num{60.8}}}
    & \dimmed{\num{11.6}}
    & \dimmed{\groupbest{\num{60.2}}}
    & \dimmed{\num{72.6}}
    & \dimmed{–} \\
    \bottomrule
  \end{tabularx}
  \caption{Post-training benchmark scores of \model{2}, \model{1}, and comparable open models (part I/II).
    The first two rows give each model's total and active parameters in billions, with bars on a common scale.
    \textbf{Bold is the best score among \acs*{MoE} models}, \underline{underlined is best in class}, \dimmedsquared{incomplete or non-percentage scores are greyed out}.
    Averaging is performed by language and over the category averages, with details given in the caption of \cref{tab:post-training-evals-2}.
  }%
  \label{tab:post-training-evals}
\end{table}

\begin{table}[!tbp]
  \centering
  \scriptsize
  \setlength{\tabcolsep}{3pt}
  \setlength{\aboverulesep}{0pt}
  \setlength{\belowrulesep}{0pt}
  \renewcommand{\arraystretch}{1.15}
  \begin{tabularx}{\linewidth}{>{\raggedright\arraybackslash}X!{\color{AABorder}\vrule width 0.4pt}rr!{\color{AABorder}\vrule width 0.4pt}rrrrr!{\color{AABorder}\vrule width 1.5pt}rrr!{\color{AABorder}\vrule width 1.5pt}rr!{\color{AABorder}\vrule width 1.5pt}rr}
    \toprule
    Type
    & \multicolumn{7}{c!{\color{AABorder}\vrule width 1.5pt}}{\acs*{MoE}}
    & \multicolumn{3}{c!{\color{AABorder}\vrule width 1.5pt}}{\acs*{MoE}}
    & \multicolumn{2}{c!{\color{AABorder}\vrule width 1.5pt}}{\acs*{MoE}}
    & \multicolumn{2}{c}{\archlabel{Dense}} \\
    Active parameters
    & \multicolumn{7}{c!{\color{AABorder}\vrule width 1.5pt}}{3B}
    & \multicolumn{3}{c!{\color{AABorder}\vrule width 1.5pt}}{4--6B}
    & \multicolumn{2}{c!{\color{AABorder}\vrule width 1.5pt}}{12B}
    & \multicolumn{1}{c}{\dimmed{\num{27}\mbox{B}}}
    & \multicolumn{1}{c}{\dimmed{\num{70}\mbox{B}}} \\
    \cmidrule(lr){2-8} \cmidrule(lr){9-11} \cmidrule(lr){12-13} \cmidrule(lr){14-15}
    & \multicolumn{2}{c!{\color{AABorder}\vrule width 0.4pt}}{Ours}
    & \multicolumn{5}{c!{\color{AABorder}\vrule width 1.5pt}}{Baseline models}
    & \multicolumn{3}{c!{\color{AABorder}\vrule width 1.5pt}}{}
    & \multicolumn{2}{c!{\color{AABorder}\vrule width 1.5pt}}{}
    & \multicolumn{2}{c}{} \\
    \cmidrule(lr){2-3} \cmidrule(lr){4-8}
    Eval
    & \rotatebox{90}{\model{2}\hspace{3pt}}
    & \rotatebox{90}{\model{1}\hspace{3pt}}
    & \rotatebox{90}{GLM-4.7 Flash 30B-A3B\hspace{3pt}}
    & \rotatebox{90}{Nemotron 3 Nano 30B-A3B\hspace{3pt}}
    & \rotatebox{90}{Qwen3.5 35B-A3B\hspace{3pt}}
    & \rotatebox{90}{Qwen3.6 35B-A3B\hspace{3pt}}
    & \rotatebox{90}{Qwen3-Next 80B-A3B Thinking\hspace{3pt}}
    & \rotatebox{90}{Gemma 4 26B-A4B IT\hspace{3pt}}
    & \rotatebox{90}{GPT-OSS 120B\hspace{3pt}}
    & \rotatebox{90}{Mistral Small 4 119B-A6B\hspace{3pt}}
    & \rotatebox{90}{GLM-4.5 Air 106B-A12B\hspace{3pt}}
    & \rotatebox{90}{Nemotron 3 Super 120B-A12B\hspace{3pt}}
    & \dimmed{\rotatebox{90}{Qwen3.8 27B\hspace{3pt}}}
    & \dimmed{\rotatebox{90}{Apertus 70B Instruct\hspace{3pt}}} \\
    \midrule
    \addlinespace[2pt]
    \bfseries Instruction Following
    &
    &
    &
    &
    &
    &
    &
    &
    &
    &
    &
    &
    &
    &  \\
    \avg{Average (EN)}
    & \avg{\groupbest{\num{78.1}}}
    & \avg{\num{62.5}}
    & \avg{\num{64.5}}
    & \avg{\num{73.2}}
    & \avg{\num{72.7}}
    & \avg{\num{66.1}}
    & \avg{\num{60.7}}
    & \avg{\champion{\num{79.9}}}
    & \avg{\num{71.1}}
    & \avg{\num{49.8}}
    & \avg{\num{38.2}}
    & \avg{\groupbest{\num{73.7}}}
    & \dimmed{\avg{\num{81.9}}}
    & \dimmed{\avg{\num{25.5}}} \\
    IFBench (loose-prompt)
    & \groupbest{\num{78.1}}
    & \num{62.5}
    & \num{64.5}
    & \num{73.2}
    & \num{72.7}
    & \num{66.1}
    & \num{60.7}
    & \champion{\num{79.9}}
    & \num{71.1}
    & \num{49.8}
    & \num{38.2}
    & \groupbest{\num{73.7}}
    & \dimmed{\num{81.9}}
    & \dimmed{\num{25.5}} \\
    \midrule
    \bfseries Grounding / Hallucinations
    &
    &
    &
    &
    &
    &
    &
    &
    &
    &
    &
    &
    &
    &  \\
    \avg{Average (EN)}
    & \avg{\num{59.4}}
    & \avg{\num{42.7}}
    & \avg{\num{57.6}}
    & \avg{\num{55.9}}
    & \avg{\num{67.5}}
    & \avg{\champion{\num{68.3}}}
    & \avg{\num{61.0}}
    & \avg{\num{62.5}}
    & \avg{\groupbest{\num{62.6}}}
    & \avg{\num{61.5}}
    & \avg{\num{63.8}}
    & \avg{\groupbest{\num{64.4}}}
    & \dimmed{\avg{\num{67.5}}}
    & \dimmed{\avg{\num{31.5}}} \\
    SQuAD (M/A Grounding Score)
    & \champion{\num{23.4}}
    & \num{0.0}
    & \num{1.5}
    & \num{0.0}
    & \num{23.0}
    & \num{12.1}
    & \num{0.0}
    & \num{0.0}
    & \num{0.0}
    & \groupbest{\num{6.3}}
    & \groupbest{\num{0.0}}
    & \groupbest{\num{0.0}}
    & \dimmed{\num{8.3}}
    & \dimmed{\num{0.0}} \\
    SQuAD (Utility Accuracy)
    & \num{83.4}
    & \num{77.4}
    & \num{82.7}
    & \num{75.0}
    & \champion{\num{89.8}}
    & \num{88.9}
    & \num{73.5}
    & \groupbest{\num{88.4}}
    & \num{74.2}
    & \num{76.8}
    & \groupbest{\num{81.9}}
    & \num{80.8}
    & \dimmed{\num{85.6}}
    & \dimmed{\num{1.6}} \\
    RGB Closed-Book
    & \num{51.0}
    & \num{52.0}
    & \num{78.0}
    & \num{80.0}
    & \num{81.0}
    & \num{79.0}
    & \groupbest{\num{89.0}}
    & \num{79.0}
    & \num{85.0}
    & \groupbest{\num{86.0}}
    & \num{92.0}
    & \champion{\num{93.0}}
    & \dimmed{\num{73.0}}
    & \dimmed{\num{85.0}} \\
    RGB Negative (Abstention)
    & \num{85.6}
    & \num{73.9}
    & \num{79.6}
    & \num{76.9}
    & \groupbest{\num{86.6}}
    & \num{79.6}
    & \num{81.3}
    & \groupbest{\num{86.0}}
    & \num{79.6}
    & \num{82.3}
    & \champion{\num{89.0}}
    & \num{74.6}
    & \dimmed{\num{70.6}}
    & \dimmed{\num{55.9}} \\
    AA-Omniscience Non-Hallucination Rate (public set)
    & \num{44.0}
    & \num{15.0}
    & \num{3.8}
    & \num{19.0}
    & \num{11.1}
    & \champion{\num{56.7}}
    & \num{12.3}
    & \num{14.3}
    & \num{23.7}
    & \groupbest{\num{34.7}}
    & \groupbest{\num{39.0}}
    & \num{13.9}
    & \dimmed{\num{67.3}}
    & \dimmed{\num{16.4}} \\
    RGB Fact-Check (Error Correction)
    & \num{34.0}
    & \num{14.0}
    & \num{68.0}
    & \num{58.0}
    & \num{74.0}
    & \num{74.0}
    & \groupbest{\num{85.0}}
    & \num{61.0}
    & \groupbest{\num{77.0}}
    & \num{60.0}
    & \num{74.0}
    & \champion{\num{90.0}}
    & \dimmed{\num{53.0}}
    & \dimmed{\num{14.0}} \\
    FRAMES (<24k)
    & \num{71.2}
    & \num{65.7}
    & \num{67.8}
    & \num{70.9}
    & \groupbest{\num{75.9}}
    & \num{74.7}
    & \num{68.9}
    & \num{69.5}
    & \champion{\num{78.3}}
    & \num{71.9}
    & \num{70.4}
    & \groupbest{\num{74.9}}
    & \dimmed{\num{78.6}}
    & \dimmed{\num{58.9}} \\
    \dimmed{FRAMES (>24k)}
    & \dimmed{\num{73.0}}
    & \dimmed{–}
    & \dimmed{\num{73.9}}
    & \dimmed{\num{74.3}}
    & \dimmed{\champion{\num{81.5}}}
    & \dimmed{\num{78.8}}
    & \dimmed{\num{68.5}}
    & \dimmed{\num{76.6}}
    & \dimmed{\groupbest{\num{78.4}}}
    & \dimmed{\num{76.6}}
    & \dimmed{–}
    & \dimmed{\groupbest{\num{78.8}}}
    & \dimmed{\num{83.3}}
    & \dimmed{–} \\
    SealQA <24k (no distractors)
    & \num{80.7}
    & \num{56.6}
    & \num{71.7}
    & \num{69.0}
    & \champion{\num{88.3}}
    & \num{82.8}
    & \num{82.1}
    & \groupbest{\num{82.1}}
    & \num{80.7}
    & \num{73.1}
    & \num{67.6}
    & \groupbest{\num{82.8}}
    & \dimmed{\num{86.9}}
    & \dimmed{\num{35.9}} \\
    SealQA <24k (12 distractors)
    & \num{61.0}
    & \num{30.0}
    & \num{65.0}
    & \num{54.0}
    & \groupbest{\num{78.0}}
    & \num{67.0}
    & \num{57.0}
    & \champion{\num{82.0}}
    & \num{65.0}
    & \num{62.0}
    & \num{60.0}
    & \groupbest{\num{70.0}}
    & \dimmed{\num{84.0}}
    & \dimmed{\num{16.0}} \\
    \dimmed{SealQA >24k (no distractors)}
    & \dimmed{\champion{\num{100.0}}}
    & \dimmed{–}
    & \dimmed{\num{61.1}}
    & \dimmed{\num{66.7}}
    & \dimmed{\champion{\num{100.0}}}
    & \dimmed{\num{88.9}}
    & \dimmed{\num{83.3}}
    & \dimmed{\num{77.8}}
    & \dimmed{\groupbest{\num{83.3}}}
    & \dimmed{\num{77.8}}
    & \dimmed{–}
    & \dimmed{\groupbest{\num{94.4}}}
    & \dimmed{\num{94.4}}
    & \dimmed{–} \\
    \dimmed{SealQA >24k (12 distractors)}
    & \dimmed{\num{65.1}}
    & \dimmed{–}
    & \dimmed{\num{57.1}}
    & \dimmed{\num{39.7}}
    & \dimmed{\champion{\num{73.0}}}
    & \dimmed{\num{71.4}}
    & \dimmed{\num{50.8}}
    & \dimmed{\groupbest{\num{68.3}}}
    & \dimmed{\num{63.5}}
    & \dimmed{\num{49.2}}
    & \dimmed{–}
    & \dimmed{\groupbest{\num{68.3}}}
    & \dimmed{\num{82.5}}
    & \dimmed{–} \\
    \midrule
    \bfseries Agentic Retrieval
    &
    &
    &
    &
    &
    &
    &
    &
    &
    &
    &
    &
    &
    &  \\
    \avg{Average (EN)}
    & \avg{\num{77.3}}
    & \avg{\num{42.7}}
    & \avg{\num{58.8}}
    & \avg{\num{71.5}}
    & \avg{\groupbest{\num{78.9}}}
    & \avg{\num{61.2}}
    & \avg{\num{50.5}}
    & \avg{\num{71.2}}
    & \avg{\groupbest{\num{76.1}}}
    & \avg{\num{66.8}}
    & \avg{\num{69.3}}
    & \avg{\champion{\num{79.1}}}
    & \dimmed{\avg{\num{83.7}}}
    & \dimmed{\avg{\num{14.0}}} \\
    \avg{Average (DE)}
    & \avg{\champion{\num{69.4}}}
    & \avg{\num{23.7}}
    & \avg{\num{49.0}}
    & \avg{\num{65.4}}
    & \avg{\num{68.4}}
    & \avg{\num{58.8}}
    & \avg{\num{61.6}}
    & \avg{\num{66.2}}
    & \avg{\groupbest{\num{67.7}}}
    & \avg{\num{65.4}}
    & \avg{\num{62.9}}
    & \avg{\groupbest{\num{67.9}}}
    & \dimmed{\avg{\num{73.5}}}
    & \dimmed{\avg{\num{26.5}}} \\
    MuSiQue (EN)
    & \num{77.3}
    & \num{42.7}
    & \num{58.8}
    & \num{71.5}
    & \groupbest{\num{78.9}}
    & \num{61.2}
    & \num{50.5}
    & \num{71.2}
    & \groupbest{\num{76.1}}
    & \num{66.8}
    & \num{69.3}
    & \champion{\num{79.1}}
    & \dimmed{\num{83.7}}
    & \dimmed{\num{14.0}} \\
    \dimmed{Honeypot (EN/DE)}
    & \dimmed{\champion{\num{80.8}}}
    & \dimmed{\num{25.3}}
    & \dimmed{\num{69.4}}
    & \dimmed{\num{60.2}}
    & \dimmed{\num{77.5}}
    & \dimmed{\num{74.3}}
    & \dimmed{\num{13.5}}
    & \dimmed{\groupbest{\num{75.4}}}
    & \dimmed{\num{66.6}}
    & \dimmed{\num{68.1}}
    & \dimmed{–}
    & \dimmed{\groupbest{\num{68.8}}}
    & \dimmed{\num{85.0}}
    & \dimmed{–} \\
    Agentic Wiki QA (DE)
    & \champion{\num{69.4}}
    & \num{23.7}
    & \num{49.0}
    & \num{65.4}
    & \num{68.4}
    & \num{58.8}
    & \num{61.6}
    & \num{66.2}
    & \groupbest{\num{67.7}}
    & \num{65.4}
    & \num{62.9}
    & \groupbest{\num{67.9}}
    & \dimmed{\num{73.5}}
    & \dimmed{\num{26.5}} \\
    \midrule
    \bfseries Industry RAG
    &
    &
    &
    &
    &
    &
    &
    &
    &
    &
    &
    &
    &
    &  \\
    \avg{Average (EN)}
    & \avg{\champion{\num{89.7}}}
    & \avg{\num{53.9}}
    & \avg{\num{76.0}}
    & \avg{\num{61.2}}
    & \avg{\num{87.0}}
    & \avg{\num{86.0}}
    & \avg{\num{62.7}}
    & \avg{\num{79.4}}
    & \avg{\groupbest{\num{82.8}}}
    & \avg{\num{74.9}}
    & \avg{\groupbest{\num{82.2}}}
    & \avg{\num{80.3}}
    & \dimmed{\avg{\num{93.2}}}
    & \dimmed{\avg{\num{23.9}}} \\
    \avg{Average (DE)}
    & \avg{\num{67.5}}
    & \avg{\num{42.7}}
    & \avg{\num{60.9}}
    & \avg{\num{46.1}}
    & \avg{\champion{\num{70.0}}}
    & \avg{\num{65.8}}
    & \avg{\num{31.1}}
    & \avg{\num{49.4}}
    & \avg{\groupbest{\num{64.2}}}
    & \avg{\num{53.4}}
    & \avg{\groupbest{\num{63.4}}}
    & \avg{\num{57.6}}
    & \dimmed{\avg{\num{80.2}}}
    & \dimmed{\avg{\num{14.4}}} \\
    Semiconductors (EN)
    & \champion{\num{80.4}}
    & \num{35.3}
    & \num{64.7}
    & \num{39.2}
    & \num{79.4}
    & \num{79.4}
    & \num{41.2}
    & \num{63.7}
    & \groupbest{\num{70.6}}
    & \num{62.7}
    & \groupbest{\num{72.5}}
    & \num{69.6}
    & \dimmed{\num{89.2}}
    & \dimmed{\num{11.8}} \\
    German Public Sector (DE)
    & \num{75.0}
    & \num{54.0}
    & \num{69.5}
    & \num{65.5}
    & \champion{\num{80.0}}
    & \num{72.0}
    & \num{29.5}
    & \num{66.5}
    & \groupbest{\num{77.0}}
    & \num{50.0}
    & \num{69.5}
    & \groupbest{\num{78.0}}
    & \dimmed{\num{89.0}}
    & \dimmed{\num{8.0}} \\
    \dimmed{Aerospace (DE)}
    & \dimmed{\num{58.9}}
    & \dimmed{\num{14.1}}
    & \dimmed{\num{50.4}}
    & \dimmed{\num{44.1}}
    & \dimmed{\champion{\num{62.3}}}
    & \dimmed{\num{59.0}}
    & \dimmed{\num{48.1}}
    & \dimmed{\groupbest{\num{58.1}}}
    & \dimmed{\num{48.1}}
    & \dimmed{\num{47.0}}
    & \dimmed{–}
    & \dimmed{\groupbest{\num{54.9}}}
    & \dimmed{\num{73.8}}
    & \dimmed{–} \\
    Automotive Supplier (EN)
    & \champion{\num{99.0}}
    & \num{72.4}
    & \num{87.2}
    & \num{83.3}
    & \num{94.6}
    & \num{92.6}
    & \num{84.2}
    & \groupbest{\num{95.1}}
    & \num{95.0}
    & \num{87.1}
    & \groupbest{\num{92.0}}
    & \num{91.0}
    & \dimmed{\num{97.3}}
    & \dimmed{\num{35.9}} \\
    Industrial Drive Technology (DE)
    & \champion{\num{60.0}}
    & \num{31.4}
    & \num{52.3}
    & \num{26.8}
    & \champion{\num{60.0}}
    & \num{59.5}
    & \num{32.7}
    & \num{32.3}
    & \num{51.4}
    & \groupbest{\num{56.8}}
    & \groupbest{\num{57.3}}
    & \num{37.3}
    & \dimmed{\num{71.4}}
    & \dimmed{\num{20.9}} \\
    \midrule
    \bfseries Long Context
    &
    &
    &
    &
    &
    &
    &
    &
    &
    &
    &
    &
    &
    &  \\
    \dimmed{LongBench Pro}
    & \dimmed{\num{64.5}}
    & \dimmed{\num{0.0}}
    & \dimmed{–}
    & \dimmed{\num{53.2}}
    & \dimmed{\num{70.3}}
    & \dimmed{\champion{\num{70.8}}}
    & \dimmed{\num{63.8}}
    & \dimmed{\groupbest{\num{64.3}}}
    & \dimmed{–}
    & \dimmed{\num{56.4}}
    & \dimmed{–}
    & \dimmed{\groupbest{\num{62.9}}}
    & \dimmed{\num{76.9}}
    & \dimmed{–} \\
    \dimmed{AA-LCR}
    & \dimmed{\num{68.3}}
    & \dimmed{\num{0.0}}
    & \dimmed{\num{49.7}}
    & \dimmed{\num{42.3}}
    & \dimmed{\num{66.3}}
    & \dimmed{\champion{\num{69.7}}}
    & \dimmed{\num{46.3}}
    & \dimmed{\groupbest{\num{68.3}}}
    & \dimmed{–}
    & \dimmed{\num{52.3}}
    & \dimmed{–}
    & \dimmed{\groupbest{\num{67.0}}}
    & \dimmed{\num{81.3}}
    & \dimmed{–} \\
    \bottomrule
  \end{tabularx}
  \caption{ Post-training benchmark scores of \model{2}, \model{1}, and comparable open models (part II/II).
    Continuation of \cref{tab:post-training-evals} with the Instruction Following, Grounding, Agentic Retrieval, Industry \ac*{RAG} and Long Context rows.
    The models are grouped by architecture and sorted by active parameters.
    In each row, the \textbf{best score among the compared \acs*{MoE} models is set bold}, and \underline{the best score of a group is underlined}.
    The dense models activate several times as many parameters per token, so they are \dimmedsquared{greyed out} and are unmarked.
    To ensure a fair comparison, we exclude from the averages, and \dimmedsquared{set in grey}, every row in which a model other than Apertus \num{70}B Instruct has no score.
    Likewise, we exclude the AA-Omniscience Index, whose range is not a percentage, Honeypot, which is bilingual, and the individual \acs*{BFCL} v4 splits.
    The Overall rows are the unweighted mean of the category averages of their language.
  }
  \label{tab:post-training-evals-2}
\end{table}

\subsubsection{Real-World and Customer Use Cases}

Most public benchmarks fail to capture specialised sector needs.
For example, popular information retrieval benchmarks require a specific harness and internet search ability, and do not test harness choices, specialised corpora and data ingestion.
We, therefore, work on (1) measuring contextualised performance in customer-specific use cases, such as automotive, aerospace, German public sector, semiconductors, and industrial drive technology as well as (2) building environments that address these specifics directly, and thus reflect real-world use cases, in particular for agentic \ac{RAG} applications.

Our results show strong performance in distribution on publicly available corpora and synthetic multi-hop \ac{RAG} tasks: \model{2} scores highest among the plotted models on Honeypot and ties the best of them on Agentic Wiki QA (DE) (\cref{fig:rl-spider-retrieval}).
Our retrieval environments, together with feedback from the customer-proxy evaluations, contribute to improving the model's agentic \ac{RAG} capabilities and domain-specific logic without ever training on customer data.
Our final scores on the customer-proxy evaluations provide evidence that the retrieval environments reflect the specifics of production setups (harness, data ingestion) as well as real-world data distributions.

\subsubsection{Hallucination Reduction}

\begin{figure}[tb]
  \centering
  \includegraphics{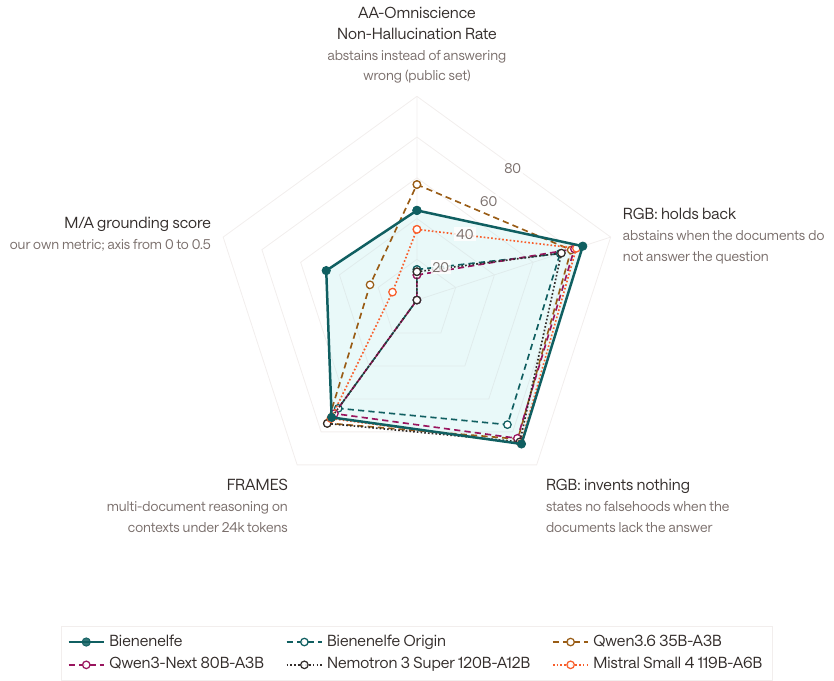}
  \caption{
    Grounding and hallucination benchmark scores (\%) of the released \model{2} checkpoint, \model{1},
    the three open models of \cref{fig:intro-spider} and Qwen3-Next 80B-A3B. The AA-Omniscience non-hallucination rate is one minus the share of the questions not answered correctly that the model answered wrongly instead of abstaining or giving a partial answer.
    The two RGB scores are measured on questions whose documents do not contain the answer.
    The M/A grounding score is our own metric for how much of the answer provably came from the document; its spoke runs from \num{0} to \num{0.5}, not from \num{0} to \num{100}.
  }
  \label{fig:rl-spider-grounding}
\end{figure}

\model{2} hallucinates less than \model{1}: of the AA-Omniscience questions it does not answer correctly, it abstains or answers partially on \qty{44}{\percent} (\model{1}: \qty{15}{\percent}), and it abstains on \qty{86}{\percent} of the RGB questions whose documents lack the answer (\model{1}: \qty{74}{\percent}).
On our own M/A grounding score, which certifies rather than estimates how much of an answer came from the document, \model{2} reaches \qty{23.4}{\percent} where \model{1} and several other models score \num{0}.
See \cref{fig:rl-spider-grounding} for more details.

%% file: sections/04-model-factory/00-model-factory.tex
\clearpage
\clearpage
\part{Model Factory}\label{sec:model-factory}

\begin{figure}[b!]
  \centering
  \includegraphics{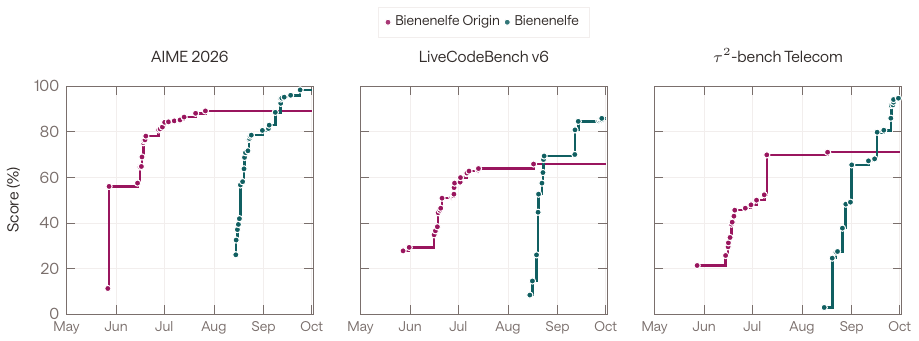}
  \caption{%
    Best score reached so far by any post-trained checkpoint of \model{1} and \model{2} on three benchmarks, taken from our internal leaderboard.
    Each point is a single evaluation of one checkpoint, released or not, so the lines can end above the scores of \cref{tab:post-training-evals}.
    \model{2} climbs faster than \model{1}.
  }
  \label{fig:hill-climb}
\end{figure}

Our Model Factory is the system we use to iterate on our training recipes; \model{1} and \model{2} are the first models it produced.
Our emphasis is not just on the current performance of \model{2}, but on how quickly that performance can be improved over time.
Poolside's Model Factory \parencite{poolside2025modelfactory} and Microsoft AI's hill-climbing machine \parencite{mai2026thinking1} place a similar emphasis.
Our hill climbing accelerates as our learnings build up and our Model Factory evolves: \model{2} climbs faster than \model{1} (\cref{fig:hill-climb}).

How quickly \model{2} improves depends on how quickly the learnings of many teams reach its training recipe.
The challenge is organisational, as more than \num{50} people, and an increasing number of agents, contribute to \model{2}.
The training recipe they contribute to has grown into a pipeline of pre-training, mid-training, long-context extension, \ac{SFT}, \ac{RL}, and evaluation, developed by different teams.
A manual training process~-- typically a collection of scripts and configs on a shared filesystem~-- does not scale to this pipeline, for three reasons.
First, every manual step is an opportunity for human error, and an error on a run of hundreds of GPUs is expensive.
Second, teams cannot easily run steps developed by other teams, so they optimise single stages in isolation, with little signal on the impact on the final model.
Third, the record of why a hyperparameter has its value, which datasets a mix holds and which recipe produced a checkpoint is scattered across chat threads, wiki pages and local notes, and is lost over time.

To solve this organisational challenge, we implement the entire training pipeline in code, turning model training into a collaborative software project.
We call this approach \ac{MTaC} \parencite{barlow2026mtac}.
By lifting the pipeline into code, we gain composability, consensus and provenance.
Composability comes from expressing manual steps as functions with typed inputs and outputs, which we build abstractions around and compose into full experiments.
Consensus comes from version control, where the \texttt{main} branch represents the team's collective best understanding of how to train a model.
Provenance comes from code comments and commit history, which encode the learnings and decisions that led to \texttt{main}.

%% file: sections/04-model-factory/01-savanna-model-training-as-code.tex
\section{\savanna{}: \Acl*{MTaC}}
\label{sec:savanna}

\savanna{}, our Model Factory, is our implementation of \ac{MTaC}.
\savanna{} implements the \model{2} training recipe end to end and provides the framework for experimenting with it.

\savanna{} owns the orchestration of the \model{2} pre- and post-training runs and their resumes, ablations and sweeps, checkpointing and model merging, evaluation, lineage, abstractions over clusters and storage, artefact retention, and declarative model releases.

\subsection{Core Abstractions}
\label{sec:savanna-core-abstractions}

\savanna{} runs are declarative.
A single TOML file describes a pre- or post-training run and its evaluation strategy end to end.
The following is a simplified example of our \model{2} medium-scale pre-training recipe on \texttt{main}:

\begin{toml}
[pre.train]
model_architecture_config = "bienenelfe_ablation_scale_m"
gpu_count = { a100 = 256, h100 = 128, b200 = 64 }
training_steps = 10342
dataset = "pre-training-mix:v9"
tokenizer = "bienenelfe-tokenizer:v14"
checkpoint_interval = 100
intermediate_eval_interval = 1000
intermediate_eval_merge_window = 4

[pre.eval]
benchmarks = [
    {name = "mmlu_bpb"},
    {name = "gsm8k_de_bpb", num_fewshot = 8},
]

[mid.train]
# ...
\end{toml}

Behind this declarative interface is the pipeline code, which composes a number of core abstractions that absorb accidental complexity.

\savanna{}'s \texttt{Checkpoint} abstraction encapsulates the storage, retention, versioning, lineage and format conversion (\ac{DCP}/\acl*{HF}) of model checkpoints.
Consuming or producing a \texttt{Checkpoint} connects its lineage, so users and their agents can navigate from a checkpoint to the configuration and commit that produced it, the datasets it was trained on, and its evaluation results.

Benchmarks range from static question sets to agentic environments, where the model acts over many turns with tools in a sandbox.
Each needs its own evaluation backend, inference settings, model deployment, scoring and aggregation.
\savanna{}'s \texttt{Benchmark} abstraction encapsulates these together with the storage and lineage of the results, so users declare the benchmarks they are interested in and \savanna{} reports the scores.

While a run trains, \savanna{} registers each intermediate checkpoint as it appears, merges it with the preceding checkpoints of the configured window, and evaluates the result.
This gives us signal over time, and the early signal lets us save GPU hours by stopping runs that are not promising.

We operate multiple GPU clusters, each with different hardware and software stacks, and we want to run the same recipe on all of them.
\savanna{}'s \texttt{Cluster} abstraction encapsulates the differences in GPU type, workload manager, storage systems, and other cluster-specific details, so a user only needs to name the cluster they want to run on when they dispatch their job.

To improve the \model{2} recipe, we run experiments, which are programmatic invocations of (subsets of) the pipeline with customisations.
We merge the learnings from an experiment into \texttt{main}, making them part of subsequent \model{2} iterations.

\subsection{Continuous Integration}
\label{sec:savanna-ci}

With everyone iterating on the same codebase, \texttt{main} has to work at all times, so that anyone can branch off it, run an experiment and expect it to succeed.
A conflict between two changes surfaces only when they are integrated, and the cost of resolving it grows with the size of both, so we practise trunk-based development and land changes on \texttt{main} often and in small increments \parencite{fowler2020branching} with targeted testing.

Every night, our Model Factory trains and evaluates a small model end to end.
This catches regressions in the training or evaluation codebases, which are versioned dependencies of \savanna{} that automation bumps throughout the day as they release; a bump of the training codebase also runs training-specific tests before it merges.

\subsection{Infrastructure}
\label{sec:savanna-infrastructure}

A user dispatches a training job from their branch, in which they have defined their experiment.
The job submits an execution to Flyte \parencite{flyte}, a Kubernetes-native workflow engine with durable execution, retries, caching and a UI to visualise runs.
Every job pins versions of the code and of its artefacts (data, checkpoints, tokenisers), which we store immutably in a registry, so runs are hermetic.
Our primary clusters run on Kubernetes for its flexibility, modularity and ecosystem.
We use Kueue \parencite{kueue} for priority-based queueing and topology-aware scheduling to ensure jobs land within a single InfiniBand domain.
We observe cluster state, job status and logs, queue load and storage pressure via Grafana.
As with our pipeline, our clusters are managed as code.

\subsection{Agents in the Pipeline}
\label{sec:savanna-agents}

Agents increasingly write the code and run the experiments behind \model{2}.
To contribute, they need visibility into the state of training, a bounded space of actions they can take unsupervised, and measurable objectives to hill climb, each of which \ac{MTaC} provides.

Visibility follows from every experiment being code pushed to the repository: agents see what is running and can trace it to the code that defines it and the reasoning behind that code.
Given the lineage, logs and metrics the factory exposes, and instructions written for them, agents autonomously review proposed experiments and triage failed runs.

To bound the action space, the factory wraps storage, clusters and registries in abstractions that only permit safe operations, so researchers and agents can act freely within these limits.
The registry, for instance, adds new artefact versions but never modifies existing ones (\cref{sec:savanna-infrastructure}), and launches on a cluster can be gated by an independent review agent.
These bounds let agents act with growing autonomy.

Objectives are quantities we define across experiments, which agents then hill climb.
An example is GPU utilisation in evaluation: we deploy hundreds of models each week, each experiment with its own benchmarks and inference settings.
This heterogeneity leaves room for idle GPUs, slow start-up and failed requests.
Agents monitor these deployments, trace inefficiencies and propose fixes.

Agents working on the pipeline also encounter limitations that can go unnoticed when they act autonomously, such as instructions that fall short, steps they have to work around or known compromises they have to make.
They either report these limitations as issues in the factory's repository, or mark them with \texttt{\# XXX} comments in the code, leaving hints for future agents to pick up and hence steering how the factory evolves.
Agents also analyse their own failures and propose changes to their instructions and harnesses that improve accuracy and reduce time and cost.
Through these loops, experiments on \model{2} also improve the factory.

%% file: sections/04-model-factory/02-savanna-in-bienenelfe-training.tex
\section{\savanna{} in \model{2} Training}
\label{sec:savanna-bienenelfe}

Every \model{2} experiment was orchestrated by \savanna{}.
The following examples show where \savanna{}'s features proved particularly useful.

\paragraph{Pre-Training Mix Experiment}
To set the pre-training mix, the pre-training data team trained thousands of small proxy models, each on a different mix, and fitted a mixing law to their scores (\cref{sec:data-mixing}).
These proxy models drew from the same data pool and ran through the same pre-training and evaluation pipeline as production, both as defined on \texttt{main}.
\savanna{} allowed us to implement and launch this as a single experiment, where each proxy training and evaluation task was queued independently such that it could fill the gaps between the larger jobs on the cluster.

\paragraph{Pre-Training Run of \model{2}}
We launched the \model{2} pre-training on a protected branch from \texttt{main}.
Over the course of training, \savanna{} automatically evaluated a merge of the last \num{4} checkpoints every \num{2000} steps.
This allowed us to monitor training progress continuously and ensure the run stays on track.
\savanna{} handles training and evaluation resumes automatically from the correct checkpoint while preserving lineage and metrics, making restarts still appear as a single run.
This reduces complexity of mid-run improvements, such as adapting the logging and profiling frequency, which in \model{2} pre-training raised the throughput by about \qty{3.7}{\percent} (\cref{fig:bienenelfe-tps}).
Running from this point-in-time branch allowed us to keep the training recipe stable, only changing it with carefully reviewed diffs applied via pull request.
The run recovered from infrastructure failures automatically (\cref{sec:fault-tolerance}).

\paragraph{Mid-Training Mix Ablations}
To choose the mid-training mix, we trained a proxy model on each candidate mix (\cref{sec:mid-evals}).
\savanna{} allowed us to compose a flash-\ac{SFT} stage onto each proxy model and evaluate it on the \ac{SFT} benchmarks, giving us a richer post-training signal alongside the pre-training evaluations.

%% file: sections/05-summary.tex
\clearpage
\clearpage
\part{Summary}
\label{part:summary}

\subsection*{Model Overview}

\model{2} is the second model after \model{1}, an English--German \ac{MoE} with \num{\modelTwoTotalParams}\mbox{B} total and \num{\modelTwoActiveParams}\mbox{B} active parameters per token, released as open weights under the Apache~2.0 licence.
It was trained on nearly \num{24}T tokens across pre-training, mid-training and long-context extension, supports contexts of up to \maxContextLengthShort{} tokens, and was built end to end by Aleph Alpha.

In the three months between \model{1} and \model{2}, we changed almost every stage of the pipeline, i.e.\ pre-training data curation, attention mechanism, sparsity, load balancing, tokeniser, the length of each training stage, an expanded and remixed \ac{SFT} data pool, and the \ac{RL} environments and infrastructure.

\pgfplotstableread[col sep=comma, row sep=\\, format=inline]{%
  de_best_name,de_fewer_best\\%
  GPT-5,11.2\\%
}\compressionFacts
We make several choices which set \model{2} apart from others for German.
Its tokeniser, a \num{128000}-token vocabulary trained with UniBPE, a hybrid of \ac{BPE} and Unigram-loss merge scoring, needs \compressionPercent{de_fewer_best} fewer tokens for German text than \compressionName{de_best_name}, the best of the nine tokenisers of other models we evaluated, while staying close to it in English.
More than a fifth of its pre-training tokens are German, with open German datasets complemented by more than \num{2}T German tokens that we curate from the web with our own pipeline or generate synthetically.
In post-training, \ac{SFT} cold-starts reasoning in German across all domains, so that the model reasons in German rather than only answering in it.

The \model{2} architecture keeps the cost of a token low: \num{40} of its \num{50} blocks use \ac{SWA} with \ac{RoPE}, the other \num{10} use full attention without positional encoding, and each \ac{MoE} layer routes a token to \num{6} of its \num{384} routed experts, in addition to one shared expert.
Post-training starts with \ac{SFT}, whose released model averages two runs of about \sftSoupValue{tokens_4000}B tokens each.
It ends with \ac{RL} on more than \num{1.2}M internally curated tasks covering reasoning, tool use, instruction following, code and retrieval.

\subsection*{Capabilities}

On the English and German evaluation aggregates, \model{2} scores above every compared \ac{MoE} model of its size (\num{3}B active parameters).
It also matches or exceeds the \ac{MoE} models with \num{12}B active parameters on math and competitive coding.
Compared with \model{1}, it improves substantially in both languages and scores at least as high on every shared benchmark, with the exception of RGB Closed-Book.
In both languages, \model{2} achieves the highest aggregate among the \ac{MoE} models we compared.
In German, its closest \ac{MoE} model is GPT-OSS 120B. The dense Qwen3.8 27B, which activates nearly eight times as many parameters per token, scores higher in both languages.
Among the open models evaluated, \model{2} lies on the Pareto frontier of quality and serving cost in both languages (\cref{fig:quality-cost}).

On the five proxy Industry \ac{RAG} evaluations, which run over unseen corpora and questions and which we monitored during \ac{RL} development, \model{2} performs strongly in both English and German, ahead of every \ac{MoE} baseline in English and of every \ac{MoE} baseline except Qwen3.5 35B-A3B in German.
Hallucination falls markedly relative to \model{1}.
On English multiple-choice tasks with an added \enquote{I don't know} option, \basemodel{2} distributes its probability mass towards abstention and correctness more than any other comparably-sized model, and is second only to Nemotron 3 Super 120B-A12B Base.
Compared with \model{1}, the post-trained \model{2} abstains far more often instead of answering wrongly, holds back more often when documents lack the answer, and invents fewer falsehoods.

As a base model, \model{2} scores above or ties every compared model of its size in both languages, and is the best-performing model in its weight class up to \num{5}B active parameters.
Its overall aggregates are second only to Nemotron 3 Super 120B-A12B Base, which has more than three times as many active parameters; GLM-4.5 Air, also with \num{12}B active parameters, scores lower in both languages.
Compared with \basemodel{1}, \basemodel{2} improves substantially in both languages.

\subsection*{Model Factory}

In the course of developing the \model{} family, the speed at which we made well-informed decisions on model size and design, training data mixes, and benchmarks set our pace more than compute did.
Our Model Factory is built around that constraint, which required automation.
It consists of a centralised repository, Savanna, shared by all training teams and was designed to scalably and reliably run hundreds of reproducible, concurrent experiments, together with a common ablation protocol, as well as comparison methods to drive modelling choices.
As inputs to this central repository, we have stage- and function-specific repositories owned or shared by subteams.
Because we develop and integrate the whole pipeline in-house, from data curation to \ac{RL}, each of our lessons learned becomes a trackable change to our shared internal recipe.

\subsection*{Conclusion}

One of the most durable results of this work is neither a checkpoint nor a training pipeline; it is a team in Europe that has trained two \model{} generations in a short amount of time and knows how to fail forward and learn.
We are excited to continue building sovereign models while incorporating learnings from the past and the feedback we receive.
For instance, we see improving German reasoning, reducing hallucinations, and extending multilingual capabilities to further European languages as strong areas of interest.
\clearpage

%% file: sections/06-legal-due-diligence-example-deep-dive-into-copyright-law.tex
\clearpage
\part*{Legal Due Diligence Example: Deep Dive Into Copyright Law}
\label{part:legal}

Legal conformity is one of our core values, especially for model training.
Our data handling and model training are built in compliance with applicable regulations, in particular the EU AI Act, the \acl*{GPAI} (\ac{GPAI}) Code of Practice under the EU AI Act (cf.\ below), the \ac{GDPR} (cf.\ measures described in this Tech Report), and the \acs{DSM-Directive} and national copyright laws.
Copyright laws are a core aspect of our focus to build regulatory compliant technology, which is why we go even beyond obligations contained in the applicable legal framework.
In the following section, we illustrate how we operate to achieve this goal by using an aspect of our copyright law due diligence as an example:

As a \ac{GPAI} Model developer training its models in the EU, European legislation sets the legal starting point and framework for our work.
With regard to Copyright Law, the Text and Data Mining exception (Article 4(3) \acs{DSM-Directive}) in European and national Copyright law is of central importance and is directly referenced by Article 53(1)(c) of the EU AI Act alongside the obligations of \ac{GPAI} Model Providers.
According to Article 53(1)(c) of the EU AI Act, identification of and compliance with rights reservations expressed under the Text and Data Mining exception shall furthermore be illustrated in the copyright policy of \ac{GPAI} Model Providers.
The \ac{GPAI} Code of Practice chapter on copyright sets out concrete commitments.
\ac{GPAI} Model Providers can demonstrate compliance with the EU AI Act's obligations by adhering to the \ac{GPAI} Code of Practice (cf.\ Article 53(4) EU AI Act).
Aleph Alpha was part of the working group, which drafted the \ac{GPAI} Code of Practice and signed it.
By being a signatory to the \ac{GPAI} Code of Practice, we committed to adhere to and respect the commitments set out therein.

We have assessed the legal requirements under copyright law and the commitments contained in the \ac{GPAI} Code of Practice.
We then decided to implement an additional layer of legal assessments with regard to the use of training datasets compiled, curated and made available by third parties (\enquote{Third-Party Datasets}).
The \ac{GPAI} Code of Practice does not contain explicit commitments regarding the use of Third-Party Datasets.
Rather, it solely states that its commitments do not affect the application of copyright provisions to protected content scraped or crawled from the internet by third parties and used by the model developer.
The lack of explicit commitment can harm the trust people have in this technology.
We believe that applying a comparable standard to Third-Party Datasets as is expected for direct crawling activities under the \ac{GPAI} Code of Practice is important to help build trustworthy technology.
This is why we assess Third-Party Datasets against the \ac{GPAI} Code of Practice commitments for direct crawling using all reasonably available information on such datasets before we decide to use them for training to validate whether the entity compiling and making available the data has complied with copyright law.

This assessment entails the following steps:
\begin{enumerate}
  \item \textbf{Legal Data Source:} We investigate the dataset creation methods of the third party based on reasonably available information.
  We assess the source of the data contained in the dataset on possible indications of copyright infringements and whether such data was made available on the initial source in compliance with applicable copyright laws.
  If we find indications for infringements on the data source level, we exclude the corresponding dataset from our model development.
  \item \textbf{Adherence to Opt-Outs and Licence Terms on the Data Source:} In a next step we assess whether the content of the data source may be crawled and used for AI training.
  In this step, we assess if the entity crawling and compiling the data adhered to appropriate instructions such as instructions expressed in accordance with the Robots Exclusion Protocol as specified in the \ac{IETF} \ac{RFC} No.\ 9309 (\enquote{robots.txt}).
  For this assessment we collect information on the crawling behaviour of the third party's crawlers or individually assess these files.
  If we find indications that the entity crawling the data did not adhere to an appropriate opt-out, we exclude such datasets from our training.
  Furthermore, in this step we assess potential licence terms and other restrictions of such crawled content expressed in natural language.
\end{enumerate}

These steps go beyond the commitments contained in the \ac{GPAI} Code of Practice and are supplemented by additional technical measures we implement during data collection and curation, which are set out in the relevant sections in this report.
By doing so, we raise the bar for transparent and compliant AI Model developers and emphasise the importance of building trustworthy technology by respecting fundamental rights such as copyrights beyond the legal requirements.
\clearpage

%% file: sections/07-contributors.tex
\section*{Contributors}
\backmatterentry{Contributors}
\label{sec:contributors}

Within each team, contributors are listed alphabetically by last name.

\medskip
\noindent\textbf{Pre-Training Data}\\
Thomas F. Burns, Niko D{\"u}rr, Fabian Gebhart, Bastian Harren, Andreas Hartel, Peter Rodenkirch Nfissi, Ivo Schaper, Lukas Schott, Mohamed Soliman, Denis Wa{\ss}mann

\medskip
\noindent\textbf{Pre-Training Efficiency}\\
Steffen Hirschmann, Jordan Sassoon, Michael W{\"u}thrich

\medskip
\noindent\textbf{Pre-Training Architecture and Training}\\
Fabien C. Y. Benureau, Jonas Knupp, Jiaze Li, Sohir Maskey, Pit Neitemeier, Philipp Scholl, Eric Schreiber, Alessio Serra, Volker Stampa, Sascha Wirges

\medskip
\noindent\textbf{Evaluations}\\
Markus Klein, Alejandro Petry Pacheco, Martin Reinhardt, Raziye Sari, Frank Schneider, Prabhu Teja Sivaprasad

\medskip
\noindent\textbf{Supervised Fine-Tuning}\\
Niklas Finken, Garwin Lechner, Dylan Rodriquez, Leonard Salewski, Simon Thel, Irina Vidal Migall{\'o}n

\medskip
\noindent\textbf{Environments}\\
Max Hammer, Murali Kondragunta, Philipp Siedler

\medskip
\noindent\textbf{Agentic \acs*{RAG} and Real-World Applications}\\
Duncan Blythe, Pranav Ragupathy, Till Speicher, Alexei Zacharov

\medskip
\noindent\textbf{Hallucination Reduction}\\
Augustijn de Boer, Bj{\"o}rn Deiseroth, B{\'e}ni Bal{\'a}zs Egressy, Johann Higl, Jurek Leonhardt, Omar Mehio, Letitia Parcalabescu

\medskip
\noindent\textbf{Reinforcement Learning}\\
Artur Baranowski, Felix Berkenkamp, Amrita Bhattacherjee, Lukas Bl{\"u}baum, Bastian Boll, Niklas Freymuth, Pablo Iyu Guerrero, Serge Thilges

\medskip
\noindent\textbf{Model Factory}\\
Moritz Althaus, Michael Barlow, Mitja Kleider, Dominik L{\^e}-Kellner

\medskip
\noindent\textbf{Legal}\\
Andreas Jagusch, Pablo Schumacher

\medskip
\noindent\textbf{Technical Leadership}\\
Yasser Jadidi, Ilhan Scheer, Samuel Weinbach

%% file: sections/08-acknowledgements.tex
\section*{Acknowledgements}
\backmatterentry{Acknowledgements}

We would like to thank all Aleph Alpha teams that continuously tested our models and gave feedback.
We would especially like to thank the inference, chat, and system evaluation teams for their tireless support with internal deployments and quick fixes to enable company-wide testing of the model.

We thank Verda for supplying the GPU compute infrastructure behind this work: they provision and operate the bare-metal fleet and Kubernetes clusters, and maintain the cluster services on our behalf.
Their responsiveness in adapting the platform to our requirements throughout the training of \model{1} and \model{2} was invaluable.

We are especially grateful to Paul Chang, Riccardo Mereu, and Daniel Obolensky for their significant contributions to improving pre- and post-training efficiency.
Their efforts were key in achieving a performant model.

We would also like to express our particular thanks to Vishal Verma, Nikolai Syrjälä, and Ananth Mahadevan for their substantial engineering contributions, and the rest of the Verda team for their support throughout this work.

\clearpage

%% file: sections/09-appendix/00-appendix.tex
\clearpage
\part*{Appendix}\label{part:appendix}

%% file: sections/09-appendix/A-quality-cost-analysis.tex
\section{Quality-Cost Analysis}\label{app:quality-cost}
Benchmark scores alone do not capture the computational cost of serving a model.
\Cref{fig:quality-cost} compares aggregate model quality scores with estimated decode text throughput per GPU. We use high-concurrency decode throughput as a proxy for efficiency in workloads where decoding dominates, e.g.\ during \ac{RL}.

\paragraph{Quality Measurement}
We directly use the \emph{Overall} rows of \cref{tab:pre-training-evals,tab:post-training-evals}.

\paragraph{Throughput Measurement}
We benchmark models with vLLM on a single node with eight NVIDIA B200 GPUs, using synthetic prompts.
Models use FP8 weights and an FP8 \ac{KV} cache, except GPT-OSS, which retains its native MXFP4 weight format.
Each baseline is served from its publisher's FP8 release where one exists and quantised to FP8 at load otherwise.
The FP8 cache setting covers only the attention \ac{KV} cache.
The recurrent state of the hybrid models keeps the precision their publishers set: FP32 for the Mamba-2 state of Nemotron 3 and the Gated DeltaNet state of Qwen3.5, Qwen3.6 and Qwen3.8, as their model configurations declare and as NVIDIA recommends for accuracy \parencite{nvidia2026nemotron3superfp8}, and BF16 for Qwen3-Next, whose configuration sets none.
Because this state is not quantised, an FP8 \ac{KV} cache frees less memory on these models than on attention-only ones \parencite{vllm2026issue55196}.

For each model and context length, we evaluate every eligible combination of data, tensor and, for \ac{MoE} models, expert parallelism in our eight-GPU search space.
We measure each layout at a concurrency near its \ac{KV}-cache limit and report the highest decode throughput.
This gives each model its best measured layout under the same hardware budget.

\begin{figure}[tb]
  \centering
  \includegraphics{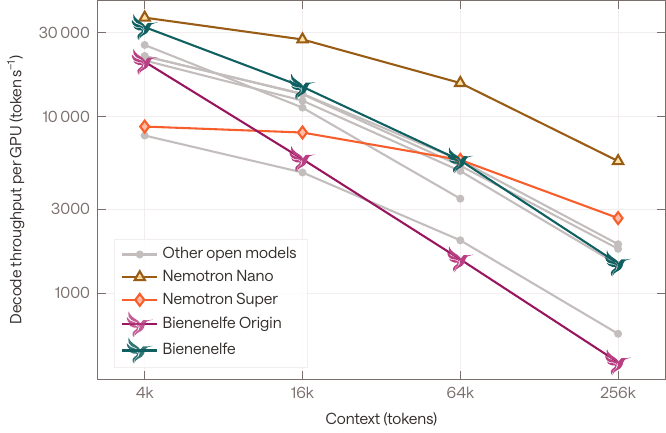}
  \caption{Decode throughput per GPU of the fastest layout at each context length.
    The other open models are Qwen3.5 35B-A3B, Qwen3.6 35B-A3B, Qwen3.8 27B, Qwen3-Next 80B-A3B Thinking and GPT-OSS 120B.
  }
  \label{fig:decode-context}
\end{figure}

For \cref{fig:quality-cost}, all models of a training stage are charged the throughput at one shared context length, close to the average context their evaluations decode at.
A sample's context grows from its prompt to its prompt plus completion while it decodes, so we take the prompt plus half the completion, average it over a model's benchmarks weighted by their decoded tokens, and then over the models in the figure.
This gives about \num{1000} tokens for the pre-training evaluations, none of which averages more than \num{1700}, and about \num{16000} tokens for the post-training evaluations, with individual models between \num{9700} and \num{27000}.
We report the throughput at \num{4}k tokens for base models and at \num{16}k tokens for post-trained models, since we deem \num{1}k tokens unrepresentative of common workloads.
\Cref{fig:decode-context} shows the decode throughput of each model's fastest layout at every measured context length.

We express throughput as estimated UTF-8 bytes per second per GPU by multiplying measured decode tokens per second by each model's bytes per token.
We calculate separate English and German factors by applying each model's tokeniser to the same plain-text prompts from the pre-training evaluation suite, then use the factor matching each plot's language.
Because these factors come from prompts rather than generated text, they are an approximation, particularly for post-trained reasoning outputs.

Benchmark scores of the baselines come from their BF16 checkpoints, apart from GPT-OSS which is in MXFP4, while their throughput is measured with FP8 weights and an FP8 \ac{KV} cache.
The figure therefore assumes that FP8 serving preserves their scores.

%% file: sections/09-appendix/B-model-architecture.tex
\section{Model Architecture}
\label{app:architecture}

\subsection{Tokeniser}
\label{app:tokenizer}

\subsubsection{Morphology}
\label{sec:appendix-tokenizer-morphology}
\pgfplotstableread[col sep=comma, row sep=\\, format=inline]{%
  others,en_prec_bienenelfe,en_prec_bpe,en_prec_best,en_order_bienenelfe,en_order_bpe,en_order_best,de_prec_bienenelfe,de_prec_bpe,de_prec_best,de_order_bienenelfe,de_order_bpe,de_order_best,k128_en_prec_bienenelfe,k128_en_prec_best,vocab_low,vocab_high,prec_drop,order_change\\%
  9,50.0,36.4,42.2,82.5,72.6,75.3,65.2,60.2,48.7,79.1,73.6,71.9,50.0,38.8,248077,262144,6.0,1.1\\%
}\morphologyFacts

We evaluate morphological alignment on two sets of \num{1000} words, in English and German, drawn at regular intervals from frequency-ranked word lists, excluding proper nouns and abbreviations.
For each word we annotate the positions where a cut between morphemes is appropriate, allowing alternative segmentations where several are defensible; a language model proposed the segmentations, and a person has reviewed the first \num{100} words of each language without changing any.
We tokenise each word as it appears after a space in running text: English words in lowercase and capitalised form, German words in their majority casing.

\begin{table}[tb]
  \centering
  \small
  \setlength{\tabcolsep}{4pt}
  \begin{tabular}{lrrrrrrr}
    \toprule
    &  & \multicolumn{3}{c}{English} & \multicolumn{3}{c}{German} \\
    \cmidrule(lr){3-5}\cmidrule(lr){6-8}
    Tokeniser
    & Vocabulary
    & Precision (\unit{\percent})
    & Order (\unit{\percent})
    & Fertility
    & Precision (\unit{\percent})
    & Order (\unit{\percent})
    & Fertility \\
    \midrule
    \multicolumn{8}{l}{\emph{Native vocabulary size}} \\
    \model{2}
    & \num{128000}
    & \textbf{50.0}
    & \textbf{82.5}
    & 1.65
    & \textbf{65.2}
    & \textbf{79.1}
    & 1.87 \\
    \acs*{BPE}
    & \num{128000}
    & 36.4
    & 72.6
    & 1.63
    & 60.2
    & 73.6
    & 1.89 \\
    \midrule
    GPT-5        & \num{200019} & 40.1 & 74.8 & 1.76 & 47.3 & 71.5 & 2.56 \\
    Tekken       & \num{131072} & 38.6 & 74.0 & 1.92 & 46.2 & 71.9 & 2.68 \\
    Qwen3-Next   & \num{151669} & 37.9 & 75.3 & 1.83 & 38.3 & 69.1 & 3.20 \\
    Qwen3.5--3.8 & \num{248077} & 38.4 & 75.3 & 1.80 & 44.0 & 70.2 & 2.60 \\
    Gemini       & \num{262144} & 42.2 & 74.2 & 1.54 & 46.2 & 69.4 & 2.59 \\
    DeepSeek V4  & \num{129280} & 37.6 & 73.0 & 1.78 & 41.0 & 68.3 & 3.02 \\
    Kimi K3      & \num{163586} & 37.2 & 74.4 & 1.83 & 37.2 & 67.2 & 3.41 \\
    GLM 5.3      & \num{154856} & 38.1 & 75.3 & 1.82 & 40.8 & 69.6 & 2.93 \\
    EuroLLM      & \num{128000} & 36.7 & 72.7 & 2.09 & 48.7 & 71.9 & 2.56 \\
    \midrule
    \multicolumn{8}{l}{\emph{Truncated to the first \num{128000} tokens}} \\GPT-5&\num{128021}&38.8&74.7&1.92&44.5&71.2&2.78\\
    Tekken&\num{128000}&38.7&74.0&1.93&46.0&71.8&2.70\\
    Qwen3-Next&\num{128026}&37.9&75.3&1.83&38.0&69.1&3.23\\
    Qwen3.5--3.8&\num{128033}&37.9&75.3&1.83&38.0&69.1&3.23\\
    Gemini&\num{128000}&38.7&74.0&1.87&41.2&68.7&3.02\\
    Kimi K3&\num{128002}&36.4&74.4&1.93&36.6&67.1&3.54\\
    GLM 5.3&\num{128036}&37.9&75.3&1.83&38.0&69.1&3.23\\
    \bottomrule
  \end{tabular}
  \caption{Morphological alignment of the \model{2} tokeniser and ten other tokenisers on \num{1000} English and \num{1000} German words, each at its own vocabulary size and truncated to its first \num{128000} tokens.
    \emph{Precision} is the share of word-internal token boundaries that fall on a morpheme boundary.
    \emph{Order} denotes the probability that a tokeniser merges a position inside a morpheme before a morpheme boundary of the same word (\qty{50}{\percent} is chance).
    \emph{Fertility} denotes the number of tokens per word.
    The highest precision and order of each column are set in bold.
    \acs*{BPE} is plain \acl*{BPE} trained on the same corpus with the same vocabulary size as \model{2}.
    \emph{Tekken} is Mistral's tokeniser \parencite{mistral2024nemo}, also used by Apertus and Nemotron; \emph{Qwen3.5--3.8} is the tokeniser shared by Qwen3.5, Qwen3.6 and Qwen3.8; and \emph{Gemini} the tokeniser shared by Gemma~4 and Gemini \parencite{gemma2026gemma4technicalreport}.
    GPT-5, DeepSeek V4 and EuroLLM denote the tokenisers of these models \parencite{openai2023tiktoken,deepseekai2026deepseekv4,martins2024eurollm}.
    Every difference to \model{2} in precision and order is significant (\qty{99}{\percent} paired bootstrap intervals over words).
  }
  \label{tab:tokenizer-morphology}
\end{table}

We then compare \model{2}'s tokenisation against \ac{BPE} trained on the same corpus and other tokenisers.
\Cref{tab:tokenizer-morphology} lists precision and order as two morphological alignment measures and the fertility (number of tokens per word) for every tokeniser at its own vocabulary size and with every vocabulary truncated to \num{128000} tokens.
In German, \morphologyPercent{de_prec_bienenelfe} of \model{2}'s word-internal token boundaries fall on morpheme boundaries, against \morphologyPercent{de_prec_best} for the best of \morphologyValue{others} other tokenisers and \morphologyPercent{de_prec_bpe} for \ac{BPE}.
UniBPE also increases English morphological alignment, with \model{2} achieving \morphologyPercent{en_prec_bienenelfe} against \morphologyPercent{en_prec_best} for the best other tokeniser and \morphologyPercent{en_prec_bpe} for \ac{BPE}.
We also study the order of the merges: a perfectly morphological tokeniser would join every position inside a morpheme before joining two morphemes together.
We measure the probability that, picking at random one gold morpheme boundary and one other position inside the same word, the tokeniser merges the other position first.
Here, \qty{50}{\percent} indicates random behaviour while higher values indicate better morphological alignment.
\model{2} scores \morphologyPercent{en_order_bienenelfe} in English and \morphologyPercent{de_order_bienenelfe} in German, against \morphologyPercent{en_order_bpe} and \morphologyPercent{de_order_bpe} for \ac{BPE}, and at most \morphologyPercent{en_order_best} and \morphologyPercent{de_order_best} for the other tokenisers.

Truncating the larger vocabularies to \num{128000} tokens changes the merge order by at most \morphologyValue[1]{order_change} points, so the order is insensitive to vocabulary size.
In contrast, the boundary precision drops by up to \morphologyValue{prec_drop} points for the largest vocabularies (\morphologyValue{vocab_low} to \morphologyValue{vocab_high} tokens), whose late tokens cover whole words.
At an equal size of \num{128000} tokens, \model{2}'s lead in English increases, with \morphologyPercent{k128_en_prec_bienenelfe} against \morphologyPercent{k128_en_prec_best} for the best other tokeniser, and it stays ahead of all other tokenisers on both measures.
All differences in precision and order between \model{2} and the other tokenisers are significant (\qty{99}{\percent} paired bootstrap intervals over words).

\FloatBarrier

\subsection{QK Normalisation}
\label{app:qk-norm}
\begin{figure}[tb]
  \centering
  \includegraphics{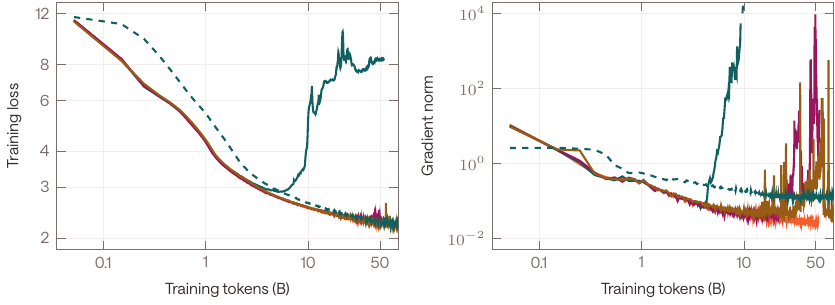}
  \includegraphics{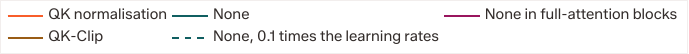}
  \caption{Training loss and gradient norm at scale M with and without QK normalisation.
    Both are averaged over bins of \num{0.1}B tokens, over the first \num{53}B pre-training tokens.
    Each configuration is a single run from scratch at the baseline learning rates of M unless stated otherwise.
  }
  \label{fig:qk-norm}
\end{figure}

Queries and keys are normalised per head with \ac{RMS} normalisation \parencite{henry2020querykey,zhang2019rmsnorm} before \ac{RoPE}:
\begin{equation}
  \hat q_i=\gamma_q\odot\frac{q_i}{\operatorname{RMS}(q_i)},
  \qquad
  \hat k_j=\gamma_k\odot\frac{k_j}{\operatorname{RMS}(k_j)},
  \qquad
  \operatorname{RMS}(x)=\sqrt{\tfrac{1}{d_h}\lVert x\rVert_2^2+\varepsilon},
  \label{eq:qk-norm}
\end{equation}
with learned gains \(\gamma_q,\gamma_k\in\mathbb R^{d_h}\).
The normalisation bounds the attention logits \(\hat q_i^\top\hat k_j/\sqrt{d_h}\) by the gains, so they cannot grow with the norms of the query and key projections.
\textcite{wortsman2024smallscale} show that the growth of these logits destabilises training at high learning rates already in small models, and that QK normalisation prevents it.
We verify this at scale M (\cref{tab:proxy-models}), where all runs share data, recipe and model apart from the treatment of the attention logits (\cref{fig:qk-norm}).
Without QK normalisation, the model at the baseline learning rates of M diverges early in training and does not recover.
With QK normalisation removed from the full-attention blocks only, the model trains, but its gradient norm spikes by several orders of magnitude in the second half of the run.
An adapted QK-Clip \parencite{kimiteam2025k2} for \ac{GQA} with threshold \num{100} in place of QK normalisation also shows gradient-norm spikes.
Without QK normalisation, the model trains stably at \num{0.1} times these learning rates, and then reaches a higher loss than the model with QK normalisation throughout the run.
With QK normalisation, training stays stable across the learning rates and batch sizes of our sweeps (\cref{sec:bienenelfe-hyperparameter-transfer}).

\FloatBarrier

\subsection{Model-Shape Evaluation Suite}
\label{app:model-shape-metrics}

For the initial model-shape sweeps, accuracy mean averages \num{17} metrics: \ac{ARC}, HellaSwag, \ac{PIQA}, HumanEval, \ac{MBPP} and MATH Minerva in English and German, English \ac{MMLU}, \ac{MMLU}-Pro, \ac{GSM8K} and TriviaQA and German Global \ac{MMLU}.
We use normalised \acl*{LL} accuracy for multiple-choice tasks, completion accuracy for \ac{GSM8K} and TriviaQA, code completion accuracy including errors for HumanEval and \ac{MBPP} and exact-flex accuracy for MATH Minerva.
\Ac{BPB} mean averages the four HumanEval and \ac{MBPP} \acs{BPB} metrics in English and German.

\FloatBarrier

\subsection{Sparsity Sweep}
\label{app:sparsity-sweep}

\begin{figure}[htbp]
  \centering
  \includegraphics{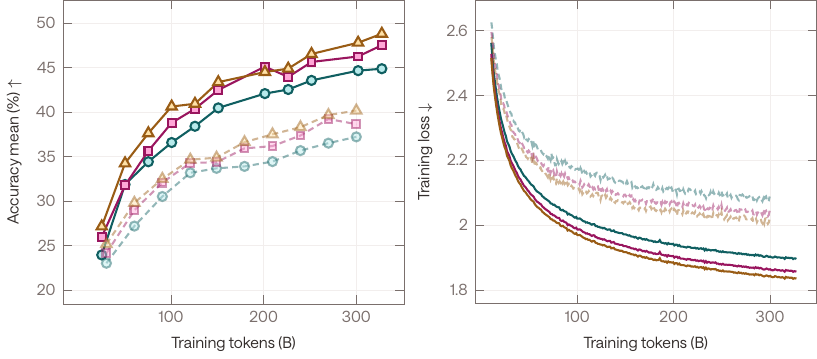}
  \includegraphics{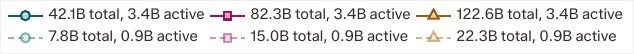}
  \caption{Aggregate benchmark accuracy and training loss against training budget for both sparsity sweeps.
    Solid curves are the configurations with \num{3.4}\mbox{B} active parameters, dashed and faded curves those with \num{0.9}B; colour encodes the expert count, so each pair shares a hue.
    The points at \num{0.9}\mbox{B} active parameters use the scaling-law learning rate \(\eta_0\).
    Both scales average the same \num{17} accuracy benchmarks.
  }\label{fig:model-shape-sparsity-evals}
\end{figure}

The training loss and accuracy of the sparsity sweep of \cref{sec:scaling-sparsity} at \num{3.4}B and \num{0.9}B active parameters are shown in \cref{fig:model-shape-sparsity-evals}.
The number of long requests that each configuration at \num{3.4}B active parameters fits on common accelerator setups is listed in \cref{tab:expert-capacity-deployment}.

\begin{table}[htbp]
  \centering
  \scriptsize
  \setlength{\tabcolsep}{3.5pt}
  \begin{tabular}{rrrrrr}
    \toprule
    Total parameters & \(2\times\)H100 & \(4\times\)H100 & \(1\times\)B200 & \(2\times\)B200 & \(2\times\)B300 \\
    \midrule
    \num{42.1}\mbox{B}  & \best{\num{33}} & \best{\num{82}} & \best{\num{46}} & \best{\num{107}} & \best{\num{171}} \\
    \num{82.3}\mbox{B}  & \num{18}        & \num{67}        & \num{31}        & \num{93}         & \num{157} \\
    \num{122.6}\mbox{B} & \num{3}         & \num{52}        & \num{16}        & \num{78}         & \num{142} \\
    \bottomrule
  \end{tabular}
  \caption{Estimated maximum number of concurrent \recommendedContextLengthShort{}-token requests in FP8, with evenly sharded weights and \acs*{KV} cache, a \qty{10}{\percent} fragmentation margin and \qty{6}{\giga\byte} of activation workspace per GPU (H100: \qty{80}{\giga\byte}, B200: \qty{192}{\giga\byte}, B300: \qty{288}{\giga\byte}).
    The largest number of each column is set in bold.
  }
  \label{tab:expert-capacity-deployment}
\end{table}

\FloatBarrier

\subsection{Model Shape at \num{80}\mbox{B} Total Parameters}
\label{app:model-shape}

With the total size fixed by the sparsity sweep of \cref{sec:scaling-sparsity}, two ablations set the shape of the \model{2} model: the granularity of its experts (\cref{app:shape-ablation}) and the width of its residual stream (\cref{app:shared-representation-width}).

\subsubsection{Expert Granularity}
\label{app:shape-ablation}

We compare three expert configurations of about \num{80}B total parameters (\cref{tab:shape-ablation}).
The \model{2} configuration has width \num{2560} and routes each token to \num{6} of \num{384} experts of hidden dimension \num{512}.
The finer-grained configuration follows the expert count of Qwen3-Next \parencite{qwen2025qwen3next}, with width \num{2048} and \num{12} of \num{512} experts of the same hidden dimension.
The coarser-grained configuration is the \num{82.3}B configuration of the sparsity sweep, with width \num{2048} and \num{6} of \num{256} experts of hidden dimension \num{1024}.
All three train for \num{327.6}B tokens and are compared at their last checkpoint.

The \model{2} configuration reaches the lowest loss, the highest accuracy and the lowest \ac{BPB}.
The finer-grained configuration comes close in loss and accuracy, and the coarser-grained configuration falls behind both.

\begin{table}[htbp]
  \centering
  \small
  \setlength{\tabcolsep}{4pt}
  \begin{tabular}{lrrrrrrrr}
    \toprule
    Configuration
    & Width
    & Heads
    & Experts
    & Top-\(k\)
    & Expert hidden
    & Loss \(\downarrow\)
    & Acc. (\unit{\percent}) \(\uparrow\)
    & \acs*{BPB} \(\downarrow\) \\
    \midrule
    \model{2}               & 2560 & 48 & 384 & 6  & 512  & \textbf{1.847} & \textbf{38.0} & \textbf{0.486} \\
    Finer-grained experts   & 2048 & 32 & 512 & 12 & 512  & 1.848          & 37.6          & 0.495 \\
    Coarser-grained experts & 2048 & 32 & 256 & 6  & 1024 & 1.860          & 35.6          & 0.497 \\
    \bottomrule
  \end{tabular}
  \caption{Expert configurations of about \num{80}\mbox{B} total parameters after \num{327.6}\mbox{B} tokens.
    The loss is the training loss averaged over the last \num{2}B tokens.
    Accuracy and \acs*{BPB} are the aggregates of the last checkpoint; the best value of each column is bold.
  }
  \label{tab:shape-ablation}
\end{table}

\FloatBarrier

\subsubsection{Residual Width}
\label{app:shared-representation-width}

We compare residual streams of different widths at \num{0.9}B active parameters and approximately matched training \acp{FLOP}, with the matrix shapes of the attention and expert layers adapted to the width.
All configurations use a sequence length of \num{16}k tokens.
The result is not separately validated at \num{3.4}B active parameters because changing \(d_{\mathrm{model}}\) would require a new matched-\ac{FLOP} sweep that retunes the attention and expert widths.

The residual width \(d_{\mathrm{model}}\) determines the dimensionality of the shared representation read and updated by both attention and \ac{MoE} blocks.
Increasing \(d_{\mathrm{model}}\) raises the parameter and compute cost of the associated projections, residual-activation memory and communication volume.
Under approximately matched training \acp{FLOP}, a wider residual stream therefore comes with fewer attention heads of the same head dimension and with narrower experts (\cref{tab:shared-width-shapes}).

\begin{table}[htbp]
  \centering
  \small
  \setlength{\tabcolsep}{4pt}
  \begin{tabular}{rrrrcrr}
    \toprule
    \(d_{\mathrm{model}}\)&Heads&Expert hidden&Dense \acs*{FFN} hidden&Learning rate&Loss \(\downarrow\)&Accuracy (\%) \(\uparrow\)\\
    \midrule
    640  & 20 & 480 & 3840 & \(\sqrt{2}\)   & 2.190          & 31.2 \\
    1024 & 16 & 384 & 3072 & \(\sqrt{2}\)   & 2.132          & 34.6 \\
    1600 & 12 & 288 & 2304 & \(1/\sqrt{2}\) & \textbf{2.109} & \textbf{34.9} \\
    \bottomrule
  \end{tabular}
  \caption{Matrix shapes and results of the three residual widths at matched training \acsp*{FLOP}, each at the learning rate with the highest mean accuracy, given as a multiple of the scaling-law learning rate \(\eta_0\).
    The head dimension, the \acs*{KV} heads and the \num{50} layers are the same for all three.
    Loss is the training loss averaged over the last \num{2}B tokens; accuracy is the mean of the accuracy metrics of the model-shape suite (\cref{app:model-shape-metrics}) at the last checkpoint, and the best value of each column is bold.
  }
  \label{tab:shared-width-shapes}
\end{table}

We compare residual widths of \num{640}, \num{1024} and \num{1600} at a fixed depth of \num{50} layers.
Each configuration is trained for \num{150}B tokens using learning rates of \(\eta_0/\sqrt{2}\), \(\eta_0\) and \(\sqrt{2}\eta_0\), where \(\eta_0\) is predicted by the learning-rate scaling law of \model{1}.
For each width, we select the learning rate with the highest mean evaluation accuracy and report all metrics from the corresponding run.

After learning-rate selection, training loss and mean accuracy improve with residual width (\cref{tab:shared-width-shapes}).
Most of the accuracy improvement occurs between widths \num{640} and \num{1024}, with a smaller gain from \num{1024} to \num{1600}.
Within the evaluated compute budget, these results favour a wider residual stream with fewer attention heads and narrower experts.

\FloatBarrier

\subsection{Leading Dense Layers and Routing in the First Layers}
\label{app:leading-dense}

\model{1} uses dense \acp{FFN} in its first two blocks, a common choice against load-balancing instabilities in the first layers \parencite{dai2024deepseekmoe,deepseekai2024deepseekv3}.
Our ablation of the \model{2} architecture suggested that \ac{EQB} makes them unnecessary: over \num{327.6}B tokens, the model without leading dense layers reaches a slightly lower loss and \ac{BPB} and a higher accuracy at a similar load balance (\cref{tab:leading-dense}).
\model{2} therefore uses \ac{MoE} layers in all \num{50} blocks.
The diagnostics of the full \model{2} run below call this outcome into question: its first two \ac{MoE} layers make almost no use of their routed experts.

\paragraph{Routing in the First Layers}
Balance metrics can look healthy while the bias overrides the router's own preferences, which leaves the routing effectively random.
We therefore track, for every \ac{MoE} layer, counted from zero, two balance metrics and three metrics of the bias's effect.
With the expert load distribution \(F_e(\mathcal B)=c_e(\mathcal B)/(K|\mathcal B|)\), the normalised scores \(P_{t,e}=S_{t,e}/\sum_{j}S_{t,j}\) and the entropy \(\mathcal H(v)=-\sum_{e=1}^{E}v_e\log v_e\), the balance metrics are the effective dead-expert fraction and the normalised router entropy \parencite{krishnamurthy2023improving}:
\begin{align}
  \text{effective dead-expert fraction}\quad
  D(\mathcal B) & =1-\frac{1}{E}\exp\!\big(\mathcal H(F(\mathcal B))\big),
  \label{eq:dead}                                                                  \\[4pt]
  \text{normalised router entropy}\quad
  H(\mathcal B) & =\frac{1}{|\mathcal B|}\sum_{t\in\mathcal B}\mathcal H(P_{t,:}),
  \qquad
  \widehat H(\mathcal B)=\frac{H(\mathcal B)}{\log E}.
  \label{eq:router-entropy}
\end{align}
With \(\mathcal I^{\mathrm u}_t=\TopKset(Z_{t,:},K)\) the set the router selects without the bias, and \(\mathcal I_t\) of \cref{eq:moe-router} the set actually routed to, the metrics of the bias's effect are
\begin{align}
  \text{top-\(K\) intervention rate}\quad
  I(\mathcal B)                & =\frac{1}{|\mathcal B|}\sum_{t\in\mathcal B}
  \left(1-\frac{\big|\mathcal I^{\mathrm u}_t\cap\mathcal I_t\big|}{K}\right),
  \label{eq:intervention}                                                     \\[4pt]
  \text{intrinsic top-\(K\) mass}\quad
  M^{\mathrm u}(\mathcal B)    & =\frac{1}{|\mathcal B|}\sum_{t\in\mathcal B}
  \sum_{e\in\mathcal I^{\mathrm u}_t} P_{t,e},
  \label{eq:intrinsic-mass}                                                   \\[4pt]
  \text{selected top-\(K\) mass}\quad
  M^{\mathrm{sel}}(\mathcal B) & =\frac{1}{|\mathcal B|}\sum_{t\in\mathcal B}
  \sum_{e\in\mathcal I_t}P_{t,e}.
  \label{eq:selected-mass}
\end{align}

\begin{figure}[tb]
  \centering
  \includegraphics{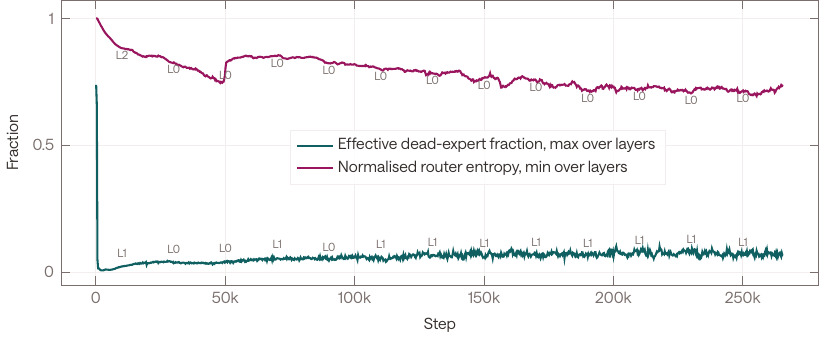}
  \caption{The effective dead-expert fraction \(D\) of \cref{eq:dead} and normalised router entropy \(\widehat H\) of \cref{eq:router-entropy} of \model{2} pre-training.
    The effective dead-expert fraction is the maximum over the \num{50} layers, the normalised router entropy the minimum.
    The tags name the layer holding the extremum at that point.
  }\label{fig:router-load-metrics}
\end{figure}

\begin{figure}[tb]
  \centering
  \begin{subfigure}{\linewidth}
    \centering
    \includegraphics{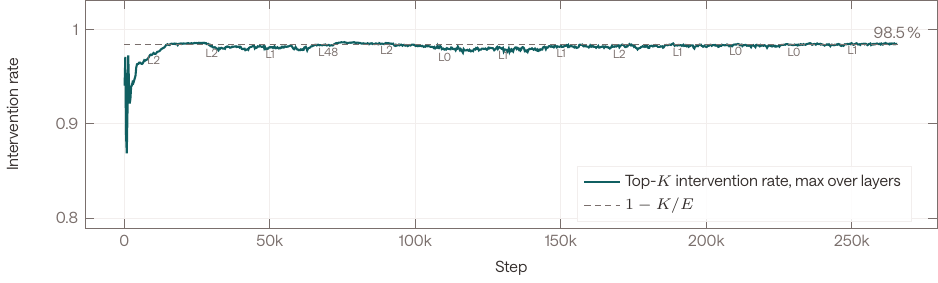}
    \caption{Top-\(K\) intervention rate \(I\), maximum over layers.}\label{fig:router-intervention}
  \end{subfigure}
  \begin{subfigure}{\linewidth}
    \centering
    \includegraphics{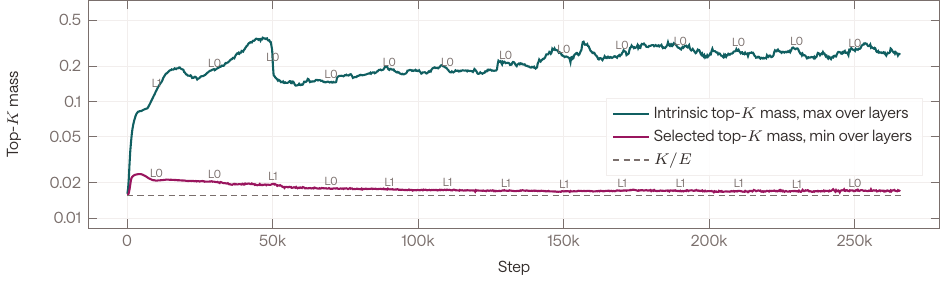}
    \caption{Intrinsic top-\(K\) mass \(M^{\mathrm u}\), maximum over layers, and selected top-\(K\) mass \(M^{\mathrm{sel}}\), minimum over layers.}\label{fig:router-mass}
  \end{subfigure}
  \caption{The routing diagnostics of \cref{eq:intervention,eq:intrinsic-mass,eq:selected-mass} of \model{2}, each at the layer where it is most extreme.
    The tags name the layer holding the extremum at that point, and the dashed lines mark a uniform choice: an intervention rate of \(1-K/E\) in~(\subref{fig:router-intervention}) and a mass of \(K/E\) in~(\subref{fig:router-mass}).
  }\label{fig:router-intervention-mass}
\end{figure}

In the \model{2} run, the first two layers look healthy on the balance metrics (\cref{fig:router-load-metrics}): their effective dead-expert fraction is the highest of all layers but moderate, and their router entropy is the lowest, a strong routing preference.
The bias, however, overrides that preference almost entirely (\cref{fig:router-intervention-mass}).
Their top-\(K\) intervention rate reaches \(\qty{98.5}{\percent}\approx 1-K/E\) towards the end of training, and while their intrinsic top-\(K\) mass is the highest of all layers, their selected top-\(K\) mass is the lowest, more than an order of magnitude smaller and close to the uniform value \(K/E\).

\begin{figure}[tb]
  \centering
  \includegraphics{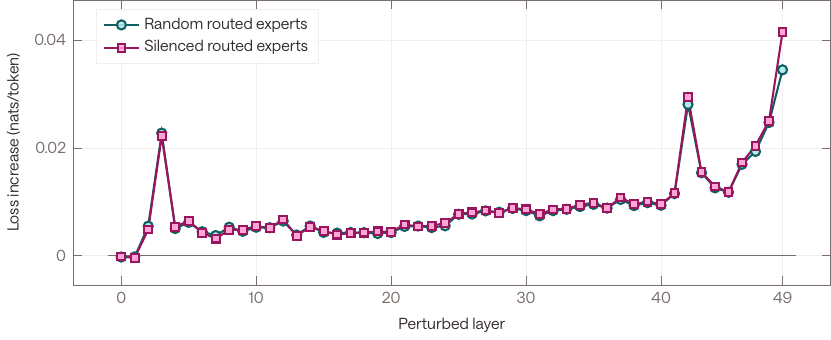}
  \caption{Loss increase from perturbing the routing of one \acs*{MoE} layer of \model{2} at step \num{225000}, all other layers untouched, on \num{65536} held-out tokens against a baseline of \num{1.850} nats per token.
    Random routed experts sends each token to uniformly drawn experts; silenced routed experts removes the routed path's output and keeps the shared expert.
  }\label{fig:routing-perturbation}
\end{figure}

To test whether their routed experts matter, we perturb the routing of one \ac{MoE} layer at a time at the step-\num{225000} checkpoint and measure the loss on \num{65536} held-out tokens, leaving the other \num{49} layers unchanged, similar to \textcite{riquelme2021scaling} and \textcite{jin2026llalmoe}.
\emph{Random routed experts} replaces each token's selected experts with a uniform random draw and tests whether the routing decision matters; \emph{silenced routed experts} removes the routed output, keeping the shared expert, and tests whether the routed experts contribute at all.
Only for layers~0 and~1 are both perturbations free, with loss changes indistinguishable from zero (\cref{fig:routing-perturbation}), so the model does not rely on their routed experts.
Layer~48 also has a high intervention rate, but perturbing it raises the loss clearly, so a high intervention rate alone does not mean that the routed experts are unused.

We tried fixing this issue during \ac{SFT}, where load balancing is disabled, by densifying layers~0 and~1, concatenating the shared expert and the six routed experts with the lowest biases into a dense \ac{FFN} of hidden width \num{3584}, keeping their pre-trained weights.
This gave similar training loss and no clear downstream improvement, so \model{2} keeps these layers as \ac{MoE} layers.
The \num{327.6}B-token ablation did not reveal this behaviour, as it emerged only after longer training.
Resolving this issue remains future work.

\FloatBarrier

\subsection{Pre-Training Infrastructure}
\label{sec:infrastructure-details}

To efficiently support large sharding degrees at our model scale, we make use of well-established optimisations.
As a general reference, see for example \textcite{tazi2025ultrascale}.
For PyTorch~\parencite{ansel2024pytorch} FSDP2 modules, these are:
\begin{description}
  \item[\path{set_separate_reduce_scatter_group()}] Dedicated NCCL process group for reduce-scatters, so they overlap with all-gathers instead of serialising behind them.
  \item[\path{set_requires_all_reduce()} and \path{set_is_last_backward()}] Microbatch-aware collectives that defer the inter-replica all-reduce and FSDP2's end-of-backward finalisation to the last microbatch.
  \item[\path{set_reshard_after_backward()}] Keeps the token embeddings and the first \num{40} blocks unsharded after the backward pass between the two gradient-accumulation steps, which removes the need to all-gather weights starting with the second forward step.
  We force-reshard before the optimiser.
  \item[\path{set_reduce_scatter_max_input_buffers()}] Three reduce-scatter input buffers (about \qty{800}{\mega\byte} each at this scale), to hide infrequent, non-deterministic communication latency that would otherwise be exposed to the critical path.
\end{description}

As the \ac{EP} backend, we use DeepEP~v2 with the ElasticBuffer API~\parencite{deepep2025}.
We wrap \texttt{buffer.dispatch()} and \texttt{buffer.combine()} into PyTorch custom ops, so they seamlessly integrate into a fully PyTorch-compiled transformer layer.
To reduce device memory consumption, we store no activation inside the \ac{MoE} \ac{FFN} and recompute them in the backward pass.
Every token is routed to \(K\) experts, so the size of each intermediate activation is scaled up by a factor of \(K\); at the same time the arithmetic intensity of a grouped matrix-multiply operation that underlies the \ac{MoE} \ac{FFN} is lower than that of a dense layer.
This makes the recomputation tractable.
To further reduce the memory consumption, we reuse the same ElasticBuffer across all layers and also recompute the dispatch in the backward pass.

\begin{figure}[t]
  \centering
  \includegraphics{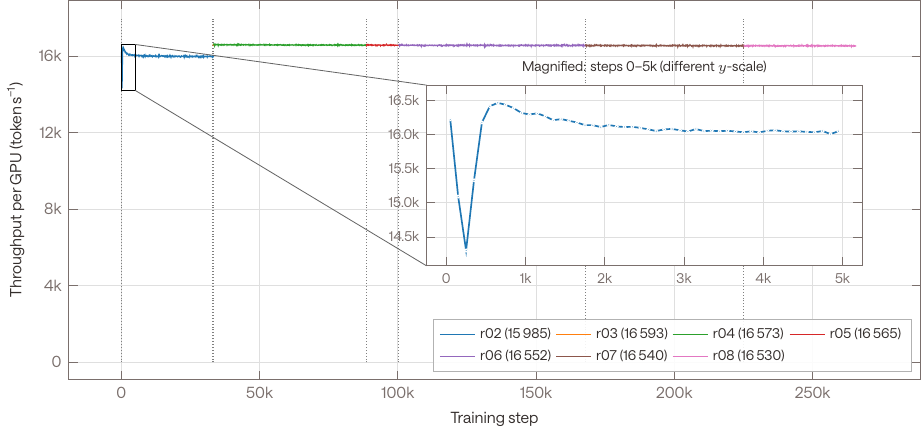}
  \caption{Training throughput of the \model{2} pre-training runs broken down into their constituent runs r02--r08.
    The plot shows the median throughput in tokens per second and GPU over bins of \num{100} steps.
    We do this to smooth out dips caused by our profiling, logging and checkpointing; the plot shows that there are no expert-parallel-related degradations over time.
    Dotted vertical lines mark restart boundaries.
    The legend values are per-run medians.
    The inset magnifies the first \num{5000} steps.
    The increase of about \qty{3.7}{\percent} from r02 to r03 comes from decreasing the logging and profiling frequency.
  }
  \label{fig:bienenelfe-tps}
\end{figure}

To balance the workload across all expert-parallel ranks, we rely on our \ac{MoE} load-balancing methodology as described in~\cref{sec:load-balancing}.
The per-GPU throughput over time in tokens per second is visualised in~\cref{fig:bienenelfe-tps}.
The throughput is stable after an initial dip.
The dip reflects the time until the routing first stabilises, such that the load balancing can be effective.

\subsubsection{Straggler Detection}\label{sec:straggler-detection}

Our straggler detection approach is threefold: (1)~We do passive health monitoring via \ac{DCGM}. (2)~We actively measure the compute time of each transformer block and compare them online. (3)~Most intrusively, and thus offline, we take PyTorch profiles on all GPUs and match collective communication kernels across the profiles to figure out when each GPU joined each collective and what throughput it observed.
We use three distinct approaches because they can be used online or offline and detect different kinds of stragglers or straggler reasons, like compute or communication.

To obtain online runtime measurements, we follow the approach of \textcite{jiang2024megascale} and time each transformer block during a forward pass with CUDA events.
We inject these events through the \texttt{register\_forward\_pre\_hook} and \texttt{register\_forward\_hook} methods of \texttt{torch.nn.Module}.
To avoid fully draining the GPU work queue when evaluating the CUDA events, we use double-buffering.
This means that each training step evaluates the CUDA events of its preceding step.
We explicitly gather the runtimes on GPU/process 0 to get more detailed statistics than just the average runtime.

For detecting complicated compute or communication straggler reasons, we use a profiling-based approach.
Our main training run writes a PyTorch profile once every 503 steps (about \qty{50}{\minute}; 101 steps before r03), and we match the collective communication kernels between the profiles of all ranks and compare their arrival time and observed throughput.
Take, for example, one \ac{FSDP} weight all-gather.
If the wall-clock duration of such an all-gather is significantly shorter on one rank than on others, it likely arrived later at the collective than all other ranks.
The matching of these collectives across the profiles will tell us if this is the case.
One possible reason for a late arrival could be that the process is a compute straggler; the profile will also show us this reason.
If all ranks communicate at a rate considerably slower than the maximum throughput, a communication straggler might be present.

\begin{figure}[t]
  \centering
  \resizebox{0.49\linewidth}{!}{%
    \includegraphics{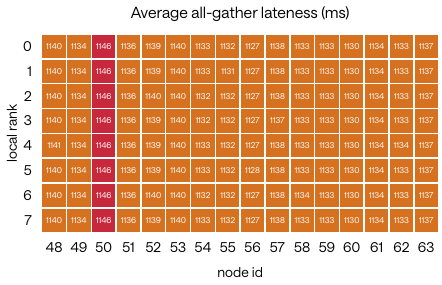}%
  }\hfill{}
  \resizebox{0.49\linewidth}{!}{%
    \includegraphics{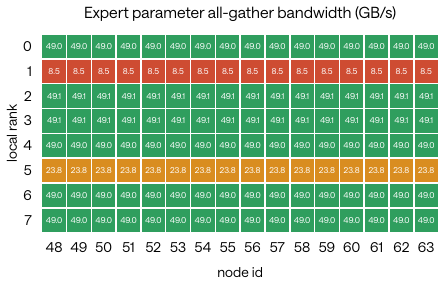}%
  }
  \caption{Average all-gather (over all process groups) lateness per node and local rank (left) and expert parameter all-gather bandwidth (right) for nodes 48--63 (\ac*{DP}/model replica 3).
    The intersection is local ranks 1 and 5 on node 50.
    These visualisations stem from a profile of a timestep, where the PCIe links of \acsp*{NIC} 1 and 5 on node 50 were downtrained, i.e.\ the GPU-to-\acs*{NIC} communication speed was degraded.
  }
  \label{fig:all-gather-heatmaps}
\end{figure}

We identify communication stragglers by comparing the collective communication arrival times and throughput across all parallelism dimensions separately.
Our main training run ran with \ac{EP}~\num{8}, \ac{FSDP}~\num{16}, \ac{DP}~\num{6}.
This gives us \emph{four} different communication groups for each rank.
Take rank~0 as an example.
It communicates with the following partners: (1)~with ranks~1--7 expert parallel (all-to-all), (2)~with ranks~8, 16, 24,~\ldots,~120 during expert parameter all-gathers/reduce-scatters, (3)~with ranks~1--127 during dense parameter all-gathers/reduce-scatters, and (4)~with ranks~128, 256,~\ldots,~640 during data-parallel all-reduces.
If the InfiniBand speed of rank~0 were reduced, we would notice this in groups~(2)--(4).
An example of such an analysis is visualised in \cref{fig:all-gather-heatmaps}.
With this methodology, we were able to pinpoint every communication straggler that occurred during the main pre-training run.

\subsubsection{Fault Tolerance}
\label{sec:fault-tolerance}

We observed \num{38} unplanned interruptions during the net \num{377000} GPU hours pre-training run.
This yields a rate of \num{0.1} failures per \num{1000} GPU hours.
\Cref{tab:failure-reasons} lists the reasons for the observed failures.
The single most frequent hardware failure reported by \ac{DCGM} was \enquote{GPU has fallen off the bus} (Xid 79).
Xids are the identifiers of the errors that the NVIDIA driver reports to the kernel, and Xid 79 \enquote{GPU has fallen off the bus} indicates that the driver attempts to but cannot access the GPU over PCIe \parencite{nvidia2026xid}.

\begin{table}
  \centering
  \small
  \begin{tabular}{lrr}
    \toprule
    Reason & Number of occurrences & Details \\
    \midrule
    GPU / node hardware fault       & \num{10} & Xid 79, 94, 145, 154 \\
    Object-storage read stall (S3)  & \num{9}  & Timeout (cURL error code 28) during \texttt{GetObject} \\
    Bad link                        & \num{2} \\
    Checkpoint-save timeout         & \num{1}  & \ac{DCP} checkpoint write to CephFS timed out \\
    Not investigated / inconclusive & \num{16} & \\
    \midrule
    Total & \num{38} & \\
    \bottomrule
  \end{tabular}
  \caption{All failures that occurred during the main pre-training run, classified by root cause.
    The most frequent attributed cause was a GPU or node hardware fault (for example NVLink dropped or \acs*{NIC} problems).
    Our system classified these automatically, with verification by humans.
    Some failures could not be attributed, and we thus list them as \enquote{inconclusive}.
  }
  \label{tab:failure-reasons}
\end{table}

Our fault-tolerance setup combines frequent checkpointing with automatic recovery.
This setup is the most compatible with our current cluster orchestration, which requires reboots after hardware failures.

We checkpoint every \num{250} training steps.
For failures that are uncorrelated to the checkpointing, this means that we need to recompute on average \num{125} steps.
With a step time of about \qty{6}{\second} that makes \qty{12.5}{\minute}.
As shown in \cref{tab:recovery-times}, the observed recompute time happens to be lower than expected.

Our cluster orchestration automatically detects faulty nodes and swaps them out of the main training run for automatic reboot and/or manual inspection.
The Model Factory (cf.\ \cref{sec:model-factory}) classifies failures as application- or system-level errors and retries them within their separate budgets.
The majority of the time is spent in a grace period for the remaining nodes to checkpoint their state.
Together, this took on average \qty{29.4}{\minute} as shown in \cref{tab:recovery-times}.
This average includes every failure.
Initially, we swapped out the nodes by hand, resulting in downtimes of \qtyrange{30}{60}{\minute}.
After that, we put the automatic cluster recovery in place and the average downtime decreased drastically.
We still include these numbers in the calculation of the average of \qty{29.4}{\minute}.

The third component in our fault handling is the application startup.
This is relatively fixed at \qty{7.6}{\minute} and includes launching the application, rendezvous of all GPUs, \texttt{torch.compile} as well as checkpoint loading.

\begin{table}
  \centering
  \small
  \begin{tabular}{@{}l>{\raggedright\arraybackslash}p{0.55\linewidth}rr@{}}
    \toprule
    Step & Explanation & Average time & Total time \\
    \midrule
    Downtime
    & Hot-swap of faulty node, automatic job restart, Kubernetes scheduling
    & \qty{29.4}{\minute}
    & \qty{18.6}{\hour} \\
    Startup
    & Distributed launch, rendezvous, \texttt{torch.compile}, checkpoint loading
    & \qty{7.6}{\minute}
    & \qty{4.8}{\hour} \\
    Recompute
    & Recomputation of training steps between last checkpoint and failure
    & \qty{8.2}{\minute}
    & \qty{5.2}{\hour} \\
    \midrule
    Total &  & \qty{45.3}{\minute} & \qty{28.7}{\hour} \\
    \bottomrule
  \end{tabular}
  \caption{Time for automatic recovery of the \num{38} failures.
    After the first couple of failures, we increased checkpoint frequency to lower the \enquote{recompute} time reported in the table.
    The average nonetheless includes all samples and is therefore higher than the time we observed during most of the run.
    The totals are computed from unrounded values.
  }
  \label{tab:recovery-times}
\end{table}

\FloatBarrier

%% file: sections/09-appendix/C-optimisation.tex
\section{Optimisation}
\label{app:optimisation}

\subsection{Training Batch Construction}
\label{sec:batch-construction}

Training batches should represent the dataset distribution.
\Acl*{SGD} relies on each microbatch gradient being an unbiased, low-variance estimate of the gradient over the whole training distribution.
That estimate is unbiased only if the examples in a batch are drawn independently from that distribution \parencite{bottou2018optimization}.
Insufficient shuffling violates this \parencite{bottou2012sgdtricks}, and so do long documents that spread over consecutive batches.
Both have two consequences: (1) within a batch, correlated examples shrink the effective batch size, and (2) across batches, the mixture the model sees drifts over time, so the model is chasing a moving target and fits disproportionately what it saw most recently.
A third problem is cutting documents: a document longer than the sequence length has to be cut into pieces, and every cut separates tokens from context they could otherwise attend to.
The following subsections address these problems.

\subsubsection{Data Loading Pipeline}\label{sec:data-loading-pipeline}
The training mix is assembled offline according to the mix of each stage, which Olmix \parencite{chen2026olmix} selects for pre-training (\cref{sec:pre-training-data}) and MergeMix \parencite{wang2026mergemix} for mid-training (\cref{sec:mid-training}).
Our materialisation job writes the resulting mix into parquet files with row groups of \num{4096} rows and emits a row index alongside the data.
An additional index file fixes the order in which documents are visited, at the granularity of row groups.
Dataloader workers load entire row groups and iterate over all documents in each loaded row group sequentially.
This approach keeps the number of S3 GET requests low and avoids overloading the storage system.
Since the order is materialised once, every rank reads the same permutation, and a run can be reproduced or resumed from the index alone.
At training time we build on the standard PyTorch dataloader with several worker processes per rank.
The document stream is split so that each rank sees a disjoint slice of the index and each worker of a rank a disjoint slice of that, and every worker then runs the same per-worker pipeline:
\begin{enumerate}
  \item reads its row groups of raw text;
  \item unrolls them into individual documents;
  \item tokenises each document online and appends the end-of-sequence token;
  \item splits documents longer than the sequence length into chunks with the \ac{RSDB} described below;
  \item bin-packs chunks into sequences of exactly the sequence length;
  \item converts each packed sequence into input, label and position tensors.
\end{enumerate}
Positions restart at zero at every document boundary within a packed sequence, which is how the attention backends recover the document mask.
The dataloader collates the sequences of one worker into a microbatch, and its full state, from the position in the index down to the buffer contents of every worker, is checkpointed so that a resumed run continues the exact same data order.

\subsubsection{Chunking and Packing}\label{sec:chunking-packing}
The common way to fill sequences of a fixed length \(L\) concatenates the tokenised documents, separated by end-of-sequence tokens, and splits the stream at every multiple of \(L\) \parencite{ding2024fewertruncations}.
This concat-and-chunk packing has two problems.
First, it cuts documents that would fit into one sequence: almost every sequence boundary falls inside a document, so a document shorter than \(L\) is split whenever it straddles a boundary, and the tokens after the cut lose the context before it.
\textcite{ding2024fewertruncations} report better downstream scores when these cuts are avoided.
Second, a document many times longer than \(L\) fills several consecutive sequences on its own.
A document of \num{200000} tokens at a sequence length of \num{16}k occupies \num{13} consecutive sequences.
With sequence-order packing, those land in a handful of adjacent microbatches, so for a few optimiser steps the gradient is dominated by one document, one domain and one writing style.
Across the whole run every token is still seen once, but the per-step batches no longer follow the dataset distribution.
Shuffling documents before chunking does not address this: the pieces of one document still come out back to back.
The \ac{RSDB} addresses the second problem and the bin packer the first.

\paragraph{\actitle{RSDB}}
We break this locality during data loading with an \ac{RSDB}, following \textcite{singh2026arceetrinitylargetechnical}.
Each dataloader worker keeps a buffer of \num{4096} tokenised documents together with a read offset per document.
To produce the next piece, the worker draws a buffered document uniformly at random, emits the next contiguous chunk of at most one sequence length starting at that document's offset, and advances the offset.
A document leaves the buffer when its last chunk has been emitted, and the slot is refilled from the input stream.
A document of \(k\) chunks is therefore drawn \(k\) times over the lifetime of the buffer.
Its chunks are separated in expectation by roughly one buffer's worth of chunks from other documents.
Within a document the chunks keep their original order; only their spacing in the training stream changes.
\textcite{singh2026arceetrinitylargetechnical} report a smoother training loss with the \ac{RSDB} than with sequential chunking, and we observe the same.

\paragraph{Bin Packing}
The chunks then go into the bin packer, which fills each sequence of exactly the sequence length from a pool of \num{1024} pending chunks.
The packer always anchors a sequence on the oldest chunk in its pool, which bounds how long any chunk can wait, and fills the remaining capacity with a subset-sum search over the rest.
If no subset fills the sequence exactly, the packer takes the smallest subset that overfills it and drops the tokens beyond the sequence length.
At sequence length \(L\), a document of \(n\) tokens needs \(\lceil n / L \rceil - 1\) cuts.
The \ac{RSDB} makes exactly these cuts, and the packer adds a cut only when no subset fills a sequence exactly.
Since the packer can only reorder chunks within its pool of \num{1024}, while the \ac{RSDB} separates the chunks of one document by several thousand chunks from other documents, pieces that the \ac{RSDB} emitted far apart stay far apart, and a long document's chunks end up spread over many microbatches rather than concentrated in a few.
Positions restart at zero at every chunk, so the model sees the pieces of a long document as independent contexts, but the ordering no longer concentrates them in time.

%% file: sections/09-appendix/D-data-and-training-stages.tex
\section{Data and Training Stages}
\label{app:data-stages}

\subsection{\Acl*{CC} Internal Pipeline}
\label{sec:cc-internal-pipeline}

This appendix describes the stages of the \ac{CC} pipeline of \cref{sec:cc-pipeline} in more detail, and \cref{fig:cc-funnel} shows the documents each stage received and removed.

\subsubsection{Heuristic Filters}
\label{sec:heuristic-filters}

\begin{figure}[tb]
  \centering
  \includegraphics{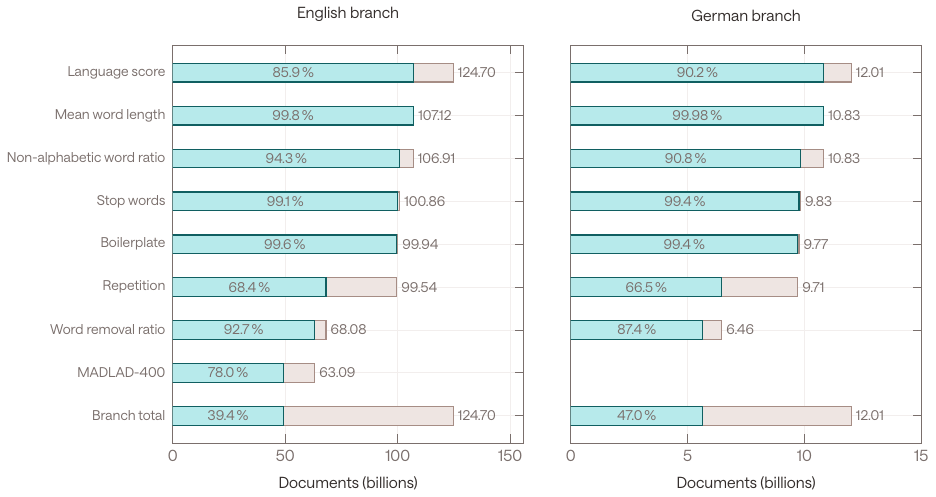}
  \includegraphics{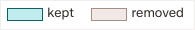}
  \caption{Documents entering each heuristic filter step of the English and German branches, split into the documents the step kept and removed, summed over all processed \acs*{CC} snapshots.
    The label in the kept segment is the step's yield, the label at the end of the bar the documents it received in billions.
    Line modifiers drop almost no documents and are omitted, and the MADLAD-400 filter runs in the English branch only.
  }
  \label{fig:cc-heuristic-yield}
\end{figure}

The heuristic filters are based on and extend those of Dolma 3 \parencite{olmo2026olmo3}, which derive from \ac{DCLM} \parencite{li2024dclm} and RefinedWeb \parencite{penedo2023refinedweb}.
Every extracted document passes the language-agnostic steps of \cref{tab:cc-heuristic-agnostic}, and the language identification routes it to an English or a German branch, which applies the steps of \cref{tab:cc-heuristic-thresholds}.
\Cref{fig:cc-heuristic-yield} shows the yield of each step.

\begin{table}[tb]
  \centering
  \small
  \begin{tabular}{@{}l>{\raggedright\arraybackslash}p{0.65\linewidth}@{}}
    \toprule
    Step & Drop when \\
    \midrule
    Non-null check & the extracted text is missing or empty \\
    \midrule
    Character count
    & fewer than \num{150} characters \\
    Word count
    & fewer than \num{50} or more than \num{100000} words \\
    Symbol ratio
    & more than \num{0.1} symbols per word, where a symbol is a hash sign or an ellipsis \\
    Bullet lines
    & more than \qty{90}{\percent} of lines start with a bullet character, an asterisk or a hyphen \\
    Ellipsis lines
    & more than \qty{30}{\percent} of non-empty lines end with an ellipsis or \enquote{read more} \\
    Uppercase lines
    & more than \qty{50}{\percent} of non-empty lines have more than half of their characters in uppercase \\
    \midrule
    Block list & the host is a blocked domain, the URL is a blocked page, or the URL contains a banned word \\
    \midrule
    Language identification & the fastText top label over the first \num{5000} bytes is neither English nor German \\
    \bottomrule
  \end{tabular}
  \caption{Language-agnostic heuristic filter steps, in pipeline order.
    The word-count, symbol-ratio, bullet-line and ellipsis-line steps are Gopher document filters \parencite{rae2021gopher} at their default thresholds.
  }
  \label{tab:cc-heuristic-agnostic}
\end{table}

\begin{table}[tb]
  \centering
  \small
  \begin{tabular}{@{}l>{\raggedright\arraybackslash}p{0.3\linewidth}>{\raggedright\arraybackslash}p{0.15\linewidth}>{\raggedright\arraybackslash}p{0.2\linewidth}@{}}
    \toprule
    Step & Filters / Modifiers & English & German \\
    \midrule
    Language score
    & fastText confidence score below
    & \num{0.65}
    & \num{0.821} \\
    Mean word length
    & mean word length in bytes outside
    & \numrange{3}{10}
    & \numrange{0}{15} \\
    Non-alphabetic word ratio
    & share of words without an alphabetic character above
    & \num{0.20}
    & \num{0.226} \\
    Stop words
    & fewer than two occurrences of a word from
    & \num{8} Gopher words
    & \num{15} FineWeb2 words \\
    Boilerplate
    & more than \qty{40}{\percent} of paragraphs contain a phrase from
    & \num{13} English phrases
    & \num{16} German and the \num{13} English \\
    Repetition
    & duplicate line fraction above
    & \num{0.30}
    & \num{0.263} \\
    & duplicate paragraph fraction above
    & \num{0.30}
    & \num{0.30} \\
    & characters in duplicate lines or paragraphs above
    & \num{0.20}
    & \num{0.20} \\
    & most frequent 2-, 3-, 4-gram fraction above
    & \num{0.20}, \num{0.18}, \num{0.16}
    & \num{0.169}, \num{0.153}, \num{0.137} \\
    & repeated 5- to 10-gram fraction above
    & \numrange{0.15}{0.10}, step \num{0.01}
    & \num{0.141}, \num{0.134}, \num{0.126}, \num{0.117}, \num{0.108}, \num{0.098} \\
    Line modifiers
    & digit-only lines, counter lines such as \enquote{3 likes}, lines with fewer than two words
    & \multicolumn{2}{l}{shared} \\
    Substring line modifier
    & lines of at most ten words that contain one of these substrings
    & \enquote{items in cart}, \enquote{Read more...}, \enquote{Sign-in}
    & \enquote{Artikel im Warenkorb}, \enquote{Mehr lesen...}, \enquote{Weiterlesen...}, \enquote{Anmelden}, \enquote{Einloggen} \\
    Word removal ratio
    & share of words removed by the line modifiers above
    & \num{0.05}
    & \num{0.05} \\
    MADLAD-400 rules 2 and 5
    & fewer than five sentences, or at least \qty{20}{\percent} questionable sentences
    & applied
    & not applied \\
    \bottomrule
  \end{tabular}
  \caption{Heuristic filter steps of the English and German branches, in pipeline order, with the threshold of each branch.
    English thresholds follow Gopher; German thresholds and the German stop-word list follow the FineWeb2 \texttt{deu\_Latn} configuration.
    The paragraph and character thresholds are the Gopher defaults in both branches.
    The English branch restricts the MADLAD-400 filter \parencite{kudugunta2023madlad} to rules 2 and 5, following the ablations of Dolma 3.
  }
  \label{tab:cc-heuristic-thresholds}
\end{table}

A cleanup pass runs on every dataset of the pre-training, mid-training, and long-context pools.
It drops empty and degenerate documents, collapses runs of newlines in a random share of the documents to at most three in code and math sources and two elsewhere, and caps horizontal whitespace at \num{16} characters within a line and \num{256} characters of indentation.

\subsubsection{Exact Deduplication}
\label{sec:exact-dedup}

Exact deduplication removes every document whose UTF-8 text, without normalisation, has the same \num{128}-bit xxHash \parencite{collet2026xxhash} as an earlier document, first within each snapshot and then across all snapshots in chronological order, so the earliest copy remains \parencite{lee2022deduplicating}.
Each kept document records three counts: its copies within its snapshot, its copies in the corpus, and the number of snapshots that contain a copy.
The copy count correlates with document quality \parencite{fang2025datasets}, and TxT360 \parencite{txt360} and FineWeb2 \parencite{penedo2025fineweb2} upsample by it.
The counts thus leave the decision of how often to repeat a document to the mixing stage, without rerunning deduplication.
The global pass stores only the hash table with the counts and the location of each document's text.
A new snapshot therefore merges into the table at a cost proportional to the table and the snapshot, independent of the number of earlier snapshots.
We apply no URL deduplication, since one URL can hold distinct documents across crawls.

\subsubsection{Fuzzy Deduplication}\label{sec:fuzzy-dedup}

Fuzzy deduplication uses MinHash \parencite{broder1997resemblance} with locality-sensitive hashing.
Documents are normalised as in RefinedWeb and CCNet \parencite{wenzek2020ccnet}, with every digit replaced by a placeholder, and shingled into token 5-grams.
The signatures have \num{26} bands of \num{11} rows, which pair two documents with probability one half at a Jaccard similarity of \num{0.72} and \qty{99}{\percent} at \num{0.85}.
This budget is smaller than those of RefinedWeb and \ac{DCLM}, since every cluster of candidate pairs is verified on the Jaccard similarity of token 3-grams at a threshold of \num{0.80}.
The verification is exhaustive for clusters below \num{500} documents and uses a second MinHash pass with \num{200} bands of \num{31} rows for larger ones.
Each kept document records the size of its cluster.
The stage runs per language in \num{30} English and \num{6} German shards, to which the snapshots are dealt round-robin in chronological order, so near-duplicates in different shards remain.
It removes \qty{24.9}{\percent} of the English and \qty{33.3}{\percent} of the German documents.

\subsubsection{Substring Deduplication}
\label{sec:substring-dedup}

Substring deduplication follows the suffix-array method of \textcite{lee2022deduplicating}, which keeps one occurrence of every repeated byte sequence of at least \num{500} bytes and cuts the others out of their documents.
We keep the occurrence earliest in a shard's byte order, fuse adjacent removal ranges that cover at least \qty{95}{\percent} of the fused span, and drop a document with fewer than \num{100} bytes left.
The stage runs per language in \num{41} English shards of at most three snapshots and \num{7} German shards of \num{16} to \num{19} snapshots.
It removes \qty{20.4}{\percent} of the English and \qty{29.6}{\percent} of the German bytes.
The cuts join the text on either side without a check for coherence, and cut documents skip the heuristic filters.
\subsection{Mid-Training Details}
\label{app:mid-details}

This section gives the parts of \cref{sec:mid-training} that the main text summarises or omits: the ablation setup, the mixing method, more detailed ablation results, the language sweep, decontamination, the transition from the pre-training mix, and the production run.

\subsubsection{Ablation Setup}
\label{sec:mid-ablations}

\subparagraph{Proxy Model and Budgets}
The proxy model has \num{384} experts and a hidden size of \num{1024}.
Its pre-training horizon of \num{3.05}T tokens matches the \num{256} \acl*{TPP} of production, \num{20}T tokens over \num{78.1}B total parameters.
Every ablation loads the proxy pre-training checkpoint at step \num{166000}, after \num{2.79}T tokens or \qty{91}{\percent} of the matched horizon, and mid-trains for \num{5000} steps at \num{16.8}M tokens per step (\num{84}B) or \num{23842} steps (\num{400}B).
The mixes of the \num{84}B runs were materialised for a \num{90}B budget.

\subparagraph{Merged Checkpoints}
\label{sec:ablation-merges}
At \num{84}B the M10 merge averages the ten checkpoints saved every \num{500} steps, which is the whole run.
At \num{400}B, M10 averages ten checkpoints \num{2100} steps (\num{35}B tokens) apart from step \num{5000} to the endpoint, and M4 averages four checkpoints \num{4200} steps (\num{70}B tokens) apart ending at the endpoint.

\subparagraph{Aggregate Score}
\label{sec:ablation-final-score}
The benchmarks of the aggregate score come from the pre-training suite (\cref{sec:pre-training-eval-scores}).
The general-knowledge mean averages \ac{MMLU}~\parencite{hendrycks2021mmlu}, \ac{MMLU}-Pro~\parencite{wang2024mmlupro}, \ac{ARC}~\parencite{clark2018arc}, HellaSwag~\parencite{zellers2019hellaswag}, \ac{PIQA}~\parencite{bisk2020piqa}, and TriviaQA~\parencite{joshi2017triviaqa} in English, and Global \ac{MMLU}~\parencite{singh2025globalmmlu}, \ac{ARC}, HellaSwag, and \ac{PIQA} in German.
The math mean averages \ac{GSM8K}~\parencite{cobbe2021gsm8k} and MATH~\parencite{hendrycks2021math}, and the code mean HumanEval~\parencite{chen2021humaneval} and \ac{MBPP}~\parencite{austin2021mbpp}, each in both languages.
HumanEval is affected by contamination (\cref{sec:mid-decon}), so the tables also report the aggregate score without HumanEval, which drops the English and German HumanEval from the code means, leaves \ac{MBPP} as the code benchmark, and keeps the rest of the formula unchanged.

\subparagraph{\Acs*{SFT} Recipe and Eval Noise}
The \ac{SFT} recipe of this section is \num{1000} steps at a context of \num{256}k tokens with a global batch of \num{64} sequences and a peak learning rate of \num{0.5} of pre-training.
Its data is one mix of \num{26} datasets, separate from the \model{2} \ac{SFT} mix (\cref{sec:sft}), with \qty{54.5}{\percent} chat, \qty{20.7}{\percent} math, \qty{10.8}{\percent} science, and the rest agentic, instruction following, German, and the model's identity data.
The \ac{SFT} models are scored on the following post-training benchmarks.
The public benchmarks included: AIME 2026 \parencite{maa2026aime} and its German translation, MATH-500 \parencite{lightman2023verify}, GPQA Diamond \parencite{rein2023gpqa}, IFEval
\parencite{zhou2023ifeval}, IFBench \parencite{pyatkin2025ifbench}, LongBench Pro %
\parencite{chen2026longbenchpro} at two context lengths, HumanEval+ \parencite{liu2023evalplus}, %
and LiveCodeBench \parencite{jain2024livecodebench}, as well as two further benchmarks.
Two are internal German benchmarks, one for science and mathematics and KulturBank for cultural knowledge (\cref{app:sft-kulturbank}).
A majority of the benchmarks have an eval-noise model estimated from \num{20} replicate \acp{SFT} that differ in seed only.
The model gives each score a \qty{95}{\percent} interval from the binomial standard error of \textcite{miller2024errorbars}, with an effective sample size fitted to the replicates, so the interval depends on the score and on the benchmark.
It also gives a pooled score, the mean over the calibrated benchmarks, with the variances summed and the degrees of freedom pooled by Welch-Satterthwaite \parencite{welch1947generalization,satterthwaite1946approximate}.
The interval describes how precisely one score is known.
Whether two runs differ is a test of their difference against the noise of both runs, which resolves a difference of about \num{1.4} times one half-width.
Two disjoint intervals need about twice the half-width and are the more conservative test, so two runs with overlapping intervals can still differ.

\subsubsection{Mixing Method}
\label{sec:mid-mixing}
\label{app:mid-mixing-method}

Besides the mid-training pool of \cref{sec:mid-pool} and the two sampling rules of \cref{sec:mid-method}, the language sweep (\cref{sec:mid-lang-sweep}) uses a second dataset pool and a third sampling rule for the two language domains.

\subparagraph{Repetition Caps}
A dataset's mid-training repetition cap is four epochs minus its epochs in the \num{20}T tokens of pre-training.
At the \num{400}B and \num{90}B budgets no repetition cap binds.
At the \num{5.2}T production budget the waterfill pins every dataset of \domDeInstr{} and four datasets of \domStemQA{} (\cref{sec:mid-method}).
The weight of \domDeInstr{} is the most its repetition caps allow, and every domain receives its full weight.

\paragraph{Optimisation of the Domain Weights}
Each expert trains on the datasets of its domain under the pre-training-delivered rule.
All experts start from the same checkpoint as every other proxy ablation (\cref{sec:mid-ablations}), so their parameters can be averaged \parencite{wortsman2022soups}.
We keep the in-domain composition fixed, since optimising it with the same method requires one expert per dataset, \num{87} training runs in total.

\paragraph{Sampled Weightings}
The sampled weightings centre on the domain weights of C1 and C2, with \num{30} at concentration \num{1}, \num{30} at concentration \num{2}, and \num{40} spread over concentrations from \num{0.1} to \num{5}, the range RegMix~\parencite{liu2024regmix} samples from.
The hand-placed weightings step each of the four smallest domains, \domDeInstr{}, \domSpecialized{}, \domAgentic{}, and \domStemDocs{}, from zero to the largest weight the repetition caps of its datasets allow, with the other domains scaled along the centre.

\paragraph{Solved Weights}
The \ac{KL} penalty is the regulariser of Olmix's exact solver, for which \textcite{chen2026olmix} recommend a weight of \num{0.05}, and the production mix uses \num{0.01} (\cref{sec:mid-effect}).
The repetition-cap constraint of the solve (\cref{sec:mid-method}) limits each domain's weight to the share of the budget its datasets can fill under their repetition caps.
The constraint bounds domains only, so a dataset can still be pinned inside its domain at the production budget.
In all four projected solutions (\cref{tab:mid-solutions}), the projection changes the in-domain composition of \domDeInstr{} and \domStemQA{}, in all but S4 that of \domEnInstr{}, and in all but S6 that of \domCode{}.
The solve and the projections use the repetition caps of a \num{30}T pre-training, under which \domDeInstr{} fills at most \qty{1.4}{\percent} of \num{5}T, the weight every solution gives it (\cref{tab:mid-candidates}).
The production mix uses the repetition caps of the \num{20}T pre-training run, under which fewer datasets are pinned and \domDeInstr{} fills \qty{1.9}{\percent} of \num{5.2}T. The production mix raises \domDeInstr{} to that weight and scales the other eight domains of S5 down by a common factor (\cref{sec:mid-effect}).
\Cref{sec:mid-ablation-results} compares the six solutions to the hand-set candidates.

\begin{table}[htbp]
  \centering
  \small
  \begin{tabular}{llrl}
    \toprule
    Mix & Prior & \acs*{KL} penalty & Projection \\
    \midrule
    S1&C2&$0.00$&from \num{5}\mbox{T}\\
    S2&C2&$0.01$&--\\
    S3&C2&$0.01$&from \num{5}\mbox{T}\\
    S4&C2&$0.05$&from \num{5}\mbox{T}\\
    S5&C3&$0.01$&--\\
    S6&C3&$0.01$&from \num{5}\mbox{T}\\
    \bottomrule
  \end{tabular}
  \caption{The reference mix and \acs*{KL} penalty of each materialised optimised solution, and whether its in-domain composition is projected from a \num{5}\mbox{T} budget (\cref{sec:mid-method}).
    \Cref{tab:mid-candidates} lists their domain weights.
  }
  \label{tab:mid-solutions}
\end{table}

\subsubsection{Ablation Results}
\label{sec:mid-ablation-results}

\begin{figure}[htbp]
  \begin{minipage}[c]{0.5\linewidth}
    \centering
    \small
    \begin{tabular}{lccc}
      \toprule
      & \num{84}\mbox{B} & \multicolumn{2}{c}{\num{400}\mbox{B}} \\
      \cmidrule(lr){2-2}\cmidrule(lr){3-4}
      Mix & M10 & M10 & M4 \\
      \midrule
      C2&$60.9$&$62.2$&$61.1$\\
      C3&$60.9$&$63.6$&$62.4$\\
      production mix&$62.2$&--&$65.5$\\
      continued pre-training&--&--&$60.6$\\
      \bottomrule
    \end{tabular}
    \captionof{table}{Aggregate score of the mixes trained at both proxy budgets, at the merged checkpoints.
      The continued pre-training baseline is an M4 of the proxy pre-training run at \num{3.2}T tokens, whose four checkpoints lie \num{80}B tokens apart, against \num{70}B for the mixes.
    }
    \label{tab:mid-horizon}
  \end{minipage}\hfill
  \begin{minipage}[c]{0.47\linewidth}
    \centering
    \includegraphics{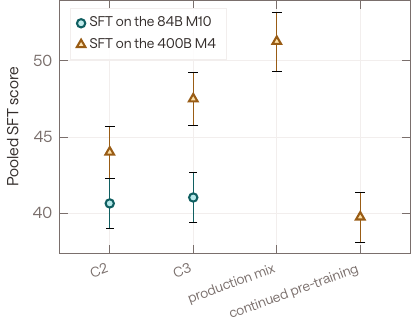}
    \caption{Pooled score of the \acs*{SFT} trained on the merges of \cref{tab:mid-horizon}, on the M10 of the smaller budget and on the M4 of the larger, with the \qty{95}{\percent} eval-noise interval as error bar.
    }
    \label{fig:mid-horizon}
  \end{minipage}
\end{figure}

\paragraph{Budget}
The pair tests of the adjacent \num{400}B runs (\cref{sec:mid-lessons}) resolve a difference of \num{2.4} points between continued pre-training and C2, \num{2.5} between C2 and C3, and \num{2.6} between C3 and the production mix, and each of the three gaps exceeds its threshold at \(p < \num{0.01}\) (\cref{tab:mid-horizon,fig:mid-horizon}).

\paragraph{Ranking}
The ranking of \cref{tab:mid-ranking} covers every \num{84}B run, the budget at which every mix was trained.
These runs are the three hand-set candidates, the transition run of C1, the domain-weight sweep, the language sweep, the production mix, and the \model{1} mid-training mix, an \num{80}/\num{20} blend of high-quality English web data and its translation into German.
Runs whose mixes differ only in the projection (S2 and S3, S5 and S6), in the transition (C1 and C1 transition), or in the sampling rule of the language domains (L1 and L2, L3 and L4) score within \num{0.6} points of each other on the merge, the spread against which we read the ranking.
S3 leads the hand-set candidates by \numrange{1.2}{2.0} points, beyond that spread.
The production mix ranks fifth of \num{18}, within \num{0.7} points of the four runs above it.

\begin{figure}[htbp]
  \centering
  \includegraphics{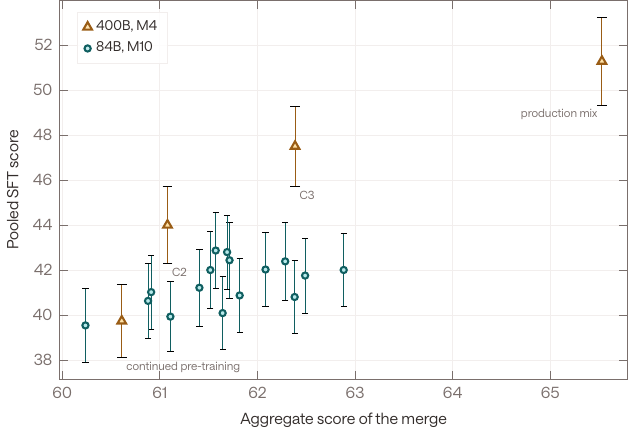}
  \caption{Pooled score of the \acs*{SFT} on each merge against the aggregate score of the merge, with the \qty{95}{\percent} eval-noise interval as error bar, for the M4 merges of the four \num{400}\mbox{B} runs and the M10 merges of the \num{16} runs at \num{84}\mbox{B} with an \acs*{SFT}.
  }
  \label{fig:mid-merge-sft}
\end{figure}

\paragraph{Transfer from Merge to \Acs*{SFT}}
For the four \num{400}B runs of \cref{sec:mid-lessons} and the \num{16} runs at \num{84}B with an \ac{SFT}, \Cref{fig:mid-merge-sft} plots the pooled \ac{SFT} score against the aggregate score of the merge the \ac{SFT} starts from.
At \num{84}B the eval noise alone has a standard deviation of \num{0.84} points, against \num{1.04} for the \ac{SFT} scores, so two thirds of their variance is noise.
Of the \num{120} pair tests among these \acp{SFT}, \num{14} clear the noise, against \num{6} expected by chance, and each of them sets one of the seven highest \acp{SFT} against one of the three lowest.

\begin{table}[htbp]
  \centering
  \footnotesize
  \setlength{\tabcolsep}{4pt}
  \begin{tabular}{lcccccc}
    \toprule
    & \multicolumn{2}{c}{M10} & \multicolumn{2}{c}{M10 w/o HumanEval} & \multicolumn{2}{c}{\acs*{SFT}} \\
    \cmidrule(lr){2-3}\cmidrule(lr){4-5}\cmidrule(lr){6-7}
    Mix & score & rank & score & rank & score & \(\pm\) \\
    \midrule
    S3&$62.9$&$1$&$61.9$&$1$&$42.0$&$1.6$\\
    L6&$62.5$&$2$&$61.7$&$2$&$41.8$&$1.7$\\
    S2&$62.4$&$3$&$61.0$&$7$&$40.8$&$1.6$\\
    S6&$62.3$&$4$&$61.0$&$6$&$42.4$&$1.8$\\
    production mix&$62.2$&$5$&$61.2$&$3$&--&--\\
    L3&$62.1$&$6$&$60.8$&$11$&$42.0$&$1.7$\\
    S1&$61.8$&$7$&$60.6$&$12$&$40.9$&$1.7$\\
    L4&$61.7$&$8$&$61.2$&$4$&$42.4$&$1.7$\\
    S5&$61.7$&$9$&$61.1$&$5$&$42.8$&$1.6$\\
    C1&$61.6$&$10$&$61.0$&$8$&$40.1$&$1.6$\\
    L2&$61.6$&$11$&$60.8$&$10$&$42.9$&$1.7$\\
    L1&$61.5$&$12$&$61.0$&$9$&$42.0$&$1.7$\\
    L5&$61.4$&$13$&$60.1$&$15$&$41.2$&$1.7$\\
    S4&$61.1$&$14$&$60.3$&$13$&$39.9$&$1.6$\\
    C1 transition&$61.1$&$15$&$59.7$&$18$&--&--\\
    C3&$60.9$&$16$&$59.8$&$17$&$41.0$&$1.6$\\
    C2&$60.9$&$17$&$60.2$&$14$&$40.6$&$1.7$\\
    \model{1} mid-training mix&$60.2$&$18$&$59.9$&$16$&$39.5$&$1.6$\\
    \bottomrule
  \end{tabular}
  \caption{Every proxy run at the smaller budget, ordered by the aggregate score of its M10 merge, with the score and rank of the merge and of the merge without HumanEval, and the pooled score of the \acs*{SFT} on the merge with the half-width of its \qty{95}{\percent} eval-noise interval.
    C1 transition is C1 entered through the transition of \cref{sec:transition-index}, and every other run switches directly from the pre-training mix.
    The merges of C2, C3, and the production mix average the first ten checkpoints of their \num{400}B runs.
    Dashes mark the two runs without an \acs*{SFT}.
  }
  \label{tab:mid-ranking}
\end{table}

\paragraph{Domain Weights}
The domain weights covary within each sweep, so a correlation of a domain weight with a score (\cref{tab:mid-weight-correlations}) describes this set of mixes and includes the effect of the domains that covary with it.
The aggregate score and the math means rise with \domEnInstr{} and fall with \domEnWeb{} and \domSpecialized{}.
TriviaQA rises with \domEnWeb{}, at \(r = \num{0.81}\), and HellaSwag (DE) and the German general-knowledge mean rise with \domDeWeb{}, at \num{0.84} and \num{0.82}.
The English general-knowledge mean barely moves with \domEnWeb{} (\cref{sec:mid-lessons}), so the other English knowledge benchmarks fall on average as TriviaQA rises with it.

\begin{table}[htbp]
  \centering
  \footnotesize
  \setlength{\tabcolsep}{3pt}
  \begin{tabular}{lrrrrrrrrrr}
    \toprule
    Domain
    & \rotatebox{90}{final}
    & \rotatebox{90}{w/o HumanEval}
    & \rotatebox{90}{knowledge EN}
    & \rotatebox{90}{knowledge DE}
    & \rotatebox{90}{math EN}
    & \rotatebox{90}{math DE}
    & \rotatebox{90}{\acs*{MBPP} EN}
    & \rotatebox{90}{\acs*{MBPP} DE}
    & \rotatebox{90}{TriviaQA}
    & \rotatebox{90}{HellaSwag DE} \\
    \midrule
    \domEnWeb
    & \num{-0.66}
    & \num{-0.62}
    & \num{-0.11}
    & \num{0.55}
    & \num{-0.67}
    & \num{-0.72}
    & \num{-0.32}
    & \best{\num{-0.33}}
    & \best{\num{0.81}}
    & \num{0.61} \\
    \domDeWeb
    & \num{-0.57}
    & \num{-0.45}
    & \num{0.09}
    & \best{\num{0.82}}
    & \num{-0.63}
    & \num{-0.65}
    & \num{-0.17}
    & \num{-0.32}
    & \num{0.75}
    & \best{\num{0.84}} \\
    \domEnInstr
    & \best{\num{0.73}}
    & \best{\num{0.67}}
    & \best{\num{0.22}}
    & \num{-0.59}
    & \best{\num{0.76}}
    & \best{\num{0.73}}
    & \num{0.29}
    & \best{\num{0.44}}
    & \best{\num{-0.78}}
    & \num{-0.68} \\
    \domDeInstr
    & \num{0.47}
    & \num{0.55}
    & \num{0.03}
    & \num{-0.14}
    & \num{0.38}
    & \num{0.36}
    & \best{\num{0.52}}
    & \num{0.03}
    & \num{-0.30}
    & \num{-0.07} \\
    \domCode
    & \num{0.41}
    & \num{0.26}
    & \num{0.01}
    & \best{\num{-0.85}}
    & \num{0.53}
    & \num{0.57}
    & \num{0.02}
    & \num{-0.02}
    & \num{-0.48}
    & \best{\num{-0.92}} \\
    \domAgentic
    & \num{0.06}
    & \num{0.20}
    & \num{-0.20}
    & \num{0.36}
    & \num{-0.09}
    & \num{0.00}
    & \num{0.29}
    & \num{0.28}
    & \num{-0.28}
    & \num{0.47} \\
    \domStemDocs
    & \num{0.05}
    & \num{0.03}
    & \best{\num{-0.33}}
    & \num{-0.08}
    & \num{-0.14}
    & \num{0.06}
    & \num{0.16}
    & \num{0.18}
    & \num{-0.05}
    & \num{0.05} \\
    \domStemQA
    & \num{0.48}
    & \num{0.48}
    & \num{0.13}
    & \num{-0.38}
    & \num{0.58}
    & \num{0.52}
    & \num{0.19}
    & \num{0.28}
    & \num{-0.71}
    & \num{-0.47} \\
    \domSpecialized
    & \best{\num{-0.71}}
    & \best{\num{-0.73}}
    & \num{-0.17}
    & \num{0.50}
    & \best{\num{-0.77}}
    & \best{\num{-0.73}}
    & \best{\num{-0.39}}
    & \num{-0.31}
    & \num{0.70}
    & \num{0.62} \\
    \bottomrule
  \end{tabular}
  \caption{Pearson correlation of each domain weight with the aggregate score and its constituents across the M10 merges of the \num{17} runs on the \num{16} mixes of \cref{tab:mid-candidates}, two of them on C1 (direct switch and transition).
    The \model{1} mid-training mix is not composed from the nine domains and has no weights.
    The code columns are \acs*{MBPP} alone.
    The highest and the lowest correlation of each column are set in bold.
  }
  \label{tab:mid-weight-correlations}
\end{table}

\begin{figure}[htbp]
  \centering
  \includegraphics{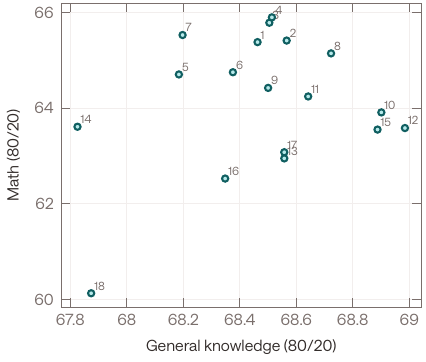}
  {}
  \includegraphics{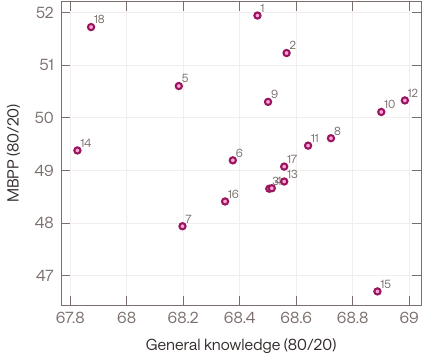}
  \caption{General knowledge against math (left) and against \acs*{MBPP} (right) for the M10 merge of every proxy run in \cref{tab:mid-ranking}.
    Each run trained \num{84}B tokens.
    Each mean is weighted \num{80}/\num{20} over English and German.
    The label is the run's rank on the aggregate score in \cref{tab:mid-ranking}.
  }
  \label{fig:mid-tradeoff}
\end{figure}

\paragraph{Trade-Offs}
Across the \num{84}B runs the general-knowledge mean spans \num{1.2} points, against \num{5.8} for math and \num{5.2} for \ac{MBPP}, so the spread of the aggregate score comes from math and code (\cref{fig:mid-tradeoff}).
English math and \ac{MBPP} correlate at \(r = \num{-0.18}\), and German knowledge and math at \num{-0.73}, through HellaSwag (DE), which correlates with German math at \num{-0.72}.
A knowledge mean averages gains and losses on single benchmarks (\cref{sec:mid-ablation-knowledge}), so the means of \cref{fig:mid-tradeoff} can hide the loss of one benchmark.

\begin{figure}[htbp]
  \centering
  \includegraphics{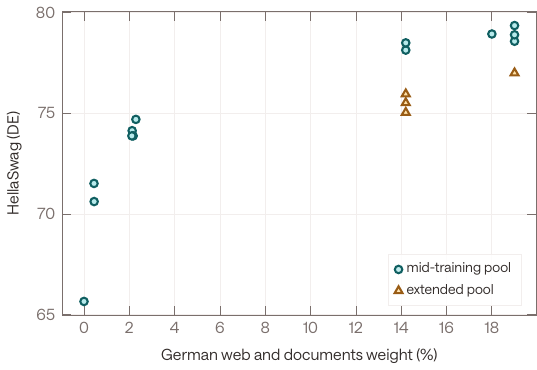}
  \caption{HellaSwag (DE) of the M10 merge of every proxy run at \num{84}\mbox{B} against the weight of \domDeWeb{} in its mix, by the dataset pool of the language domains.
    The figure omits the \model{1} mid-training mix, which has no domain weights.
  }
  \label{fig:mid-hellaswag-deweb}
\end{figure}

\paragraph{Knowledge Benchmarks}
\label{sec:mid-ablation-knowledge}
At the \num{400}B proxy budget, the production mix rises over continued pre-training by \num{3.6} points on \ac{MMLU}-Pro, \num{2.9} on \ac{PIQA}, \num{2.4} on \ac{MMLU}, \num{1.7} on Global \ac{MMLU}, and \num{1.5} on \ac{ARC}, and falls by \num{8.4} on TriviaQA, \num{7.2} on HellaSwag (DE), and \num{3.9} on HellaSwag.
On HellaSwag (DE), C2 and C3, with \qtyrange{18}{19}{\percent} \domDeWeb{} against \qty{2.1}{\percent} in the production mix, score within \num{1.6} points of continued pre-training (\cref{tab:mid-horizon-summary}).
Across the \num{84}B runs, HellaSwag (DE) rises from \num{65.7} at \qty{0}{\percent} to \num{79.4} at \qty{19}{\percent} \domDeWeb{} (\cref{fig:mid-hellaswag-deweb}), and HellaSwag rises from \num{72.6} at \qty{0}{\percent} to \num{75.2} at \qty{33}{\percent} \domEnWeb{}.
Runs on the mid-training pool at an equal domain weight of \domDeWeb{} differ by at most \num{0.9} points on HellaSwag (DE), and runs on the extended pool score lower at the same weight (\cref{sec:mid-lang-sweep}).
TriviaQA falls by \numrange{6.2}{8.4} points at \num{400}B in all three mixes, including C2 with \qty{33}{\percent} \domEnWeb{}.
At \num{84}B and equal domain weights, C1 under the natural sampling rule scores \num{2.8} points above C2 under the pre-training-delivered rule on TriviaQA, against at most \num{0.8} between runs of the language sweep that differ only in the dataset pool or the sampling rule (\cref{sec:mid-lang-sweep}).
All three mixes at \num{400}B use the pre-training-delivered rule, and continued pre-training also differs from them in sequence length, which leaves the cause of the TriviaQA loss open.

\subsubsection{Language Sweep}
\label{sec:mid-lang-sweep}

The language sweep tests whether the in-domain composition of \domEnWeb{} and of \domDeWeb{}, the two language domains, causes the losses of the production run on TriviaQA and HellaSwag (DE) (\cref{sec:mid-knowledge}).
L1 to L4 share one set of domain weights and differ only in the dataset pool and sampling rule of the two language domains (\cref{tab:mid-lang-sweep}).
These weights raise the two language domains of the production mix before its waterfill to \qty{20}{\percent} of the budget and scale the other seven domains down by a common factor.
Within the \qty{20}{\percent}, \domEnWeb{} receives \qty{5.8}{\percent} and \domDeWeb{} \qty{14.2}{\percent}, so that the English language and instruction domains together weigh twice as much as the German ones.
L5 has the domain weights of C1, and L6 its own (\cref{tab:mid-candidates}).
The extended pool adds the \num{26} language datasets of pre-training that the mid-training pool excludes, the low-quality tiers of the web crawls and the rephrased and translated German web text.
Under the optimised sampling rule, the two language domains split their budget at the per-dataset weights that the pre-training mix optimiser assigns over the dataset pool of the run.

After \ac{SFT}, none of the pair tests among L1 to L6, S5, and S6 clears the noise (\cref{sec:mid-ablations}).
At equal domain weights, runs on the extended pool score \numrange{2.4}{3.1} points lower on HellaSwag (DE) than runs on the mid-training pool (L4 against L2, L3 against L1, and L5 against C1), against at most \num{0.9} between runs on the mid-training pool at an equal domain weight of \domDeWeb{} (\cref{fig:mid-hellaswag-deweb}).
In the same pairs, the German general-knowledge mean differs by \numrange{0.9}{1.3} points.
On TriviaQA, the dataset pool or the sampling rule of the language domains alone changes the score by at most \num{0.8} points, and both together by \num{1.2}.

\begin{table}[htbp]
  \centering
  \small
  \begin{tabular}{llll}
    \toprule
    Mix & Dataset pool & Sampling & Other domains as in \\
    \midrule
    L1 & mid-training pool & natural                & S5 \\
    L2 & mid-training pool & optimised              & S5 \\
    L3 & extended pool     & pre-training-delivered & S5 \\
    L4 & extended pool     & optimised              & S5 \\
    L5 & extended pool     & pre-training-delivered & C1 \\
    L6 & extended pool     & optimised              & production mix \\
    \bottomrule
  \end{tabular}
  \caption{Dataset pool and sampling rule of the two language domains in each run of the language sweep, and the mix whose in-domain composition the other seven domains keep.
    \Cref{tab:mid-candidates} lists the domain weights.
  }
  \label{tab:mid-lang-sweep}
\end{table}

\Cref{tab:mid-candidates} lists the domain weights and in-domain composition of every mix trained at proxy scale and of the production mix.

\begin{table}[htbp]
  \centering
  \footnotesize
  \setlength{\tabcolsep}{3pt}
  \begin{tabular}{lrcccccccccll}
    \toprule
    & & \multicolumn{9}{c}{Domain weights (\%)} & \multicolumn{2}{c}{In-domain composition} \\
    \cmidrule(lr){3-11} \cmidrule(lr){12-13}
    Mix
    & B
    & \rotatebox{90}{\domEnWebShort}
    & \rotatebox{90}{\domDeWebShort}
    & \rotatebox{90}{\domEnInstrShort}
    & \rotatebox{90}{\domDeInstrShort}
    & \rotatebox{90}{\domCodeShort}
    & \rotatebox{90}{\domAgenticShort}
    & \rotatebox{90}{\domStemDocsShort}
    & \rotatebox{90}{\domStemQAShort}
    & \rotatebox{90}{\domSpecializedShort}
    & Dataset pool
    & Sampling \\
    \midrule
    C1&$400$&$33.0$&$19.0$&$10.0$&$1.0$&$10.0$&$8.5$&$6.5$&$9.0$&$3.0$&mid-training pool&natural\\
    C2&$400$&$33.0$&$19.0$&$10.0$&$1.0$&$10.0$&$8.5$&$6.5$&$9.0$&$3.0$&mid-training pool&pre-training-delivered\\
    C3&$400$&$14.0$&$18.0$&$16.2$&$1.4$&$8.0$&$16.2$&$6.0$&$16.2$&$4.0$&mid-training pool&pre-training-delivered\\
    \midrule
    S1&$90$&$0.0$&$0.0$&$31.0$&$1.4$&$42.1$&$0.0$&$0.3$&$25.1$&$0.0$&mid-training pool&pre-training-delivered\\
    S2&$90$&$3.1$&$0.4$&$31.5$&$1.4$&$29.4$&$7.8$&$10.8$&$15.5$&$0.1$&mid-training pool&pre-training-delivered\\
    S3&$90$&$3.1$&$0.4$&$30.7$&$1.4$&$29.8$&$7.9$&$10.9$&$15.7$&$0.1$&mid-training pool&pre-training-delivered\\
    S4&$90$&$20.3$&$2.3$&$14.9$&$1.4$&$27.1$&$9.7$&$13.7$&$9.6$&$1.0$&mid-training pool&pre-training-delivered\\
    S5&$90$&$1.3$&$2.1$&$31.5$&$1.4$&$16.4$&$15.5$&$6.0$&$25.5$&$0.2$&mid-training pool&pre-training-delivered\\
    S6&$90$&$1.3$&$2.2$&$31.0$&$1.4$&$16.8$&$15.9$&$6.1$&$25.1$&$0.2$&mid-training pool&pre-training-delivered\\
    \midrule
    L1&$90$&$5.8$&$14.2$&$25.9$&$1.7$&$13.5$&$12.8$&$4.9$&$21.0$&$0.2$&mid-training pool&natural\\
    L2&$90$&$5.8$&$14.2$&$25.9$&$1.7$&$13.5$&$12.8$&$4.9$&$21.0$&$0.2$&mid-training pool&optimised\\
    L3&$90$&$5.8$&$14.2$&$25.9$&$1.7$&$13.5$&$12.8$&$4.9$&$21.0$&$0.2$&extended pool&pre-training-delivered\\
    L4&$90$&$5.8$&$14.2$&$25.9$&$1.7$&$13.5$&$12.8$&$4.9$&$21.0$&$0.2$&extended pool&optimised\\
    L5&$90$&$33.0$&$19.0$&$10.0$&$1.0$&$10.0$&$8.5$&$6.5$&$9.0$&$3.0$&extended pool&pre-training-delivered\\
    L6&$90$&$5.8$&$14.2$&$24.0$&$4.0$&$12.5$&$15.4$&$4.5$&$19.4$&$0.2$&extended pool&optimised\\
    \midrule
    production mix&$5{,}196$&$1.3$&$2.1$&$31.4$&$1.9$&$16.3$&$15.4$&$5.9$&$25.4$&$0.2$&mid-training pool&pre-training-delivered\\
    \bottomrule
  \end{tabular}
  \caption{Every mid-training mix with its domain weights and its in-domain composition.
    C1 to C3 have hand-set domain weights and were materialised at \num{400}B. S1 to S6 are the optimised solutions (\cref{sec:mid-mixing}) and L1 to L6 the language sweep (\cref{sec:mid-lang-sweep}), both at \num{90}B. The production mix (\cref{sec:mid-effect}) is materialised at \num{5.2}T. Short domain names follow \cref{tab:mid-pool-domains}.
    In the language sweep, the dataset pool and sampling rule describe the two language domains (\cref{tab:mid-lang-sweep}).
  }
  \label{tab:mid-candidates}
\end{table}

\subsubsection{Decontamination}
\label{sec:mid-decon}

To reduce the risk that a benchmark score reflects memorisation of the benchmark itself, we remove from the training corpus every document that contains text from our evaluation suite.
The reference eval set contains all \num{226} benchmark tasks of our evaluation framework with all available splits (train, validation, and test), because training and validation items share the format, the answer style and potentially near-duplicates of the test items, so a model trained on them scores higher without any test instance leaking.

We detect contamination with the algorithm of the \texttt{decon} tool~\parencite{allenai2025decon}.

Multiple-choice tasks are unpacked into one entry per answer option.
Questions and answers are indexed separately as \num{5}- and \num{3}-grams over \texttt{cl100k} tokens.
Each n-gram is weighted by its \ac{IDF} across the reference set: a rare, distinctive n-gram is strong evidence of contamination, but one that appears in many evaluation instances, such as a stopword sequence or shared prompt boilerplate, is weak evidence and counts for little.

Inspection is deterministic and involves no model inference.
The inspection probes a sample of a training document's n-grams against the index.
On a hit, the inspection expands in both directions and keeps the evaluation instances that continue to match.
It scores the span as an \ac{IDF}-weighted overlap ratio of the question and of the answer, weighted towards the question, and flags a document whose combined score exceeds a threshold.
We drop a flagged document as a whole, whichever task it matched.
Short matches must be perfect, which keeps common phrases from being flagged.
We keep AI2's default parameters throughout.

Contamination late in training is memorised and inflates benchmark scores~\parencite{magar2022contamination}.
Contamination early in training is forgotten once enough training follows it~\parencite{bordt2024forget}, whereas contamination in the final stages is followed by too little training to be forgotten.
We therefore decontaminate the mid-training pool, which also feeds the long-context extension stage, and the post-training pool (\cref{sec:sft-decon}).

\begin{figure}[tb]
  \centering
  \includegraphics{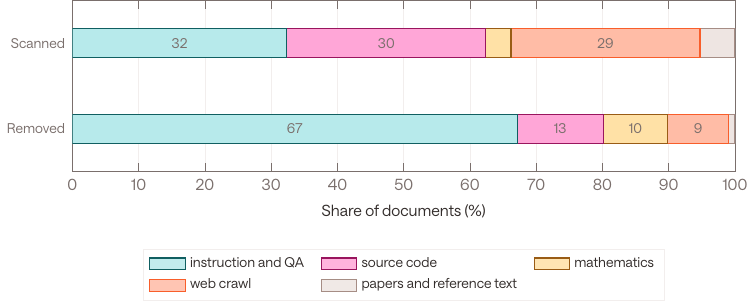}
  \caption{Composition of the inspected mid-training pool and of the documents decontamination removed from it, by class of data, for our mid-training mix.
  }
  \label{fig:decon-groups}
\end{figure}

Across all datasets from the mid-training pool, the decontamination run inspected \num{8.3}B documents and removed \num{1.5}M of them, \qty{0.018}{\percent} of documents and \qty{0.06}{\percent} of tokens.
Removed documents average \num{4856} tokens against \num{745} for the pool, about six times longer at a similar median, since we drop a document completely once any of its spans matches, and long documents such as committed benchmark files or multi-item synthetic records offer many spans.
The aggregates depend on which datasets the mid-training pool holds and are indicative only.
Contamination is concentrated in a few classes of data, as \cref{fig:decon-groups} shows.
Synthetic instruction and question-answering data holds \qty{32}{\percent} of the inspected documents but \qty{67}{\percent} of the removals, and mathematics datasets have the highest rate at \qty{0.05}{\percent} of their documents.
Source code contributes \qty{13}{\percent} of the removals, most of them structured data files in which benchmark datasets were committed to repositories verbatim.
Web crawl text contributes \qty{9}{\percent} at a rate of \qty{0.006}{\percent}, and scientific papers, PDF scrapes, and curated reference and legal text together contribute under \qty{1}{\percent}.

On the benchmark side, \ac{MMLU}~\parencite{hendrycks2021mmlu}, MATH~\parencite{hendrycks2021math}, and \ac{GSM8K}~\parencite{cobbe2021gsm8k} account for \qty{70}{\percent} of all matched (document, evaluation instance) pairs, \ac{MMLU}-Pro~\parencite{wang2024mmlupro} for a further \qty{10}{\percent}, and \ac{ARC}~\parencite{clark2018arc} and a small collection of other benchmarks (including HellaSwag~\parencite{zellers2019hellaswag} and SQuAD~\parencite{rajpurkar2016squad}) for most of the remainder.

Counted per question rather than per pair, \cref{tab:decon-coverage} gives the share of each benchmark's evaluated questions that the detector matched in at least one document of the pool, in any prompt variant of the benchmark.
\ac{GSM8K}, HumanEval and MATH are matched almost completely, the knowledge and reasoning benchmarks in part, and \ac{MBPP} and TriviaQA in about a tenth of their questions only.
These shares count only the questions the detector matched, and they understate how much of each benchmark the removal took out of the pool.
A flagged document is dropped as a whole, so a file that holds a whole benchmark is removed once the detector matches a single one of its questions, and all its other questions go with it.
Counting every question that appears verbatim in the removed documents gives far higher shares than the detector matched (\cref{tab:decon-coverage}): \qty{91}{\percent} against \qty{56}{\percent} of the \ac{MMLU} questions and \qty{87}{\percent} against \qty{11}{\percent} of the TriviaQA questions~\parencite{gebhart2026decon}.

\begin{table}[tb]
  \centering
  \scriptsize
  \renewcommand{\arraystretch}{1}
  \setlength{\tabcolsep}{5pt}
  \begin{tabular}{llrrr}
    \toprule
    Benchmark & Split & Questions & Found & Share (\unit{\percent}) \\
    \midrule
    \acs*{GSM8K}    & test       & \num{1319}  & \num{1319} & \num{100.0} \\
    HumanEval       & test       & \num{164}   & \num{163}  & \num{99.4} \\
    MATH            & test       & \num{5000}  & \num{4698} & \num{94.0} \\
    SQuAD           & validation & \num{10205} & \num{7671} & \num{75.2} \\
    \acs*{ARC}      & test       & \num{3548}  & \num{2209} & \num{62.3} \\
    \acs*{MMLU}     & test       & \num{14042} & \num{7879} & \num{56.1} \\
    \acs*{MMLU}-Pro & test       & \num{12032} & \num{4797} & \num{39.9} \\
    HellaSwag       & test       & \num{10003} & \num{2981} & \num{29.8} \\
    TriviaQA        & validation & \num{7993}  & \num{908}  & \num{11.4} \\
    \acs*{MBPP}     & test       & \num{500}   & \num{56}   & \num{11.2} \\
    \bottomrule
  \end{tabular}
  \caption{Share of each benchmark's questions that at least one document of the mid-training pool matched before removal, in the split our evaluation scores.
    A question counts once however many documents and prompt variants of the benchmark match it.
  }
  \label{tab:decon-coverage}
\end{table}

Much of this contamination is templated: a benchmark question followed by rewritten options, a stated answer, and a worked explanation, which is what synthetic instruction data generated from benchmark exemplars looks like.

To measure the effect of decontamination, we compared four mid-training runs of the proxy model after the same number of steps~\parencite{gebhart2026decon}.
A run on a mix drawn from the pool before decontamination, at its natural contamination rate of about \qty{0.04}{\percent} of tokens, and a run on a mix drawn from the decontaminated pool agree to within their checkpoint-to-checkpoint variation, averaged over the last five evaluations.
The naturally contaminated run also blends pre-training data into its first \num{40}B tokens, so it is a looser comparison.
At natural dilution and this scale, the contamination is not measurable.
This does not make decontamination optional.
The contaminated share of a natural mix is roughly the same at any scale, but the expected number of times a contaminated document is seen is not.
As the mix grows, each domain's sampling factor rises towards its repetition cap, and the domains upsampled first, math, code and synthetic instruction data, are the ones that hold the contaminated documents.
Two further runs show what such exposure does.
An upper-bound run adds every document removed from the pool back into the decontaminated proxy mix once, \qty{8.2}{\percent} of its tokens.
Averaged over the last five evaluations, it inflates HumanEval by \num{45} points, \ac{MMLU} by \num{28} and TriviaQA by \num{25}, halves the bits per byte on the benchmark text, and leaves the long-context suites unchanged.
A run at target-scale exposure adds each removed document back as often as the production run would have seen it, \qty{1.6}{\percent} of the proxy mix, and still inflates HumanEval by \num{19} points, TriviaQA and MATH by \num{17}, and \ac{MMLU} and \ac{GSM8K} by \num{7}.
For HumanEval, \ac{MMLU} and TriviaQA, whose removed documents are mostly copies of evaluated items, the gain is memorisation and not capability.
For \ac{GSM8K} and MATH it is in part in-distribution learning from their training splits: the run at target-scale exposure recovers \qty{70}{\percent} and \qty{140}{\percent} of the upper-bound accuracy gain, but only \qty{58}{\percent} and \qty{46}{\percent} of the drop in \ac{BPB}.
A benchmark score is evidence of capability only if leakage is ruled out by construction rather than argued away after the fact.

The benchmark scores over the mid-training run show signs of contamination that the method misses, in particular reference solutions that appear without their prompt: \num{1926} documents carrying a HumanEval reference solution, covering \num{100} of the \num{164} problems, passed through the production run~\parencite{gebhart2026decon} (\cref{sec:mid-humaneval}).

\subsubsection{Transition into the Mid-Training Mix}
\label{sec:transition-index}

The step from the pre-training mix to the mid-training mix changes the data distribution at a constant learning rate.
Prior work studies such changes of distribution in continual pre-training and mid-training \parencite{gupta2023rewarm, parmar2024reuse, olmo2026olmo3}, and retaining a share of the previous data limits the forgetting they cause \parencite{ibrahim2024continual, wang2025learningdynamics, abbes2026replay, liu2025midtraining}.

We therefore interpolate between the two mixes over the window of \(H\) tokens of \cref{sec:mid-effect}, in which the token share of the pre-training mix decays linearly as \[ w(t) = 1 - \frac{t}{H}, \qquad 0 \le t \le H, \] so that the pre-training mix contributes \(H/2\), \num{50}B tokens, to the window.
The loader index realises the schedule by interleaving the bulk-shuffled indices of the two mixes, in bulks of \num{4096} rows.
It places the \(i\)-th pre-training bulk, with token midpoint \(m_i\), where the cumulative pre-training tokens \(t - t^2/(2H)\) reach \(m_i\), that is, at \[ t_i = H \left(1 - \sqrt{1 - \frac{2 m_i}{H}}\right).
\] These bulks come from the part of the pre-training index beyond the \num{20}T tokens of the pre-training run, so their documents lie beyond the model's pre-training data.
The schedule holds up to whole bulks of \num{4096} rows, and the pre-training documents in the window remain potentially contaminated, since we do not decontaminate the pre-training pool.

\subsubsection{Production Run}
\label{sec:mid-production-run}

\begin{figure}[htbp]
  \centering
  \includegraphics{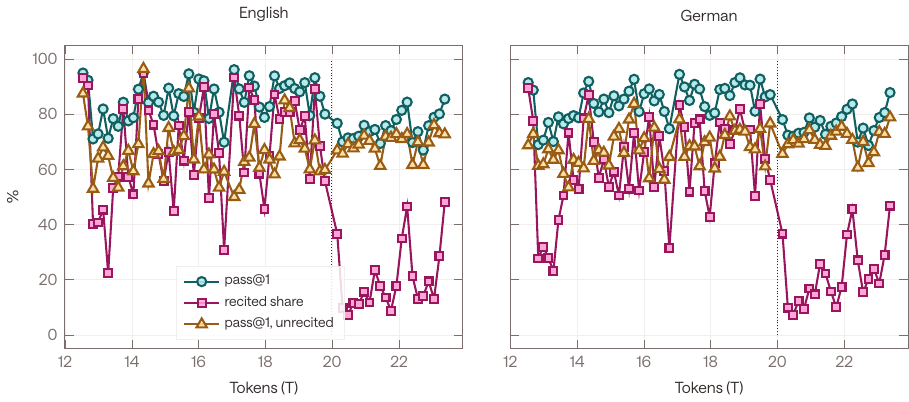}
  \caption{HumanEval pass@1, the share of recited completions, and pass@1 on the problems without a recited completion for every HumanEval evaluation of pre-training and mid-training, against the tokens of the chain.
    The dotted line marks the start of mid-training.
  }
  \label{fig:mid-recitation}
\end{figure}

\paragraph{HumanEval Scores and Recitation}
\label{sec:mid-humaneval}
During pre-training, English HumanEval moves by more than \num{25} points from one evaluated merge to the next (\cref{fig:mid-recitation}).
For every HumanEval evaluation of the production run, we compare each completion with the reference solution shipped with the benchmark, after dropping comments and collapsing whitespace, and count it as recited at a similarity of \num{0.95} or more.
During pre-training, \qtyrange{22}{95}{\percent} of all completions are recited, and pass@1 follows the recited share linearly, with a Pearson correlation of \num{0.90} over the \num{96} evaluations, \num{48} per language.
Mid-training on the decontaminated mix lowers the recited share to below \qty{10}{\percent} of completions within the first \num{0.3}T tokens, and from \num{0.3}T to \num{0.8}T pass@1 stays at \numrange{70}{72} in English and \numrange{72}{73} in German.
Throughout mid-training, pass@1 on the problems a merge does not recite stays at \numrange{61}{76} in English and \numrange{61}{79} in German, against \numrange{50}{96} and \numrange{53}{84} on a shifting set of problems during pre-training.
The fall of pass@1 at the start of mid-training thus coincides with the fall of the recited share, and the pre-training HumanEval scores are not comparable with those of a decontaminated model.
The recited solutions come from the pre-training mix, which remains potentially contaminated (\cref{sec:mid-decon}).
The recitation that returns in later mid-training merges is a checkpoint-level state, independent of the exposure to HumanEval text in the mid-training data.

\paragraph{Knowledge Benchmarks}
\label{sec:mid-knowledge}
From the pre-training M4 to the mid-training M4 of the production run, scores rise by \num{2.9} points on \ac{MMLU}-Pro and \num{1.2} on \ac{ARC}, fall by \num{6.8} on TriviaQA, \num{5.6} on HellaSwag (DE), and \num{2.6} on HellaSwag, and change by at most \num{0.7} on the other five knowledge benchmarks (\cref{tab:mid-production}).
The M20 merge at the end of mid-training lifts the English and German knowledge means by \num{1.6} and \num{1.0} points over the M4 merge, and the final base model scores within \num{0.1} points of it on both.
The loss on HellaSwag (DE) is as predicted by the proxy runs, and the cause of the TriviaQA loss remains open (\cref{sec:mid-ablation-knowledge}).

\begin{table}[htbp]
  \centering
  \footnotesize
  \setlength{\tabcolsep}{2pt}
  \begin{tabular}{lcccccccccccccccccc}
    \toprule
    &
    &
    & \multicolumn{2}{c}{Tokens (T)}
    & \multicolumn{2}{c}{Score}
    & \multicolumn{2}{c}{Knowledge}
    & \multicolumn{2}{c}{Math}
    & \multicolumn{2}{c}{Code}
    & \multicolumn{6}{c}{Benchmarks (EN unless marked)} \\
    \cmidrule(lr){4-5}\cmidrule(lr){6-7}\cmidrule(lr){8-9}\cmidrule(lr){10-11}\cmidrule(lr){12-13} \cmidrule(lr){14-19}
    Stage
    & Step
    & M
    & stage
    & total
    & aggregate
    & w/o HE
    & EN
    & DE
    & EN
    & DE
    & EN
    & DE
    & HE
    & Pro
    & \acs{ARC}
    & TQA
    & HS
    & HS DE \\
    \midrule
    pre-training&$262000$&M4&$19.78$&$19.78$&$77.7$&$74.8$&$80.7$&$89.6$&$79.4$&$71.4$&$70.3$&$82.3$&$80.0$&$55.1$&$94.0$&$82.7$&$84.5$&$91.4$\\
    mid-training&$33600$&M4&$3.38$&$23.37$&$80.4$&$77.6$&$79.8$&$88.2$&$84.0$&$75.2$&$75.9$&$84.4$&$85.5$&$58.0$&$95.2$&$75.9$&$81.9$&$85.8$\\
    mid-training&$34000$&M20&$3.42$&$23.41$&$81.0$&$79.1$&$81.4$&$89.2$&$85.1$&$76.5$&$75.7$&$80.3$&$83.0$&$61.3$&$95.8$&$77.3$&$83.2$&$86.4$\\
    long-context&$1000$&M20&$0.20$&$23.63$&$81.6$&$78.9$&$81.4$&$89.3$&$85.0$&$77.2$&$76.8$&$85.0$&$86.2$&$61.5$&$95.7$&$77.5$&$83.1$&$86.9$\\
    \bottomrule
  \end{tabular}
  \caption{The production run on the pre-training suite at the last evaluated merge of pre-training, mid-training, and long-context extension, whose M20 merge is the final base model.
    Every intermediate evaluation was on an M4 merge, so the M4 rows are the comparable ones.
    Mid-training and long-context training start from the raw checkpoints at pre-training step \num{264750} and mid-training step \num{34200}.
    The columns are the aggregate score, the aggregate score without HumanEval (HE), the general-knowledge, math, and code means per language, HumanEval (HE), and the knowledge benchmarks with the largest changes (Pro is \acs*{MMLU}-Pro, TQA is TriviaQA, HS is HellaSwag, and HS DE is HellaSwag (DE)).
    M is the merge window.
    Tokens are those of the stage at that step and of the whole chain.
  }
  \label{tab:mid-production}
\end{table}

\paragraph{Post-Training Benchmarks after \Acs*{SFT}}
The \acp{SFT} on the mid-training endpoint and on the final base model agree within their eval-noise intervals on every benchmark with a noise model (\cref{tab:mid-production-sft}).
Long-context extension thus changes none of these benchmarks beyond the eval noise.
Against the \ac{SFT} on the \num{400}B proxy run of the production mix (\cref{fig:mid-horizon}), the final base model gains \numrange{13}{44} points per benchmark, a gap that reflects model size and training length at the same mix.

\begin{table}[htbp]
  \centering
  \footnotesize
  \begin{tabular}{lcccccc}
    \toprule
    & \multicolumn{3}{c}{Mid-training M20} & \multicolumn{3}{c}{Final base model} \\
    \cmidrule(lr){2-4}\cmidrule(lr){5-7}
    Benchmark & score & low & high & score & low & high \\
    \midrule
    AIME 2026&$71.3$&$63.3$&$78.6$&$72.7$&$64.9$&$79.9$\\
    AIME 2026 DE&$63.3$&$52.0$&$74.0$&$73.5$&$62.9$&$83.0$\\
    GPQA Diamond CoT&$66.2$&$58.4$&$73.5$&$66.2$&$58.4$&$73.5$\\
    IFBench&$50.3$&$45.8$&$54.9$&$50.0$&$45.5$&$54.5$\\
    LongBenchPro&$47.7$&--&--&$49.3$&--&--\\
    LongBenchPro 128k&$37.4$&--&--&$38.0$&--&--\\
    HumanEval+&$92.7$&$89.4$&$95.4$&$92.1$&$88.7$&$94.9$\\
    LiveCodeBench v6&$66.7$&$62.4$&$70.9$&$67.4$&$63.1$&$71.5$\\
    KulturBank&$48.0$&--&--&$48.5$&--&--\\
    \bottomrule
  \end{tabular}
  \caption{Post-training benchmark scores of the \acs*{SFT} on the mid-training M20 merge at step \num{34000} and on the final base model, with the \qty{95}{\percent} eval-noise interval where the benchmark has a noise model.
    Both \acsp*{SFT} use the recipe of \cref{sec:mid-ablations} with a peak learning rate of \num{0.3} of pre-training (\cref{tab:sft-ablation-settings}) and are separate from the \model{2} \acs*{SFT} (\cref{sec:sft}).
    They ran a smaller benchmark suite than the proxy \acsp*{SFT}, without MATH-500 and IFEval, so they have no pooled score comparable with the proxy \acsp*{SFT}.
    The intervals are those of the proxy-scale noise model, which we assume transfers to this model size.
  }
  \label{tab:mid-production-sft}
\end{table}

\subsection{Full Long-Context Results of the Mix Ablation}
\label{app:lc-ablation-results}

\Cref{tab:lc-ablation-full} lists the long-context legs of the mix ablation at every evaluated context length, for the merged checkpoint of every branch, together with the mean over the evaluated lengths.

\begin{table}[p]
  \centering
  \scriptsize
  \renewcommand{\arraystretch}{1}
  \begin{tabular}{llrrrrrrrr}
    \toprule
    Benchmark
    & Mix
    & \num{4}\mbox{k}
    & \num{8}\mbox{k}
    & \num{16}\mbox{k}
    & \num{32}\mbox{k}
    & \num{64}\mbox{k}
    & \num{128}\mbox{k}
    & \num{256}\mbox{k}
    & Mean \\
    \midrule
    RULER NIAH
    & s = 0
    & 0.983
    & 0.981
    & 0.964
    & 0.967
    & 0.926
    & 0.947
    & 0.878
    & 0.949 \\
    & s = 1/4
    & 0.989
    & 0.984
    & 0.969
    & 0.974
    & 0.940
    & 0.954
    & 0.897
    & 0.958 \\
    & s = 1/3
    & \textbf{0.994}
    & \textbf{0.988}
    & \textbf{0.975}
    & \textbf{0.979}
    & \textbf{0.942}
    & \textbf{0.968}
    & \textbf{0.921}
    & \textbf{0.967} \\
    & control
    & 0.992
    & 0.976
    & 0.964
    & 0.968
    & 0.917
    & 0.936
    & 0.860
    & 0.945 \\
    \midrule
    RULER VT
    & s = 0
    & 0.978
    & 0.956
    & 0.926
    & 0.960
    & \textbf{0.960}
    & 0.916
    & 0.892
    & 0.941 \\
    & s = 1/4
    & \textbf{0.992}
    & 0.974
    & 0.956
    & 0.954
    & 0.940
    & \textbf{0.940}
    & \textbf{0.952}
    & 0.958 \\
    & s = 1/3
    & 0.976
    & \textbf{0.978}
    & 0.956
    & 0.964
    & 0.954
    & 0.936
    & 0.900
    & 0.952 \\
    & control
    & 0.984
    & \textbf{0.978}
    & \textbf{0.974}
    & \textbf{0.974}
    & 0.952
    & 0.934
    & 0.940
    & \textbf{0.962} \\
    \midrule
    RULER QA
    & s = 0
    & 0.685
    & 0.660
    & 0.655
    & 0.665
    & 0.630
    & 0.590
    & 0.515
    & 0.629 \\
    & s = 1/4
    & \textbf{0.700}
    & 0.660
    & 0.655
    & 0.650
    & \textbf{0.645}
    & 0.570
    & 0.575
    & 0.636 \\
    & s = 1/3
    & 0.695
    & \textbf{0.685}
    & 0.645
    & \textbf{0.680}
    & 0.625
    & \textbf{0.605}
    & \textbf{0.605}
    & \textbf{0.649} \\
    & control
    & 0.685
    & 0.675
    & \textbf{0.665}
    & 0.655
    & 0.630
    & 0.585
    & 0.545
    & 0.634 \\
    \midrule
    RULER WE
    & s = 0
    & 0.956
    & 0.564
    & 0.522
    & \textbf{0.474}
    & 0.462
    & 0.458
    &
    & 0.573 \\
    & s = 1/4
    & \textbf{0.970}
    & \textbf{0.667}
    & 0.533
    & 0.436
    & 0.462
    & 0.445
    &
    & 0.586 \\
    & s = 1/3
    & 0.954
    & 0.570
    & 0.527
    & 0.465
    & 0.466
    & \textbf{0.466}
    &
    & 0.575 \\
    & control
    & 0.963
    & 0.654
    & \textbf{0.555}
    & 0.460
    & \textbf{0.476}
    & 0.445
    &
    & \textbf{0.592} \\
    \midrule
    HELMET JsonKV
    & s = 0
    &
    & 0.990
    & \textbf{1.000}
    & 0.990
    & \textbf{1.000}
    & 0.950
    &
    & 0.986 \\
    & s = 1/4
    &
    & \textbf{1.000}
    & \textbf{1.000}
    & \textbf{1.000}
    & 0.990
    & 0.970
    &
    & 0.992 \\
    & s = 1/3
    &
    & 0.990
    & \textbf{1.000}
    & \textbf{1.000}
    & \textbf{1.000}
    & \textbf{0.980}
    &
    & \textbf{0.994} \\
    & control
    &
    & \textbf{1.000}
    & 0.990
    & \textbf{1.000}
    & \textbf{1.000}
    & \textbf{0.980}
    &
    & \textbf{0.994} \\
    \midrule
    HELMET RagQA
    & s = 0
    &
    & 0.613
    & 0.603
    & 0.576
    & \textbf{0.558}
    &
    &
    & 0.587 \\
    & s = 1/4
    &
    & 0.608
    & 0.585
    & 0.585
    & 0.550
    &
    &
    & 0.582 \\
    & s = 1/3
    &
    & \textbf{0.633}
    & \textbf{0.635}
    & \textbf{0.615}
    & 0.520
    &
    &
    & \textbf{0.601} \\
    & control
    &
    & 0.605
    & 0.581
    & 0.586
    & 0.556
    &
    &
    & 0.582 \\
    \midrule
    HELMET Rerank
    & s = 0
    &
    & 0.466
    & 0.331
    & 0.285
    & 0.155
    & \textbf{0.158}
    &
    & 0.279 \\
    & s = 1/4
    &
    & 0.480
    & 0.348
    & 0.260
    & 0.167
    & 0.118
    &
    & 0.275 \\
    & s = 1/3
    &
    & \textbf{0.489}
    & \textbf{0.363}
    & \textbf{0.297}
    & \textbf{0.194}
    & 0.113
    &
    & \textbf{0.291} \\
    & control
    &
    & 0.471
    & 0.356
    & 0.251
    & 0.154
    & 0.119
    &
    & 0.270 \\
    \midrule
    HELMET LongQA InfBench QA
    & s = 0
    &
    & 0.157
    & 0.185
    & 0.265
    & 0.300
    & 0.329
    &
    & 0.247 \\
    & s = 1/4
    &
    & 0.145
    & 0.208
    & 0.289
    & 0.301
    & 0.358
    &
    & 0.260 \\
    & s = 1/3
    &
    & 0.163
    & 0.233
    & 0.293
    & \textbf{0.321}
    & \textbf{0.371}
    &
    & \textbf{0.276} \\
    & control
    &
    & \textbf{0.176}
    & \textbf{0.236}
    & \textbf{0.299}
    & 0.303
    & 0.331
    &
    & 0.269 \\
    \midrule
    HELMET LongQA InfBench Choice
    & s = 0
    &
    & \textbf{0.460}
    & 0.420
    & 0.500
    & \textbf{0.590}
    & 0.590
    &
    & 0.512 \\
    & s = 1/4
    &
    & 0.440
    & 0.440
    & \textbf{0.510}
    & 0.560
    & 0.570
    &
    & 0.504 \\
    & s = 1/3
    &
    & 0.440
    & 0.440
    & 0.500
    & 0.570
    & 0.590
    &
    & 0.508 \\
    & control
    &
    & 0.440
    & \textbf{0.460}
    & 0.490
    & 0.570
    & \textbf{0.620}
    &
    & \textbf{0.516} \\
    \bottomrule
  \end{tabular}
  \caption{Long-context results of the mix ablation for the merged checkpoint of every branch, by benchmark and context length.
    RULER \acs*{NIAH}, \acs*{VT} and \acs*{WE} report recall and RULER \acs*{QA} substring exact match.
    HELMET JsonKV and LongQA InfBench Choice report exact match, RagQA substring exact match, Rerank \acl*{NDCG} (\acs*{NDCG}) at~\num{10}, and LongQA InfBench QA ROUGE-L F1.
    An empty cell is a length the leg was not evaluated at, the mean runs over the evaluated lengths, and the best value of every column within a benchmark is bold.
  }
  \label{tab:lc-ablation-full}
\end{table}

\FloatBarrier

%% file: sections/09-appendix/E-base-model-evaluation.tex
\section{Base Model Evaluation}
\label{app:base-evaluation}
\subsection{Baseline Models}
\label{app:baselines}

Every model column of \cref{tab:pre-training-evals,tab:post-training-evals} is served with vLLM \parencite{kwon2023vllm} on one node of eight NVIDIA B200 GPUs, and \cref{tab:baseline-models} lists each model's checkpoint and deployment.
Parallelism is given as \acf{TP} and \acf{DP} ranks, with \ac{EP} marking expert-parallel \ac{MoE} layers; the window is the served context length.
Sampling is given as temperature over top-p; \enquote{default} takes the sampling parameters of the checkpoint's generation config.
SWE-bench Verified additionally runs on a dedicated single-replica engine per model.

\begin{table}[htbp]
  \centering
  \scriptsize
  \setlength{\tabcolsep}{2pt}
  \def\UrlBreaks{\do\/\do\-\do\.\do\_}%
  \begin{tabular}{l>{\raggedright\arraybackslash}p{0.20\linewidth}lrll>{\raggedright\arraybackslash}p{0.14\linewidth}}
    \toprule
    Model & Checkpoint & Reference & Window & Parallelism & Sampling & Notes \\
    \midrule
    \bfseries Pre-training table &   &             &      &         &           & \\
    \basemodel{2}                & – & this report & 260k & TP2 DP4 & 0.6 / 0.6 & \\
    \basemodel{1}                & – & this report & 64k  & TP1 DP8 & 0.6 / 0.6 & \\
    Apertus 70B Base&\href{https://huggingface.co/swiss-ai/Apertus-70B-2509}{\path{swiss-ai/Apertus-70B-2509}}&\parencite{apertus2025}&64k&TP2 DP4&0.6 / 0.6&\\
    Olmo 3 7B Base&\href{https://huggingface.co/allenai/Olmo-3-1025-7B}{\path{allenai/Olmo-3-1025-7B}}&\parencite{olmo2026olmo3}&64k&TP1 DP8&0.6 / 0.6&\\
    Olmo 3 32B Base&\href{https://huggingface.co/allenai/Olmo-3-1125-32B}{\path{allenai/Olmo-3-1125-32B}}&\parencite{olmo2026olmo3}&64k&TP1 DP8&0.6 / 0.6&\\
    Nemotron 3 Nano 30B-A3B Base&\href{https://huggingface.co/nvidia/NVIDIA-Nemotron-3-Nano-30B-A3B-Base-BF16}{\path{nvidia/NVIDIA-Nemotron-3-Nano-30B-A3B-Base-BF16}}&\parencite{nvidia2025nemotron3nano}&64k&TP1 DP8 EP&0.6 / 0.6&\\
    Nemotron 3 Super 120B-A12B Base&\href{https://huggingface.co/nvidia/NVIDIA-Nemotron-3-Super-120B-A12B-Base-BF16}{\path{nvidia/NVIDIA-Nemotron-3-Super-120B-A12B-Base-BF16}}&\parencite{nvidia2026nemotron3super}&64k&TP2 DP4&0.6 / 0.6&\\
    Qwen3.5 35B-A3B Base&\href{https://huggingface.co/Qwen/Qwen3.5-35B-A3B-Base}{\path{Qwen/Qwen3.5-35B-A3B-Base}}&\parencite{qwen2026qwen35}&64k&TP1 DP8 EP&0.6 / 0.6&static YaRN factor 4\\
    GLM-4.5 Air 106B-A12B Base&\href{https://huggingface.co/zai-org/GLM-4.5-Air-Base}{\path{zai-org/GLM-4.5-Air-Base}}&\parencite{glm45team2025glm45}&64k&TP2 DP4 EP&0.6 / 0.6&\\
    Gemma 4 26B-A4B Base&\href{https://huggingface.co/google/gemma-4-26B-A4B}{\path{google/gemma-4-26B-A4B}}&\parencite{gemma2026gemma4technicalreport}&64k&TP1 DP8&1.0 / 0.95&\\
    \bfseries Post-training table &   &             &      &         &            & \\
    \model{2}                     & – & this report & 256k & TP1 DP8 & default    & reasoning effort \texttt{high} \\
    \model{1}                     & – & this report & 64k  & TP1 DP8 & 0.6 / 0.95 & \\
    Qwen3.6 35B-A3B&\href{https://huggingface.co/Qwen/Qwen3.6-35B-A3B}{\path{Qwen/Qwen3.6-35B-A3B}}&\parencite{qwen2026qwen36}&256k&TP1 DP8 EP&default&\\
    Qwen3-Next 80B-A3B Thinking&\href{https://huggingface.co/Qwen/Qwen3-Next-80B-A3B-Thinking}{\path{Qwen/Qwen3-Next-80B-A3B-Thinking}}&\parencite{qwen2025qwen3next}&256k&TP1 DP8 EP&default&\\
    Nemotron 3 Super 120B-A12B&\href{https://huggingface.co/nvidia/NVIDIA-Nemotron-3-Super-120B-A12B-BF16}{\path{nvidia/NVIDIA-Nemotron-3-Super-120B-A12B-BF16}}&\parencite{nvidia2026nemotron3super}&256k&TP2 DP4&default&\\
    Nemotron 3 Nano 30B-A3B&\href{https://huggingface.co/nvidia/NVIDIA-Nemotron-3-Nano-30B-A3B-BF16}{\path{nvidia/NVIDIA-Nemotron-3-Nano-30B-A3B-BF16}}&\parencite{nvidia2025nemotron3nano}&256k&TP1 DP8 EP&default&\\
    Qwen3.8 27B&\href{https://huggingface.co/Qwen/Qwen3.8-27B}{\path{Qwen/Qwen3.8-27B}}&\parencite{qwen2026qwen38}&256k&TP1 DP8&default&\\
    Mistral Small 4 119B-A6B&\href{https://huggingface.co/mistralai/Mistral-Small-4-119B-2603}{\path{mistralai/Mistral-Small-4-119B-2603}}&\parencite{mistral2026small4}&256k&TP2 DP4&default&FP8 checkpoint; reasoning effort \texttt{high}\\
    Gemma 4 26B-A4B IT&\href{https://huggingface.co/google/gemma-4-26B-A4B-it}{\path{google/gemma-4-26B-A4B-it}}&\parencite{gemma2026gemma4technicalreport}&256k&TP1 DP8 EP&1.0 / 0.95&thinking enabled; text only\\
    GPT-OSS 120B&\href{https://huggingface.co/openai/gpt-oss-120b}{\path{openai/gpt-oss-120b}}&\parencite{openai2025gptoss}&128k&TP8 DP1&default&native MXFP4 checkpoint; reasoning effort \texttt{high}\\
    Apertus 70B Instruct&\href{https://huggingface.co/swiss-ai/Apertus-70B-Instruct-2509}{\path{swiss-ai/Apertus-70B-Instruct-2509}}&\parencite{apertus2025}&64k&TP2 DP4&0.8 / 0.9&completions capped at \num{16384} tokens\\
    Qwen3.5 35B-A3B&\href{https://huggingface.co/Qwen/Qwen3.5-35B-A3B}{\path{Qwen/Qwen3.5-35B-A3B}}&\parencite{qwen2026qwen35}&256k&TP1 DP8 EP&1.0 / 0.95&\\
    GLM-4.5 Air 106B-A12B&\href{https://huggingface.co/zai-org/GLM-4.5-Air}{\path{zai-org/GLM-4.5-Air}}&\parencite{glm45team2025glm45}&128k&TP4 DP2 EP&0.6 / 1.0&\\
    GLM-4.7 Flash 30B-A3B&\href{https://huggingface.co/zai-org/GLM-4.7-Flash}{\path{zai-org/GLM-4.7-Flash}}&\parencite{zai2026glm47}&198k&TP1 DP8&1.0 / 0.95&\\
    \bottomrule
  \end{tabular}
  \caption{Checkpoint and deployment of every model of the pre- and post-training benchmark tables.
    Hugging Face checkpoints are linked.
    The configurations are read from the release definitions at the commit that rendered the model card.
    A model without a public reference is internal, and \enquote{this report} marks our own checkpoints.
  }
  \label{tab:baseline-models}
\end{table}

%% file: sections/09-appendix/F-i-don-t-know-evaluations.tex
\section{\enquote{I Don't Know} Evaluations}
\label{app:idk-evals}

This appendix complements \cref{sec:pre-training-eval-scores} with evaluations of how models distribute probability when they may abstain.

\paragraph{Motivation}
\textcite{kalai2025} argue that language models hallucinate partly because most benchmarks use binary grading, under which guessing always scores at least as well as abstaining.
They propose \enquote{open rubric} evaluations that state the scoring in the prompt, with wrong answers penalised by \(L=t/(1-t)\) points for a correctness threshold \(t\).
Our ternary score (\num[retain-explicit-plus]{+1} correct, \num{0} abstain, \num{-1} wrong) is the \(t=0.5\), \(L=1\) case of that scheme, although we do not state the rubric to the model.
The distributional correctness score (DCS) \parencite{burns2026when} goes further and scores the model's whole probability distribution over the answer options plus an \enquote{I don't know} (IDK) option: \(\mathrm{DCS}:=(p_c-p_w)(1-p_{\mathrm{IDK}})\), where \(p_c\), \(p_w\), and \(p_{\mathrm{IDK}}\) are the probability masses placed on correct, wrong, and IDK options, respectively.
DCS lies in \([-1,1]\) and abstaining scores \num{0}.
\textcite{burns2026when} reports that on half of \num{12} tested benchmarks, every one of the six models tested had negative DCS, so the negative averages below are expected.
Confidence-weighted accuracy (C-Accuracy) is \(p_c\) when the argmax option is correct and \num{0} otherwise.
It assumes that base-model probabilities are informative, and \textcite{kadavath2022} found that this largely holds for well-formatted multiple-choice questions.
\textcite{kumaran2026} give causal evidence that language models use internal confidence to decide whether to abstain, and steer internal confidence signals up or down to change how often models abstained.

\paragraph{Setup}
We evaluate \basemodel{2}, the checkpoints after its pre-training and mid-training stages, and the following baseline models of \cref{sec:pre-training-eval-scores}: \basemodel{1}, Olmo 3 7B, Olmo 3 32B, Apertus 70B, Nemotron 3 Nano 30B-A3B, and Qwen3.5 35B-A3B, together with GLM-4.5 Air 106B-A12B and Nemotron 3 Super 120B-A12B. We use \num{7} English IDK variants.
Each task appends an IDK option to an existing multiple-choice task and is scored by log-likelihood.
As in the main evaluations, all runs and metrics use eval-framework \parencite{eval_framework}.
The source tasks are:
\begin{itemize}
  \item \textbf{Grade-school science.} \ac{ARC}.
  \item \textbf{Commonsense.} GoldenSwag \parencite{chizhov2025goldenswag} (a filtered HellaSwag subset), \ac{PIQA}, and Winogrande \parencite{sakaguchi2020winogrande}.
  \item \textbf{Knowledge and expert questions.} \ac{MMLU}, \acs{MMLU}-Pro, and GPQA.
\end{itemize}

\paragraph{Results}
\Cref{tab:idk-means} gives the mean over the \num{7} tasks.
For readability, all scores are the raw metric times \num{100}.
Higher scores are better.
The largest model, Nemotron 3 Super 120B-A12B, is first on all metrics.
\basemodel{2} is second on C-Accuracy and DCS. A paired Bayesian bootstrap \parencite{rubin1981} of the per-question scores gives a posterior probability above \num{0.9999} that \basemodel{2} has a higher average DCS than every model besides Nemotron 3 Super 120B-A12B.

\begin{table}[tb]
  \centering
  \small
  \setlength{\tabcolsep}{3.5pt}
  \begin{tabular}{@{}lrrrr@{}}
    \toprule
    Model & Accuracy & C-Accuracy & Ternary & DCS \\
    \midrule
    \basemodel{1}
    & \num{37.07 \pm 0.65}
    & \num{22.53 \pm 0.36}
    & \num{27.59 \pm 1.33}
    & \num{-24.01 \pm 0.49} \\
    \model{2} (pre-training)
    & \num{41.08 \pm 0.63}
    & \num{26.25 \pm 0.38}
    & \num{36.42 \pm 1.29}
    & \num{-19.57 \pm 0.54} \\
    \model{2} (mid-training)
    & \num{42.19 \pm 0.63}
    & \num{27.21 \pm 0.38}
    & \num{38.78 \pm 1.29}
    & \num{-18.37 \pm 0.56} \\
    \basemodel{2}
    & \num{45.33 \pm 0.61}
    & \underLine{\num{29.84 \pm 0.40}}
    & \num{44.45 \pm 1.24}
    & \underLine{\num{-13.67 \pm 0.59}} \\
    Olmo 3 7B
    & \num{35.76 \pm 0.65}
    & \num{21.93 \pm 0.35}
    & \num{24.22 \pm 1.35}
    & \num{-24.84 \pm 0.48} \\
    Olmo 3 32B
    & \num{40.03 \pm 0.62}
    & \num{24.96 \pm 0.35}
    & \num{36.87 \pm 1.27}
    & \num{-19.84 \pm 0.49} \\
    Apertus 70B
    & \num{37.80 \pm 0.64}
    & \num{24.80 \pm 0.35}
    & \num{33.12 \pm 1.30}
    & \num{-23.32 \pm 0.49} \\
    Nemotron 3 Nano 30B-A3B
    & \num{43.37 \pm 0.61}
    & \num{28.56 \pm 0.36}
    & \num{43.31 \pm 1.25}
    & \num{-17.71 \pm 0.51} \\
    Qwen3.5 35B-A3B
    & \underLine{\num{45.74 \pm 0.59}}
    & \num{29.58 \pm 0.36}
    & \underLine{\num{47.20 \pm 1.20}}
    & \num{-14.53 \pm 0.52} \\
    GLM-4.5 Air 106B-A12B
    & \num{43.93 \pm 0.60}
    & \num{28.92 \pm 0.37}
    & \num{45.72 \pm 1.21}
    & \num{-15.20 \pm 0.53} \\
    Nemotron 3 Super 120B-A12B
    & \best{\num{51.20 \pm 0.57}}
    & \best{\num{34.99 \pm 0.39}}
    & \best{\num{53.44 \pm 1.14}}
    & \best{\num{-7.44 \pm 0.59}} \\
    \bottomrule
  \end{tabular}
  \caption{Mean \(\pm\) standard error over the \num{7} IDK tasks, times \num{100}.
    In each column, \textbf{bold} is best and \underline{underline} is second.
  }
  \label{tab:idk-means}
\end{table}

\acs{MMLU}-Pro gives the clearest single-task result. %
\basemodel{2} scores \num{-20.23}, ahead of every comparison model except Nemotron 3 Super 120B-A12B (\num{-11.02}).
The other models score between \num{-36.33} (GLM-4.5 Air 106B-A12B) and \num{-58.31} (Olmo 3 7B), with Apertus 70B and \basemodel{1} also at the bottom (\num{-57.56} and \num{-57.99}).
\acs{MMLU}-Pro has \num{10} options instead of \ac{MMLU}'s \num{4}, so it is harder to score well by guessing.
DCS adds a penalty for probability left on the wrong options, and \basemodel{2} moves more of that probability onto IDK or the correct option than any other model tested except Nemotron 3 Super.

%% file: sections/09-appendix/G-reasoning-mode-at-long-context.tex
\section{Reasoning Mode at Long Context}
\label{app:posttraining-long-context}

\begin{figure}[H]
  \centering
  \includegraphics{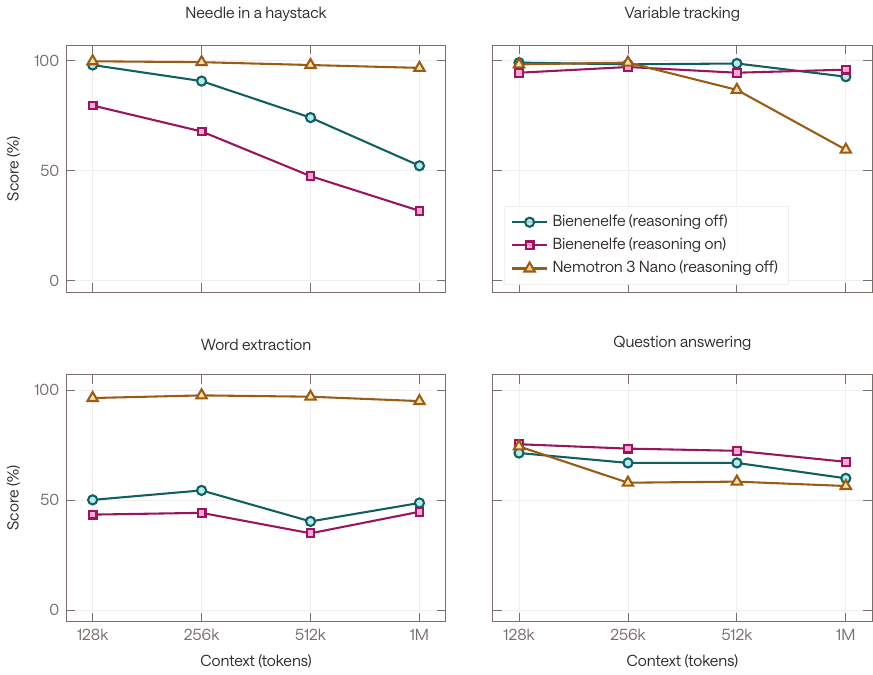}
  \caption{RULER scores of the post-trained \model{2} checkpoint with reasoning off and on, against Nemotron 3 Nano evaluated with thinking off.
    \model{2} is prompted through its chat template.
    Reasoning off allows up to \num{1}k generated tokens and reasoning on up to \num{16}k.
  }
  \label{fig:lc-ruler-1m}
\end{figure}

We evaluate the post-trained \model{2} checkpoint on the RULER tasks of \cref{sec:long-context-extrapolation} with reasoning on and off (\cref{fig:lc-ruler-1m}).
Reasoning off raises \ac{NIAH} recall, while reasoning on scores higher on \ac{QA} at the longest context lengths.
The Qwen3 technical report also finds lower RULER scores in thinking mode and suggests that extra reasoning can interfere with retrieval \parencite{yang2025qwen3}.
The reasoning-on run allows up to \num{16}k generated tokens and the reasoning-off run up to \num{1}k, so this comparison does not isolate the effect of reasoning alone.
Nemotron 3 Nano is evaluated with thinking off, as recommended for this task.
At \num{1}M tokens, \model{2} with reasoning off scores above Nemotron 3 Nano on \ac{VT} and \ac{QA}, but below it on \ac{NIAH} and \ac{WE}.

%% file: sections/09-appendix/H-supervised-fine-tuning.tex
\section{\actitle{SFT} (\acs*{SFT})}
\label{app:sft}

\suppressfloats[t]

\subsection{Long-Context \acs*{SFT}
  Data Ablation}%
\label{app:sft-lc-ablation}
\suppressfloats[t]
To test whether our long-context data improve reasoning over long documents, we fine-tune the proxy model with the ablation recipe (\cref{app:sft-ablation-recipes}) with and without them (\cref{tab:sft-lc-ablation}).
\Ac{VT} gains significantly.
The gains on AA-LCR (\(p = \num{0.19}\)) and LongBench Pro (\(p = \num{0.065}\)) are not significant, while LongBench Pro gains at every length and more at longer contexts.
Single-passage and single-key retrieval change little.

\begin{table}[htbp]
  \centering
  \small
  \begin{tabular}{lrlrlr}
    \toprule
    Benchmark & Base & \(\pm\) & Long-context mix & \(\pm\) & \(\Delta\) \\
    \midrule
    \multicolumn{6}{l}{\emph{Long context, \unit{\percent}}}\\RULER \acs*{QA} 128k (substring EM)&51.5&6.9&\textbf{53.0}&6.9&\(+1.5\)\\
    RULER \acs*{VT} 128k (recall)&17.6&6.8&\textbf{32.8}&9.0&\(+15.2\)\rlap{*}\\
    HELMET JsonKV 128k (substring EM)&\textbf{22.0}&8.2&21.0&8.0&\(-1.0\)\\
    AA-LCR (accuracy)&25.0&8.5&\textbf{33.0}&9.3&\(+8.0\)\\
    \midrule
    \multicolumn{6}{l}{\emph{LongBench Pro EN, \unit{\percent}}}\\Overall (8k--128k)&30.2&3.2&\textbf{33.5}&3.3&\(+3.2\)\\
    8k&38.6&7.8&\textbf{39.4}&7.6&\(+0.7\)\\
    16k&34.0&7.2&\textbf{34.8}&7.4&\(+0.9\)\\
    32k&30.0&7.2&\textbf{33.6}&7.5&\(+3.7\)\\
    64k  & 23.2 & 6.8 & \textbf{28.8} & 7.2 & \(+5.6\) \\
    128k & 25.3 & 6.9 & \textbf{30.7} & 7.4 & \(+5.4\) \\
    \bottomrule
  \end{tabular}
  \caption{Effect of the long-context \acs*{SFT} data on the proxy model with long-context extension, trained for \num{1000} steps (\num{16.8}\mbox{B} tokens).
    The long-context mix adds a \qty{5}{\percent} share of the synthesised long-context conversations.
    We evaluate on the long-context benchmarks of pre-training and post-training (\cref{tab:pre-training-ruler,tab:post-training-benchmarks}), on a subset of the RULER and HELMET tasks.
    Single runs; \(\pm\) is the half-width of the \qty{95}{\percent} \acs*{CI}, and * marks \(p < \num{0.05}\) in a paired test.
    The better arm is set in bold.
  }
  \label{tab:sft-lc-ablation}
\end{table}

\FloatBarrier

\subsection{Impact of German Reasoning Data}
\label{app:sft-german-share}
To measure how much German \ac{SFT} data the model needs to reason in German, we fine-tune \basemodel{2} with the ablation recipe of \cref{tab:sft-ablation-settings} on mixes that upsample all German data by the same factor, and report each score with its \qty{95}{\percent} \ac{CI} over the \num{30} problems (\cref{tab:german-share}).

\begin{table}[H]
  \centering
  \small
  \setlength{\tabcolsep}{4pt}
  \begin{tabular}{lcccccccccc}
    \toprule
    & \multicolumn{2}{c}{\textbf{German share} (\%)} & \multicolumn{6}{c}{\textbf{German AIME 2026} (\%)} & \multicolumn{2}{c}{\textbf{English AIME 2026} (\%)} \\
    \cmidrule(lr){2-3}\cmidrule(lr){4-9}\cmidrule(lr){10-11}
    &
    &
    & \multicolumn{2}{c}{score}
    & \multicolumn{2}{c}{correct \& German}
    & \multicolumn{2}{c}{loops}
    & \multicolumn{2}{c}{score} \\
    \cmidrule(lr){4-5}\cmidrule(lr){6-7}\cmidrule(lr){8-9}\cmidrule(lr){10-11}
    \textbf{Run}
    & all
    & math
    & mean
    & \qty{95}{\percent} \acs*{CI}
    & mean
    & \qty{95}{\percent} \acs*{CI}
    & mean
    & \qty{95}{\percent} \acs*{CI}
    & mean
    & \qty{95}{\percent} \acs*{CI} \\
    \midrule
    de-0
    & 0.0
    & 0.0
    & \textbf{\num[reset-text-series=false]{70.2}}\sigstars{**}
    & \ci{60.5}{79.9}
    & \num{42.3}\sigstars{*}
    & \ci{34.2}{50.3}
    & \textbf{\num[reset-text-series=false]{2.7}}\sigstars{***}
    & \ci{1.0}{4.4}
    & \num{64.0}
    & \ci{53.5}{74.4} \\
    de-1
    & 2.2
    & 0.4
    & \num{57.1}
    & \ci{44.8}{69.4}
    & \num{57.1}
    & \ci{44.8}{69.4}
    & \num{22.5}
    & \ci{13.3}{31.7}
    & \num{71.3}
    & \ci{61.6}{80.9} \\
    de-4
    & 8.2
    & 1.6
    & \num{59.0}
    & \ci{45.3}{72.6}
    & \num{59.0}
    & \ci{45.3}{72.6}
    & \num{18.3}
    & \ci{10.3}{26.4}
    & \textbf{\num[reset-text-series=false]{78.1}}
    & \ci{67.9}{88.3} \\
    de-16
    & 26.5
    & 5.0
    & \num{65.8}\sigstars{**}
    & \ci{54.6}{77.0}
    & \textbf{\num[reset-text-series=false]{65.8}}\sigstars{***}
    & \ci{54.6}{77.0}
    & \num{17.9}\sigstars{*}
    & \ci{9.4}{26.4}
    & \num{76.0}
    & \ci{66.2}{85.9} \\
    \midrule
    \multicolumn{11}{l}{\emph{Upsampled vs.\ no German data}} \\de-16 \(-\) de-0&&&\num[print-implicit-plus]{-4.4}&\ci{-10.6}{1.8}&\num[print-implicit-plus]{23.5}\sigstars{***}&\ci{14.8}{32.3}&\num[print-implicit-plus]{15.2}\sigstars{**}&\ci{6.8}{23.6}&\num[print-implicit-plus]{12.1}\sigstars{**}&\ci{5.6}{18.5}\\
    \bottomrule
  \end{tabular}
  \caption{Effect of the German share in the \acs*{SFT} mix on AIME 2026 (avg@16), with \qty{95}{\percent} \aclp*{CI} over the \num{30} problems.
    Runs sample every German group \(\times0\), \(\times1\), \(\times4\) or \(\times16\).
    Correct \& German counts rollouts that are correct and in German, and Loops rollouts that end in a repetition loop.
    The last row is the paired difference of de-16 and de-0.
    The best estimate of each column is set in bold.
    Stars mark Holm-corrected paired \(t\)-tests against de-1, or against zero in the last row: \(^{*}p < \num{0.05}\), \(^{**}p < \num{0.01}\), \(^{***}p < \num{0.001}\).
  }
  \label{tab:german-share}
\end{table}

\FloatBarrier

\pgfplotstableread[col sep=comma, row sep=\\, format=inline]{%
  validation_positions,rho_math_500_general0,rho_math_500_general33,math500_low,math500_high\\%
  20.0,0.32,0.26,84,91\\%
}\sftMergeFacts

\subsection{\Acs*{SFT} MergeMix Sweep}
\label{app:sft-merge-mix}
\begin{figure}[htbp]
  \centering
  \includegraphics{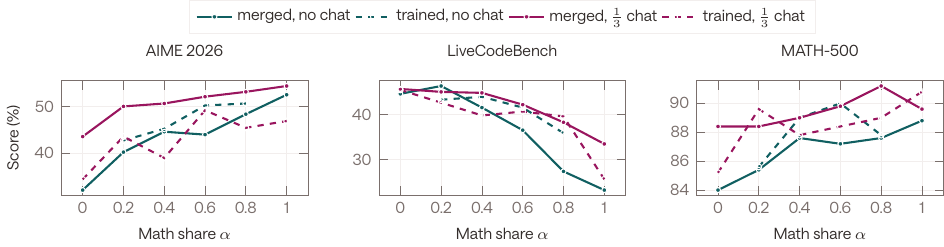}
  \caption{Scores of the merged (solid) and trained (dashed) weightings in the two math and code series, over the math share \(\alpha\) of the math and code part.
  }%
  \label{fig:sft-merge-validation}
\end{figure}

Following \textcite[Section~4.2]{wang2026mergemix}, we compare merged and trained mixes on math and code before the full sweep.
We train a math, a code and a chat specialist from the ablation base model of \cref{sec:mid-ablations}, each for \sftMergeValue{validation_positions}B packed tokens.
They train on all math, all code and all general chat datasets of the \ac{SFT} mix, each dataset sampled in proportion to its unique tokens (natural sampling, \cref{sec:mid-method}).
A weighting gives each specialist a share, and the shares sum to one.
For every weighting, we evaluate its merge, the specialists averaged with these shares, and a model trained for the same budget on data sampled with the same shares.
We use two series of weightings.
In the first, the math share \(\alpha\) runs from \num{0} to \num{1} in steps of \num{0.2}, and code takes the rest.
In the second, chat has a share of \(\frac{1}{3}\), and math and code split the remaining \(\frac{2}{3}\) in the ratio \(\alpha\) to \(1-\alpha\).
The single specialists at the ends of the first series are the same model either way, so the first series compares \num{4} weightings and the second \num{6}.
For each benchmark\footnote{The benchmarks are AIME 2026 \parencite{maa2026aime}, LiveCodeBench \parencite{jain2024livecodebench} and MATH-500 \parencite{lightman2023verify}.} we compute the Spearman correlation between merged and trained scores in each series (\cref{tab:sft-merge-validation,fig:sft-merge-validation}).
Merging and training order the weightings alike on AIME 2026 and LiveCodeBench.
They disagree on MATH-500 (\sftMergeValue{rho_math_500_general0} and \sftMergeValue{rho_math_500_general33}), where every weighting scores between \sftMergeValue{math500_low} and \sftMergePercent{math500_high}.
On this benchmark, the scores barely move with the weighting, so neither ordering carries information.

\begin{table}[htbp]
  \centering
  \small
  \begin{tabular}{lcccccc}
    \toprule
    Benchmark
    & \(\rho\), no chat
    & \(\rho\), \(\frac{1}{3}\) chat
    & Spread, no chat
    & Spread, \(\frac{1}{3}\) chat
    & Min
    & Max \\
    \midrule
    AIME 2026&$0.80$&$0.77$&$7.9$&$14.8$&$32.1$&$54.4$\\
    LiveCodeBench&$0.80$&$0.94$&$8.1$&$20.0$&$23.1$&$46.5$\\
    MATH-500&$0.32$&$0.26$&$4.4$&$5.6$&$84.0$&$91.2$\\
    \bottomrule
  \end{tabular}
  \caption{Agreement between merging and data mixing on math and code.
    \(\rho\) is the Spearman correlation between merged and trained scores, over the \num{4} weightings of the series without the chat specialist and the \num{6} of the series with it.
    Spread is the range of the trained scores in a series, in points; Min and Max bound all merged and trained scores, in percent.
  }%
  \label{tab:sft-merge-validation}
\end{table}

\Cref{tab:sft-merge-trained-weights} gives the cluster shares of every weighting trained as a data mix.
Within a cluster, the datasets keep their proportions in the reference mix.

\begin{table}[htbp]
  \centering
  \footnotesize
  \setlength{\tabcolsep}{4pt}
  \begin{tabular}{lrrrrrrrr}
    \toprule
    Cluster
    & \shortstack[r]{Baseline\\weighting}
    & \shortstack[r]{\(\kappa = 50\)\\draw 15}
    & \shortstack[r]{Refinement\\draw 13}
    & \shortstack[r]{Refinement\\draw 7}
    & \shortstack[r]{\(\kappa = 15\)\\draw 6}
    & \shortstack[r]{Refinement\\draw 5}
    & \shortstack[r]{Gradient-\\boosted trees}
    & \acs*{GP} \\
    \midrule
    Math I
    & \num{9.8}
    & \num{8.2}
    & \num{0.7}
    & \num{3.2}
    & \num{15.6}
    & \num{1.6}
    & --
    & -- \\
    Math II
    & \num{6.9}
    & \num{4.7}
    & \num{3.4}
    & --
    & \num{1.1}
    & \num{1.6}
    & --
    & -- \\
    Math III
    & \num{2.3}
    & \num{2.4}
    & --
    & \num{2.7}
    & --
    & --
    & --
    & -- \\
    Math proofs
    & \num{1.7}
    & \num{1.9}
    & --
    & --
    & --
    & --
    & --
    & -- \\
    Code I
    & \num{6.0}
    & \num{7.2}
    & \num{4.2}
    & \num{2.1}
    & --
    & --
    & --
    & -- \\
    Code II
    & \num{3.4}
    & --
    & \num{1.1}
    & --
    & \num{1.9}
    & \num{2.9}
    & --
    & -- \\
    Software engineering I
    & \num{4.6}
    & \num{7.9}
    & \num{11.3}
    & \num{2.2}
    & \num{2.5}
    & \num{3.7}
    & --
    & \num{2.8} \\
    Software engineering II
    & \num{2.1}
    & \num{3.4}
    & --
    & \num{1.3}
    & --
    & --
    & \num{1.0}
    & \num{2.6} \\
    Science I
    & \num{2.8}
    & \num{6.1}
    & \num{13.1}
    & \num{11.7}
    & \num{13.3}
    & \num{15.6}
    & \num{15.4}
    & \num{11.1} \\
    Science II
    & \num{3.6}
    & \num{1.1}
    & --
    & \num{7.2}
    & \num{1.3}
    & --
    & \num{9.4}
    & \num{11.6} \\
    General chat
    & \num{1.9}
    & \num{1.0}
    & --
    & --
    & --
    & \num{1.1}
    & \num{2.9}
    & \num{5.8} \\
    Instruction following I
    & \num{1.6}
    & \num{1.3}
    & \num{3.6}
    & --
    & --
    & --
    & \num{5.1}
    & \num{5.3} \\
    Instruction following II and safety
    & \num{2.1}
    & \num{3.2}
    & --
    & \num{0.7}
    & \num{1.1}
    & --
    & \num{6.5}
    & \num{6.8} \\
    Tool calling
    & \num{1.7}
    & \num{2.9}
    & \num{7.3}
    & \num{4.7}
    & \num{4.2}
    & \num{8.3}
    & \num{10.8}
    & \num{4.3} \\
    Interactive agents
    & \num{2.3}
    & \num{2.0}
    & \num{13.3}
    & \num{15.2}
    & --
    & \num{10.6}
    & \num{2.6}
    & \num{2.0} \\
    Terminal I
    & \num{1.8}
    & \num{7.8}
    & \num{1.6}
    & \num{3.2}
    & \num{0.9}
    & \num{4.8}
    & \num{7.1}
    & \num{9.0} \\
    Terminal II
    & \num{2.0}
    & \num{2.5}
    & \num{0.7}
    & --
    & \num{3.5}
    & --
    & --
    & \num{4.2} \\
    Long context
    & \num{20.4}
    & \num{9.4}
    & \num{9.5}
    & \num{24.7}
    & \num{23.8}
    & \num{37.8}
    & \num{16.1}
    & \num{26.1} \\
    German I
    & \num{12.2}
    & \num{8.2}
    & --
    & --
    & \num{11.4}
    & --
    & --
    & \num{5.7} \\
    German II
    & \num{10.9}
    & \num{18.7}
    & \num{30.2}
    & \num{20.9}
    & \num{19.2}
    & \num{12.0}
    & \num{22.9}
    & \num{2.6} \\
    \bottomrule
  \end{tabular}
  \caption{Cluster shares of the weightings trained as data mixes (\cref{tab:sft-merge-trained}), in percent of the packed tokens, with the weightings in the order of that table.
    A dash marks a cluster the sampler pruned below \qty{0.5}{\percent}.
  }
  \label{tab:sft-merge-trained-weights}
\end{table}

\FloatBarrier

\subsection{\acs*{SFT} Ablation Recipes}%
\label{app:sft-ablation-recipes}
Ablations train with a shorter recipe than the \model{2} \ac{SFT} runs, from either \basemodel{2} at target scale or the mid-trained proxy model (\cref{tab:sft-ablation-settings}).

\begin{table}[htbp]
  \centering
  \small
  \begin{tabular}{l >{\raggedright\arraybackslash}p{0.62\linewidth}}
    \toprule
    Setting & Value \\
    \midrule
    Base checkpoint
    & \basemodel{2} or mid-trained proxy model (\cref{sec:mid-ablations}) \\
    Peak learning rate
    & \num{0.3} (base model) or \num{0.5} (proxy) of pre-training in every group \\
    Context length
    & \recommendedContextLengthShort \\
    Batch size
    & \num{64} sequences, \num{16.8}\mbox{M} tokens per step \\
    Steps
    & \num{1000}, \num{16.8}\mbox{B} training tokens \\
    Learning-rate schedule
    & \ac{WSD}, warmup over \num{50} steps, linear decay over the last \num{200} steps to a tenth of the peak \\
    Parallelism
    & \acs*{FSDP} over \num{64} (base model) or \num{32} (proxy) GPUs \\
    Load balancing
    & Disabled (\cref{sec:load-balancing}) \\
    \bottomrule
  \end{tabular}
  \caption{Training settings of the \ac*{SFT} ablation recipe, for the two base checkpoints it starts from.
    A step covers a quarter of the tokens of a step of the \model{2} runs (\cref{tab:sft-settings}), and the recipe trains for a quarter of their steps.
  }
  \label{tab:sft-ablation-settings}
\end{table}

\begin{figure}[htbp]
  \centering
  \includegraphics{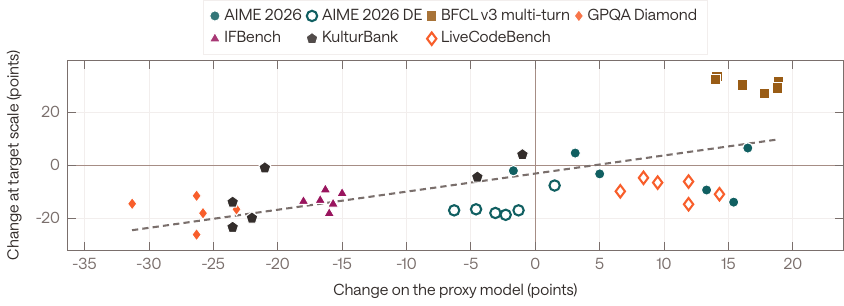}
  \caption{Score changes of \sftProxyValue{mixes} \acs*{SFT} configurations after \acs*{SFT} of the proxy model and of a model of the size of \model{2}: the change of every later configuration against the first on each benchmark, in points.
    A point in the upper-right or lower-left quadrant changes in the same direction at both scales.
    The dashed line is the least-squares fit of the target change on the proxy change, with \(R^2\) of \sftProxyValue{changes_r2} and a slope of \sftProxyValue{changes_slope}.
  }
  \label{fig:sft-proxy-transfer}
\end{figure}

\pgfplotstableread[col sep=comma, row sep=\\, format=inline]{%
  mixes,means_rho,changes,same_sign,changes_r2,changes_slope\\%
  7,0.00,45,33,0.43,0.69\\%
}\sftProxyFacts
To test whether \ac{SFT} ablations on the proxy model carry over to target scale, we train \sftProxyValue{mixes} \ac{SFT} configurations on both models, each with the identical dataset mix at both scales.
The configurations differ in the data mix, the assignment of reasoning effort and the loss masking.
The proxy predicts the direction in which a change of the configuration moves a benchmark at target scale in \sftProxyValue{same_sign} of \sftProxyValue{changes} cases (\cref{fig:sft-proxy-transfer}), but not the ranking of configurations by mean score (Spearman \(\rho = \sftProxyValue{means_rho}\)).

\FloatBarrier

\subsection{KulturBank}
\label{app:sft-kulturbank}
KulturBank is an in-house benchmark of \num{200} German questions on general knowledge about Germany.
The questions cover eight topics with \num{25} questions each: food and drink, geography, history, culture, nature, sport, technology and economy.
The model answers from its own knowledge, without tools or retrieved context.
It reasons briefly in German and then gives the answer as a short term in \texttt{\textbackslash boxed\{\}}.
Of the questions, \num{16} rest on a false premise and have no correct answer; for these, the model explains the error and answers \texttt{\textbackslash boxed\{Falsche Prämisse\}}.
If the model does not know an answer, the prompt asks it to answer \texttt{\textbackslash boxed\{Unbekannt\}} instead of guessing.
A response counts as correct if the boxed term matches one of the accepted answers of the question.

%% file: sections/09-appendix/I-reinforcement-learning.tex
\section{\actitle{RL}}
\label{app:rl}

\subsection{Environments}
\label{app:rl-environments}

\begin{table}[tbp]
  \centering
  \footnotesize
  \begin{tabular*}{\textwidth}{@{\extracolsep{\fill}}lS[table-format=6.0]rrS[table-format=1.5, group-digits=integer]@{}}
    \toprule
    Environment & {Train rows} & Steps & Tokens & {Weight} \\
    \midrule
    \multicolumn{5}{@{}l}{\emph{Math and Reasoning}} \\
    math                                             & 107064 & 1 & \num{128}k & 0.11555 \\
    math (de)                                        & 105514 & 1 & \num{128}k & 0.01095 \\
    reasoning                                        & 29191  & 1 & \num{64}k  & 0.04425 \\
    math proofs                                      & 8638   & 1 & --         & 0.00745 \\
    \addlinespace
    \multicolumn{5}{@{}l}{\emph{Code Generation}} \\
    code                                          & 10601 & 1 & \num{128}k & 0.11302 \\
    code (de)                                     & 11815 & 1 & \num{128}k & 0.01644 \\
    \addlinespace
    \multicolumn{5}{@{}l}{\emph{Software Engineering and Terminal Use}} \\
    SWE                                                                 & 184886 & 100 & \num{128}k & 0.25102 \\
    terminal use                                                        & 247193 & 100 & \num{128}k & 0.10952 \\
    \addlinespace
    \multicolumn{5}{@{}l}{\emph{Tool Use}} \\
    customer service                       & 7353  & 100 & \num{10000} & 0.02870 \\
    enterprise assistant                   & 81815 & 20  & \num{32}k   & 0.04776 \\
    enterprise assistant (de)              & 5215  & 20  & \num{32}k   & 0.02191 \\
    \addlinespace
    \multicolumn{5}{@{}l}{\emph{Retrieval and Long-Context Question Answering}} \\
    retrieval                                                                   & 16594 & 100 & \num{32}k   & 0.04151 \\
    long context                                                                & 1221  & 1   & \num{10000} & 0.00745 \\
    \addlinespace
    \multicolumn{5}{@{}l}{\emph{Instruction Following and Structured Output}} \\
    instruction following                                                     & 88761 & 1 & \num{10000} & 0.15849 \\
    \addlinespace
    \multicolumn{5}{@{}l}{\emph{Identity and Safety}} \\
    model identity                                    & 3012 & 1 & \num{2000} & 0.00734 \\
    copyright                                         & 1011 & 1 & \num{32}k  & 0.00328 \\
    copyright (de)                                    & 962  & 1 & \num{32}k  & 0.00328 \\
    safety refusal                                    & 4520 & 1 & \num{32}k  & 0.00110 \\
    political balance                                 & 6266 & 1 & --         & 0.00548 \\
    \addlinespace
    \multicolumn{5}{@{}l}{\emph{Hallucination Reduction}} \\
    hallucination reduction                               & 288333 & 1 & -- & 0.00550 \\
    \bottomrule
  \end{tabular*}
  \caption{Environments of the \acs*{RL} run with the number of rows in the train split, the maximum number of steps per sequence, the completion-token budget per sequence, and their sampling weight, grouped as in \cref{fig:rl-environment-mix}.
    Budgets come in five tiers, \num{2000}, \num{10000}, \num{32}k, \num{64}k and \num{128}k tokens, set per environment for B300 nodes; on A100 and H100 nodes every budget above \num{32}k is capped at \num{32}k to fit their shorter context.
    Environments with a dash have no completion-token budget and are only bound by the context limit of \num{256}k tokens.
    The weights are the sampling weights of the \model{2} \acs*{RL} run.
  }
  \label{tab:rl-environment-list}
\end{table}

\subsubsection{Reward Structure by Environment}
\label{app:rl-environments-rewards}

\paragraph{Code Generation}
The extracted program runs in a sandbox against every test case of the task with a \qty{20}{\second} timeout per test.
Its output is compared with the expected output, with a tolerance for floating-point numbers and case-insensitive yes/no answers, so a solution is not failed for whitespace or formatting alone.
The reward is the fraction of test cases passed.

\paragraph{Software-Engineering Agents}
The policy works in a per-task container through a coding harness with a shell tool, and its changes are extracted as a candidate patch.
The patch is applied in a fresh container and the task's tests are run: the reward is the fraction of fail-to-pass tests that pass, or zero if any pass-to-pass test fails.
Sequences are capped at \num{100} steps.

\paragraph{Software-Engineering Repair}
The search-and-replace edits are compared with the gold patch by the judge model, without executing tests; the reward is the judge's score, and a non-empty solution block is the \emph{Solution block} signal.

\paragraph{Terminal Tasks}
Terminal tasks are graded by the corpus's own test script, which runs \texttt{pytest} over the final machine state and writes a test report; the reward is the fraction of tests that pass.

\paragraph{Customer-Service Environment}
The reward compares the final database state with the expected one, checks the required actions and judges natural-language assertions about the conversation.

\paragraph{Retrieval-Augmented Question Answering}
The policy searches a document corpus with the tools of \cref{tab:retrieval-tools} and gives its final answer in the format that the prompt sets out.
The reward is the best token F1 of that answer against the gold answers, and a sequence that never calls a search tool scores zero.
The \emph{Boxed answer}, \emph{Answer format} and \emph{Citations} signals of \cref{tab:env-format-signals} score the form of the answer.
\Cref{sec:appendix-rag} gives the step caps and further details of the environment.

Where an environment has a token budget, exhausting it scores zero in the tool-use environments and triggers grading in the software-engineering environments.

\subsubsection{Long-Context Environment}\label{app:long-context-mcq-environment}
For every document, GLM-5.2 \parencite{zai2026glm52} writes two questions with six options, one to four of them correct.
A seeded plan per document fixes which part of the document each question draws on (beginning, middle or several distant passages) and how many options are correct, so the distribution of question shapes is not left to the generator.
A question has to combine facts from several passages, must not be answerable from the conversation turns alone, and uses true statements from elsewhere in the document as wrong options.

Each question is answered eight times by a \model{2} checkpoint at low reasoning effort and eight times by GLM-5.2, and enters the training set only if GLM-5.2 solves it at least once and \model{2} at most six times; where \model{2}'s attempts agree on a different answer or solve it at most once, Kimi K2.6 \parencite{moonshot2026kimik26} rules on the disputed options without seeing the key.
The reward is \num{1} only when the letters on the final \texttt{Answer:} line are exactly the correct set, and an empty response scores \num{0} instead of being dropped, so reasoning past the token budget on a long document is penalised.

\subsubsection{Hallucination Reduction Environments}\label{app:merlin-arthur-environments}

The hallucination reduction environments implement a Merlin--Arthur protocol \parencite{waeldchen2024merlin,deiseroth2025bounding} on various \ac{RAG} training datasets.
Arthur is the model that we train, i.e.\ the policy \(\pi_\theta\), and Merlin and Morgana are masking algorithms.
For a seed with context \(x\), question \(q\), and gold answer \(a\), they choose masks \(m \in \{0,1\}^C\) over the \(C\) maskable units (sentences) of \(x\).
The masks are ensured to hide similar amounts of context with some tolerance.
The masks optimise the following objective, where \(x \odot m\) denotes a context \(x\) in which parts have been masked out by \(m\):
\begin{equation}
  \hat{m}_{\text{Mer}}, \hat{m}_{\text{Mor}} = \operatorname*{arg\,max}_{m_{\text{Mer}}, m_{\text{Mor}}} \log \pi_\theta(a \mid x \odot m_{\text{Mer}}, q) + \log \bigl(1 - \pi_\theta(a \mid x \odot m_{\text{Mor}}, q)\bigr),
\end{equation}
such that \(|k_{\text{Mer}} - k_{\text{Mor}}| \le \delta\), where \(k = |m|\) is the number of units a mask hides.
Moreover, each mask hides at least one unit and leaves at least one visible, and the two masked contexts must differ.
We approximate both masks by first ranking the sentences by how much masking each one individually changes the log-probability of \(a\), teacher-forced on the same inference engine that produces the rollouts, and then removing the top (for Morgana) or bottom (for Merlin) \(k\).
During training, we recompute the masks from the policy at every step, so Morgana follows the features that make Arthur prone to hallucinations; no second model or human annotation is involved in generating these abstention examples.
Since Merlin and Morgana hide the same number of sentences within an adjustable tolerance, Arthur cannot tell them apart by how much is missing.

Arthur's training relies on rewards obtained from completions on the question \(q\) paired with Merlin, Morgana, and full (unmasked) contexts.
A completion \(y\) receives a reward \(R(y)\) that rewards Arthur for doing what its role asks and penalises a wrong answer on the full context and Merlin's:
\begin{equation*}
  R(y) \approx
  \begin{cases}
    1  & \text{if \(y\) answers \(a\) on the full context or Merlin's,}        \\
    -1 & \text{if \(y\) gives another answer on the full context or Merlin's,} \\
    1  & \text{if \(y\) abstains on Morgana's context,}                        \\
    0  & \text{otherwise.}
  \end{cases}
\end{equation*}
We additionally influence these values by small terms for token overlap with \(a\) and for abstaining on the full context or Merlin's.
An answer counts as correct on an exact match with the gold answer or one of its aliases; otherwise, an \ac{LLM} judge model decides whether it matches and whether it abstains.

Each of the three versions of a seed is rolled out \num{8} times, and each role is centred on its own mean; the advantage of rollout \(j\) is computed as
\begin{equation*}
  A_j \propto R(y_j) - \bar R_{\text{role}(j)},
\end{equation*}
where \(\bar R_{\text{role}(j)}\) is the mean training reward of the rollouts of its role scaled by a fixed weight per role and by the quality of the masks (for Merlin and Morgana only).
Arthur cannot gain advantage over correct answers by abstaining, since answers (and not abstentions) are expected on the full context or on Merlin’s.
A separate environment poses unanswerable questions with their full context and rewards only abstention.

The Merlin--Arthur protocol also gives us a hallucination measurement.
How often the model answers correctly on Merlin's context is its \emph{completeness}, and how often it refuses on Morgana's is its \emph{soundness}.
Together, they yield our \textbf{M/A grounding score}, a lower bound on how much of the answer provably came from the document.
The lower bound makes the grounding score conservative: zero means that it cannot be certified that the document was used, not that it went unused.
More details on the grounding score and the Merlin--Arthur protocol can be found in our paper \parencite{deiseroth2025bounding}.

\subsubsection{Retrieval-Augmented Question Answering}

\label{sec:appendix-rag}

Many customer applications answer questions from documents within the domain of their organisation, so the model has to search a corpus over several steps with the specific tools the application offers and answer from the retrieved text.
Three environments train this behaviour, all three are part of the \ac{RL} mix (\cref{sec:retrieval_environments}), and \path{retrieval} and \path{retrieval-de} also provide the rollouts that cold-start \ac{SFT} (\cref{sec:retrieval-sft}).
\path{retrieval} poses multi-hop questions over English data, \path{retrieval-de} poses German questions over German data, and \path{retrieval-honeypot}, referred to as Honeypot, poses synthetic multi-hop questions over specialised corpora.
This section describes the task, the tools they use, the question sets, the randomisation applied to each episode, the infrastructure that runs the environments during training, and the evaluation and results.

\paragraph{Task and Tools}
In each episode, the policy receives a question and a set of tools, searches the corpus, reads the documents it finds and gives a final answer in the format that the prompt specifies.
The answer is graded by F1 score based on word overlap with the gold answer and its aliases, together with the format signals of \cref{app:rl-environments-rewards}.
\Cref{tab:retrieval-tools} lists the tools, which fall into three groups: search tools that return passages, filter tools that restrict a search to documents with given metadata, and document access tools that read a document once a search has found it.
The retrieval tools take a query, a result count \(k\) (unless the session fixes it) and an optional \texttt{filters} argument that restricts them to documents matching given filter-column values; \texttt{grep} takes a regular expression and a match limit.
Each training episode draws a step budget of \num{10}, \num{25}, \num{50} or \num{100} steps with equal probability and states it in the prompt.
Validation episodes run with a fixed budget of \num{100} steps in \path{retrieval} and \path{retrieval-de} and \num{50} steps in \path{retrieval-honeypot}.

\begin{table}[tb]
  \centering
  \small
  \begin{tabular}{@{}>{\raggedright\arraybackslash}p{0.22\linewidth}>{\raggedright\arraybackslash}p{0.74\linewidth}@{}}
    \toprule
    Tool & Description \\
    \midrule
    \multicolumn{2}{@{}l}{\emph{Search tools}} \\
    \texttt{vector\_search}
    & Dense embedding search; returns the top-\(k\) chunks with score, document id, title and text. \\
    \texttt{lexical\_search}
    & BM25 keyword search over the same chunks. \\
    \texttt{hybrid\_search}
    & Merges the dense and lexical results by reciprocal-rank fusion. \\
    \texttt{grep}
    & Regular-expression search like \mbox{\texttt{grep -n}} over the documents BM25 retrieves for the pattern. \\
    \addlinespace
    \multicolumn{2}{@{}l}{\emph{Filter tools}} \\
    \texttt{list\_filter\_values}
    & Lists the values of a filter column with their document counts, most common first. \\
    \texttt{find\_documents}
    & Finds documents by exact filter values without a query; returns their ids, titles and filter values. \\
    \addlinespace
    \multicolumn{2}{@{}l}{\emph{Document access tools}} \\
    \texttt{open\_document}
    & Reads a document by id, in pieces with an offset or whole up to \num{8000} or \num{16000} characters. \\
    \texttt{open\_pages}
    & Reads up to four fixed-size pages (\num{2000} or \num{4000} characters) of a document by filename. \\
    \texttt{open\_window}
    & Reads \num{2000} or \num{4000} characters centred on a character position from a search hit. \\
    \texttt{read\_lines}
    & Reads numbered lines from a line given by a search hit, \num{50} by default and \num{200} at most. \\
    \bottomrule
  \end{tabular}
  \caption{Tools of the retrieval environments, grouped by role.
    The filter tools are served only when the corpus declares filter columns, and \texttt{hybrid\_search} and \texttt{grep} need a BM25 index over the document chunks.
  }
  \label{tab:retrieval-tools}
\end{table}

\paragraph{Factual Questions}
The questions of \path{retrieval} are taken from the training split of the multi-hop question-answering dataset MuSiQue~\parencite{trivedi2022musique}.
They are answered over an English pool, served through one of several index bundles that differ in their \acf*{ETL} parameters: the embedding model and chunk size for vector search, and the BM25 parameters for lexical search.
Before training, we apply Kimi-K2.7-Code~\parencite{moonshot2026kimik27code} to roll out every question in the environment, and these rollouts drive a filtering and repair pass.
A question is dropped when one of its supporting articles is missing from the corpus, when Gemma-4-31B~\parencite{gemma2026gemma4technicalreport} answers it correctly without retrieval, or when an \ac{LLM} judge labels it malformed, ambiguous or unsupported on the basis of its rollout.
Where the judge considers a question repairable, we apply GLM-5.2~\parencite{zai2026glm52} to rewrite an outdated gold answer or an underspecified question, and again use GLM-5.2 to check the rewritten question and answer.
The training split is restricted to an allow-list of questions that are not in the \ac{SFT} pipeline, and the test split is excluded from training in both \ac{SFT} and \ac{RL}; \cref{sec:retrieval-sft} describes the rollouts used in \ac{SFT}.
\path{retrieval-de} questions are translated and filtered versions of questions from MuSiQue, HotpotQA~\parencite{yang2018hotpotqa} and 2WikiMultihopQA~\parencite{ho2020twowiki}, processed by the same pipeline and answered over a German pool.

\paragraph{Honeypot Questions}
Honeypot covers nine specialised corpora (industrial product documentation, aerospace certification, financial regulation, EU and German law, scientific reports and economic development publications) that broadly cover our customers' related domains of interest, and every question is served over its own corpus.
Its questions are generated following the Context-1 recipe~\parencite{bashir2026context1}.
For each corpus, seed phrases taken mostly from document titles, supplemented by a hand-written list, are issued to the search tools, facts are extracted from the retrieved passages, and a question is generated for each fact, with that fact as its answer.
Each fact is then located at further positions in the corpus, and new facts from the surrounding text give follow-up questions in the same way.
Chaining an initial question with its follow-ups gives a multi-hop task in which answering hop \(k\) requires the answer to hop \(k-1\).
A hop is kept only if a search for its clue together with the previous answer finds the target document and a search for the clue alone does not, so that each hop depends on the one before.
Each corpus also declares document-level filter columns, and the search tools accept a filter on them, so that every retrieved document satisfies it.
The collection holds training and test questions over all nine corpora, and none of the training questions entered the \ac{SFT} pipeline.

\paragraph{Randomisation per Episode}
A deployed retrieval application fixes its tools, its prompt and its index, and these choices differ between applications.
To keep the model independent of any one setup, each episode draws its own configuration.
It serves one or more of the search tools and a single document access tool over one of the available index bundles of the corpus.
Its prompt preamble comes from a pre-configured set and fixes the answer format, the answer contents, whether tool use is required and the step budget.
Each session also samples a name, parameter names and a result format for every search tool, and the name of the document-id parameter (\texttt{document\_id}, \texttt{doc\_id} or \texttt{ref}).
The document access tool is always \texttt{read\_lines} when \texttt{grep} is served, and search results carry whichever address the access tool takes: a document id, a filename and page, a character position or a line number.

\paragraph{Infrastructure}
The retrieval tools in production applications rely on a set of services: a tool-calling server, an embedding-model server, a relational database, a vector-search engine and a lexical search index.
The environment mimics the production scenario by offering the policy the same tools, but simplifies their implementation to reduce the burden on the training infrastructure.
A tool handler inside the rollout process serves the tools of \cref{tab:retrieval-tools}, with the filter tools only where the corpus declares filter columns, and dispatches each tool call as a direct function call on a retrieval engine embedded within the process.
Therefore, the only traffic leaving the process is generation requests to the policy's inference endpoint.

For vector search, the index bundle fixes the query encoder: Qwen3-Embedding-0.6B~\parencite{zhang2025qwen3embedding} for every corpus, bge-large-en-v1.5~\parencite{xiao2023cpack} for the English pool and BGE-M3~\parencite{chen2024bgem3} for the German pool.
The encoder runs on the CPU inside the environment process through the \texttt{sentence-transformers} library~\parencite{reimers2019sentencebert}.
An asynchronous batcher coalesces the queries that parallel rollouts send at the same time into batched encodes on a small pool of search-worker threads.
All three encoders have fewer than \num{1}B parameters, which makes CPU-based deployments feasible at sufficient throughput.
Lexical search uses a memory-mapped BM25S index~\parencite{lu2024bm25s} over the same chunks as vector search.

\paragraph{Evaluation on Industry \acs*{RAG} Benchmarks} To check that agentic \ac{RAG} performance generalises beyond MuSiQue and Honeypot, we built the five Industry \ac{RAG} benchmarks of \cref{tab:customer-proxy-benchmarks}.
They are in-house benchmarks, not publicly available, and each is a proxy for a deployed customer assistant in a public or industrial use case; their results are the Industry \ac{RAG} rows of \cref{tab:post-training-evals-2}.
The benchmarks share no documents or questions with the \ac{SFT} and \ac{RL} data, although some Honeypot corpora come from the same domains.
Their questions are generated synthetically for the benchmarks and enter neither the \ac{SFT} nor the \ac{RL} data.
Gains on these benchmarks therefore reflect how well the training and the randomisation of tools, prompts and index bundles transfer to out-of-distribution real-world corpora and questions.
Their scores over the training runs of September 2026 are shown in \cref{fig:hill-climb-retrieval}.

\begin{table}[tb]
  \centering
  \footnotesize
  \begin{tabular}{@{}>{\raggedright\arraybackslash}p{0.18\linewidth}>{\raggedright\arraybackslash}p{0.33\linewidth}>{\raggedright\arraybackslash}p{0.43\linewidth}@{}}
    \toprule
    Name & Corpus & Tasks \\
    \midrule
    Semiconductors
    & Manufacturer manuals and datasheets for microcontrollers, power-management chips and transceivers.
    & Engineering questions on the documented chips, with reference answers. The set mixes questions on parameter values, feature availability, errata, procedures and design advice, and each question names its chip family. \\
    \addlinespace
    German Public Sector
    & German public-administration documents: laws, legal documents, IT-governance decisions and reports.
    & German fact questions on the documents, with reference answers. Each answer is one fact from one document group, such as an amount, a count or a legal obligation. \\
    \addlinespace
    Aerospace
    & Public aviation regulations and accident reports, mixed with energy, economy and news documents.
    & German questions on aviation regulation, certification, flight-simulator standards and accident investigations, with reference answers. \\
    \addlinespace
    Automotive Supplier
    & Synthetic product-development problem reports with tracker metadata, such as status, severity and discipline.
    & Questions on the reports and their metadata, with reference answers. Most questions filter reports by fields such as status, severity, author or date, and others search report content or count reports. \\
    \addlinespace
    Industrial Drive Technology
    & Manufacturer product documentation for motors, brakes, gear units and drive electronics.
    & German service questions on the documented products, with reference answers. The set mixes product-information and troubleshooting questions, and some questions give a serial number instead of the product name. \\
    \bottomrule
  \end{tabular}
  \caption{Corpus and task sources of the five Industry \acs*{RAG} benchmarks, in-house proxies for deployed customer assistants.}%
  \label{tab:customer-proxy-benchmarks}
\end{table}

\FloatBarrier

\subsection{Harness Tools}
\label{app:rl-harness-tools}

Some of the coding harnesses that are used have dependencies that might need to be installed at runtime.
The Terminus~2 harness of Terminal-Bench \parencite{merrill2026terminalbench} drives the agent through a \texttt{tmux} session inside the sandbox, and most task images do not ship \texttt{tmux}.
Installing it at rollout time would need network access in every sandbox and would repeat the installation thousands of times per optimiser step.
The executor therefore adds a statically linked \texttt{tmux} binary as a read-only bottom layer of every sandbox's OverlayFS union (\cref{sec:rl-sandboxes}), so every sandbox sees \texttt{tmux} without any change to its image.
The layer places its files at the paths where images keep their own, so an image that ships \texttt{tmux} shadows the layer's copy and the layer only fills gaps.
Static linking keeps the binary independent of the task image's C library.

\FloatBarrier
\subsection{Per-Token Logit-Gradient Norms}
\label{app:gradient-norms}

By the chain rule, the derivative of the policy-gradient loss term of \cref{eq:rl-btv-loss} splits as
\begin{equation*}
  \nabla_\theta \tilde{\ell}_t^{\mathrm{PG}} = \frac{\partial \tilde{\ell}_t^{\mathrm{PG}}}{\partial \mathbf{z}}\,
  \frac{\partial \mathbf{z}}{\partial \theta}
\end{equation*}
i.e.\ into the loss gradient with respect to the training policy logits \(\frac{\partial \tilde{\ell}_t^{\mathrm{PG}}}{\partial \mathbf{z}}\) and the model's Jacobian \(\partial \mathbf{z}/\partial \theta\).
All other quantities do not depend on \(\theta\) or are non-differentiable.

\subsubsection{Infinity-Norm of the Logit Gradient}\label{sec:rl-inf-norm}
Since applying the \ac{BTV} mask truncates the gradient, assume that the sampled token is not masked, i.e.\ it has \(k_t = 1\).
Then, \cref{eq:rl-btv-loss} simplifies to \(\tilde{\ell}_t^{\mathrm{PG}} = -A_t\,r_t\).
This policy-gradient term is linear in the training policy probability of the sampled token through the importance ratio, allowing us to determine the logit gradient norm in closed form.

\paragraph{Log-Softmax Jacobian}
We define \(\mathbf{p} = \operatorname{softmax}(\mathbf{z}) = \pi_\theta(\cdot\mid s_t)\) as the training probabilities with entries \(p_v\) over indices \(v\) of the token vocabulary.
Let \(w\) be a different vocabulary index, and \(y_t\) denote the sampled token, i.e.\ \(p_{\theta,t} = p_{y_t}\).
Splitting the logit gradient further around the normalised log-probabilities \(\log\mathbf{p}\) yields a partial derivative of the loss with respect to the log-probabilities, and a Jacobian of the log-softmax that maps the unnormalised logits to the log-probabilities.
The derivative of the log-softmax can be expressed as
\begin{equation*}
  \frac{\partial \log p_v}{\partial z_w} = \delta_{vw} - p_w,
\end{equation*}
where \(\delta_{vw}\) is the Kronecker delta.
Since the loss depends on the logits only through the log-probability \(\log p_{\theta,t}\) of the sampled token, we only need the corresponding row \( \nabla_{\mathbf{z}} \log p_{\theta,t} \) of the Jacobian.
This row is given as
\begin{equation}
  \mathbf{g}_t \coloneq \nabla_{\mathbf{z}} \log p_{\theta,t} = \mathbf{u}_t - \mathbf{p},
  \qquad
  g_{t,v} =
  \begin{cases}
    1 - p_{\theta,t}, & v = y_t,    \\
    -p_v,             & v \neq y_t.
  \end{cases}
  \label{eq:lse-g}
\end{equation}
where \(\mathbf{u}_t\) is the one-hot vector of \(y_t\).

\paragraph{Infinity-Norm of the Likelihood-Ratio Gradient}
Since \(\partial r_t/\partial \log p_{\theta,t} = r_t\), the log-derivative trick yields the gradient of the loss with respect to the sampled normalised log-probability as
\begin{equation*}
  \frac{\partial \tilde{\ell}_t^{\mathrm{PG}}}{\partial \log p_{\theta,t}} = -A_t\,r_t,
\end{equation*}
which is a scalar that multiplies the log-softmax Jacobian row \cref{eq:lse-g} in the logit gradient.
Using the homogeneity of the \(\infty\)-norm, the norm of the logit gradient is therefore
\begin{equation*}
  \bigl\lVert\nabla_{\mathbf{z}} \tilde{\ell}_t^{\mathrm{PG}} \bigr\rVert_\infty = \Bigl\lVert \frac{\partial \tilde{\ell}_t^{\mathrm{PG}}}{\partial \log p_{\theta,t}} \mathbf{g}_t \Bigr\rVert_\infty = \lvert A_t\rvert\,r_t\,\lVert\mathbf{g}_t\rVert_\infty.
\end{equation*}
Since the probabilities \(p_v\) are normalised, the entries of \(\mathbf{g}_t\) off the sampled token satisfy
\begin{equation*}
  \max_{v\neq y_t} \lvert - p_v \rvert \le \sum_{v \neq y_t} p_v = 1 - p_{\theta,t},
\end{equation*}
meaning that the maximum is always attained at the coordinate of the sampled token.
Thus the \(\infty\)-norm of the log-softmax Jacobian row simplifies to \(\lVert\mathbf{g}_t\rVert_\infty = 1 - p_{\theta,t}\), and the norm of the logit gradient is
\begin{equation*}
  \bigl\lVert\nabla_{\mathbf{z}} \tilde{\ell}_t^{\mathrm{PG}} \bigr\rVert_\infty = \lvert A_t\rvert\,r_t\,(1 - p_{\theta,t}) = \lvert A_t\rvert\,\frac{p_{\theta,t}\,(1 - p_{\theta,t})}{p_{\mu,t}} \eqcolon \lvert A_t\rvert\,G_t,
\end{equation*}
which is \cref{eq:rl-gradient-norm}.

\subsubsection{1- and 2-Norm Bounds}\label{sec:rl-norm-bounds}
Similarly, the \(1\)-norm of the logit gradient is
\begin{equation*}
  \bigl\lVert\nabla_{\mathbf{z}} \tilde{\ell}_t^{\mathrm{PG}} \bigr\rVert_1 = \lvert A_t\rvert\,r_t\, \bigl( \lvert 1 - p_{\theta,t} \rvert + \sum_{v \neq y_t} p_v \bigr)= 2\,\lvert A_t\rvert\,G_t,
\end{equation*}
and the \(2\)-norm is
\begin{equation*}
  \bigl\lVert\nabla_{\mathbf{z}} \tilde{\ell}_t^{\mathrm{PG}} \bigr\rVert_2 = \lvert A_t\rvert\,r_t\,\sqrt{(1 - p_{\theta,t})^2 + \sum_{v \neq y_t} p_v^2},
\end{equation*}
which is bounded by \((1 - p_{\theta,t})\sqrt{2}\,\lvert A_t\rvert\,r_t\).
Since \(\lVert\mathbf{x}\rVert_\infty \le \lVert\mathbf{x}\rVert_2 \le \lVert\mathbf{x}\rVert_1\) in general, the \(\infty\)-norm is also a lower bound on the \(2\)-norm.

\subsubsection{Pushforward Analysis}
\label{sec:rl-norm-pushforward}
The scaling \(c_t\) of \cref{sec:rl-norm-bound} bounds the gradient on the logits as a means to limit the change in training policy probability.
We next show this relationship by analysing a direct gradient step on the logits, as if they were free parameters, and the resulting effect on the probabilities.
In training, the step reaches the logits through the model's Jacobian \(\partial\mathbf{z}/\partial\theta\) and sums over all tokens of the batch, so this analysis describes the contribution of a single token.

To first order, a gradient step \(\delta\mathbf{z} = -\eta\,\nabla_{\mathbf{z}}\tilde{\ell}_t^{\mathrm{PG}}\) with step size \(\eta\) on the logits changes the probabilities through the Jacobian of the softmax,
\begin{equation*}
  \delta\mathbf{p} = \mathbf{J}\,\delta\mathbf{z},
  \qquad
  \mathbf{J} \coloneq \frac{\partial\mathbf{p}}{\partial\mathbf{z}} = \operatorname{diag}(\mathbf{p}) - \mathbf{p}\mathbf{p}^\top.
\end{equation*}
Inserting \(\nabla_{\mathbf{z}}\tilde{\ell}_t^{\mathrm{PG}} = -A_t\,r_t\,\mathbf{g}_t\) and \(\mathbf{p}^\top\mathbf{g}_t = p_{\theta,t} - \lVert\mathbf{p}\rVert_2^2\), the probability of the sampled token changes by
\begin{equation*}
  \delta p_{\theta,t} = \eta\,A_t\,r_t\,p_{\theta,t}\bigl(1 - 2p_{\theta,t} + \lVert\mathbf{p}\rVert_2^2\bigr) = \eta\,A_t\,r_t\,p_{\theta,t}\,\lVert\mathbf{g}_t\rVert_2^2.
\end{equation*}
With the \(2\)-norm bounds \((1 - p_{\theta,t})^2 \le \lVert\mathbf{g}_t\rVert_2^2 \le 2\,(1 - p_{\theta,t})^2\), this change satisfies
\begin{equation*}
  \eta\,\lvert A_t\rvert\,G_t\,p_{\theta,t}(1 - p_{\theta,t}) \le \lvert\delta p_{\theta,t}\rvert \le 2\,\eta\,\lvert A_t\rvert\,G_t\,p_{\theta,t}(1 - p_{\theta,t}),
\end{equation*}
and is thus proportional to \(G_t\) up to a factor of at most \(2\).
The scaling \(c_t\) clips \(G_t\) at \(G_{\max}\), and since \(p_{\theta,t}(1 - p_{\theta,t}) \le \tfrac{1}{4}\), a single token changes its own probability to first order by at most
\begin{equation*}
  \lvert c_t\,\delta p_{\theta,t}\rvert \le 2\,\eta\,\lvert A_t\rvert\,G_{\max}\,p_{\theta,t}(1 - p_{\theta,t}) \le \eta\,\lvert A_t\rvert\,G_{\max}/2.
\end{equation*}
\FloatBarrier
\subsection{Training Recipe Details}
\label{app:rl-training-recipe}

\subsubsection{FP8 Quantisation-Aware Training}
\label{app:rl-fp8}

\textcite{infixai2025unifiedfp8} find that the training--inference mismatch comes from the quantisation itself rather than from the FP8 kernels.
Following this finding, the trainer of \cref{sec:rl-fp8} applies the same FP8 rounding as the inference engine but computes in BF16, without FP8 kernels.
Even so, the median training--inference mismatch \(\lvert\log r_t\rvert\) is roughly three times that of BF16 serving.

The inference engine stores weights in FP8 (E4M3) in blocks of \numproduct{128 x 128}, each with an FP32 scale, quantises activations dynamically in groups of \numproduct{1 x 128}, and keeps the \ac{KV} cache in FP8.
The embeddings, the normalisation layers, the router gates and the \ac{LM} head remain in BF16, and both policies keep the \ac{LM} head's logits in FP32, which \textcite{minimax2025m1} find reduces the training--inference mismatch.
In the forward pass, the trainer rounds the weights and input activations of all other linear layers and of the routed experts to FP8 with the same block geometry and scales, and rounds the queries, keys and values to FP8 with unit scales, as the inference engine does.
\ac{RMS} normalisation \parencite{zhang2019rmsnorm} bounds the norm of its output by its largest gain times the square root of its dimension, for any input.
The QK normalisation \parencite{henry2020querykey} therefore bounds every query and key element, and \ac{RoPE} \parencite{su2024roformer}, where applied, preserves this bound as a rotation.
The normalisation before the attention block bounds the values together with the value weights.
For \model{2}, these bounds lie below the E4M3 maximum of \num{448} \parencite{micikevicius2022fp8}, so no input can overflow.
The backward pass uses the straight-through estimator \parencite{bengio2013estimating}.

\subsubsection{Staleness at Scale}
\label{app:rl-staleness}

The loop of rollout generation and optimiser steps (\cref{sec:rl-async}) runs at the speed of its slowest part, the trainer, inference or saturated environment resources such as sandboxes, and is balanced when inference produces groups as fast as the trainer consumes them.
Staleness is counted in optimiser steps, but rollouts take wall-clock time.
The staleness of a group is therefore roughly its rollout duration divided by the step time of the trainer, so any speedup of the trainer at a fixed generation latency makes groups staler.
Suppose a balanced loop takes \qty{2}{\minute} per optimiser step and a code rollout takes \qty{60}{\minute}, so code groups arrive with a staleness of \num{30}.
Doubling both the trainer and the inference GPUs at the same batch size halves the step time to about \qty{1}{\minute} and doubles inference throughput, so the loop stays balanced.
The decode speed per request is unchanged, however, so a code rollout still takes \qty{60}{\minute} and now arrives with a staleness of \num{60}.
In some of our runs, code groups exceeded the staleness limit in this way after we scaled up the run.
Scaling the batch size with the GPU count also keeps the step time, and hence the staleness, constant.
\FloatBarrier
\subsection{\acs*{RL} Hyperparameters}
\label{app:rl-hyperparameters}

This section lists the hyperparameters of the main \ac{RL} run.
\Cref{tab:rl-hparams-algorithm} covers the algorithmic parameters of \cref{sec:rl-environment-mix,sec:rl-training-recipe}, and \cref{tab:rl-hparams-technical} covers hardware, precision and optimisation.
\Cref{tab:rl-opsd-failures} lists the failure classes of the targeted \ac{OPSD}, and \cref{app:rl-opsd} gives its hyperparameters.

\begin{table}[htbp]
  \centering
  \small
  \begin{tabular}{@{}ll@{}}
    \toprule
    Parameter & Value \\
    \midrule
    \multicolumn{2}{@{}l}{\itshape Rollouts} \\
    Group size \(\lvert\mathcal{G}\rvert\)       & \num{8} \\
    Groups per batch \(B\)                       & \num{256} \\
    Sampling                                     & temperature \num{1}, top-\(p\) \num{1}, no top-\(k\) \\
    Maximum rollout steps                        & \num{100} \\
    Reasoning-effort weights (low, medium, high) & \num{0.1}, \num{0.1}, \num{0.8} \\
    Probability of preserving reasoning traces   & \num{0.5} \\
    \midrule
    \multicolumn{2}{@{}l}{\itshape Curriculum} \\
    Target bucket weights \(c_b\) (hard, trainable, solved) & \num{0.08}, \num{0.9}, \num{0.02} \\
    Pass rate of hard and solved prompts                    & \(< \num{1e-5}\) and \num{1} \\
    Default pass rate                                       & \num{0.5} \\
    Policy updates until \(\gamma_b\) saturates             & \num{50} \\
    \midrule
    \multicolumn{2}{@{}l}{\itshape Group filtering} \\
    Reward standard deviation below which rewards count as equal & \num{1e-3} \\
    Replacement attempts after a dropped group                   & \num{2} \\
    Maximum staleness                                            & \num{42} policy updates \\
    \midrule
    \multicolumn{2}{@{}l}{\itshape Policy loss} \\
    \acs*{BTV} thresholds \(\epsilon_+\), \(\epsilon_-\) & \num{0.2}, \num{0.2} \\
    Squared log-ratio coefficient \(\tau_{\mathrm{KL}}\) & \num{1e-3} \\
    Gradient-norm bound \(G_{\max}\)                     & \num{2} \\
    \midrule
    \multicolumn{2}{@{}l}{\itshape Format signals} \\
    Aggregation norm \(p\)                         & \num{2} \\
    Global multiplier \(m_{\min}\)                 & \num{0.5} \\
    Minimum factor \(f_k\)                                       & \num{0.5}, \num{0} for
    \path{non_empty_output}                                                                                \\
    Completion-budget threshold (low, medium, high) & \num{0.2}, \num{0.4}, \num{0.8} of the budget \\
    Completion-budget sigmoid steepness             & \num{6} \\
    \bottomrule
  \end{tabular}
  \caption{Algorithmic hyperparameters of the main \acs*{RL} run.
    Symbols follow \cref{sec:rl-environment-mix,sec:rl-training-recipe}.
  }
  \label{tab:rl-hparams-algorithm}
\end{table}

\begin{table}[htbp]
  \centering
  \small
  \begin{tabular}{@{}ll@{}}
    \toprule
    Parameter & Value \\
    \midrule
    \multicolumn{2}{@{}l}{\itshape Hardware and parallelism} \\
    GPUs                                                     & NVIDIA B300, \num{256} on \num{32} nodes \\
    Trainer                                                  & \num{96} GPUs, \acs*{FSDP} without further parallelism \\
    Policy inference                                         & \num{144} single-GPU vLLM replicas \\
    Judge inference                                          & \num{16} single-GPU vLLM replicas, frozen \\
    Rollout threads per inference replica                    & \num{45} \\
    Share of inference capacity used for validation          & \qty{5}{\percent} \\
    \midrule
    \multicolumn{2}{@{}l}{\itshape Precision and memory} \\
    Inference weights and \acs*{KV} cache                & FP8 \\
    Trainer                                              & FP8 quantisation-aware training \\
    \acs*{LM} head                                       & FP32 for both policies \\
    Exported checkpoints                                 & BF16 and FP8 \\
    Activation checkpointing                             & full \\
    \midrule
    \multicolumn{2}{@{}l}{\itshape Optimisation} \\
    Optimiser steps
    & \num{1000} \\
    Attention and expert weights
    & Muon, spectral, Nesterov \\
    Embedding, norm scales, \acs*{LM} head and router
    & Adam, \(\beta_1 = 0.9\), \(\beta_2 = 0.95\), \(\epsilon = \num{1e-15}\) \\
    Weight decay
    & none \\
    Learning rate: attention and expert weights
    & \num{1e-5} \\
    Learning rate: embedding and norm scales
    & \num{2e-5} \\
    Learning rate: \acs*{LM} head and router
    & \num{2.5e-7} \\
    Learning-rate warmup
    & \num{50} steps, linear \\
    Learning-rate decay ratio
    & \num{0} \\
    Global gradient-norm clipping
    & \num{1.0} \\
    \acs*{MoE} load-balancing loss
    & none \\
    \midrule
    \multicolumn{2}{@{}l}{\itshape Sequences and asynchrony} \\
    Context length                                           & \num{256}k tokens \\
    Packed tokens per trainer rank                           & \num{393216} \\
    Weight push                                              & after every optimiser step \\
    Replay buffer capacity                                   & \num{512} groups in \num{8} shards \\
    \midrule
    \multicolumn{2}{@{}l}{\itshape Sandboxes} \\
    Executors
    & one per node on \num{31} nodes, \num{112} CPUs each, shared at \texttt{nice} \num{10} \\
    Writable layer
    & \texttt{tmpfs}, \qty{8}{\gibi\byte} per sandbox \\
    Address space per command
    & \qty{16}{\gibi\byte} \\
    Processes per sandbox
    & \num{4096}, then the sandbox is killed \\
    Command timeout
    & \qty{600}{\second} \\
    Output per command
    & \qty{1}{\mebi\byte} per stream \\
    Network
    & loopback; egress proxy with an allowlist where dependencies are installed \\
    \bottomrule
  \end{tabular}
  \caption{Technical and optimisation hyperparameters of the main \acs*{RL} run.
    The optimiser groups follow \cref{tab:bienenelfe-optimizer}.
  }
  \label{tab:rl-hparams-technical}
\end{table}

\FloatBarrier
\begingroup
\footnotesize
\setlength{\tabcolsep}{4pt}
\newcommand{\familylabel}[2]{\multirow{#1}{*}{\rotatebox[origin=c]{90}{#2}}} \newcommand{\categorylabel}[2]{\multirow{#1}{*}{\rotatebox[origin=c]{90}{\scriptsize\shortstack{#2}}}}
\begin{longtable}{@{}cc p{0.28\textwidth}p{0.52\textwidth}@{}}
  \toprule
  Family & Category & Failure class & Gist of the remedial hint \\
  \midrule
  \endfirsthead
  \toprule
  Family & Category & Failure class & Gist of the remedial hint \\
  \midrule
  \endhead
  \bottomrule
  \endfoot
  \bottomrule
  \caption{Failure classes of the targeted \acs*{OPSD}, grouped by the categories of \cref{fig:rl-opsd,fig:rl-opsd-detections} and, within a category, in order of priority.
    The hints are shortened for readability.
  }
  \label{tab:rl-opsd-failures} \\
  \endlastfoot
  \familylabel{16}{Tool use} & \categorylabel{8}{invalid tool call}
  & \path{unparseable_tool_call}
  & A tool call is a single valid JSON object inside the tool-call tags that matches the tool's schema. \\
  &
  & \path{wrong_argument_type}
  & Arguments use the exact JSON types the tool declares. \\
  &
  & \path{tool_contract_error}
  & Call tools with exactly the declared argument names and types. \\
  &
  & \path{entity_already_exists}
  & Re-read the task for the exact identifier and for whether the item is to be created or updated. \\
  &
  & \path{precondition_failed}
  & Check the entity's current state and complete only the prerequisite steps the task requires. \\
  \cmidrule(l){2-4}
  & \categorylabel{8}{invented tool or reference}
  & \path{unknown_tool_name}
  & Only the tools listed in the system prompt can be called, and the hint lists their signatures. \\
  &
  & \path{entity_not_found}
  & Take identifiers from the task or earlier tool results, or look them up with a search tool. \\
  &
  & \path{unknown_url}
  & Open only URLs that an earlier search result shows verbatim, or that result's number, and never guess a URL. \\
  &
  & \path{result_out_of_range}
  & A result number is the bracketed index of a search result, counted over all searches so far. \\
  \midrule
  \familylabel{27}{Terminal} & \categorylabel{10}{syntax errors}
  & \path{unparseable_tool_call}
  & A tool call is one JSON object with escaped string arguments, one call per block. \\
  &
  & \path{key_name_with_newline}
  & A key name such as \texttt{C-c} acts as that key only without a trailing newline. \\
  &
  & \path{missing_tool_call}
  & Every response must contain a tool call through the tool-call channel. \\
  &
  & \path{shell_quoting}
  & Shell commands must be complete and correctly quoted, and code with its own quotes belongs in a file. \\
  &
  & \path{inline_script_syntax_error}
  & Check code passed inline to the interpreter for typos and altered quotes, and put multi-line code in a file. \\
  &
  & \path{edited_file_syntax_error}
  & A file the model wrote earlier that does not parse breaks whatever runs or imports it, so fix it first. \\
  \cmidrule(l){2-4}
  & \categorylabel{10}{wrong tool or target}
  & \path{unknown_tool_name}
  & Only the offered tools can be called, and the hint lists their signatures. \\
  &
  & \path{tool_contract_error}
  & Call tools with exactly the declared argument names and types. \\
  &
  & \path{edit_mismatch}
  & The text to replace must match the current file exactly and once, so re-read the file first. \\
  &
  & \path{noop_edit}
  & An edit that reproduces the original text changes nothing. \\
  &
  & \path{invalid_line_range}
  & Line numbers must exist in the current version of the file. \\
  &
  & \path{create_on_existing_file}
  & Edit an existing file instead of creating it again. \\
  &
  & \path{no_changes_submit}
  & Grading uses the changed repository files, and nothing has been changed yet. \\
  &
  & \path{git_unavailable}
  & Work inside a git repository and keep copies of files before editing them. \\
  \cmidrule(l){2-4}
  & \categorylabel{7}{output or\\time limits}
  & \path{typed_into_busy_terminal}
  & Keystrokes reach the shell only once its prompt is the last line on screen, so wait or interrupt first. \\
  &
  & \path{typed_ahead_in_batch}
  & Keystrokes after a command that outlives its duration feed that command, so give long commands enough time. \\
  &
  & \path{command_timeout}
  & Bound long commands with a timeout, or shrink the work instead of rerunning the command unchanged. \\
  &
  & \path{output_truncated}
  & Long tool outputs are cut, so read files in slices and filter command output. \\
\end{longtable}
\endgroup
\FloatBarrier
\subsection{Further \acs*{OPSD} Analysis}
\label{app:rl-opsd}
\pgfplotstableread[col sep=comma, row sep=\\, format=inline]{%
  warmup,early_end,window,smooth_window,tokens_early,tokens_late\\%
  25,75,50,25,0.10,0.034\\%
}\rlOpsdFacts
\pgfplotstableread[col sep=comma, row sep=\\, format=inline]{%
  envs,detectors,bin,edge,early_first,early_last,late_first,late_last,groups_early,groups_late,fail_early,fail_late,fail_step_drop,fail_falls,share_terminal_limits_late\\%
  8,24,25,50,0,49,951,1000,13.0,7.2,4.21,2.56,31,6,59\\%
}\rlOpsdCountsFacts
\pgfplotstableread[col sep=comma, row sep=\\, format=inline]{%
  tool_use_envs\\%
  3\\%
}\rlOpsdTracesFacts

The main run weights the \ac{OPSD} loss of \cref{eq:rl-objective} with a coefficient \(\lambda_{\mathrm{OPSD}}=\num{1}\), uses the forward \ac{KL} divergence over the full vocabulary and flags at most \num{5} steps per sequence.
It selects the all-wrong groups, which the group-relative objective drops for low reward variance and whose highest environment reward is below \num{0.2}; \cref{tab:rl-opsd-failures} lists its failure classes.
Counts come from the rollout traces, where all \rlOpsdCountsValue{detectors} detectors score each assistant turn of each training sequence of the \rlOpsdCountsValue{envs} \acs*{OPSD} environments.
The trainer logs only the first matching detector on at most \num{5} turns per \acs*{OPSD} group, so its logs serve only for the shares of trained groups and tokens, the teacher agreement and the entropy.
Early and late values span the first and last \rlOpsdCountsValue{edge} steps on the traces and, to compensate for staleness, steps \rlOpsdValue{warmup} to \rlOpsdValue{early_end} and the last \rlOpsdValue{window} steps in the logs.

\begin{figure}[tbp]
  \centering
  \includegraphics[width=\linewidth]{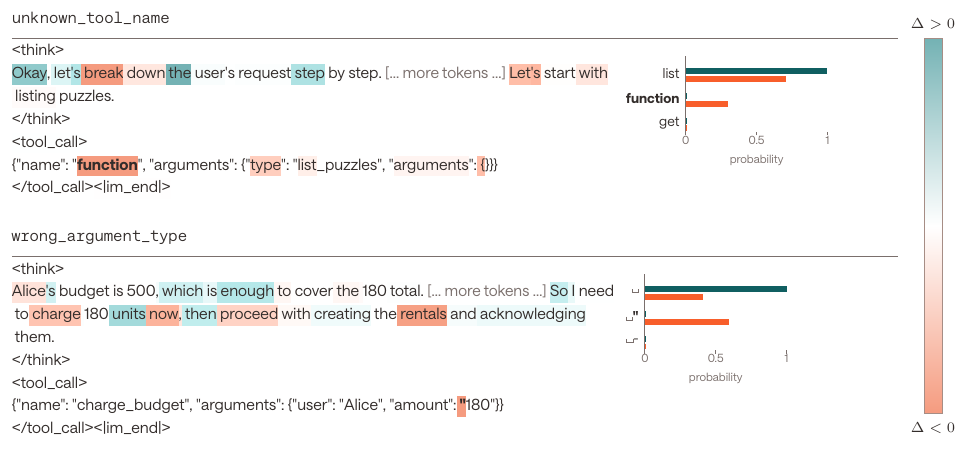}
  \caption{ Teacher versus student on a flagged turn for two different failure classes.
    Token colour is given by \(\Delta = \log p_{\text{teacher}} - \log p_{\text{student}}\) of the turn the student sampled, rescored under the hinted context and the bare context.
    The hint down-weights a token with \(\Delta < 0\) and up-weights one with \(\Delta > 0\).
    The reasoning is shortened.
  }
  \label{fig:rl-opsd-hint-examples}
\end{figure}

\begin{figure}[tbp]
  \centering
  \includegraphics{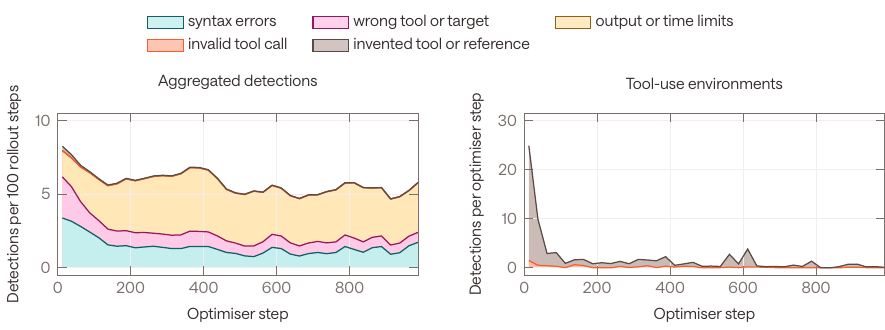}
  \caption{ Failures that the targeted \acs*{OPSD} detects over the \model{2} \acs*{RL} run, counted on its rollout traces in bins of \rlOpsdCountsValue{bin} steps.
    Left: Detections per \num{100} rollout steps in all \rlOpsdCountsValue{envs} \acs*{OPSD} environments.
    When controlling for rollout steps, error detections decrease during training, though not monotonically.
    Right: Detections over the course of training for the \rlOpsdTracesValue{tool_use_envs} tool-use environments.
    In contrast to the terminal environments (\cref{fig:rl-opsd}), the tool-use environments quickly converge towards almost no detections.
  }
  \label{fig:rl-opsd-detections}
\end{figure}

\paragraph{Detected Failures}
\Cref{tab:rl-opsd-envs} shows how detections per training sequence fall in \rlOpsdCountsValue{fail_falls} of the \rlOpsdCountsValue{envs} environments, with the average decreasing from \rlOpsdCountsValue{fail_early} to \rlOpsdCountsValue{fail_late}.
The left of \Cref{fig:rl-opsd-detections} shows that detections per rollout step also fall, by \rlOpsdCountsPercent{fail_step_drop}, though not monotonically.
The right of \Cref{fig:rl-opsd-detections} shows how, over the course of training, errors related to invented tools and references rapidly decrease and eventually are trained away.
In contrast, output or time limits rise, making up \rlOpsdCountsPercent{share_terminal_limits_late} of the late detections.
Nearly all of them are \path{output_truncated}, which fires when the harness cuts a long command output, and they
rise in the software-engineering environments, where the curriculum keeps drawing long tasks on large repositories.
\Cref{tab:rl-opsd-envs} shows how \ac{OPSD}-annotated sequences consistently receive lower environment rewards on average, suggesting that the error detection successfully finds sequences where the model struggles.
\Cref{fig:rl-opsd-hint-examples} visualises the differences between teacher and policy token probabilities for two exemplary flagged turns, finding that the teacher focuses on the turn's error in both cases.

\begin{figure}[tbp]
  \centering
  \includegraphics{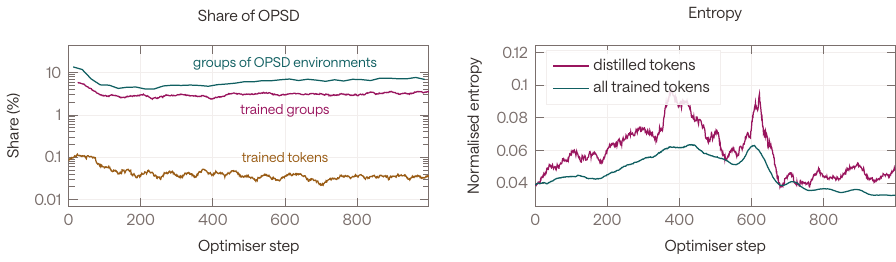}
  \caption{ Left: Total share of \acs*{OPSD} groups in eligible environments, total share of \acs*{OPSD} groups among all groups added to the replay buffer, and total share of training tokens using the distillation loss.
    All three shares drop at the start of training and then stay relatively consistent.
    Right: Normalised entropy compared between all trained tokens and distilled tokens.
    Distilled tokens consistently have a higher entropy than average tokens.
    Values are averaged over \rlOpsdValue{smooth_window} steps.
  }
  \label{fig:rl-opsd-signal}
\end{figure}

\paragraph{Share of the Training Signal and Entropy}
The left of \cref{fig:rl-opsd-signal} shows how the share of training groups in the \acs*{OPSD} environments that become \acs*{OPSD} groups falls from \rlOpsdCountsPercent{groups_early} to \rlOpsdCountsPercent{groups_late} during training, and how only \rlOpsdPercent{tokens_early} to \rlOpsdPercent{tokens_late} of tokens are trained with the distillation signal.
The right of \cref{fig:rl-opsd-signal} overlays the entropy of the \acs*{OPSD} tokens with the entropy of all trained tokens, as shown in the left panel of \cref{fig:rl-dynamics}.
Both follow the same rough shape, but the normalised entropy stays higher on the distilled tokens.

\begin{table}[tbp]
  \centering
  \small
  \begin{tabular}{llrrrcc}
    \toprule
    &  & \multicolumn{3}{c}{Detections per sequence} & \multicolumn{2}{c}{Reward by detection} \\
    \cmidrule(lr){3-5}\cmidrule(lr){6-7}
    Environment
    & Family
    & Steps \rlOpsdCountsValue{early_first}--\rlOpsdCountsValue{early_last}
    & Steps \rlOpsdCountsValue{late_first}--\rlOpsdCountsValue{late_last}
    & Change (\unit{\percent})
    & With
    & Without \\
    \midrule
    General agent        & Tool use & 0.16 & 0.17 & \cellcolor{AABrandPrimary!1!white}\num{3} & 0.57 & 0.68 \\
    Customer-service     & Tool use & 1.45 & 0.02 & \cellcolor{AADarkTeal!50!white}\num{-99}  & 0.64 & 0.73 \\
    Workplace assistant  & Tool use & 0.04 & 0.03 & \cellcolor{AADarkTeal!17!white}\num{-38}  & 0.59 & 0.78 \\
    Nemotron terminal    & Terminal & 5.09 & 0.66 & \cellcolor{AADarkTeal!50!white}\num{-87}  & 0.68 & 0.74 \\
    SWE-Gym              & Terminal & 5.21 & 3.05 & \cellcolor{AADarkTeal!19!white}\num{-41}  & 0.42 & 0.44 \\
    SWE-smith            & Terminal & 4.41 & 4.74 & \cellcolor{AABrandPrimary!3!white}\num{7} & 0.73 & 0.87 \\
    SWE-smith non-Python & Terminal & 5.11 & 4.47 & \cellcolor{AADarkTeal!5!white}\num{-13}   & 0.78 & 0.92 \\
    TMAX                 & Terminal & 5.72 & 1.04 & \cellcolor{AADarkTeal!50!white}\num{-82}  & 0.72 & 0.88 \\
    \bottomrule
  \end{tabular}
  \caption{Targeted \acs*{OPSD} per environment of the \model{2} \acs*{RL} run.
    Detections are the failures the detectors flag per training sequence over the first and the last \rlOpsdCountsValue{edge} optimiser steps, and the change is the late value relative to the early one.
    The reward is the mean environment reward of the training sequences with and without a detected failure over all steps.
    In most environments, detections decrease over the course of training, while sequences with detected errors receive a lower average reward.
  }
  \label{tab:rl-opsd-envs}
\end{table}

\clearpage
\ifbienenelfesectionleftout\acsetup{pages/display=none}\fi
\begingroup
\setlist[description]{itemsep=0.2\AABase}
\printacronyms[name={Acronyms}, preamble={\backmatterentry{Acronyms}}]
\endgroup
\clearpage